\documentclass{article}

\usepackage[preprint]{neurips_2026}

\usepackage{amsmath,amsfonts,bm}

\def\eqref#1{equation~\ref{#1}}

\def\1{\bm{1}}

\DeclareMathAlphabet{\mathsfit}{\encodingdefault}{\sfdefault}{m}{sl}
\SetMathAlphabet{\mathsfit}{bold}{\encodingdefault}{\sfdefault}{bx}{n}

\usepackage[utf8]{inputenc}
\usepackage[T1]{fontenc}
\usepackage{url}
\usepackage{booktabs}
\usepackage{amsfonts}
\usepackage{nicefrac}
\usepackage{microtype}
\usepackage{xcolor}
\usepackage{float}
\usepackage[normalem]{ulem}

\usepackage{amsmath}
\usepackage{amssymb}
\usepackage{mathtools}
\usepackage{amsthm}

\usepackage{graphicx}
\usepackage{wrapfig}
\usepackage[font=small,labelfont=bf]{caption}
\usepackage{subcaption}
\usepackage{multicol}
\usepackage{multirow}
\usepackage{tabularx}
\usepackage{array}
\usepackage{colortbl}

\usepackage{enumitem}
\usepackage{listings}
\usepackage{listingsutf8}
\usepackage[ruled,vlined,linesnumbered]{algorithm2e}
\usepackage{minted}

\usepackage[most]{tcolorbox}

\usepackage[compact]{titlesec}

\usepackage[pagebackref=true,colorlinks=true]{hyperref}

\definecolor{scalarDark}{HTML}{7B2D26}
\definecolor{scalarMid}{HTML}{9E3A32}
\definecolor{scalarLight}{HTML}{C4847E}
\definecolor{scalarBg}{HTML}{FCF3F1}
\definecolor{scalarBorder}{HTML}{E8CCC8}
\definecolor{scalarLink}{HTML}{9E3A32}

\hypersetup{
  urlcolor=scalarLink,
  linkcolor=scalarLink,
  citecolor=scalarDark,
}

\usepackage{xspace}

\usepackage{pifont}
\newcommand{\cmark}{\ding{51}}
\newcommand{\xmark}{\ding{55}}

\renewcommand{\citet}[1]{\citep{#1}}

\definecolor{verylightgrey}{rgb}{0.95,0.95,0.95}
\definecolor{darkgrey}{rgb}{0.25,0.25,0.25}

\newtcolorbox{promptbox}[2][]{
  colback=verylightgrey,
  colframe=black,
  coltitle=white,
  colbacktitle=darkgrey,
  boxrule=0.5pt,
  arc=5mm,
  outer arc=5mm,
  leftrule=1pt,
  rightrule=1pt,
  toprule=1pt,
  bottomrule=1pt,
  left=10pt,
  right=10pt,
  top=10pt,
  bottom=10pt,
  boxsep=0pt,
  title={\centering\strut#2\strut},
  fonttitle=\bfseries\large,
  #1
}

\usepackage[most]{tcolorbox}
\usepackage{xcolor}

\definecolor{skillblue}{HTML}{2F6B9A}
\definecolor{skillbluebg}{HTML}{EDF5FB}
\definecolor{skillamber}{HTML}{B56A10}
\definecolor{skillamberbg}{HTML}{FFF5E6}
\definecolor{skillgreen}{HTML}{287A50}
\definecolor{skillgreenbg}{HTML}{ECF8F1}
\definecolor{skillred}{HTML}{A94442}
\definecolor{skillredbg}{HTML}{FCEEEE}
\definecolor{bankheader}{HTML}{3F4145}

\newtcolorbox{skillbankbox}{
  enhanced,
  breakable,
  colback=white,
  colframe=bankheader,
  colbacktitle=bankheader,
  coltitle=white,
  fonttitle=\bfseries,
  title={Abridged evolved skill bank used on \texttt{catapult} seed 45},
  boxrule=0.8pt,
  arc=3mm,
  left=5pt,
  right=5pt,
  top=5pt,
  bottom=5pt,
  before skip=6pt,
  after skip=6pt
}

\newtcolorbox{skillentry}[3]{
  enhanced,
  colback=#2,
  colframe=#1,
  boxrule=0.5pt,
  borderline west={3pt}{0pt}{#1},
  arc=1mm,
  left=5pt,
  right=4pt,
  top=3pt,
  bottom=3pt,
  before skip=2pt,
  after skip=2pt,
  title={#3},
  coltitle=#1,
  fonttitle=\bfseries\scriptsize,
  attach title to upper={\quad}
}

\newcommand{\umasscicslogo}{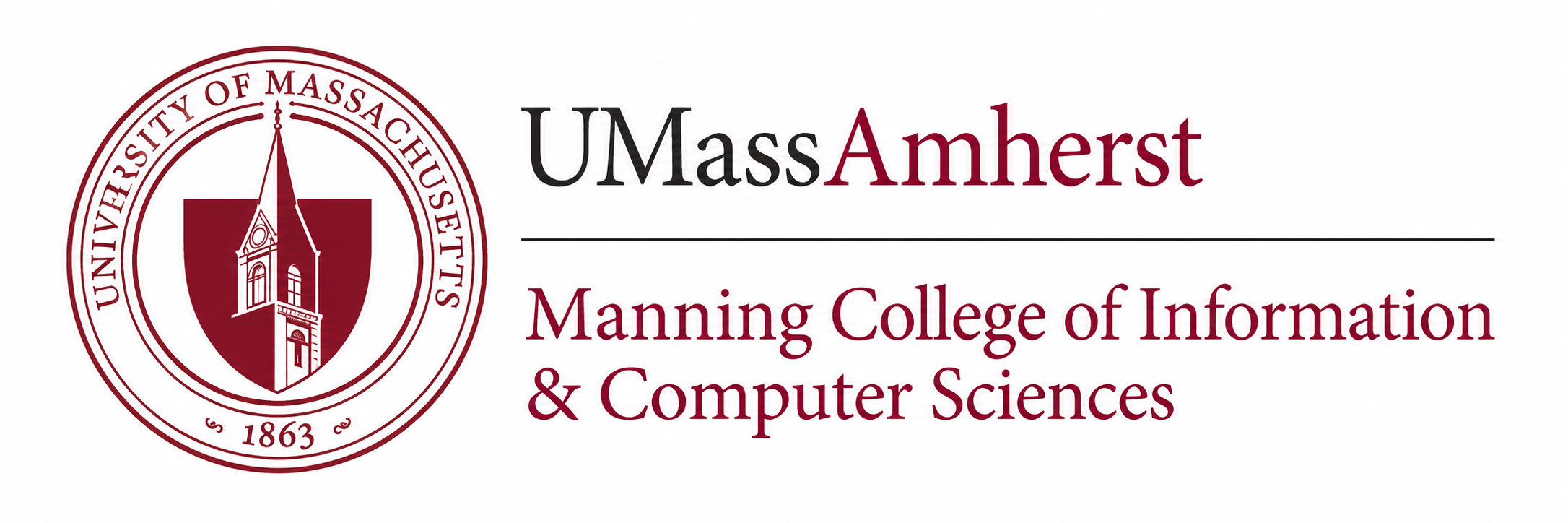}
\newcommand{\scalarlablogo}{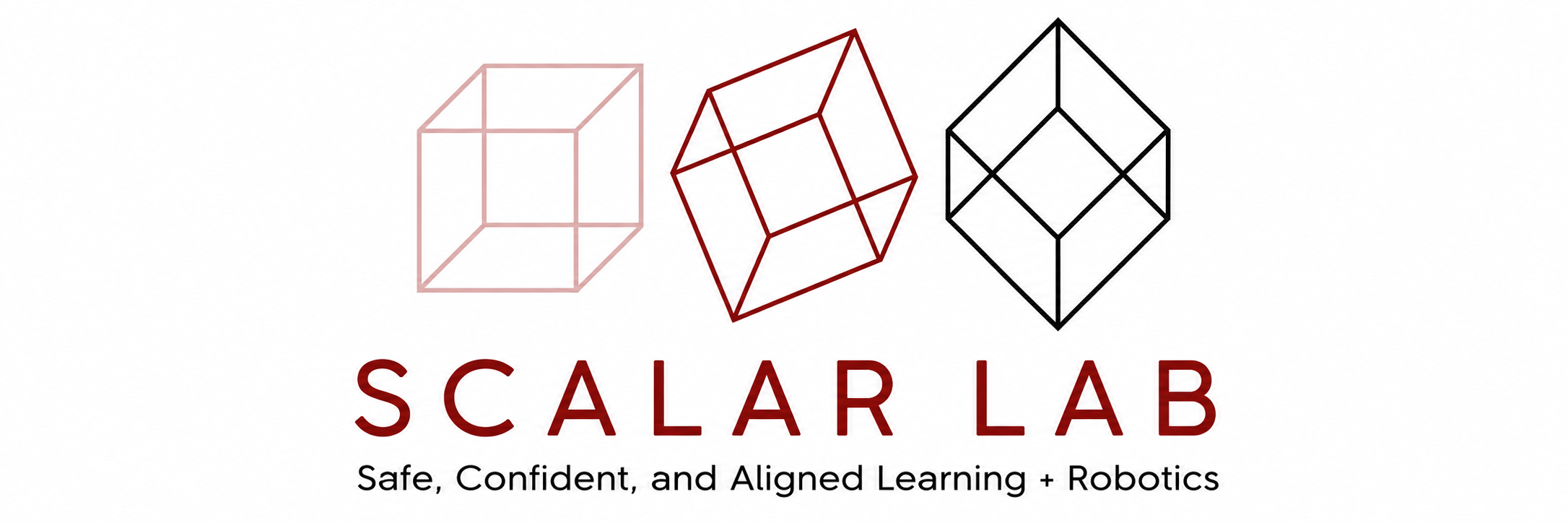}

\newcommand{\HExAWebsiteURL}{https://general-exp-3-continual-learning-agent.github.io/HeXA/}
\newcommand{\HExACodeURL}{https://github.com/General-Exp-3-Continual-Learning-Agent/HeXA-Hierarchical-Experimentalist-Agents}
\newcommand{\InterphyreWebsiteURL}{https://sankaranv.com/interphyre/}
\newcommand{\InterphyreCodeURL}{https://github.com/General-Exp-3-Continual-Learning-Agent/HeXA-Hierarchical-Experimentalist-Agents/tree/main/interphyre}

\newcommand{\cardlogobox}[1]{%
  \begin{minipage}[c][1.75cm][c]{0.26\linewidth}
    \centering
    \includegraphics[
      height=1.5cm,
      width=\linewidth,
      keepaspectratio
    ]{#1}
  \end{minipage}%
}

\newtcolorbox{HExAheadercardbox}{
  enhanced,
  breakable,
  colback=scalarBg,
  colframe=scalarBorder,
  boxrule=0.5pt,
  arc=11pt,
  outer arc=11pt,
  left=10pt,
  right=10pt,
  top=10pt,
  bottom=1pt,
  width=\textwidth,
  before skip=0.05in,
  after skip=1.5em,
}

\newenvironment{HExAheadercard}{%
  \begin{HExAheadercardbox}

  \begin{center}
    {\LARGE\bfseries Hierarchical Experimentalist Agents\par}

    \vspace{0.9em}

    {\bfseries
    Abhranil Chandra \quad
    Utsav Dhanuka \quad
    Varun Gandhi \quad
    Sankaran Vaidyanathan \quad
    Scott Niekum
    \par}

    \vspace{0.35em}

    {University of Massachusetts Amherst\par}

    \vspace{0.55em}

    {\small
    \href{\HExAWebsiteURL}{HExA Website}
    \hspace{0.5em}$|$\hspace{0.5em}
    \href{\HExACodeURL}{HExA Code}
    \hspace{0.5em}$|$\hspace{0.5em}
    \href{\InterphyreWebsiteURL}{Interphyre Website}
    \hspace{0.5em}$|$\hspace{0.5em}
    \href{\InterphyreCodeURL}{Interphyre Code}
    \par}
  \end{center}

  \vspace{0.75em}

  \centerline{\large\bf Abstract}
  \vspace{0.5ex}
  \noindent\ignorespaces
}{%
  \par
  \vspace{1.0em}

  \noindent
  \cardlogobox{\umasscicslogo}%
  \hfill
    \begin{minipage}[c][1.35cm][c]{0.46\linewidth}
      \centering
      {\small\bfseries Correspondence:}\\[-0.02em]
      {\small\textcolor{scalarDark}{\texttt{\{abhranilchan,sniekum\}}}}\\[-0.02em]
      {\small\textcolor{scalarDark}{\texttt{@cs.umass.edu}}}
    \end{minipage}%
  \hfill
  \cardlogobox{\scalarlablogo}%

  \end{HExAheadercardbox}
}

\begin{document}


\begin{HExAheadercard}
Large language models (LLMs) are increasingly used to act in the real world and augment human decision-making, yet most systems rely on parametric knowledge acquired through imitation, optionally improved by post-training, retrieval, or search over fixed data. This paradigm struggles in novel domains or on complex queries that require information unavailable from prior knowledge alone; knowing the laws of physics, for example, does not by itself enable reliable reasoning or long-horizon actions in a complex physical system. We argue that agents therefore need \textit{active experimentation}: the ability to gather targeted query-specific evidence, discover general principles of unseen environments, and convert interaction experience into reusable skills. We introduce \textbf{Hierarchical Experimentalist Agents (HExA)}, an in-context, experiment-centric self-improvement framework that (1) iteratively designs and refines query-relevant experiments, (2) incrementally distills experience into reusable and composable skills that accelerate experimentation within and across tasks, and (3) integrates experimental evidence to act or answer queries. HExA can work with any LLM including frontier black-box models, and requires no external supervision, oracle solutions, or offline data. We evaluate HExA on multi-step reasoning, interactive partially observable games, and physics law discovery tasks. To further evaluate when active experimentation is especially necessary, we introduce \textsc{Interphyre}, a set of hard procedural physics tasks where we enable tool-calling and intervention APIs.
Current frontier LLMs including GPT-6 Astra fail in these settings: on the most challenging Interphyre tasks, we show that HExA can enable such models to near 100\% success. Moreover, HExA improves token efficiency compared to naive agentic and memory baselines and enables generalization via hierarchical skill transfer from easier to harder tasks.




\end{HExAheadercard}

\addtocontents{toc}{\protect\setcounter{tocdepth}{-10}}

\section{Introduction}
\label{sec:intro}

Large language models (LLMs) and LLM agents have demonstrated remarkable capabilities across a wide range of knowledge-intensive and agentic tasks, from code generation and mathematical reasoning to multi-step planning~\citep{gao2025omnimath,jimenez2024swebench,yao2023react,schick2023toolformer}.
Most deployed LLM systems, however, still rely primarily on parametric knowledge acquired through imitative pretraining and post-training over large collections of human-generated data or environments, augmented by static retrieval and test-time reasoning~\citep{Claude4,gpt-5,team2024gemini,team2026qwen3,gao2023retrieval}.
This paradigm works well when the relevant information is well represented in training data or can be recovered through reasoning and in-context learning from zero- or few-shot prompts~\citep{dong2024survey,mei2507survey,dou2026clbench}.
But it breaks down in two important settings. First, an agent may encounter a novel environment whose dynamics, constraints, or effective strategies were absent from training corpus or are unknown even to humans. Second, even when the relevant general knowledge is encoded in the model's parameters, applying it reliably to a particular instance may require far more than recall and reasoning, like information that must be acquired through interaction: knowing the laws of physics, for example, does not by itself solve an experimental physics problem~\citep{silver2025welcome,ying2025assessing}.

In real world settings, given the vast space of novel tasks and queries, an agent's success likely cannot rely on pretrained parametric knowledge alone. Instead, we argue that, agents should, in principle, learn like scientists through active experimentation: this entails exploration, information-gathering, proposing falsifiable hypotheses, testing them through interaction and experiments, and reasoning over what the environment confirms or refutes~\citep{spelke2007core, chuckhypothesis, team2023human, openendedlearningteam2021openendedlearningleadsgenerally,chandra2025unifying}. This is precisely the information that parametric recall, longer chains of thought, and static retrieval cannot directly provide. Experimentation, however, is costly: solving each instance from scratch may require many interaction cycles, even when related instances share useful structure. An effective agent should therefore not only blindly re-explore, but \emph{learn from its experiments}---distilling reusable skills from prior experience and transferring them across related instances to amortize the cost of future experimentation. These observations motivate the central question of our work: 

\vspace{0em}
\begin{center}
\emph{How can LLM agents efficiently learn to reason, act, and generalize in complex, novel domains \\
via in-context experimentation, skill learning, and transfer?}
\end{center}
\vspace{0em}

\begin{figure*}[t]
  \centering

  \setlength{\abovecaptionskip}{0pt}
  \setlength{\belowcaptionskip}{0pt}

  \includegraphics[width=\linewidth]{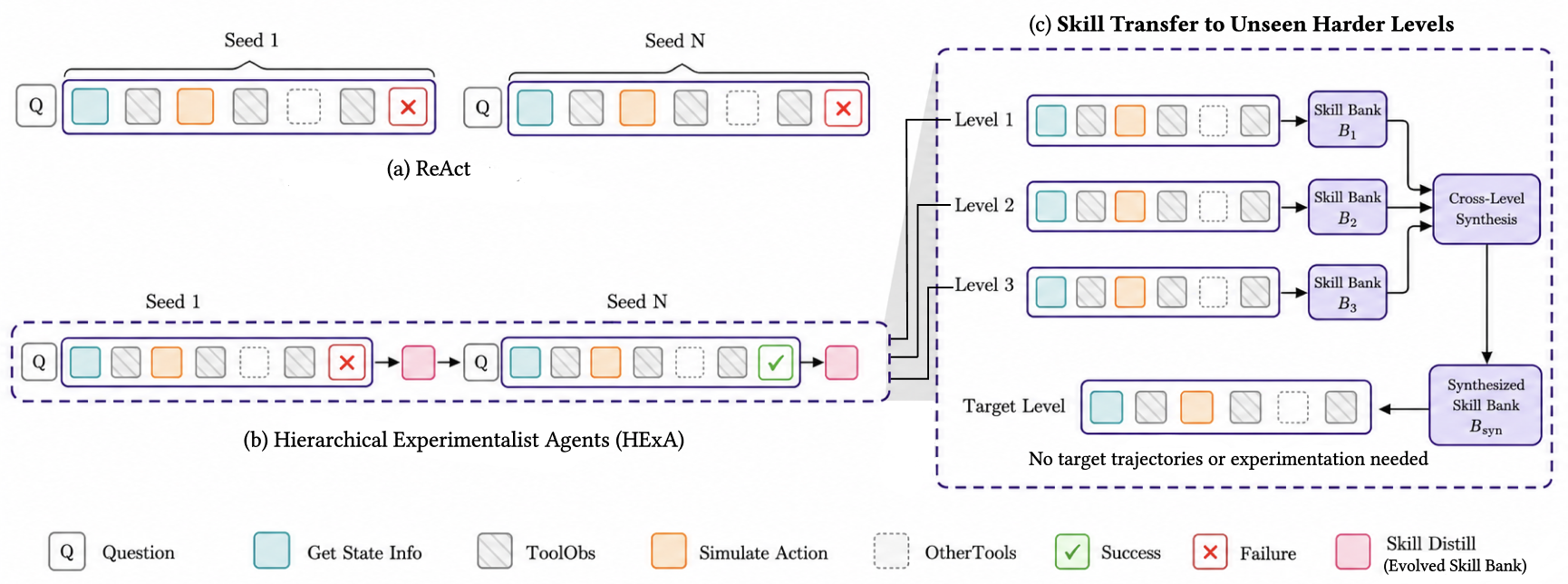}

  \caption{%
    \textbf{Overview of the \textsc{HExA} framework.}
    \textbf{(a) Baseline ReAct:} each episode is solved independently with no
    cross-episode learning. Each episode consists of alternating tool calls for
    experimentation, terminating in success~(\checkmark) or failure~(\(\times\)).
    \textbf{(b) \textsc{HExA}:} episodes are processed in successive rounds.
    After each episode, an evolver distills the trajectory into a \emph{Skill Bank}
    of reusable skills and recurring mistakes that condition subsequent episodes.
    Repeated interaction helps evolve and refine the task-specific Skill Bank.
    \textbf{(c) Transfer to unseen tasks:} each source task runs the \textsc{HExA}
    loop independently to build its Skill Bank. The resulting
    banks are synthesized by the evolver into a single generalizable Skill Bank
    for zero-shot transfer to harder target tasks, with no target trajectories
    used during bank construction.
  }
  \label{fig:hexa_main}

\end{figure*}

We thus propose \textbf{Hierarchical Experimentalist Agents (HExA)}
(Figure~\ref{fig:hexa_main}), a novel in-context reinforcement learning (ICRL) framework
that extends in-context learning along three dimensions:
(1) the actor LLM actively interacts with an environment through tools,
proposing and testing hypotheses to gather query-relevant evidence through
deliberate experimentation rather than passive observation;
(2) the evolver LLM contrasts successful and failed experimental trajectories
and distills their reusable knowledge into a persistent external skill bank;
(3) a retriever selects relevant skills from the bank and injects them into
the actor's context for subsequent episodes, providing stepping-stones for
further experimentation progress while amortizing the cost of exploration.
HExA is hierarchical in the operational sense that later trajectories are
generated while conditioned on previously learned skills, allowing subsequent
skills to refine, compose, delimit, or supersede earlier ones. Beyond
task-specific strategies, this process can therefore form higher-level
experimental skills that generalize across new seeds and configurations and
transfer to structurally different tasks. Unlike trained ICRL approaches that first train a model on offline interaction histories to implement a learning algorithm in context~\citep{laskin2022context,moeini2025survey} or skill based post-training methods~\citep{xia2026skillrl,lu2026skill0}, \textsc{HExA} requires
no reward or action labeled offline training data, additional meta-training, or parameter
updates to the actor. 
Instead, its learning experience is self-generated online by the current
skill-conditioned policy: the evolving bank guides hypothesis-driven
experimentation, successful and failed trajectories update the skill bank, that then conditions the next round. Thus, learning and data collection form a closed in-context self-improvement loop: what the agent has already learned changes not only its subsequent actions, but also which experiments it performs, which evidence and skills it collects next (Section~\ref{sec:HExA}).



Recent benchmarks for multi-step reasoning tasks ~\citep{tower_of_hanoi,mathgap,zebralogic,matchstick1} and interactive partially observable games ~\citep{liu2025gem} and analysis of data unseen during pretraining ~\citep{dou2026clbench}---focus to some extent on evaluating the in-context learning ability of frontier LLMs, but rely mostly on static reasoning and factual analysis. We evaluate HExA on these easier task setups as well. But we instantiate and formalize HExA for a broader setting: tasks where solving an instance requires active experimentation to discover new knowledge or gather targeted, instance-specific evidence, rather than recall and reasoning alone. This includes both settings where the model already knows the relevant general
principles but must experiment to apply them precisely, and genuinely
unfamiliar domains where useful knowledge must be acquired through interaction
with unseen open-ended environments~\citep{openendedlearningteam2021openendedlearningleadsgenerally,hughes2024open}. Physical reasoning is a natural testbed for this harder setting, since models often have the relevant principles yet cannot apply them at the required precision without active experimentation. To study and evaluate this general setting concretely, we introduce Interphyre (Section \ref{sec:results_setup} and Appendix~\ref{app:InterPhyre}), a procedural 2D physics reasoning benchmark for LLM agents, built on top of PHYRE \citep{bakhtin2019phyre}, with programmatic experimentation tool interfaces via scene inspection, simulation, interventions, and intermediate-dynamics
observations.We additionally evaluate HExA on DiscoverPhysics~\citep{wiemann2026discoverphysicsbenchmarkingllmsoutofthebox}, another recent benchmark testing agent experimentation ability. Together, these allow us to evaluate how HExA improves capability LLM agents for self-improvement and generalization via hypotheses driven interactive learning and hierarchical skill composition in complex unseen experiment-centric tasks, all in a non-parametric in-context setup.

In summary, our main contributions are:
\begin{enumerate}[leftmargin=*, nosep]
    \item We formalize an \textbf{experiment-centric problem setting} and introduce \textbf{Interphyre}, a benchmark to better evaluate and improve generalist LLM agents for this problem setting specifically.
    
    \item We propose \textbf{HExA}, 
    in which an LLM actor, evolver and retriever, jointly work to design experiments, test hypotheses, distill new knowledge and experience into composable hierarchical skills that evolve over interactions and generalize to unseen tasks.
   \item HExA improves both frontier and smaller LLMs, including black-box
models. On the two hardest \textsc{Interphyre} levels, where zero-shot success is
near zero, HExA reaches \textbf{83\%} and \textbf{96\%} with Claude
Sonnet~4.6, and \textbf{100\%} and \textbf{76\%} with GPT-6-Astra. Skills
synthesized from easier levels transfer zero-shot to harder ones, raising
\texttt{catapult} success from 11\% (ReAct) to 44\% (\textbf{+33} points).
Beyond InterPhyre, HExA is best in 9 of 10 task-actor settings across interactive
games, reasoning \& planning, and \textsc{DiscoverPhysics}.
\end{enumerate}

\section{Related Works}
\label{app:related_works}

\textbf{Learning from Interaction.}
LLM agents improve decision-making via \emph{parametric fine-tuning}, such as reinforcement learning \cite{ouyang2022training, rafailov2023direct, Guo_2025}, or \emph{non-parametric adaptation}. While parametric updates yield strong performance, they often incur high computational costs and risk catastrophic forgetting or over-specialization \cite{ziegler2019fine, shi2025continual, luo2025empirical}, driving the shift toward in-context agents as a more efficient alternative. \textsc{HExA}, our training-free framework, directly targets this challenge by self-improving LLM agents via in-context experimentation and self-evolution of reusable hierarchical skills.

\textbf{Skills for LMs.}
Skills are natural-language modules that capture reusable procedural knowledge for augmenting LMs at inference time \citep{a-skills-1, a-skills-2}, and skill augmentation has been validated across agent tasks \citep{si2023spokenwoz}, including coding \citep{ma2026scalingcodingagentsatomic} and web navigation \citep{wang2026webxskillskilllearningautonomous}.
Earlier skill libraries were built largely through human annotation \citep{li2026organizingorchestratingbenchmarkingagent, li2026skillsbenchbenchmarkingagentskills, liang2026skillnetcreateevaluateconnect, wang2026skillorchestralearningrouteagents}, which is effective but hard to scale in context-learning settings where contexts are long, technical, and domain-specific.
Recent work has therefore moved toward automated skill construction.
AutoSkill \citep{yang2026autoskill} extracts reusable behaviors from interaction traces as lifelong, versioned artifacts, while AutoRefine \citep{qiu2026autorefine} turns agent trajectories into reusable expertise.
CoEvoSkills \citep{zhang2026coevoskillsselfevolvingagentskills} builds multi-file skill packages through generator--verifier co-evolution; EvoSkill \citep{alzubi2026evoskill} performs failure-driven refinement into structured skill folders; and SkillX \citep{wang2026skillx} distills agent trajectories into a hierarchical skill knowledge base refined through execution feedback.
However, these methods typically rely on external feedback signals, such as execution feedback, ground-truth comparison, or task-completion rewards, to evaluate and improve skill quality \citep{ma2026skillclawletskillsevolve}, which are unavailable in context-learning scenarios without automatic feedback.
A separate line of work internalizes skills through weight updates: SKILL0 \citep{lu2026skill0} uses in-context reinforcement learning (RL) to absorb skills, while SkillRL \citep{xia2026skillrlevolvingagentsrecursive} builds a skill bank via RL-guided distillation from stronger teacher trajectories.
These approaches require parameter access, limiting their use with closed-source models and sacrificing the interpretability of natural-language skill documents.
We therefore propose \textsc{HExA} as a self-improving framework that discovers and evolves skills through in-context active experimentation.

\section{Hierarchical Experimentalist Agents: Learning via Experimentation In-Context}
\label{sec:HExA}

\setlength{\emergencystretch}{2em}


\setlength{\intextsep}{2pt}
\setlength{\columnsep}{10pt}

\begin{wrapfigure}[15]{r}{0.35\linewidth}   
\vspace{-\intextsep}
\begin{algorithm}[H]
\LinesNotNumbered
\caption{\textsc{HExA}}
\label{alg_1}
\DontPrintSemicolon
\scriptsize
\SetInd{0.5em}{0.7em}
\KwIn{Actor $\pi_{\mathrm{act}}$, evolver $\pi_{\mathrm{ev}}$, level $\ell$,\\
\hspace*{2.6em}seeds $\mathcal{S}_\ell$, rounds $R$, episodes/round $x$}
\BlankLine
\tcp{off2on warm start}
$\mathcal{K}_0 \gets \pi_{\mathrm{ev}}(\emptyset,\mathcal{T}_0)$
\textbf{or} $\mathcal{K}_0\gets\emptyset$\;
\BlankLine
\For{$n=1,\ldots,R$}{
  \tcp{top-$M$ skills + top-$N$ mistakes}
  $\mathcal{G}_n \gets
  \textsc{Retriever}(\mathcal{K}_{n-1},\ell,M,N)$\;
  \For{$i=1+x(n-1),\ldots,xn$}{
    \tcp{task + tools + skills}
    $\tau_i \gets
    \pi_{\mathrm{act}}(\cdot\mid\ell,s_i,\mathcal{G}_n)$\;
    \tcp{outcome $\times$ efficiency}
    $r(\tau_i)\gets\textsc{Reward}(\tau_i)$\;
  }
  \tcp{evolver: old bank + new trajectories + rewards}
  $\mathcal{K}_n \gets
  \pi_{\mathrm{ev}}(\mathcal{K}_{n-1},\mathcal{T}^{(n)})$\;
}
\KwRet{$\mathcal{K}_R$}
\end{algorithm}
\end{wrapfigure}

We now formalize \textsc{HExA}'s problem setup and self-improvement objective, and describe the
actor--evolver--retriever loop, experimentation feedback, hierarchical skill-bank, and cross-task transfer mechanism. Figure~\ref{fig:hexa_main} gives the
high-level framework. We use the \texttt{catapult} level of \textsc{InterPhyre} as a running example; all environment and task details are provided
in Appendix~\ref{app:InterPhyre} and ~\ref{app:external_envs}.



\paragraph{Problem setup.}
We consider a task family $\ell$ with instances
$\{(\ell,s_j)\}$, where each $s_j$ specifies an environment configuration
(e.g., geometry, parameters, and initial conditions). 
For each family $\ell$ we
hold a set of experimentation seeds $\mathcal{S}_\ell=\{s_1,\ldots,s_{|\mathcal{S}_\ell|}\}$,
on which the agent experiments and the skill bank is evolved, and a disjoint set
$\mathcal{S}^{\mathrm{ev}}_\ell$ on which the resulting bank is evaluated. 
The actor interacts with each instance through an environment-provided
tool interface for experimentation $\mathcal{F}$, which may expose scene inspection, simulation,
interventions and hypotheses testing, and task-specific analysis
(Appendix~\ref{app:tools}). Each episode has at most $T$ interaction turns.
At turn $t$, the actor observes the history
$h_t=(a_0,o_0,\ldots,a_{t-1},o_{t-1})$ and selects a tool call
{\setlength{\abovedisplayskip}{0pt}
\setlength{\belowdisplayskip}{0pt}
\begin{equation}
\label{eq:turn}
a_t \sim
\pi_{\mathrm{act}}
\!\left(\cdot\mid\ell,s_j,\mathcal{G},h_t;\mathcal{F}\right),
\qquad
o_t=\textsc{Env}(a_t),
\end{equation}}
where $\mathcal{G}$ is a skill context retrieved once at the beginning of the
episode. The episode terminates after a terminal
action, a successful simulation, or the $T$-turn is over, yielding trajectory
$\tau_j=h_{T_j}$ with $T_j\leq T$. a binary task $\textsc{Goal}_\ell$ a binary task scores, tells the terminal simulation outcome. Writing $\textsc{Sim}(\tau_j)$ for the terminal simulator state
associated with trajectory $\tau_j$, success is
{\setlength{\abovedisplayskip}{0pt}
\setlength{\belowdisplayskip}{0pt}
\begin{equation}
\label{eq:success_main}
y_j(\pi_{\mathrm{act}};\ell,\mathcal{G})
=
\mathbb{I}\!\left[
\textsc{Goal}_\ell\!\left(\textsc{Sim}(\tau_j)\right)
\right],
\qquad
\tau_j\sim
\pi_{\mathrm{act}}(\cdot\mid\ell,s_j,\mathcal{G}).
\end{equation}}
Equation~\ref{eq:success_main} defines the task without method-specific assumptions: any
agent can be evaluated against it, and skill-free baselines are the case where $\mathcal{G}=\emptyset$.

\paragraph{In-context self-improvement objective.}
Equation~\ref{eq:success_main} is method-agnostic: skill-free agents have
$\mathcal{G}=\emptyset$, while \textsc{HExA} improves the same frozen actor by
evolving only the external skill state that conditions it.
Let $\mathcal{D}_{\ell}$ denote the distribution of future instances and
$\mathcal{G}_{\mathcal{K},\ell}
=\textsc{Retriever}(\mathcal{K},\ell,M,N)$.
For fixed $\pi_{\mathrm{act}}$, we define
{\setlength{\abovedisplayskip}{0pt}
\setlength{\belowdisplayskip}{0pt}
\begin{equation}
\label{eq:objective}
J_{\ell}(\mathcal{K})
=
\mathbb{E}_{s\sim\mathcal{D}_{\ell}}
\left[
y\!\left(
\pi_{\mathrm{act}};
\ell,
\mathcal{G}_{\mathcal{K},\ell}
\right)
\right].
\end{equation}}
\textsc{HExA} keeps $\pi_{\mathrm{act}}$ fixed and evolves $\mathcal{K}$ from
interaction on $\mathcal{S}_{\ell}$ to increase $J_{\ell}$.
Thus, self-improvement occurs entirely through context:
$\mathcal{K}_R$ is useful when the same frozen actor performs better with
$\mathcal{K}_R$ than with $\mathcal{K}=\emptyset$ on unseen instances.

\paragraph{The \textsc{HExA} loop.}
\textsc{HExA} alternates three components: an \emph{actor}
$\pi_{\mathrm{act}}$ that experiments in the environment, an \emph{evolver}
$\pi_{\mathrm{ev}}$ that distills the resulting trajectories into
$\mathcal{K}$, and a \emph{retriever} that selects skill bank entries for the
actor's next episodes. Algorithm~\ref{alg_1} repeats this loop for
$R=\lceil|\mathcal{S}_\ell|/x\rceil$ rounds as shown in Appendix figure~\ref{fig:hexa_loop}. At round $n$, the retriever constructs
$\mathcal{G}_n=\textsc{Retriever}(\mathcal{K}_{n-1},\ell,M,N)$.
The actor then runs on a fresh batch of at most $x$ experimentation seeds
under the same $\mathcal{G}_n$, producing trajectories
$\mathcal{T}^{(n)}$ with rewards. Importantly, $\mathcal{T}^{(n)}$ is generated under
$\mathcal{K}_{n-1}$ rather than drawn from a fixed corpus, so each bank update
changes the policy that collects the next round of experience. The evolver updates
$\mathcal{K}_n=\pi_{\mathrm{ev}}(\mathcal{K}_{n-1},\mathcal{T}^{(n)})$.
No model weights change; the evolving bank is the persistent state carried
across rounds. $\mathcal{K}_0$ may be empty or warm-started from skill-free
actor trajectories, as ablated in Section~\ref{sec:instantiation} (see Algorithm~\ref{alg:HeXa_algo_app} for all variants).
After the final round, $\mathcal{K}_R$ is frozen: the evolver is not invoked
during held-out evaluation on $\mathcal{S}^{\mathrm{ev}}_\ell$ (Algorithm~\ref{alg:hexa_inference} gives the full inference procedure).

\begin{figure}[h]
  \centering
  \includegraphics[width=0.75\linewidth]{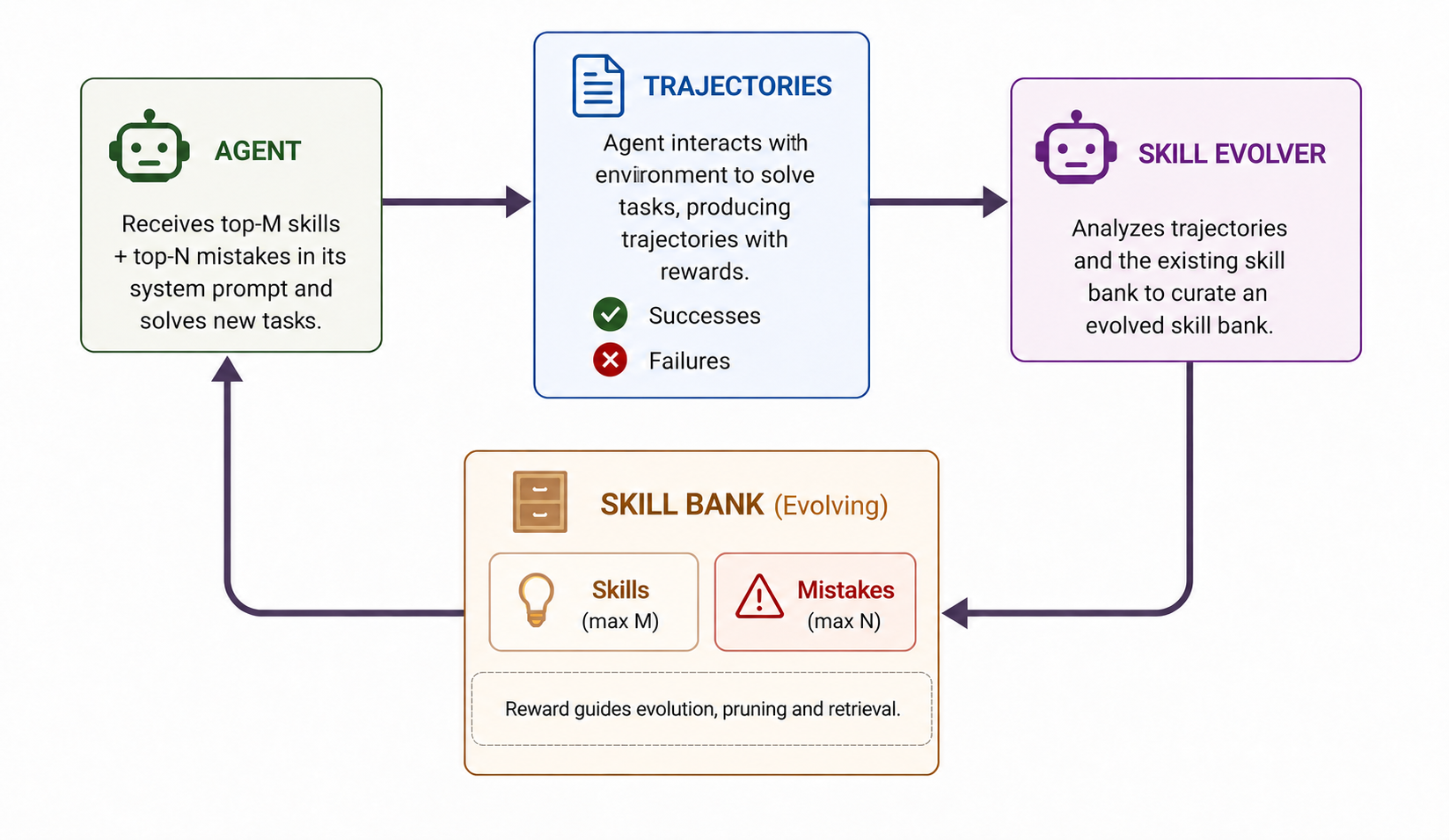}
  \caption{%
    \textbf{The \textsc{HExA} actor-evolver loop.}
    The actor receives distilled skills in its system prompt and runs on
    physics puzzles, producing trajectories annotated with rewards
    (successes~\cmark, failures~\xmark).
    The Skill Evolver analyzes these trajectories and curates the
    evolving Skill Bank: skills (max~M) are merged, pruned, or promoted
    by reward signal, and common mistakes (max~N) are updated accordingly.
    Each skill carries a generation index tracking its lineage across
    evolution rounds (Eq.~\ref{eq:weight}).
    The updated bank is injected back into the actor's system prompt for
    the next round, closing the loop.
  }
  \label{fig:hexa_loop}
\end{figure}

\paragraph{Skill-augmented actor.}
At the beginning of each episode, the actor receives the task description,
tools $\mathcal{F}$, and a skill context retrieved from the current bank:
{\setlength{\abovedisplayskip}{0pt}
\setlength{\belowdisplayskip}{0pt}
\begin{equation}
\label{eq:skill-augmented_main}
\mathcal{G}
=
\textsc{Retriever}(\mathcal{K},\ell,M,N),
\qquad
\tau_j
\sim
\pi_{\mathrm{act}}(\cdot\mid\ell,s_j,\mathcal{G}).
\end{equation}}
The retriever selects at most the top-$M$ strategy skills and top-$N$ mistake
records according to the scores described below and inserts them as a bounded
textual prefix. This changes only the actor's context: its architecture,
parameters, and decoding procedure remain unchanged.

\paragraph{Trajectory reward.}
\label{para:traj_reward}
Each trajectory may receive a scalar reward
$r(\tau)\in[-1,+1]$, which provides the evolver with evidence about trajectory
quality. The signal can come from any available scorer---for example, a
programmatic environment reward, human feedback, an LLM judge, or a learned
reward model. In our \textsc{InterPhyre} experiments, the reward combines task
outcome and interaction efficiency; its exact definition is given in
Appendix~\ref{sec:reward}.
\textsc{HExA} can also operate without explicit scalar reward labels, in which
case the evolver assesses trajectory quality from the trajectories themselves.

\paragraph{Skill distillation.}
After round $n$, 
the evolver receives the current
bank $\mathcal{K}_{n-1}$ and trajectory batch $\mathcal{T}^{(n)}$, with each
trajectory represented by its
\texttt{(thought, tool\_call, observation)} sequence and reward, and returns the updated bank 
\begin{equation}
\label{eq:evolve}
\mathcal{K}_n \;=\; \pi_{\mathrm{ev}}\!\left(\mathcal{K}_{n-1},\, \mathcal{T}^{(n)}\right),
\qquad
\left|\mathcal{K}^{\mathrm{sk}}_{n}\right| \le M_{\max},
\quad
\left|\mathcal{K}^{\mathrm{mis}}_{n}\right| \le N_{\max}
\end{equation}
The evolver contrasts high- against low-reward trajectories and emits two
complementary forms of knowledge: \emph{strategy skills} $\mathcal{K}^{\mathrm{sk}}$, which
state what to do and when, and \emph{mistake records} $\mathcal{K}^{\mathrm{mis}}$, which
state what recurrently goes wrong and how to recover from it. Each strategy skill is a
structured record defined by a
\emph{title}, a \emph{principle} (the insight gained from testing the proposed hypothesis
via experimentation), a \emph{when-to-apply} condition, an optional demonstrative
\emph{example}, the \emph{source seed} it was derived from, and a \emph{reward score}
defined below. For example, the \texttt{catapult} skill \texttt{cat\_\allowbreak ev\_\allowbreak
6\_\allowbreak 001} (Appendix~\ref{app:trajectories:bank}): title ``low
drop height maximises range; binary-search $y$ first,'' principle ``drop height $y$ is the
primary range control, with $y\!\approx\!-0.5$ giving maximum range and a $\pm0.25$ change
shifting the landing point by $1.5$--$2.0$ units,'' applicability ``start at $y{=}-0.5$,
then binary-search toward $y{=}0$ when overshooting.'' The record is supported by four
source seeds and carries a reward score of $0.93$.

\emph{Mistake records} are triples
$\mu=(\delta,\rho,\alpha)$ pairing a recurring failure $\delta$ with its inferred root
cause $\rho$ and a corrective strategy $\alpha$: e.g.\ $\delta$ ``fine-tuning a parameter
inside a chaotic zone where the trajectory responds non-monotonically,'' $\rho$ ``the
agent assumes a monotonic response and keeps incrementing,'' $\alpha$ ``when two
consecutive traces land more than $1.5$ units apart under a small parameter change, switch
from linear search to bisection.'' The evolver additionally recovers \emph{partial
skills}: reasoning steps that were individually correct inside an otherwise failed
trajectory, a signal unavailable from successes alone, since successes do not reveal which
intermediate steps were necessary rather than incidental. Because trajectories in later
rounds are generated while conditioned on previously learned skills, new skills can
refine, compose, or delimit earlier ones; whether this compounds into a hierarchy of
increasingly abstract principles is an empirical question we examine in
Section~\ref{sec:results}. The complete schema and evolver prompts are in
Appendix~\ref{sec:distillation}, and a full evolved \texttt{catapult} bank in
Appendix~\ref{app:trajectories:bank}.

\paragraph{Skill bank and retrieval.}
\label{sec:skill_bank_retrieval_main}
Distilled knowledge is stored in the skill bank $\mathcal{K}$, which keeps two collections
separately for each task family (e.g.\ \texttt{catapult}): a set of \emph{skills} (reusable strategies and principles) and a set of \emph{mistakes}
(recurring failure modes to avoid).
Each skill carries a \emph{reward score} $r_k$, computed from the trajectories that demonstrated or used it, as a function of the mean reward of these source trajectories ($\bar{r}_{\mathrm{src}}$; Eq.~\ref{eq:weight}). We show an example of this reward label in Section~\ref{sec:instantiation}, with the individual trajectory rewards in Appendix~\ref{sec:reward}. This reward label is used for two purposes: ranking skills during retrieval (top-$M$ by $r_k$), and guiding the evolver in deciding which skills to retain or prune when the bank is at capacity. Skills derived from fast, efficient successful trajectories carry higher scores; skills from partial reasoning within failed trajectories carry lower scores but may still encode valuable learnings and discoveries.

The bank is capped at $M_{\max}$ skills and $N_{\max}$ mistakes per task family to prevent the injected context from growing without bound and to force the evolver to prioritize the most informative knowledge. Full bank structure and retrieval details are in Appendix~\ref{app:hexa}.

The abridged skill bank shown alongside gives the information available to \textsc{HExA} \emph{before} it attempts the held-out \texttt{catapult} seed~64. Unlike raw trajectory memory, the bank organizes evidence from earlier episodes into a default hypothesis (S1), a diagnostic condition (S2), a corrective strategy (S3), and a mistake to avoid. On seed~64 these are used in that order: the default drop overshoots, the next trace collapses, the agent reads this as a chaotic cliff, bisects, and commits after six iterations (Appendix~\ref{app:trajectories:hexa_run}). The entries also show how individual observations are composed into general skills: S3 is a single record that merges a commit threshold and a bisection rule.

\definecolor{skilldark}{HTML}{454649}
\begin{tcolorbox}[
  enhanced, colback=white, colframe=skilldark, boxrule=0.7pt,
  arc=8pt, outer arc=8pt, left=6pt, right=6pt, top=6pt, bottom=6pt, boxsep=0pt,
  title={Abridged evolved skill bank used on \texttt{catapult} seed 64},
  colbacktitle=skilldark, coltitle=white, fonttitle=\bfseries\large,
  titlerule=0pt, toptitle=4pt, bottomtitle=4pt, width=\linewidth]
\footnotesize
\setlength{\parskip}{2pt}

\textit{Four of the records in the \texttt{catapult} bank after the last training
round, which is frozen and used unchanged on the held-out seed 64.}

\begin{skillentry}
  {skillblue}{skillbluebg}
  {S1: \textsc{Default} $\triangleright$ Right of the pivot, drop low}
Place the red ball at $x\!\approx\!0.3$, on the right half of the catapult arm, so
that it pushes that side down and flings the green ball rightward toward the
basket. The drop height $y$ is the main range control: $y\!\approx\!-0.5$ gives the
longest throw, $y\!\approx\!0.0$ a moderate one, and a change of $\pm0.25$ in $y$
moves the landing point by $1.5$--$2.0$ units.
\textbf{Use when:} making the first placement on a new seed; start at $x{=}0.3$,
$y{=}-0.5$ and move $y$ toward $0.0$ if the ball overshoots.
\end{skillentry}

\begin{skillentry}
  {skillamber}{skillamberbg}
  {S2: \textsc{Diagnose} $\triangleright$ Chaotic cliff}
In some regions a tiny change in the drop position causes a large swing in where the
green ball lands (for example $y\!\approx\!-0.15$ to $-0.25$ at $x{=}0.3$). The
boundary of such a cliff can be as narrow as $\Delta y{=}0.01$, so stepping through it
in small increments tends to jump straight past the target.
\textbf{Use when:} two traces whose $y$ differs only slightly land more than
$1.5$ units apart.
\end{skillentry}

\begin{skillentry}
  {skillgreen}{skillgreenbg}
  {S3: \textsc{Correct} $\triangleright$ Bisect, then commit}
Once a cliff is found, try the midpoint between the two placements that bracket it
instead of committing to either one. Trace each candidate before using a
simulation; at standard drop heights ($y\le0$), commit to \texttt{simulate} as soon
as the traced landing is within $0.8$ units of the target.
\textbf{Use when:} a cliff has been bracketed, or a trace is already within the
commit threshold.
\end{skillentry}

\begin{skillentry}
  {skillred}{skillredbg}
  {\textsc{Avoid} $\triangleright$ Linear search in a chaotic zone}
The agent assumes the landing position changes smoothly with $x$ or $y$ and keeps
nudging the parameter in small steps, when the response is in fact chaotic.
\textbf{Correction:} when two consecutive traces land more than $1.5$ units apart
under a small change, switch from linear search to bisection.
\end{skillentry}

\vspace{2pt}
\textbf{On seed 64.} The target blue ball sits at $x{=}3.50$. Following S1, the agent
traces $x{=}0.3$, $y{=}-0.5$ and the green ball overshoots to $4.78$. It raises the
drop to $y{=}0.0$ to shorten the throw, but the green ball now lands at $-0.92$,
behind the catapult. A $0.5$ change in $y$ has moved the landing by $5.7$ units, which
S2 identifies as a cliff rather than a gradient, so the agent applies S3 and bisects
to $y{=}-0.25$. The trace lands at $3.29$, $0.21$ from the target and inside the $0.8$
commit threshold; the agent simulates this placement and succeeds after six
iterations (Appendix~\ref{app:trajectories:hexa_run}).

\vspace{3pt}
{\centering
\textcolor{skillblue}{S1 default: lands $4.78$} $\to$
\textcolor{skillamber}{S2 cliff: $y{=}0.0$ lands ${-}0.92$} $\to$
\textcolor{skillgreen}{S3 bisect: $y{=}{-}0.25$ lands $3.29$} $\to$
\textcolor{skillgreen}{success}\par}

\label{promptbox:skillbank}
\end{tcolorbox}

\paragraph{Cross-task skill transfer and generalization.}
\label{sec:cross_task_transfer}
The same composable skills can transfer knowledge across task families.
Instance-specific details such as coordinates or object identities may not
transfer, but mechanism-level principles---for example, how general strategies work or contact
geometry affects impulse direction, can remain useful across different tasks and configurations. Given evolved source banks
$\mathcal{K}_{\ell_1},\ldots,\mathcal{K}_{\ell_S}$ and only a textual description of an unseen target task $\ell^*$, a evolver synthesizes, $\mathcal{K}_{\ell^*}=\pi^{\mathrm{tr}}_{\mathrm{ev}}\!\left(\{\mathcal{K}_{\ell_i}\}_{i=1}^{S},\,\mathrm{desc}(\ell^*)\right)$,
with $\mathcal{T}_{\ell^*}=\emptyset$.
The evolver (i) selects source skill that are structurally relevant, (ii) removes source-specific parameter and entity bindings,
(iii) re-grounds each retained principle in the target's entities,
constraints, and tools, and (iv) assigns a transfer score based on structural
correspondence and corroboration across source banks to help the retriever during zero-shot evaluations (Algorithm~\ref{alg:cross_task_synthesis}).
Because no target trajectories are available during synthesis, the transfer
procedure is constrained to express general mechanisms and applicability
conditions rather than target-specific coordinates or observed solutions.
The resulting $\mathcal{K}_{\ell^*}$ can be injected through the
retriever in Eq.~\ref{eq:skill-augmented_main} for target experimentation free skill transfer, or used as
$\mathcal{K}_0$ for subsequent within-task evolution.
Full synthesis prompts are provided in
Appendix~\ref{app:prompt:cross_level_full}; results appear in
Section~\ref{sec:results:transfer}.

\section{Experiments, Results and Ablations}
\label{sec:results}

We evaluate \textsc{HExA} along three questions: (i) whether skills distilled from experimentation improve a frozen actor on
held-out instances; (ii) which components of skill formation and reuse drive
these gains; and (iii) whether the resulting hierarchical skills generalize
and transfer beyond the tasks on which they were learned.



We test across reasoning, interactive games, and scientific discovery,
and use \textsc{Interphyre} as a controlled setting for detailed analysis of 
active experimentation and hierarchical skill formation.

\subsection{Experimental Setup}
\label{sec:results_setup}

\paragraph{Benchmarks and Evaluation Setup.}
We introduce and use \textsc{Interphyre} as the controlled testbed for analyzing how experimentation and skill acquisition improves self-improvement, and use three other task families to test whether the same
mechanism generalizes broadly:
multi-step reasoning with MathGAP~\citep{opedal2024mathgap},
ZebraLogic~\citep{lin2025zebralogic}, Tower of Hanoi~\citep{shojaee2025illusion},
and Matchstick Arithmetic~\citep{nasvytis2026insight};
interactive games including Wordle, Minesweeper, Mastermind, and
Maze~\citep{liu2025gem,lin2026orbit}; and physics-law discovery with
\textsc{DiscoverPhysics}~\citep{wiemann2026discoverphysicsbenchmarkingllmsoutofthebox}. Although \textsc{HExA} yields significant performance and learning-efficiency gains on
these tasks compared to baselines, these tasks are not designed to isolate the necessity of learning through active experimentation: zero-shot LLM performance is already relatively high, indicating substantial prior familiarity, and the tasks do not explicitly support experimentation-centric interaction.
\textsc{Interphyre} is a benchmark of conceptually simple but difficult interactive physical-reasoning tasks built on PHYRE level designs~\citep{bakhtin2019phyre}, with experimentation tools for active hypothesis testing.
Unless stated otherwise, skill banks are learned on training instances, frozen,
and evaluated on disjoint held-out instances for all tasks.
Splits and task-specific protocols are in Appendix~\ref{app:baselines}.

\begin{figure}[!h]
  \centering
  \includegraphics[width=\linewidth]{figures/interventions_example2.pdf}
  \caption{%
    \textbf{\textsc{Interphyre} and its properties.} (i) tasks whose solutions cannot be inferred from a static description and must be discovered through interaction and experiments; (ii) a tool-call programmatic interface for experimentation and hypothesis testing via scene inspection, interventions and intermediate-dynamics observations; and (iii) controllable difficulty through procedural generation.
  }
  \label{fig:intervention-example_main_paper_}
\end{figure}

\paragraph{Interphyre---A Testbed for Learning through Experimentation.} Unlike benchmarks that primarily ask whether an agent can solve a task or answer a query, our setting requires an environment in which the agent can deliberately interact and gather evidence: inspecting the current instance, testing interventions, observing intermediate dynamics, diagnosing failures, and reusing what it learned across future seeds and levels in an experimentation loop. \textsc{Interphyre} provides this in a controlled 2D physics domain with continuous actions and long-horizon causal structure. Each task requires placing a red ball at $(x,y,r)$, after which the simulator evaluates whether the level-specific success predicate is achieved. \textsc{Interphyre} builds on the level designs in the PHYRE benchmark \citep{bakhtin2019phyre} by introducing APIs for level editing, test-time interventions on environment parameters, and logging traces from the simulation. These APIs can be used to customize the benchmark for evaluating agents on unseen levels, or provided to an LLM for active experimentation. Figure~\ref{fig:agent_tools} illustrates the state-information and simulation tools available in Interphyre. Our experiments use the \textsc{Interphyre} API to expose a structured tool-calling interface for experimentation via interventional scene inspection, partial simulation, full rollouts, contact logs, and level-specific geometric analysis, for example as shown in Figure \ref{fig:intervention-example_main_paper_}. However, successful actions often depend on contact chains, lever mechanics, collision timing, and strategy-dependent dynamics. Even strong LLMs struggle to anticipate such interactions reliably during inference, particularly when only given the scene description. Now using these intervention API and traces recovered from \textsc{Interphyre} runs, HExA enables the LLM agent to reason over its proposed hypothesis and corresponding experimental observations, than merely guessing solutions from parametric prior knowledge. This lets the agent leverage the information acquired through exploration and interactions, to improve its performance in-context. \textsc{Interphyre} therefore serves as both a benchmark and a diagnostic environment for experimentation based methods such as \textsc{HExA}. More details in Appendix~\ref{app:InterPhyre}.

\setlength{\intextsep}{2pt}
\setlength{\columnsep}{8pt}

\begin{wraptable}{r}{0.48\linewidth}
\vspace{-\baselineskip}

\centering
\scriptsize

\setlength{\tabcolsep}{1.5pt}
\renewcommand{\arraystretch}{0.90}
\setlength{\abovecaptionskip}{2pt}
\setlength{\belowcaptionskip}{2pt}

\caption{%
Base LLM agent's (\textsc{Zero-shot} baseline) solve rate~(\%) across eight
\textsc{Interphyre} levels.
\textsuperscript{$\dagger$} Pass the Parcel and Catapult evaluated over
50 seeds; all other levels evaluated over 100 seeds.
}
\label{tab:direct_baseline_results_main}

\begin{tabular*}{\linewidth}{
@{\extracolsep{\fill}}
lcccccc
@{}
}
\toprule
& \textbf{Rand}
& \textbf{Q-3B}
& \textbf{Q-7B}
& \textbf{O-20B}
& \textbf{O-120B}
& \textbf{Sonnet} \\
\midrule

Cliffhanger
& 0  & 0.0 & 0.0 & 0.0  & 0.0  & 0.0 \\

Tipping Point
& 3  & 0.0 & 3.0 & 1.0  & 2.0  & 0.0 \\

Basket Case
& 4  & 0.0 & 2.0 & 5.0 & 13.0 & 19.0 \\

Down to Earth
& 8  & 3.0 & 7.0 & 6.0  & 34.7  & 68.0 \\

Two Body
& 4 & 0.0 & 0.0 & 0.0  & 6.0  & 14.0 \\

Falling in Place
& 1  & 0.0 & 1.0 & 3.0  & 29.0 & 33.0 \\

\underline{Pass the Parcel}
& 0 & 0.0 & 0.0 & 0.0 & 0.0 & 0.0$^{\dagger}$ \\

\underline{Catapult}
& 0 & 0.0 & 0.0 & 0.0 & 0.0 & 0.0$^{\dagger}$ \\

\bottomrule
\end{tabular*}

\vspace{-2pt}
\end{wraptable}

\paragraph{Baselines and LLM agents.}
On \textsc{Interphyre}, we evaluate GPT-6-Astra, Claude Sonnet~4.6,
Qwen-2.5-\{3B,7B\}-Instruct, and GPT-OSS-\{20B,120B\}, each with at most
25 interaction turns per seed, together with a random valid-placement policy.
Across the external benchmarks, we evaluate GPT-OSS-120B and Qwen3-8B on the
interactive games, GPT-OSS-120B on reasoning and planning, and Claude
Sonnet~5 on \textsc{DiscoverPhysics}. We compare three baseline regimes:
(i) memoryless agentic methods---\textsc{Zero-shot} performance of base LLMs,
\textsc{ReAct}~\citep{yao2023reactsynergizingreasoningacting}, and
Reflexion~\citep{shinn2023reflexion};
(ii) persistent episodic memory methods---Episodic RAG~\citep{lewis2020rag} and \textsc{MemRL}~\citep{zhang2026memrl}; and
(iii) interaction budget matched parametric adaptation with GRPO~\citep{shao2024deepseekmath}.
Retrieved memory is inserted in the same prompt as the \textsc{HExA}
skill bank for fair comparison; full configurations and details are in
Appendix~\ref{app:baselines}. 
Table~\ref{tab:direct_baseline_results_main} reports each LLM's \textsc{Zero-shot} solve rate across eight levels of Interphyre. On the two harder levels: \texttt{pass\_the\_parcel} and \texttt{catapult}, every model, including the random policy, scores $0\%$. These levels, where purely parametric reasoning is insufficient, are the primary testbed for our analysis and ablations of \textsc{HExA}. We report \emph{solve rate} (\% of seeds satisfying $\textsc{Goal}_\ell$, Eq.~\ref{eq:success_main}) and average turns per seed.




\subsection{Learning from experimentation yields efficient self-improvement via reusable hierarchical skill acquisition for in-context policy improvement}

\label{sec:results:interphyre}

\begin{figure}[h]
  \centering

  \setlength{\abovecaptionskip}{2pt}
  \setlength{\belowcaptionskip}{0pt}

  \includegraphics[width=\linewidth]{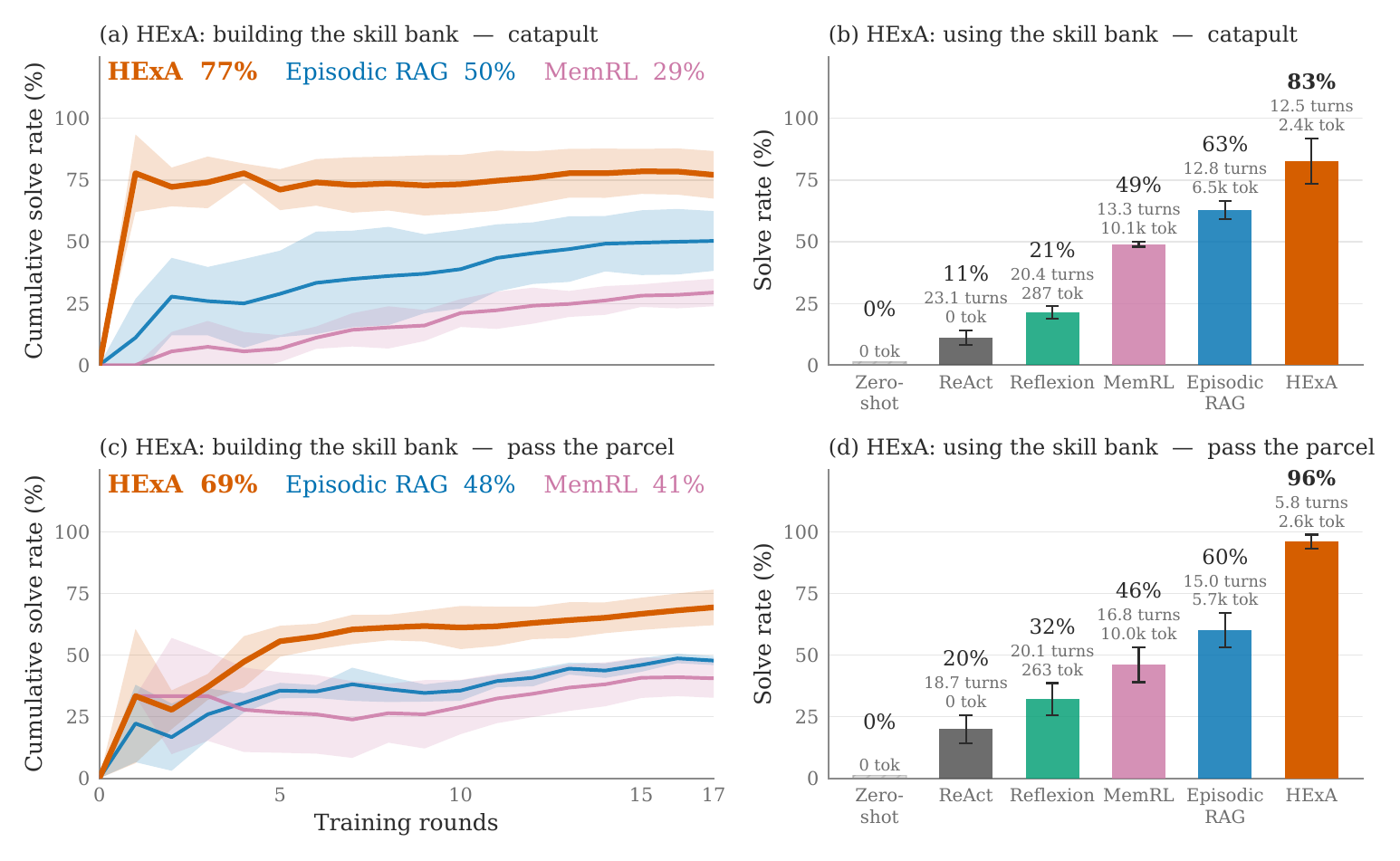}

  \caption{%
    \textbf{\textsc{HExA} uses experimentation and learned skills for higher
    success with greater efficiency.}
    Claude Sonnet~4.6 on \texttt{catapult} (top) and
    \texttt{pass\_the\_parcel} (bottom).
    \emph{Left:} cumulative solve rate while skills are built over training
    seeds.
    \emph{Right:} solve rate on held-out seeds using the final frozen skill
    bank; labels report mean turns per seed and injected tokens per episode.
    Curves and bars are means over three runs; bands and error bars denote
    $\pm1$ s.d.
    Full results are in Table~\ref{tab:catapult_cost}.%
  }

  \label{fig:build_then_freeze}
  \label{fig:astra_catapult}

  \vspace{-3pt}
\end{figure}

\emph{Can agents self-improve in-context through their own experimentation?}
We first test whether experimental experience can improve a frozen LLM's policy on unseen instances. \textsc{HExA} incrementally distills trajectories collected during experimentation into an evolving skill bank, which is then frozen and evaluated on disjoint held-out instances. On the two hardest \textsc{Interphyre} levels, \texttt{catapult} and \texttt{pass\_the\_parcel}, Claude Sonnet~4.6 achieves $0\%$ zero-shot success, while interaction alone through \textsc{ReAct} reaches only $11\%$ and $20\%$. In contrast, \textsc{HExA} achieves $82.7\%$ and $96\%$, respectively as show in Figure~\ref{fig:build_then_freeze} with more details in Table~\ref{tab:catapult_cost} and Appendix~\ref{app:results:interphyre}. The training curves show how the agent accumulates useful experimental knowledge, while held-out evaluations demonstrate its reuse as stepping stone for improvement on unseen configurations. This results persist across model scales including GPT-6-Astra, \textsc{HExA} achieves $100\%$ and $76\%$ success compared with $6\%$ and $0\%$ for the GPT-6-Astra base model, while reducing average interaction turns by approximately $45\%$ (Figure~\ref{fig:astra_ptp_catapult_main} above, and Figures~\ref{fig:app_astra_catapult},~\ref{fig:app_astra_ptp} in appendix). Similar gains with smaller open-weight models (details in Appendix Table~\ref{tab:interphyre_qwen_full}) suggest that HExA works uniformly across model families, tasks and benchmarks and improve both the capabilities and the exploration efficiency of agents.

\begin{figure}[!b]
  \centering
  \setlength{\abovecaptionskip}{2pt}
  \setlength{\belowcaptionskip}{0pt}
  \includegraphics[width=0.75\linewidth]{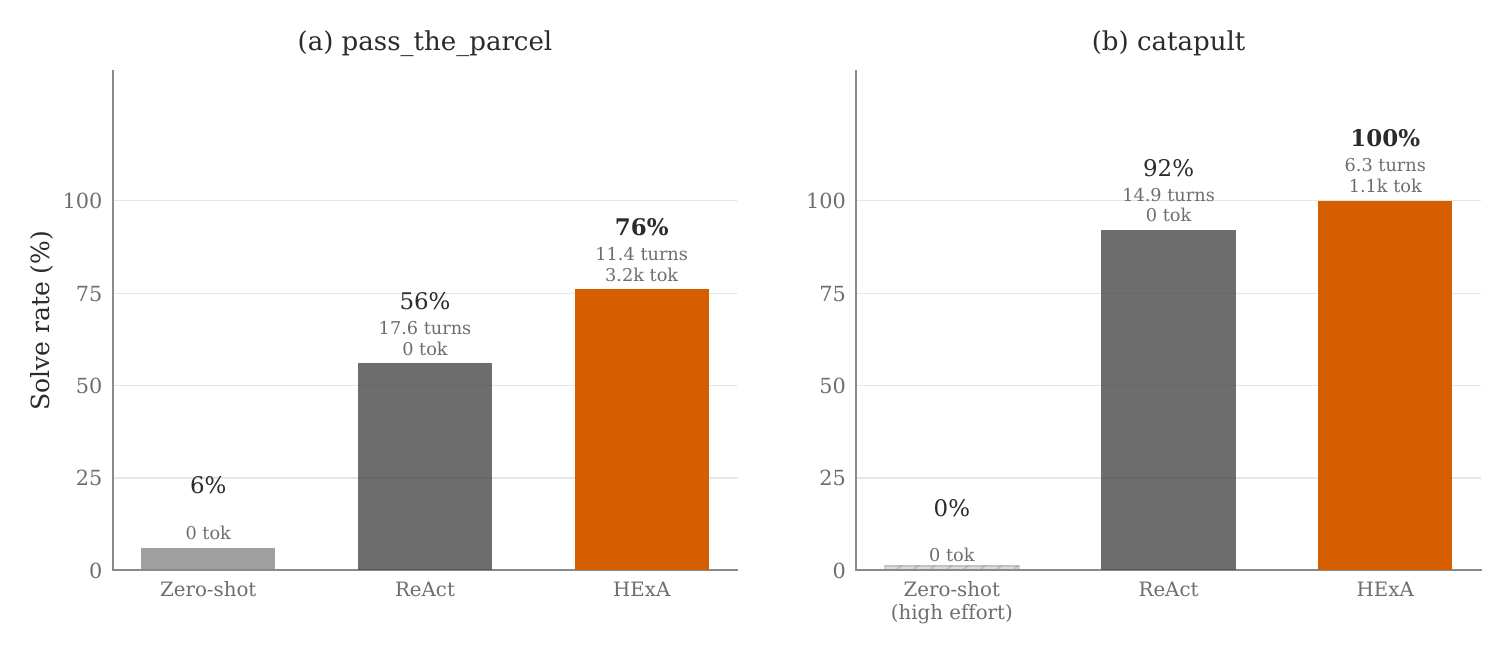}
  \caption{%
    \textbf{\textsc{HExA} lifts GPT-6-Astra past \textsc{ReAct} with fewer
    turns.}
    Solve rate on \emph{(a)} \texttt{pass\_the\_parcel} and
    \emph{(b)} \texttt{catapult} with the frozen skill bank; labels report
    mean turns per seed and injected tokens per episode.%
  }
  \label{fig:astra_ptp_catapult_main}
  \vspace{-3pt}
\end{figure}

\begin{table*}[!h]
\centering
\scriptsize
\setlength{\tabcolsep}{3pt}
\renewcommand{\arraystretch}{0.88}

\caption{%
   \textbf{\textsc{HExA} improves the actor in all ten external
  settings and is the best method in nine.}
  \emph{Left:} per-task results for each actor. We compare \textsc{HExA} with
  the actor alone (Zero-Shot) and memory baselines: Episodic RAG, which
  retrieves past rollouts, and MemRL, which retrieves utility-scored memories.
  Every memory is built on training instances, frozen, and
  evaluated on held-out instances, with all arms paired on the same instances.
  \emph{Games}: success in at least one of three episodes;
  \emph{reasoning \& planning}: accuracy on held-out items;
  \emph{\textsc{DiscoverPhysics}}: explanation score from an independent judge
  (\texttt{claude-opus-4-6}) on held-out seeds. Entries are means over three
  runs $\pm$ one standard deviation; bold marks the best mean in each row.
  \emph{Right:} the same results averaged over the tasks of each family, one
  group per actor: (a) Mastermind and Maze, (b) the four reasoning tasks,
  (c) \texttt{oscillator} and \texttt{circle}. Detailed results per environment are in 
  Appendices~\ref{app:results:games}--\ref{app:results:discovery}.
}
\label{tab:external_main}
\label{fig:external_bench}
\vspace{3pt}

\begin{minipage}[c]{0.63\textwidth}
\centering
\resizebox{\linewidth}{!}{%
\begin{tabular}{@{}llcccc@{}}
\toprule
\textbf{Task} & \textbf{Actor} & \textbf{Zero-Shot} &
\textbf{Episodic RAG} & \textbf{MemRL} & \textbf{\textsc{HExA}} \\
\midrule
\multicolumn{6}{@{}l}{\emph{Interactive games} --- \texttt{any}-episode success rate}\\
Mastermind (\texttt{any}) & \texttt{gpt-oss-120b} & .683{\scriptsize$\pm$.031} & .650{\scriptsize$\pm$.094} & .690{\scriptsize$\pm$.244} & \textbf{.883}{\scriptsize$\pm$.012} \\
Mastermind (\texttt{any}) & Qwen3-8B              & .367{\scriptsize$\pm$.072} & .492{\scriptsize$\pm$.031} & .392{\scriptsize$\pm$.043} & \textbf{.640}{\scriptsize$\pm$.082} \\
Maze (\texttt{any})       & \texttt{gpt-oss-120b} & .607{\scriptsize$\pm$.016} & .601{\scriptsize$\pm$.020} & \textbf{.667}{\scriptsize$\pm$.024} & .649{\scriptsize$\pm$.045} \\
Maze (\texttt{any})       & Qwen3-8B              & .479{\scriptsize$\pm$.019} & .483{\scriptsize$\pm$.067} & .513{\scriptsize$\pm$.032} & \textbf{.564}{\scriptsize$\pm$.010} \\
\midrule
\multicolumn{6}{@{}l}{\emph{Reasoning \& planning} --- accuracy on 100 held-out items}\\
MathGAP        & \texttt{gpt-oss-120b} & .510{\scriptsize$\pm$.006} & .560{\scriptsize$\pm$.010} & .600{\scriptsize$\pm$.015} & \textbf{.863}{\scriptsize$\pm$.009} \\
ZebraLogic     & \texttt{gpt-oss-120b} & .560{\scriptsize$\pm$.026} & .720{\scriptsize$\pm$.030} & .650{\scriptsize$\pm$.020} & \textbf{.807}{\scriptsize$\pm$.018} \\
Tower of Hanoi & \texttt{gpt-oss-120b} & .263{\scriptsize$\pm$.007} & .377{\scriptsize$\pm$.037} & .273{\scriptsize$\pm$.027} & \textbf{.520}{\scriptsize$\pm$.010} \\
Matchstick     & \texttt{gpt-oss-120b} & .737{\scriptsize$\pm$.023} & .767{\scriptsize$\pm$.009} & .703{\scriptsize$\pm$.023} & \textbf{.787}{\scriptsize$\pm$.007} \\
\midrule
\multicolumn{6}{@{}l}{\emph{\textsc{DiscoverPhysics}} --- explanation score over 15 held-out seeds}\\
\texttt{oscillator} & \texttt{claude-sonnet-5} & .516{\scriptsize$\pm$.044} & .822{\scriptsize$\pm$.008} & .780{\scriptsize$\pm$.007} & \textbf{.853}{\scriptsize$\pm$.018} \\
\texttt{circle}     & \texttt{claude-sonnet-5} & .580{\scriptsize$\pm$.012} & .580{\scriptsize$\pm$.023} & .569{\scriptsize$\pm$.025} & \textbf{.916}{\scriptsize$\pm$.034} \\
\bottomrule
\end{tabular}%
}
\end{minipage}%
\hfill
\begin{minipage}[c]{0.35\textwidth}
\centering
\includegraphics[width=\linewidth]{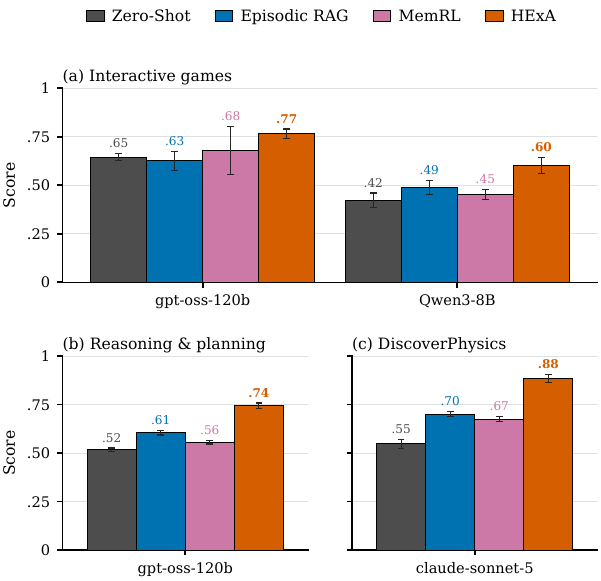}
\end{minipage}

\vspace{-5pt}
\end{table*}

\emph{Why distill experience into compositional hierarchical skills rather than naively retaining them?}
These gains cannot be explained by access to interaction alone: \textsc{ReAct} experiments on every instance but cannot reuse discoveries across episodes. Episodic RAG and \textsc{MemRL} retain prior experience, yet their retrieved trajectories and memories provide less effective guidance than \textsc{HExA}'s compact, evolving skill bank. On \texttt{catapult}, \textsc{HExA} achieves $82.7\%$ success against $62.9\%$ for Episodic RAG and $49\%$ for \textsc{MemRL}, while injecting only $2.4$k tokens compared with $6.5$k and $10.1$k, respectively. Crucially, this improvement does not require additional test-time exploration: \textsc{HExA} reduces mean interaction turns from $23.1$ to $12.5$ on \texttt{catapult} and from $18.7$ to $5.8$ on \texttt{pass\_the\_parcel} relative to \textsc{ReAct}. Accounting for both skill-bank construction and evaluation, \textsc{HExA} also incurs substantially lower total token costs than episodic-memory baselines (Table~\ref{tab:total_cost}). Thus, the benefit is not simply remembering more experience, but \emph{abstracting experimental evidence into reusable principles, applicable strategies, and failure corrections} that guide subsequent experimentation while amortizing its cost.

\emph{Does HExA generalize to other tasks and benchmarks?}
We further evaluate \textsc{HExA} across interactive games, multi-step reasoning and planning, and physics law discovery (Table~\ref{tab:external_main}). Under the same build--freeze--evaluate protocol, \textsc{HExA} improves its corresponding frozen actor in all ten task--actor settings and achieves the highest mean in nine. Notably, MathGAP accuracy increases from $51.0\%$ to $86.3\%$, Mastermind success with GPT-OSS-120B rises from $68.3\%$ to $88.3\%$, and the explanation score on the \texttt{circle} world of \textsc{DiscoverPhysics} improves from $0.580$ to $0.916$. Unlike episodic-memory baselines, which occasionally degrade performance, the distilled bank provides consistent improvements across these different tasks. Together, these findings support a broader principle: \emph{converting interaction experience into compact, reusable skills enables in-context policy improvement}, while skill-conditioned experimentation facilitates efficient adaptation to new problems.

\setlength{\intextsep}{0pt}
\setlength{\columnsep}{5pt}
\begin{wrapfigure}[19]{r}{0.48\linewidth}
  \vspace{-\intextsep}
  \centering
  \setlength{\abovecaptionskip}{3pt}
  \setlength{\belowcaptionskip}{3pt}
  \includegraphics[width=\linewidth]{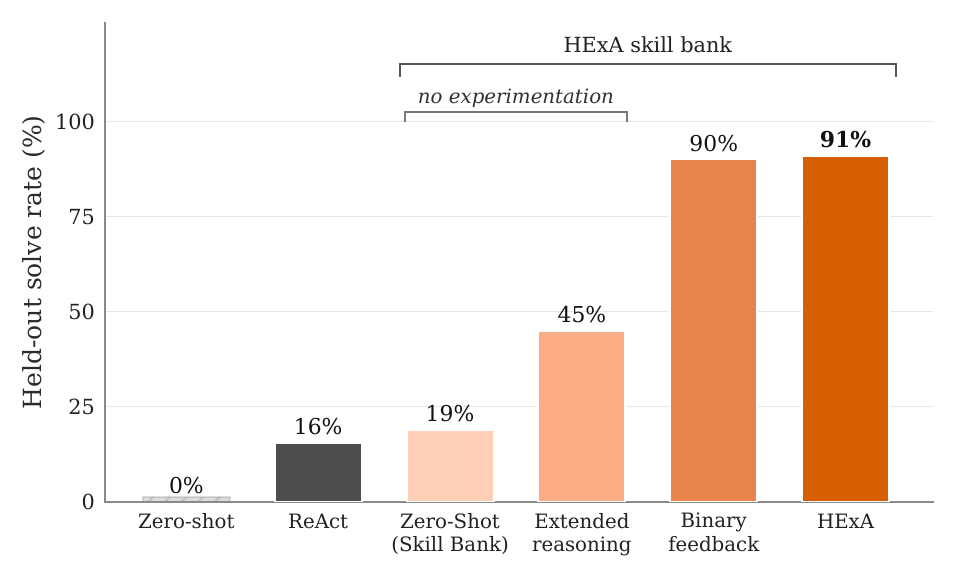}

\caption{\textbf{Effect of evaluation-time feedback.}
    Held-out solve rate for Claude Sonnet 4.6, averaged over \texttt{catapult}
    and \texttt{pass\_the\_parcel}. All orange bars use the same frozen
    \textsc{HExA} skill bank; lighter shades indicate less environment feedback.
    Without feedback, the bank yields a 19\% solve rate, rising to 45\% with
    extended reasoning. Providing only binary success or failure feedback raises
    the solve rate to 90\%. Per-level results are in Table~\ref{tab:skills_no_tools} in Appendix.}
  \label{fig:skills_no_tools_avg}
\end{wrapfigure}

\emph{Do acquired skills replace experimentation or enable better experiments?}
We disentangle these complementary roles by evaluating the same frozen skill bank with progressively more environment feedback and experimentation at inference (Figure~\ref{fig:skills_no_tools_avg}). Without any experimentation, the skill bank already improves zero-shot success from $0\%$ to $19\%$, rising to $45\%$ with extended reasoning. This shows that experimental learning produces reusable knowledge that improves the actor's reasoning even without further interaction. 
However, this setup of experiment-centric reasoning alone significantly underperform the full HEXA setup---allowing the skill-conditioned actor to also experiment and test hypotheses at inference, increasing success to $91\%$ with full feedback from simulation tools. Thus, \emph{learning skills via active experimentation and using them for further experimenting} serve distinct and complementary purposes: accumulated skills provide prior knowledge and guide the search efficiently, while new experiments with skill-conditioning acts as stepping-stones for subsequent behavior, supplying instance-specific evidence needed to further adapt the discovered knowledge to unfamiliar configurations.

\subsection{Instantiation of \textsc{HExA} and Ablations}

\label{sec:instantiation}

The skill bank in \textsc{HExA} has three orthogonal axes of instantiation
(Algorithms~\ref{alg_1} and~\ref{alg:HeXa_algo_app}): how it is \emph{initialized} before experimentation, how
it is \emph{updated} between rounds, and whether we attach interaction reward
labels to the skills.

\setlength{\intextsep}{0pt}
\setlength{\columnsep}{5pt}
\begin{wrapfigure}[20]{r}{0.48\linewidth}
  \vspace{-\intextsep}
  \centering
  \setlength{\abovecaptionskip}{3pt}
  \setlength{\belowcaptionskip}{3pt}
  \includegraphics[width=\linewidth]{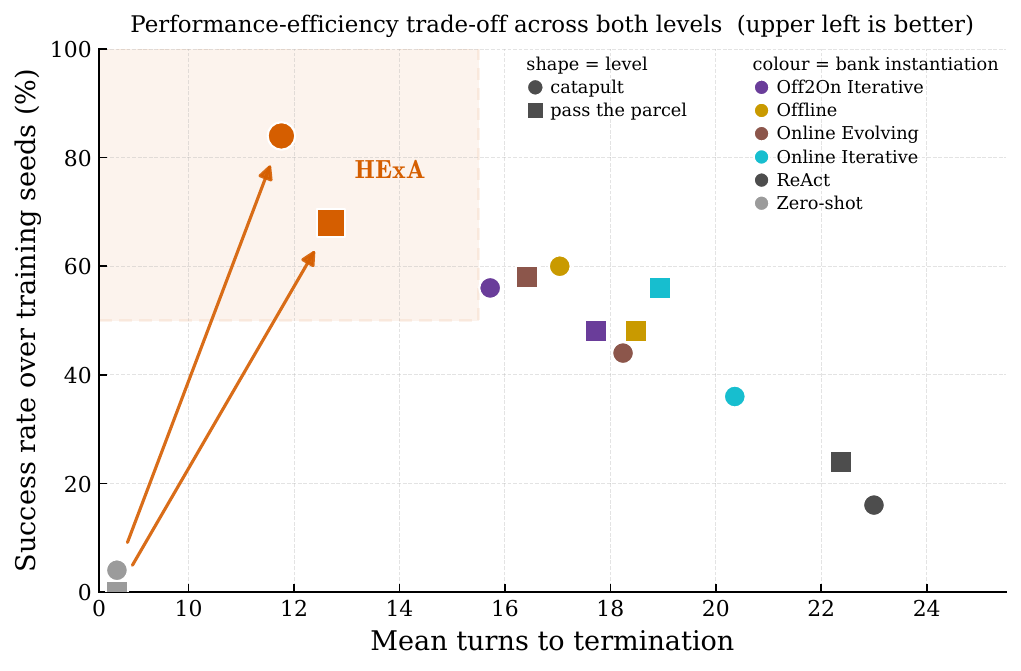}
\caption{\textbf{HExA variants accuracy--efficiency comparison.}
Solve rate vs.\ mean turns to termination for each skill-bank initialization and update strategy with ablations on Sonnet 4.6. Colour marks the strategy and shape marks the task; the upper left is better. On both levels, Off2On + Evolution is the strategy with highest accuracy and efficiency. Arrows show its gain over \textsc{Zero-shot}. Details in Table~\ref{tab:variants_compact}.}
  \label{fig:training_curves}
\end{wrapfigure}

\emph{Does skill-bank initialization and update matter?}
We compare three initialization regimes: \emph{Offline}, which distills from a warm-start batch and then freeze the skill bank; \emph{Off2On}, which uses the same
warm start but continues evolving the bank from online interaction; and
\emph{Online}, which starts from an empty bank and bootstraps skills purely from the agent's own interactions. For skill bank updates we compare \emph{Evolution}, where the evolver receives
both the accumulated bank and the newest trajectory batch, against
\emph{Iterative Replacement}, where it rebuilds the bank from the newest trajectory batch alone. Across these configurations of HExA, as shown in Figure~\ref{fig:training_curves} and Appendix
Table~\ref{tab:variants_compact}, \emph{Off2On + Evolution} gives the best
accuracy--efficiency trade-off is therefore our default
\textsc{HExA} configuration.
The update comparison is particularly informative: Evolution outperforms
Iterative Replacement on both levels.
Because round-$n$ trajectories are themselves generated while conditioned on
$\mathcal{K}_{n-1}$, retaining the previous bank allows new evidence to refine,
delimit, merge, or replace accumulated knowledge explicitly; rebuilding from
the latest batch alone discards that explicit state and conditions the update
only on the most recent trajectories.
Figure~\ref{fig:training_curves} further shows that the higher-performing
configurations also terminate in fewer turns.

\emph{How do reward labels help learning via experimentation?} \textsc{HExA} is also flexible to ingest additional experimental evidence
such as trajectory rewards, which are a function of outcome reward and efficiency
(Appendix~\ref{sec:reward}), to further guide skill evolution and retrieval. 
\setlength{\intextsep}{0pt}
\setlength{\columnsep}{10pt}
\begin{wrapfigure}[16]{r}{0.48\linewidth}
  \vspace{-\intextsep}
  \centering
  \setlength{\abovecaptionskip}{3pt}
  \setlength{\belowcaptionskip}{3pt}
  \includegraphics[width=\linewidth]{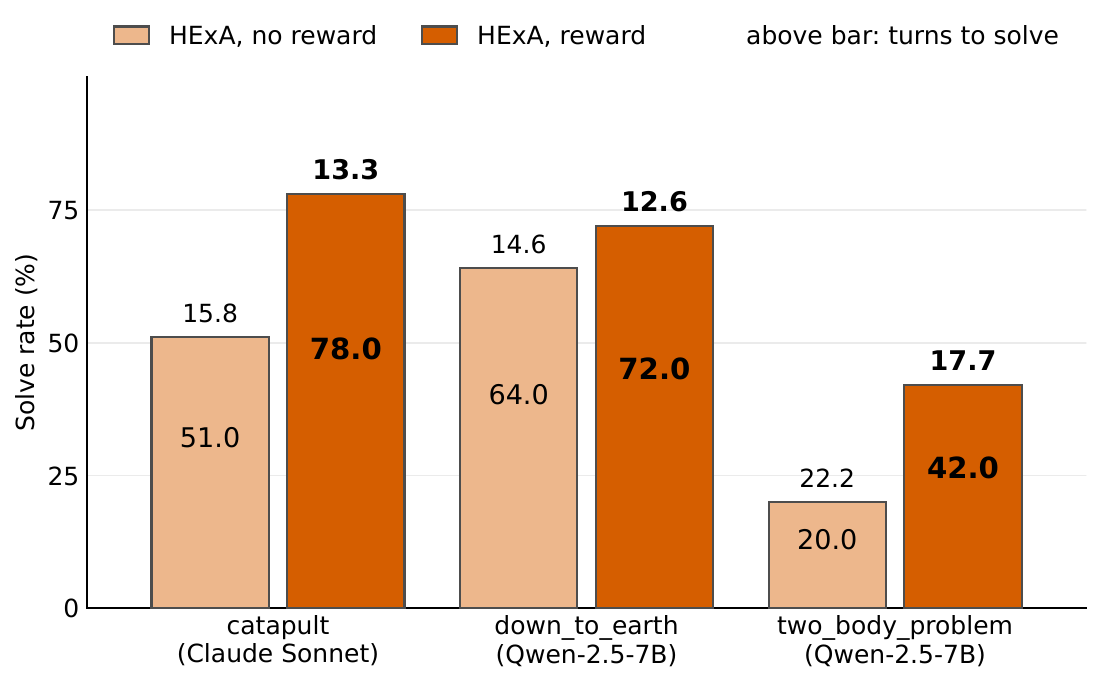}
  \caption{\textbf{} Solve rate (\%) per level; the number above each bar is the mean turns to solve. The \emph{no reward} bar treats every success and failure identically, so the evolver prunes from its own prior rather than from environment feedback. Details in Table~\ref{tab:reward_ablation}.}
\label{fig:reward_ablation}
\end{wrapfigure}
The
skill reward $r_k$ is in turn a function of the mean of individual source trajectory rewards ($\bar{r}_{\mathrm{src}}$) that invoked the skill. We define the reward in detail in Appendix~\ref{app:exp_res_final_instantiation}. HExA as a framework is agnostic to the choice of trajectory reward and skill reward and can function with no rewards. In our ablation without the reward labels, every success and failure trajectory, and every skill distilled out of them, is treated equally, and the LLM evolver implicitly updates the bank using its existing knowledge rather than environment feedback. Removing the labels costs accuracy as well as efficiency in terms of turns to solve the task on every level we ablated (details in Figure~\ref{fig:reward_ablation}, Table~\ref{tab:reward_ablation_main}). The reward labels thus provide environment-grounded feedback from the actor's own exploration that helps assign credit for effective and efficient skill acquisition, and better guides the evolver in composing, retaining or pruning skills.

\begin{figure}[!h]
  \centering

  \setlength{\abovecaptionskip}{2pt}
  \setlength{\belowcaptionskip}{0pt}

  \includegraphics[width=\linewidth]{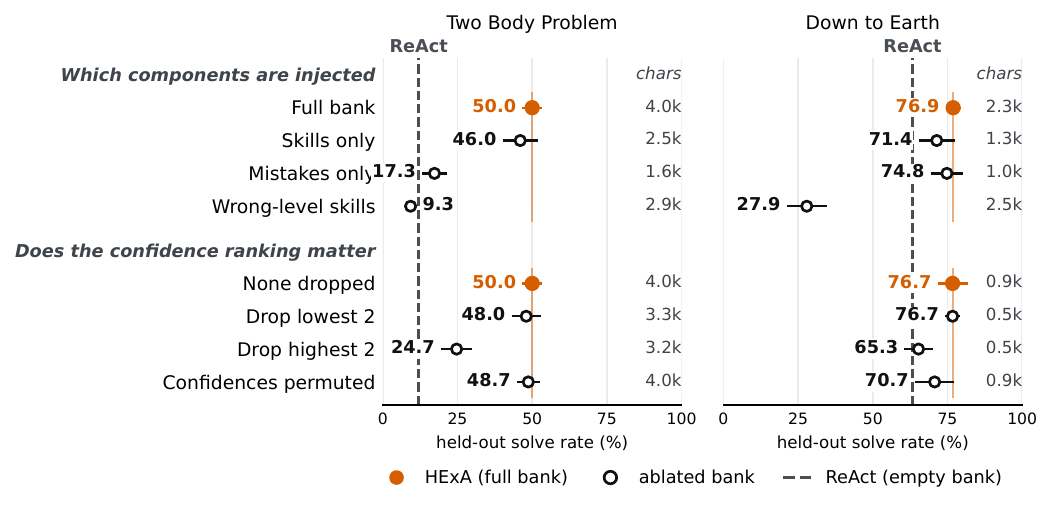}

  \caption{%
    \textbf{How the skill bank content drives the gains.} \emph{Top:} removing skills or mistakes hurts, while unnecessary wrong-level skills hurt and go below \textsc{ReAct} baseline.
    \emph{Bottom:} removing high-reward skills hurts substantially more than removing low-reward ones; permuting rewards also degrades performance.
  }

    \label{fig:bank_ablation}

  \vspace{-1pt}
\end{figure}

\emph{How does skill bank content affect results?} 
\label{sec:results:structure}Frozen-bank controls rule out the possibility that \textsc{HExA} benefits merely
from adding more or plausible-looking text to the prompt
(Figure~\ref{fig:bank_ablation}). On \texttt{two\_body\_problem}, replacing
task-specific skills with comparable-length wrong-level skills reduces success
from $50.0\%$ to $9.3\%$, near the empty-bank \textsc{ReAct} baseline
($12.0\%$); removing either skills or mistakes also degrades performance.
The learned ranking is likewise informative: on \texttt{down\_to\_earth},
dropping the two highest-reward skills lowers success from $76.7\%$ to
$65.3\%$, while dropping the two lowest leaves it unchanged; permuting
rewards also hurts despite retaining every skill. Thus, the gain depends on
task-relevant distilled content and its learned prioritization, rather than
context length alone. Full component, ranking, and context-length controls are
reported in Appendix~\ref{app:bank-content}. Thus, our default experimental setting for \textsc{HExA} is Off2On~+~Evolution with reward labeled skill bank having strategy skills, mistakes as well as general principle skills.

\subsection{Hierarchical skills help zero-shot generalization to unseen levels}
\label{sec:results:transfer}

\textsc{HExA}'s hierarchical structure arises because later trajectories are
generated while conditioned on previously learned skills, allowing subsequent
evidence to refine, delimit, combine, or replace earlier rules rather than
accumulating as isolated observations.
The transfer experiments test whether these abstractions remain useful beyond
the task on which they were learned. 
\setlength{\intextsep}{0pt}
\setlength{\columnsep}{10pt}
\begin{wrapfigure}[17]{r}{0.48\linewidth}
  \vspace{-\intextsep}
  \centering
  \setlength{\abovecaptionskip}{3pt}
  \setlength{\belowcaptionskip}{3pt}
  \includegraphics[width=\linewidth]{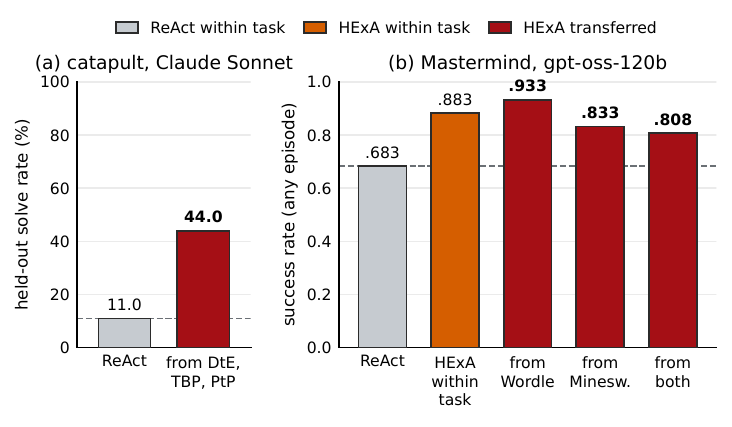}
  \caption{\textbf{HExA on transferred hierarchical skill banks improve zero-shot performance on target-tasks.}
    \emph{(a)} On \texttt{catapult}, skills synthesized from three easier levels raise Claude Sonnet to $44\%$.
    \emph{(b)} On Mastermind, transferred banks can even improve over within task HExA reaching 93\%. Full results are in Table~\ref{tab:transfer}.}
    \label{fig:transfer}
\end{wrapfigure}
The synthesized \texttt{catapult} bank makes this composition explicit:
its 9 strategy skills contain 22 references to source-level skills
($2.44$ per synthesized skill), while its 4 mistake rules contain 9 source
references ($2.25$ per synthesized mistake), details in Appendix~\ref{app:results:transfer:skill-provenance}. The transfer evolver combines lower-level knowledge about impact direction,
drop height, mass, geometric clearance, and failure recovery into
target-grounded mechanism-level skills rather than overfitting source-specific
settings. As show in Figure~\ref{fig:transfer}, these abstractions transfer purely zero-shot without any target-task trajectories or interactions: combining three easier \textsc{InterPhyre} source skill banks raises Claude Sonnet performance on \texttt{catapult} from $11\%$ to $44.0\%$.
Transfer gains also appear on other \textsc{Interphyre} levels and across
Mastermind, Maze, and \textsc{DiscoverPhysics}.
On Mastermind, HExA with the transferred bank even beats within-task \textsc{HExA}.
Full transfer results and the provenance analysis are given in
Section~\ref{sec:cross_task_transfer} and
Table~\ref{tab:transfer} in Appendix~\ref{app:results:transfer}. These results highlight how in-context learning via experimentation, enables hierarchical skill acquisition through HExA's skill evolution methods and also shows the necessity of learning reusable and transferable knowledge as compact skills as a step towards improving generalization to unseen complex domains.

\subsection{In-context policy improvement versus parametric post-training}

\label{sec:results:grpo}

To compare \textsc{HExA}'s non-parametric adaptation with RL, we
post-train Qwen-2.5-3B using GRPO on
\textsc{Interphyre} tasks (Figure~\ref{fig:grpo_comparison_main}).
In the match interaction budget setup, \textsc{HExA} reaches $23.3\%$ on
\texttt{down\_to\_earth} and $12.0\%$ on \texttt{two\_body\_problem},
while GRPO reaches $20\%$ and $6\%$ after its first training epoch.
With continued training, GRPO subsequently matches HExA scores at 3.2 and 1.5 times more interaction budget respectively.
The comparison highlights complementary adaptation timescales:
\textsc{HExA} amortizes exploration and makes newly discovered strategies immediately available through
in-context learning, whereas gradient-based RL can consolidate experience into model
weights through further optimization.
Full training curves and details are in
Appendix~\ref{app:results:grpo}.

\begin{figure}[!h]
  \centering
  \includegraphics[width=\linewidth]{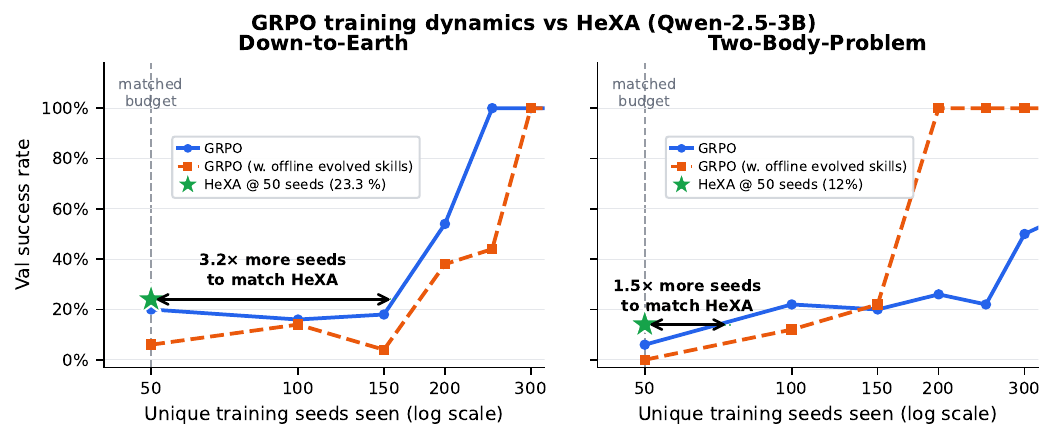}
\caption{%
  \textbf{In-context skill evolution learns faster than GRPO at the same interaction budget.}
  We compare \textsc{HExA} with GRPO fine-tuning on Qwen-2.5-3B using 50 training seeds.
  The green star marks \textsc{HExA}'s success rate after the same 50-seed budget
  (Down to Earth: $23.3\%$, Two Body Problem: $12\%$), while the dashed line marks the
  matched-budget GRPO checkpoint. In this low-data regime, \textsc{HExA} outperforms both
  GRPO variants, suggesting that reusable in-context skills provide faster early adaptation;
  with many more updates, GRPO can eventually close the gap through direct reward optimization.
}
  \label{fig:grpo_comparison_main}
\end{figure}

\section{Conclusion, Limitations, \& Future Works}

We introduced \textsc{HExA}, a non-parametric in-context RL framework in which a frozen LLM designs experiments, distills interaction experience into hierarchical skills, and reuses them across tasks without offline data or external supervision. To evaluate when active experimentation is necessary, we also introduced \textsc{InterPhyre}, a suite of hard physical reasoning tasks with tool APIs for scene inspection, intervention, and hypothesis testing. Across \textsc{InterPhyre}, \textsc{DiscoverPhysics}, multi-step reasoning, and interactive games, \textsc{HExA} outperforms agentic and memory baselines with fewer turns and tokens. On the hardest levels, frontier models such as GPT-6-Astra fail zero-shot, \textsc{HExA} reaches up to 100\% success, and its hierarchical skill acquisition mechanism, enables zero-shot  transfer to unseen harder levels. At matched budget of interactions, it also outperforms GRPO, suggesting that reusing discovered strategies through context amortizes exploration more efficiently than weight updates when data is scarce.

\textbf{Limitations and Future Work.} Skill formation eventually saturates, and additional test-time compute alone does not help directly. Skill retrieval also relies on simple engineered rewards. Promising directions include unsupervised environment design to generate more diverse learning setups, interleaving skill learning with weight updates for continual learning, and automating reward design and skill retrieval. Moreover, designing algorithms to train models to fundamentally improve in-context learning abilities and learning from interactions and experimentation is also an open problem and a direction for future works.

\section*{Acknowledgments}
This work has taken place in the Safe, Correct, and Aligned Learning and Robotics Lab (SCALAR)
at The University of Massachusetts Amherst. SCALAR research is supported in part by the NSF
(IIS-2437426), the Long-Term Future Fund, and Open Philanthropy. Scott Niekum holds concurrent
appointments as a Professor at the University of Massachusetts Amherst and as an Amazon
Scholar. This paper describes work performed at the University of Massachusetts Amherst and was
funded by a gift from Amazon.

\bibliography{iclr2027_conference}
\bibliographystyle{iclr2027_conference}


\appendix

\newpage
\addtocontents{toc}{\protect\setcounter{tocdepth}{2}}
\renewcommand{\contentsname}{Appendix Contents}
\setcounter{tocdepth}{2}
\tableofcontents

\newpage

\section{More Background and Preliminaries}


\paragraph{Tool-Augmented Large Language Model (LLM) Agents.} 
To interact with complex environments, LLMs are frequently augmented with external tools and programmatic execution capabilities. This allows them to deeply interleave reasoning with acting, enabling them to gather information, execute APIs, or manipulate simulators directly \citep{yao2022react, schick2023toolformer}. By structuring interactions as a sequence of thoughts, actions, and observations, tool-augmented agents transcend the limitations of static parametric knowledge and can adapt their strategies based on real-time environmental feedback.

\paragraph{Reinforcement Learning and MDPs.}
We model the interaction between an agent and its environment as a Markov Decision Process (MDP)~\citep{sutton1998introduction}, defined by the tuple $\mathcal{M} = \langle \mathcal{S}, \mathcal{A}, p, r, \gamma, p_0 \rangle$. Here, $\mathcal{S}$ is the state space, $\mathcal{A}$ is the action space, $p(s' \mid s, a)$ is the transition dynamics, $r(s, a)$ is the reward function, $p_0(s)$ is the initial state distribution, and $\gamma \in [0, 1)$ is the discount factor. At time $t$, the agent observes state $S_t \sim p_0$, selects an action $A_t \sim \pi(\cdot \mid S_t)$ according to its policy $\pi$, receives a reward $R_{t+1}$, and transitions to $S_{t+1}$. This generates a trajectory $\tau_t = (S_0, A_0, R_1, \dots, S_{t-1}, A_{t-1}, R_t)$. 

The core tasks in RL are \emph{policy evaluation} (estimating the expected return $v^\pi(s) = \mathbb{E}[\sum_{i=1}^\infty \gamma^{i-1} R_{t+i} \mid S_t = s]$) and \emph{control} (finding an optimal policy to maximize $J(\pi) = \sum_s p_0(s)v^\pi(s)$). By iterating between evaluating a policy and improving it, RL algorithms optimize a parameterized policy $\pi_\theta$. This optimization can occur via \emph{online pretraining} (interacting directly with the environment) or \emph{offline pretraining} (learning from a static dataset of prior interactions $\mathcal{D} = \{(s_i, a_i, r_i, s'_i)\}$).

Traditional methods for optimizing LLM policies rely on gradient-based parameter updates. For instance, Group Relative Policy Optimization (GRPO) avoids training a separate critic by using intra-group relative rewards to optimize the policy. For a given query $x$, the model samples $G$ responses $\{y^{(1)}, \dots, y^{(G)}\}$, which are scored to obtain rewards $\{R_1, \dots, R_G\}$. GRPO computes normalized advantages and updates the policy with a clipped objective:
\begin{equation}
J_{\text{GRPO}}(\theta) = \mathbb{E}_{x, \{y^{(i)}\}} \left[ \frac{1}{G} \sum_{i=1}^{G} \min \left( r_i A_i, \text{clip}(r_i, 1 - \epsilon, 1 + \epsilon) A_i \right) - \beta D_{\text{KL}}(\pi_{\theta} \| \pi_{\text{ref}}) \right]
\end{equation}

\paragraph{In-Context Reinforcement Learning (ICRL).} To circumvent the need for continuous parameter updates, In-Context Reinforcement Learning (ICRL) conditions the policy on both the current state $S_t$ and an evolving history or context $C_t$. Here, the action $A_t$ is sampled according to $\pi_{\theta}(\cdot \mid S_t, C_t)$. The pre-trained weights $\theta$ remain frozen, yet the policy can achieve high rewards in test environments that differ from those seen during pre-training. This generalization arises because the forward pass of the neural network effectively implements an RL algorithm that learns from the context $C_t$. The performance of $\pi_{\theta}$ generally improves with the length and quality of $C_t$, a phenomenon referred to as \emph{in-context improvement}~\citep{moeini2025survey}. 

\newpage


\section{Additional Methodological Details for \textsc{HExA}}
\label{app:hexa}

\begin{figure}[H]
  \centering
  \includegraphics[width=\linewidth]{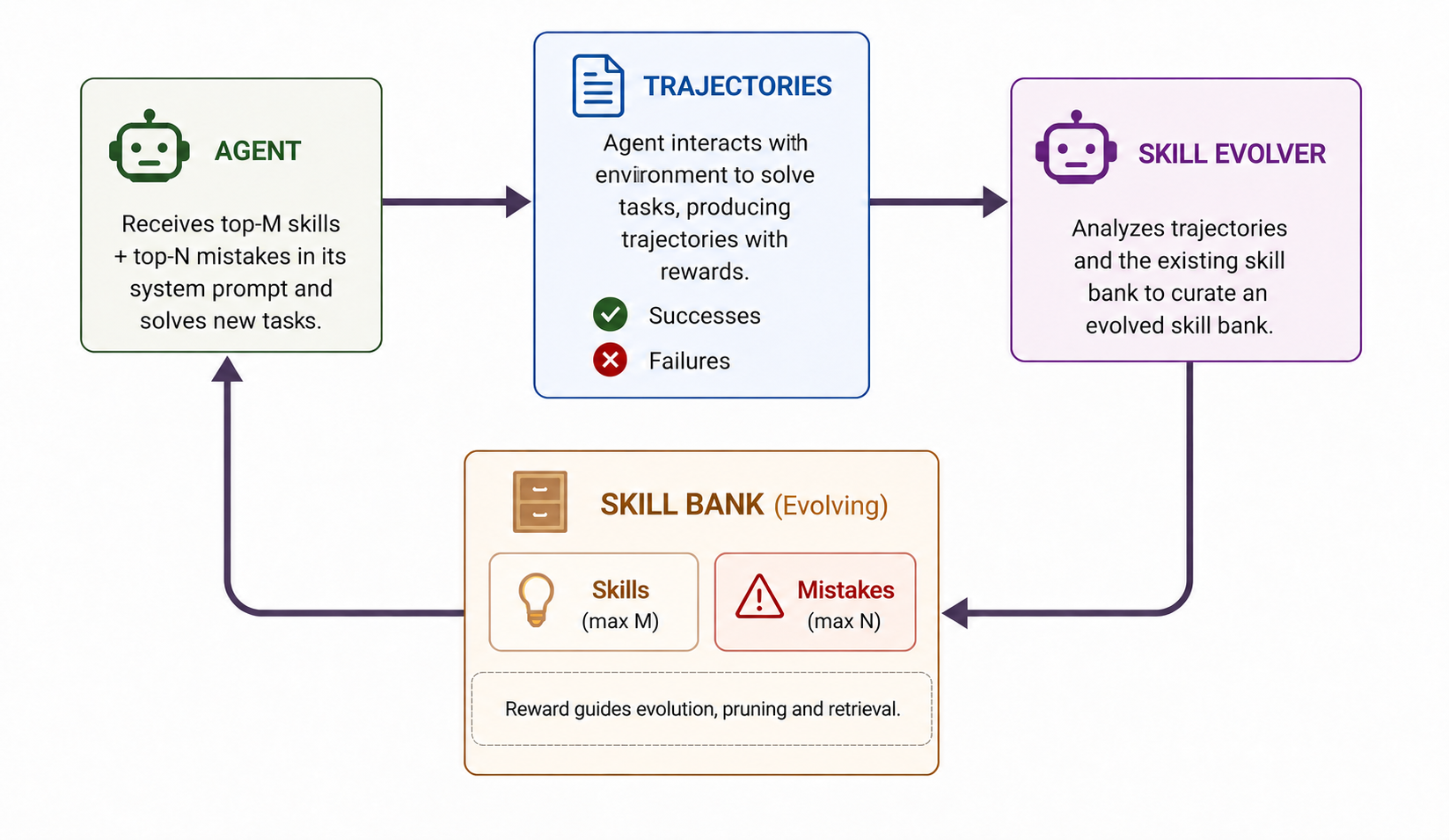}
  \caption{%
    \textbf{The \textsc{HExA} actor-evolver loop.}
    The actor receives distilled skills in its system prompt and runs on
    physics puzzles, producing trajectories annotated with rewards
    (successes~\cmark, failures~\xmark).
    The Skill Evolver analyzes these trajectories and curates the
    evolving Skill Bank: skills (max~M) are merged, pruned, or promoted
    by reward signal, and common mistakes (max~N) are updated accordingly.
    Each skill carries a generation index tracking its lineage across
    evolution rounds (Eq.~\ref{eq:weight}).
    The updated bank is injected back into the actor's system prompt for
    the next round, closing the loop.
  }
  \label{fig:hexa_loop}
\end{figure}

This appendix expands the \textsc{HExA} framework introduced in
Section~\ref{sec:HExA}. We use the notation and problem formulation defined in the
main paper and focus here on methodological details omitted there for space:
trajectory-level reward construction, the two-stage skill-distillation procedure,
skill-bank construction and retrieval, initialization and update regimes, and
cross-task skill synthesis. We additionally discuss the design choices that make
\textsc{HExA} applicable without parameter updates or externally supplied skill
annotations.

\subsection{HExA as online non-parametric in-context reinforcement learning}
\label{app:hexa_objective}

In-context reinforcement learning (ICRL) broadly refers to policy improvement
that occurs through an agent's context while its deployed model parameters
remain fixed. A common approach is to first train or meta-train a sequence
model on RL interaction histories so that, at deployment, the model can infer
how its behavior should change from the actions, observations, and rewards
appearing in its context. HExA shares the goal of policy improvement without
test-time parameter updates, but differs in where the learning mechanism and
learning state reside.

HExA requires no additional meta-training of the actor to implement a
particular RL algorithm. Instead, given a frozen tool-augmented actor
$\pi_{\mathrm{act}}$, it constructs an explicit external learning state
$\mathcal{K}$ from the actor's own online interaction experience.
For a bank $\mathcal{K}$, define
\begin{equation}
\mathcal{G}_{\mathcal{K},\ell}
=
\textsc{Retriever}(\mathcal{K},\ell,M,N),
\end{equation}
and the effective bank-conditioned policy
\begin{equation}
\pi_{\mathcal{K}}
(a_t\mid\ell,s,h_t;\mathcal{F})
:=
\pi_{\mathrm{act}}
(a_t\mid\ell,s,\mathcal{G}_{\mathcal{K},\ell},h_t;\mathcal{F}).
\label{eq:effective_policy}
\end{equation}
The parameters of $\pi_{\mathrm{act}}$ are fixed, but its effective policy
changes as $\mathcal{K}$ changes.

Let $\mathcal{D}_{\ell}$ denote the distribution of future instances of task
family $\ell$. We define the utility of a bank as the expected performance of
the policy it induces:
\begin{equation}
J_{\ell}(\mathcal{K})
=
\mathbb{E}_{s\sim\mathcal{D}_{\ell},
\,\tau\sim\pi_{\mathcal{K}}(\cdot\mid s)}
\left[
y_{\ell}(\tau)
\right],
\label{eq:hexa_full_objective}
\end{equation}
where $y_{\ell}(\tau)$ is the task-success predicate defined in
Eq.~\ref{eq:success_main}. More generally, $y_{\ell}$ can be replaced by any
task-level return or utility. Conceptually, HExA seeks a bounded external state
\begin{equation}
\mathcal{K}^{*}
\in
\arg\max_{\mathcal{K}\in\mathfrak{K}}
J_{\ell}(\mathcal{K}),
\end{equation}
where $\mathfrak{K}$ denotes the allowed class of bounded skill banks.
Unlike gradient-based RL, this optimization does not update
$\pi_{\mathrm{act}}$; improvement is represented entirely by the evolving
external state $\mathcal{K}$.

Crucially, HExA does not optimize this objective from a fixed dataset.
Its learning data are generated online by the policy induced by the current
bank. At round $n$,
\begin{equation}
\mathcal{G}_{n}
=
\textsc{Retriever}(\mathcal{K}_{n-1},\ell,M,N),
\qquad
\mathcal{T}^{(n)}
\sim
\pi_{\mathcal{K}_{n-1}},
\end{equation}
after which the evolver constructs
\begin{equation}
\mathcal{K}_{n}
=
\pi_{\mathrm{ev}}
\left(
\mathcal{K}_{n-1},
\mathcal{T}^{(n)}
\right).
\label{eq:online_bank_update}
\end{equation}
Thus the current learning state determines the actor's subsequent behavior and
therefore which evidence is collected next. Bank evolution and data collection
form a closed loop:
\[
\mathcal{K}_{n-1}
\rightarrow
\pi_{\mathcal{K}_{n-1}}
\rightarrow
\mathcal{T}^{(n)}
\rightarrow
\mathcal{K}_{n}.
\]
In HExA, these trajectories arise from active environment interaction: the
actor reasons over hypotheses, chooses tool calls or interventions, observes
their consequences, and adapts its subsequent experiments.

The trajectories are also not retained merely as additional demonstrations.
The evolver contrasts successful and unsuccessful experience and compresses it
into reusable strategy skills, mistake corrections, and partial discoveries.
Consequently, both positive and negative experience can change the effective
policy: successful trajectories provide evidence for strategies to retain,
while failures can identify invalid hypotheses, recurrent failure modes, and
correct intermediate discoveries. Because trajectories in round $n$ are
generated under skills learned before round $n$, subsequent skill evolution
can refine, compose, delimit, or replace previously acquired knowledge.

We therefore use \emph{online} to refer to the skill-construction phase:
experience collection and policy adaptation are interleaved, rather than
separated into a fixed offline dataset followed by adaptation. Held-out
evaluation is deliberately different: after the final round,
$\mathcal{K}_{R}$ is frozen and no further skill evolution occurs on
$\mathcal{S}^{\mathrm{ev}}_{\ell}$.

Finally, we define overall in-context self-improvement relative to the
unconditioned actor as
\begin{equation}
\Delta J_{\ell}^{\mathrm{HExA}}
=
J_{\ell}(\mathcal{K}_{R})
-
J_{\ell}(\emptyset).
\label{eq:hexa_improvement}
\end{equation}
For warm-started variants, the contribution of subsequent online evolution can
additionally be expressed as
\begin{equation}
\Delta J_{\ell}^{\mathrm{evolve}}
=
J_{\ell}(\mathcal{K}_{R})
-
J_{\ell}(\mathcal{K}_{0}).
\end{equation}
Positive values indicate that experience gathered by the frozen actor has
produced an external context state that improves its future effective policy.
The trajectory reward $r(\tau)$ used by HExA is an internal credit-assignment
signal for constructing this state; it should therefore be distinguished from
the task-level objective $J_{\ell}$ itself.

\subsection{Trajectory Reward}
\label{sec:reward}

Interaction trajectories differ not only in whether they solve the task, but also
in the amount and quality of experimental evidence they provide. A fast successful
trajectory provides strong evidence for an effective strategy, whereas a failed
trajectory that explores several hypotheses may still contain useful intermediate
discoveries and diagnostic information. We therefore associate each trajectory
$\tau$ with a scalar reward $r(\tau)\in[-1,+1]$ that combines
\emph{outcome} and \emph{interaction efficiency}.

For the \textsc{InterPhyre} experiments, we use the following discrete reward
categories:
\begin{equation}
\label{eq:reward}
r(\tau)=
\begin{cases}
+1.00 & \text{success},\; t \le 3,\\
+0.75 & \text{success},\; 4 \le t \le 7,\\
+0.50 & \text{success},\; 8 \le t \le 15,\\
+0.25 & \text{success},\; 16 \le t \le T,\\[3pt]
-0.50 & \text{failure},\; t \ge 10,\\
-0.75 & \text{failure},\; t < 10,
\end{cases}
\end{equation}
where $t$ is the number of interaction turns used before termination and $T$ is
the episode budget. Faster successful trajectories therefore receive larger
positive rewards. Among failures, trajectories that continue experimenting receive
a smaller penalty than trajectories that terminate early, since the former contain
more interaction evidence from which the evolver may recover partial skills or
diagnose recurring failure modes.

We use discrete rather than finely graded reward values because the reward is
consumed by an LLM evolver as evidence about trajectory quality rather than
optimized directly through gradient descent. The categories therefore encode
qualitatively distinct outcomes---fast success, slower success, extended
exploration without success, and early failure---while remaining straightforward
for the evolver to contrast~\citep{viswanathan2026discretizingrewardmodels,
gunjal2025rubrics}. As shown in Section~\ref{sec:instantiation} and Table~\ref{tab:reward_ablation_main}, reward labels are
helpful but not required: \textsc{HExA} can also evolve skills without them,
leaving trajectory-quality assessment to the evolver itself.

\subsection{Skill Distillation from Experimental Experience}
\label{sec:distillation}

At round $n$, the actor produces a batch of reward-tagged trajectories
$\mathcal{T}^{(n)}$. We partition this batch as
\[
\mathcal{T}^{(n)}
=
\mathcal{T}^{(n),+}
\cup
\mathcal{T}^{(n),-},
\]
where $\mathcal{T}^{(n),+}$ and $\mathcal{T}^{(n),-}$ denote successful and
failed trajectories, respectively. Each trajectory contains the actor's reasoning,
tool calls, resulting observations, and trajectory reward. The evolver
$\pi_{\mathrm{ev}}$ uses these records to extract complementary forms of reusable
knowledge through two stages.

\paragraph{Phase 1: Contrastive strategy extraction.}
The evolver receives the complete batch and contrasts high-reward trajectories
against lower-reward trajectories to identify which strategies, experimental
choices, or mechanistic insights distinguish effective behavior from ineffective
behavior. Rather than retaining individual action sequences, it summarizes the
relevant evidence into a small set of \emph{strategy skills}.

Each strategy skill contains:
\begin{itemize}[leftmargin=1.5em,itemsep=1pt,topsep=2pt]
    \item a short \emph{title};
    \item a natural-language \emph{principle} summarizing the mechanism or
          experimental insight;
    \item a \emph{when-to-apply} condition describing when the principle is
          relevant;
    \item an optional demonstrative \emph{example};
    \item provenance identifying the \emph{source seed(s)} supporting the skill;
          and
    \item a reward-derived score $r_k$, described in
          Section~\ref{sec:skillbank}.
\end{itemize}
The evolver is instructed to weight the evidence by trajectory quality: principles
supported by efficient successful trajectories receive stronger support than
principles inferred from marginal successes or partial evidence. In this way,
distillation operates over the \emph{contrast} between successful and unsuccessful
experimentation rather than simply summarizing successful episodes.

\paragraph{Phase 2: Mistake and partial-skill extraction.}
Successful trajectories alone do not expose all useful learning signals. A failed
trajectory can contain a correct hypothesis, a useful experimental observation, or
a valid intermediate strategy even if a later decision causes the overall episode
to fail. We therefore analyze the failed subset
$\mathcal{T}^{(n),-}$ separately.

The evolver extracts \emph{mistake records}
\[
\mu=(\delta,\rho,\alpha),
\]
where $\delta$ describes a recurring failure, $\rho$ identifies its inferred root
cause or broken causal belief, and $\alpha$ specifies an actionable correction.
These records are intended to prevent repeated exploration failures rather than
provide positive strategies directly.

The evolver also extracts \emph{partial skills}: correct intermediate reasoning or
useful discoveries embedded inside otherwise unsuccessful trajectories. For
example, an actor may correctly identify the relevant physical mechanism or a
promising parameter region but subsequently choose an ineffective adjustment.
Preserving the correct intermediate observation allows later episodes to reuse
that discovery without treating the entire failed trajectory as uninformative.

The two stages are complementary. Contrastive strategy extraction captures
reusable mechanisms associated with successful behavior, while failure analysis
captures recurring mistakes and useful intermediate discoveries that success-only
distillation would discard. Because trajectories in later rounds are themselves
generated while conditioned on previously distilled skills, newly extracted skills
can refine, delimit, combine, or replace earlier ones rather than forming an
independent list of episode summaries.

\subsection{Skill Bank Construction and Retrieval}
\label{sec:skillbank}
\label{sec:retrieval}

The distilled knowledge is stored in an external skill bank $\mathcal{K}$ indexed
by task family. The bank contains two collections:
\begin{enumerate}[leftmargin=1.5em,itemsep=1pt,topsep=2pt]
    \item $\mathcal{K}^{\mathrm{sk}}$: reusable strategy skills and principles;
    \item $\mathcal{K}^{\mathrm{mis}}$: recurring mistake patterns and their
          corrective actions.
\end{enumerate}
Unlike episodic memory, $\mathcal{K}$ does not retain complete trajectories.
Instead, it compresses repeated experimental evidence into reusable abstractions
with explicit applicability conditions.

\paragraph{Trajectory grounded skill reward.}
Each strategy skill $k$ carries a score derived from the trajectories that support
it:
\begin{equation}
\label{eq:skill_score}
r_k
=
\operatorname{clamp}\!\left(
\frac{\bar r_{\mathrm{src}}+1}{2},
\,0.1,\,
1.0
\right),
\end{equation}
where $\bar r_{\mathrm{src}}$ is the mean reward of the source trajectories from
which the skill was distilled. Skills supported by efficient successful
trajectories consequently receive higher scores, whereas skills recovered from
weaker or partial evidence can remain in the bank with lower scores. The score is
used both to prioritize skills during retrieval and to guide retention and pruning
when the bank reaches capacity.


\paragraph{Bounded skill bank.}
To keep inference-time context bounded, the bank contains at most
$M_{\max}$ strategy skills and $N_{\max}$ mistake records for each task family:
\[
\left|\mathcal{K}^{\mathrm{sk}}\right|\le M_{\max},
\qquad
\left|\mathcal{K}^{\mathrm{mis}}\right|\le N_{\max}.
\]
The capacity constraint prevents the skill context from growing with the number of
episodes and forces the evolver to consolidate, revise, or remove redundant and
less useful entries as additional evidence arrives. This distinction is central to
\textsc{HExA}: interaction experience may grow indefinitely, but the representation
made available to the actor remains a compact set of distilled skills.

\paragraph{Retrieval and prompt injection.}
At the start of an episode, the retriever constructs the skill context
\[
\mathcal{G}
=
\textsc{Retriever}(\mathcal{K},\ell,M,N)
\]
by selecting the top-$M$ strategy skills according to $r_k$ together with the
top-$N$ mistake records for the current task family. The resulting structured block
is prepended to the actor's system prompt alongside the unchanged task description
and tool documentation.

The actor is not required to execute a retrieved skill verbatim. Instead, the bank
provides reusable hypotheses, applicability conditions, and failure warnings that
the actor may use or disregard based on the current instance. Retrieval therefore
changes only the model's textual context; it does not modify model parameters,
architecture, decoding, or the environment interface. This makes the same mechanism
applicable to both open-weight models and API-served black-box LLMs.

\subsection{Initialization and Update Regimes}
\label{sec:regimes}

The core loop admits several configurations that differ along two axes:
(1) how the initial bank $\mathcal{K}_0$ is constructed, and
(2) whether knowledge accumulated in previous rounds is retained explicitly when
the bank is updated. These variants let us distinguish the benefit of warm-starting
from the benefit of cumulative skill evolution.

\paragraph{Initialization.}
We consider three regimes.

\begin{itemize}[leftmargin=1.5em,itemsep=2pt,topsep=2pt]

\item \textbf{Offline.}
A warm-start batch $\mathcal{T}_0$ of $x$ trajectories is collected from the
base actor without any skill context and distilled into
$\mathcal{K}_0=\pi_{\mathrm{ev}}(\emptyset,\mathcal{T}_0)$.
The resulting bank is then held fixed.

\item \textbf{Offline-to-Online (Off2On).}
The bank is initialized from the same type of skill-free warm-start batch, but
subsequent interaction rounds continue to update it using newly collected
trajectories.

\item \textbf{Online.}
The bank starts empty,
$\mathcal{K}_0=\emptyset$, and all skills must be bootstrapped from the actor's
subsequent online interactions.

\end{itemize}

\paragraph{Update strategy.}
For non-static banks, we consider two update rules.

\begin{itemize}[leftmargin=1.5em,itemsep=2pt,topsep=2pt]

\item \textbf{Evolution.}
The evolver receives both the current bank and the newly collected trajectory batch,
\[
\mathcal{K}_n
=
\pi_{\mathrm{ev}}
\!\left(
\mathcal{K}_{n-1},
\mathcal{T}^{(n)}
\right),
\]
and may add new skills, merge corroborating evidence, revise applicability
conditions, or remove entries contradicted by subsequent experience.

\item \textbf{Iterative Replacement.}
The evolver receives only the new trajectory batch,
\[
\mathcal{K}_n
=
\pi_{\mathrm{ev}}
\!\left(
\emptyset,
\mathcal{T}^{(n)}
\right),
\]
and reconstructs the bank from that round's trajectories. These trajectories were
generated by an actor already conditioned on $\mathcal{K}_{n-1}$, so they may
implicitly reflect earlier skills, but the previous bank itself is not supplied to
the evolver.

\end{itemize}

The Offline regime is the corresponding static case,
$\mathcal{K}_n=\mathcal{K}_{n-1}$. Crossing initialization and update choices
gives the five configurations evaluated in Section~\ref{sec:instantiation}:
Offline Static, Off2On Evolution, Off2On Iterative Replacement, Online Evolution,
and Online Iterative Replacement.

Evolution is designed to preserve useful accumulated knowledge while incorporating
new evidence. This is particularly important because iterative skill learning can
otherwise suffer from \emph{skill drift}: repeatedly rebuilding a bank from a small
recent batch may discard broadly useful principles in favor of narrow strategies
specific to the most recently observed instances. Supplying
$\mathcal{K}_{n-1}$ explicitly instead gives the evolver the opportunity to retain
a useful rule, refine its boundary conditions, merge it with corroborating
evidence, or replace it when later observations contradict it.

Algorithm~\ref{alg:HeXa_algo_app} gives the complete procedure for all
initialization and update regimes.

\begin{algorithm}[t]
\caption{\textsc{HExA}: Initialization and Update Variants}
\label{alg:HeXa_algo_app}
\DontPrintSemicolon
\small
\KwIn{
Actor $\pi_{\mathrm{act}}$,
evolver $\pi_{\mathrm{ev}}$,
task family $\ell$,
initialization $\mathcal{I}\in\{\text{empty},\text{warm-start}\}$,
update strategy $\mathcal{U}\in\{\text{evolution},\text{iterative replacement}\}$,
rounds $R$,
episodes per round $x$,
retrieval sizes $M$ and $N$
}
\KwOut{
Final Evolved skill bank $\mathcal{K}_R$
}

\tcp{Initialize the skill bank}
\uIf{$\mathcal{I}=\text{warm-start}$}{
    Collect skill-free warm-start batch $\mathcal{T}_0$ of $x$ trajectories from $\pi_{\mathrm{act}}$\;
    $\mathcal{K}_0 \gets \pi_{\mathrm{ev}}(\emptyset,\mathcal{T}_0)$\;
}
\Else{
    $\mathcal{K}_0\gets\emptyset$\;
}

\BlankLine
\tcp{Main loop: collect experience and update bank}
\For{$n=1,\ldots,R$}{
    \tcp{Retrieve top-$M$ skills and top-$N$ mistakes}
    $\mathcal{G}_n \gets \textsc{Retriever}(\mathcal{K}_{n-1},\ell,M,N)$\;

    \For{$i=1+x(n-1),\ldots,x*n$}{
        \tcp{Actor Prompt ... task + tools + skills}
        $\tau_i \gets \pi_{\mathrm{act}}(\cdot\mid\ell,s_i,\mathcal{G}_n)$\;
        \tcp{outcome $\times$ efficiency}
        $r(\tau_i)\gets\textsc{Reward}(\tau_i)$\;
    }

    \tcp{Update bank based on strategy}
    \uIf{$\mathcal{U}=\text{evolution}$}{
        \tcp{Evolve: keep old bank, incorporate new evidence}
        $\mathcal{K}_n \gets \pi_{\mathrm{ev}}(\mathcal{K}_{n-1},\mathcal{T}^{(n)})$\;
    }
    \Else{
        \tcp{Iterative Replacement: rebuild from scratch each round}
        $\mathcal{K}_n \gets \pi_{\mathrm{ev}}(\emptyset,\mathcal{T}^{(n)})$\;
    }
}

\KwRet{
$\mathcal{K}_R$
}
\end{algorithm}

Unless otherwise stated, the configuration referred to as \textsc{HExA} in our
experiments is the reward-guided Off2On~+~Evolution variant.

\subsection{Cross-Task Skill Synthesis}
\label{sec:cross_task_transfer}
\label{sec:transfer}

Within-task evolution produces banks specialized to individual task families, but
many learned principles need not be tied to the particular entities, coordinates,
or parameter values of their source task. Physical mechanisms such as momentum
transfer, contact geometry, lever mechanics, and collision order---and more general
procedural search strategies---may remain useful when the surface realization of a
task changes.

\textsc{HExA} therefore supports synthesis of a target skill bank from one or more
source banks without using target-task trajectories during bank construction.
Given evolved source banks
\[
\mathcal{K}_{\ell_1},
\ldots,
\mathcal{K}_{\ell_S}
\]
and a textual description $\mathrm{desc}(\ell^*)$ of an unseen target family
$\ell^*$, the transfer evolver produces
\[
\mathcal{K}_{\ell^*}
=
\pi^{\mathrm{tr}}_{\mathrm{ev}}
\!\left(
\{\mathcal{K}_{\ell_i}\}_{i=1}^{S},
\mathrm{desc}(\ell^*)
\right),
\qquad
\mathcal{T}_{\ell^*}=\emptyset.
\]

The synthesis procedure performs four operations:

\begin{enumerate}[leftmargin=1.5em,itemsep=2pt,topsep=2pt]

\item \textbf{Select relevant source principles.}
The evolver identifies source skills whose underlying physical or procedural
mechanisms are structurally relevant to the target.

Algorithm~\ref{alg:cross_task_synthesis} formalizes the cross-task synthesis procedure.

\begin{algorithm}[t]
\caption{\textsc{HExA}: Cross-Task Skill Synthesis}
\label{alg:cross_task_synthesis}
\DontPrintSemicolon
\small
\KwIn{
Source banks $\mathcal{K}_{\ell_1},\ldots,\mathcal{K}_{\ell_S}$,
target description $\mathrm{desc}(\ell^*)$,
transfer evolver $\pi^{\mathrm{tr}}_{\mathrm{ev}}$,
retrieval sizes $M,N$
}
\KwOut{
Synthesized target bank $\mathcal{K}_{\ell^*}$
}

\tcp{Invoke transfer evolver on source skill banks and target description}
$\mathcal{K}_{\ell^*} \gets \pi^{\mathrm{tr}}_{\mathrm{ev}}\!\left(\{\mathcal{K}_{\ell_i}\}_{i=1}^{S},\,\mathrm{desc}(\ell^*)\right)$\;

\BlankLine
\tcp{The evolver performs four operations internally:}
\tcp{(1) select source skills structurally relevant to target}
\tcp{(2) remove source-specific parameter and entity bindings}
\tcp{(3) re-ground each principle in target's entities and tools}
\tcp{(4) assign transfer score based on correspondence and corroboration}

\BlankLine
\tcp{Prepare synthesized bank for injection or initialization}
\If{zero-shot transfer}{
    \tcp{Inject directly through retriever for immediate use}
    $\mathcal{G} \gets \textsc{Retriever}(\mathcal{K}_{\ell^*},\ell^*,M,N)$\;
}
\Else{
    \tcp{Use as initialization for within-task evolution}
    $\mathcal{K}_0 \gets \mathcal{K}_{\ell^*}$\;
    \tcp{Then run standard HExA loop (Algorithm~\ref{alg:HeXa_algo_app})}
}

\KwRet{$\mathcal{K}_{\ell^*}$}
\end{algorithm}

\item \textbf{Remove source-specific bindings.}
Coordinates, object identities, and other parameterizations that are particular to
the source task are stripped away when they are not transferable.

\item \textbf{Re-ground the principle.}
The retained abstraction is expressed in terms of the entities, constraints, and
available tools of the target task.

\item \textbf{Score the transferred skill.}
The evolver assigns a transfer score according to the directness of the structural
correspondence and the extent to which the principle is corroborated across source
banks. Because no target-task trajectories exist, this score takes the place of
the reward-derived $r_k$ and is clamped to the same range $[0.1,1.0]$. It is then
used by the same retrieval mechanism that ranks within-task skills.

\end{enumerate}

The evolver is explicitly constrained not to invent target-specific coordinates or
solutions that are unsupported by the source banks and textual target description.
Transferred entries are instead expressed as general mechanisms together with
conditions specifying when they should apply. The purpose of synthesis is therefore
to re-ground reusable abstractions rather than copy source-task solutions verbatim. Algorithm~\ref{alg:cross_task_synthesis} formalizes the cross-task synthesis procedure.

The synthesized bank $\mathcal{K}_{\ell^*}$ can be used in two ways. First, it can
be injected directly through the standard retriever, allowing the actor to begin
interacting with the unseen target using transferred skills even though no
target-task trajectory was used to construct the bank. Second, it can serve as
$\mathcal{K}_0$ for a subsequent within-task \textsc{HExA} run, after which
target-specific evidence may further refine the transferred knowledge. The full
cross-task synthesis prompts are provided in
Appendix~\ref{app:prompt:cross_level_full}, and the corresponding transfer
experiments are reported in Section~\ref{sec:results:transfer}.

\begin{algorithm}[t]
\caption{\textsc{HExA}: Inference with a Learned Skill Bank}
\label{alg:hexa_inference}
\DontPrintSemicolon
\small
\KwIn{
Actor $\pi_{\mathrm{act}}$,
evolver $\pi_{\mathrm{ev}}$,
bank $\mathcal{K}^\star\in\{\mathcal{K}_R,\mathcal{K}_{\ell^*}\}$,
task family $\ell$,
evaluation seeds $\mathcal{S}^{\mathrm{ev}}_\ell$,
mode $\mathcal{V}\in\{\text{static},\text{within-seed}\}$,
turn budget $T$,
retrieval sizes $M,N$
}
\KwOut{
Evaluation trajectories and outcomes $\{(\tau_j,y_j)\}_{j=1}^{|\mathcal{S}^{\mathrm{ev}}_\ell|}$
}

\For{$j=1,\ldots,|\mathcal{S}^{\mathrm{ev}}_\ell|$}{
    \tcp{Fresh seed-local bank; $\mathcal{K}^\star$ is never modified}
    $\mathcal{K}_{j,0}\gets\mathcal{K}^\star$\;
    $h_0\gets\emptyset$\;

    \BlankLine
    \For{$t=0,\ldots,T-1$}{
        \tcp{Retrieve top-$M$ skills and top-$N$ mistakes}
        $\mathcal{G}_t\gets\textsc{Retriever}(\mathcal{K}_{j,t},\ell,M,N)$\;
        \tcp{task + tools + skills + history}
        $a_t\gets\pi_{\mathrm{act}}(\cdot\mid\ell,s_j,\mathcal{G}_t,h_t;\mathcal{F})$\;
        $o_t\gets\textsc{Env}(a_t)$\;
        $h_{t+1}\gets(h_t,a_t,o_t)$\;
        \If{$a_t$ is terminal}{
            \textbf{break}\;
        }

        \tcp{Update bank based on mode}
        \uIf{$\mathcal{V}=\text{within-seed}$}{
            \tcp{Refine from feedback; consumes no turns}
            $\mathcal{K}_{j,t+1}\gets\pi_{\mathrm{ev}}(\mathcal{K}_{j,t},h_{t+1})$\;
        }
        \Else{
            \tcp{Static: bank unchanged}
            $\mathcal{K}_{j,t+1}\gets\mathcal{K}_{j,t}$\;
        }
    }

    \BlankLine
    \tcp{Score the episode (Eq.~\ref{eq:success_main})}
    $\tau_j\gets h_{T_j}$\;
    $y_j\gets\mathbb{I}\!\left[\textsc{Goal}_\ell(\textsc{Sim}(\tau_j))=\text{true}\right]$\;
}

\KwRet{$\{(\tau_j,y_j)\}_{j=1}^{|\mathcal{S}^{\mathrm{ev}}_\ell|}$}
\end{algorithm}

\subsection{Inference with a Learned Skill Bank}
\label{sec:inference}

Within-task skill evolution (Algorithm~\ref{alg:HeXa_algo_app}) returns a frozen bank
$\mathcal{K}_R$, and cross-task skill synthesis (Algorithm~\ref{alg:cross_task_synthesis})
returns a bank $\mathcal{K}_{\ell^*}$. We write $\mathcal{K}^\star$ for either one.
At inference time the actor is run on the held-out evaluation seeds
$s_j\in\mathcal{S}^{\mathrm{ev}}_\ell$ with the same budget of at most $T$ turns per
seed as during bank construction, and success on each seed is scored by
Eq.~\ref{eq:success_main}. We consider two inference modes.

\paragraph{Static skill injection.}
The retriever constructs the skill context
$\mathcal{G}=\textsc{Retriever}(\mathcal{K}^\star,\ell,M,N)$ once at the start of the
episode, and every turn follows Eq.~\ref{eq:turn} with this fixed $\mathcal{G}$. The
bank is not modified during or between episodes.

\paragraph{Within-seed experimentation.}
This mode relaxes the assumption in Eq.~\ref{eq:turn} that $\mathcal{G}$ is held
fixed throughout the episode, enabling us to better leverage test-time compute using HExA and continue adapting from inference time experimentation as well. Each seed again starts from $\mathcal{K}^\star$, but
after every turn $t$ the evolver receives the current seed-local bank $\mathcal{K}_{j,t}$ and the
interaction history $h_{t+1}$ (the actor's tool calls and the resulting
observations so far) and refines the bank,
\[
\mathcal{K}_{j,t+1}
=
\pi_{\mathrm{ev}}
\!\left(
\mathcal{K}_{j,t},
h_{t+1}
\right),
\qquad
\mathcal{K}_{j,0}=\mathcal{K}^\star .
\]
The skill context for the next turn is retrieved from the refined bank. In this way
the agent can confirm, delimit, or discard a retrieved skill as soon as an
experiment contradicts it, and can record instance-specific discoveries (for
example, a promising parameter region) for the remaining turns. This inference time refinement does not
require extra turns, so both modes have the same budget of at most $T$ environment
interactions. Each seed starts from a fresh copy of $\mathcal{K}^\star$ and its
refined bank is discarded when the episode ends, so no information passes between
evaluation seeds and $\mathcal{K}^\star$ is unchanged.
Algorithm~\ref{alg:hexa_inference} gives both modes and our current results are evaluated using static skill injection.

\subsection{Design Principles}
\label{sec:design-principles}

The preceding components reflect four design properties of \textsc{HExA}.

\paragraph{In-context rather than parametric adaptation.}
Neither the actor nor the evolver is updated through gradient descent. Adaptation
is represented explicitly in the external skill bank and affects behavior only by
changing the context supplied to the actor. This allows \textsc{HExA} to operate
with frozen and API-served models while avoiding the computational requirements
and potential forgetting associated with repeated fine-tuning.

\paragraph{Domain-independent framework.}
The framework itself does not assume a particular environment or physical domain.
The domain-specific inputs are the task description and the tools through which the
actor interacts with the environment. The evolver consumes textual interaction
trajectories and produces structured natural-language skills. Consequently, the
same actor--evolver--retriever abstraction can be instantiated wherever an agent can
interact with an environment and convert the resulting observations into reusable
experience.

\paragraph{Reward-function flexibility.}
The particular reward function in Eq.~\ref{eq:reward} is an instantiation used for
our \textsc{InterPhyre} experiments rather than a requirement of the framework.
When an informative interaction reward is available, it can provide
environment-grounded evidence for skill credit assignment and prioritization.
When such feedback is unavailable, the evolver can still construct and update the
bank from trajectories without explicit reward labels, as evaluated in our
ablations. Thus, reward guidance supplements rather than defines the underlying
skill-learning mechanism.

\paragraph{Hierarchical and compositional skill formation.}
The skill representation is hierarchical in two operational senses. First,
distillation abstracts low-level actions, tool calls, and parameter settings into
higher-level principles with applicability conditions. Second, later trajectories
are produced while the actor already has access to skills learned in earlier
rounds. Subsequent distillation can therefore refine an earlier principle, identify
its limits, merge complementary observations, or replace a rule contradicted by
new evidence. The resulting bank is not an explicit tree; its hierarchy arises
through repeated abstraction and composition of experience across interaction
rounds. Section~\ref{sec:results:transfer} analyzes this compositional structure
and its role in transfer to unseen tasks.

\newpage
\section{Details of InterPhyre: A Testbed for Learning through Experimentation}
\label{app:InterPhyre}

\begin{figure}[h]
  \centering
  \includegraphics[width=\linewidth]{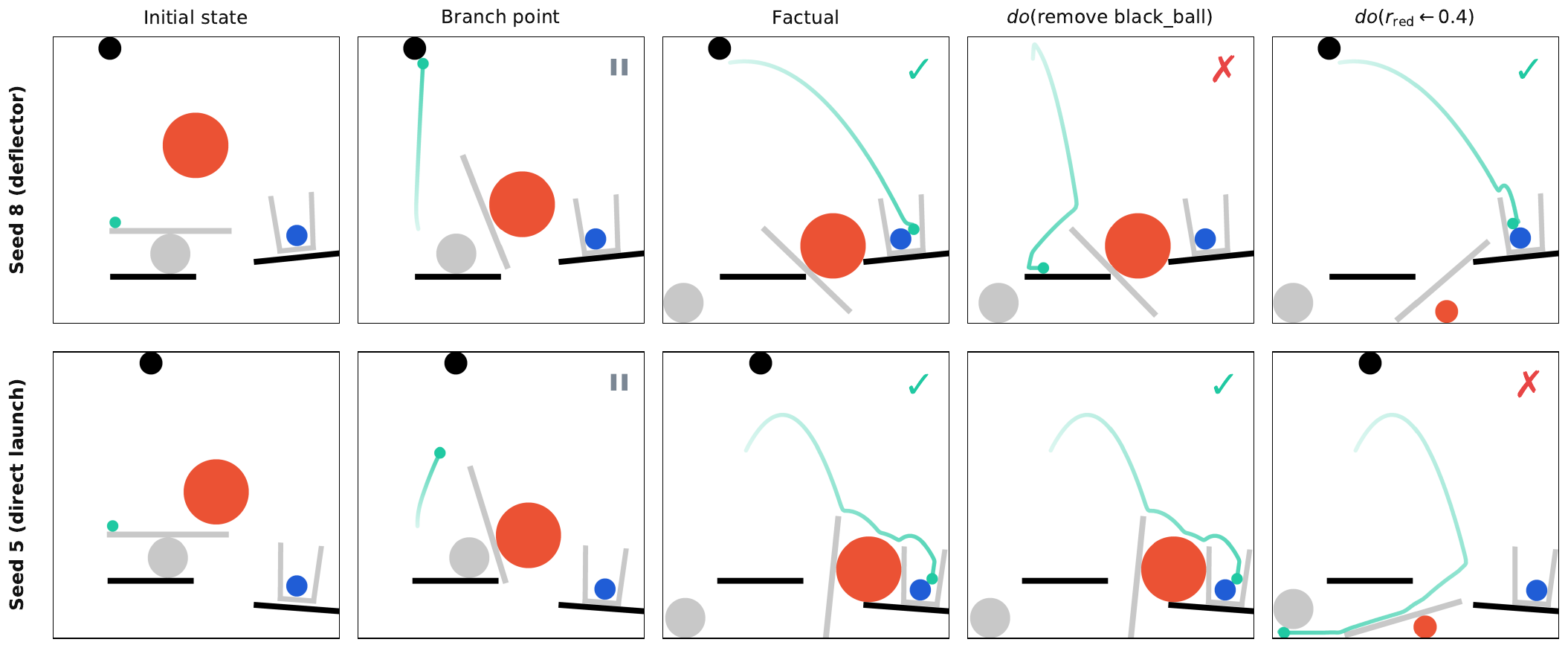}
  \caption{\textbf{\textsc{Interphyre}'s snapshot/restore API.} This branches a shared mid-trajectory state into a
    factual rollout and perturbed alternatives (Section \ref{sec:snapshot-restore}).
    Each row is one seed of the \texttt{catapult} level with a different oracle solution
    strategy.
    The leftmost column shows the initial scene for reference; the teal curve is the
    green ball's trajectory, shown up to the branch point in the \emph{Branch point}
    column and continuing through the outcome in each subsequent column.
    Top (seed\,8, deflector strategy): removing the deflector causes failure; shrinking
    the red action ball does not.
    Bottom (seed\,5, direct-launch strategy): shrinking the red ball causes failure;
    removing the deflector does not.
    The same interventions have opposite causal relevance across seeds, showing that causal
    relevance is a property of an object relative to the active strategy, not of the object
    in isolation.
    See Section \ref{sec:worked-example} for the branching protocol.}
  \label{fig:intervention-example}
\end{figure}

\subsection{Introduction}

The board-reset primitive that unlocked world-model probing in chess and Othello has no equivalent in continuous physics.
Those results relied on models trained on task-domain sequences and on the ability to specify any board state symbolically; physical state is continuous and contact-dependent, and counterfactual access requires rewinding a running simulation rather than resetting a board.
We introduce Interphyre, a 2D physics environment whose snapshot/restore API captures complete simulation state at any event-triggered branch point and branches it into paired counterfactual trajectories that diverge under controlled perturbations.
In a preliminary probing study, linear probes on Qwen3-8B residual-stream activations predict counterfactual outcomes above chance across three levels, but only when conditioned on oracle solution strategy.
On the catapult level, the same object removal is causally necessary under one strategy and irrelevant under another; pooling across strategies averages over this reversal.
The same backend supports RL generalization research: physics parameters are independent experimental axes, so a policy trained under standard conditions can be tested under altered physics without changing scene structure.
Interphyre's 25-level curriculum and open Python API are released to support interpretability work on physics-reasoning models and RL generalization research.

Existing 2D physics benchmarks (PHYRE~\citep{bakhtin2019phyre}, I-PHYRE~\citep{li2024iphyre}, Kinetix~\citep{matthews2025kinetix}) were designed to answer one question: can the agent solve the puzzle?
Two research communities now need more from physics environments.
Mechanistic interpretability of physics-reasoning LLMs requires mid-trajectory intervention and paired counterfactual rollouts from a shared physical state; without these, model activations cannot be paired with causal physical ground truth.
RL generalization and behavioral analysis require independent control over physics parameters and white-box access to simulation state; without these, a researcher cannot test whether a learned strategy transfers across physics regimes or which scene features it relies on.
No existing benchmark meets either requirement.

World-model discovery in board games relies on a property physics does not share: discrete, fully observable ground-truth states that make probe targets unambiguous and interventions exact.
Language models encode Othello boards~\citep{li2023othello} and chess positions~\citep{karvonen2024chess} as structured world models; those results used models trained on task-domain sequences, where board state was the latent variable the training objective required the model to track.
For physics-reasoning LLMs trained on general text, the analogous question is whether chain-of-thought reasoning produces structured physical representations, and answering it requires infrastructure those experiments did not need.
Physical state is continuous and contact-dependent; the probe target is the outcome of a specific contact event, not a fixed board position, and counterfactual access requires rewinding a running simulation rather than resetting a board.
The infrastructure that made board-game probing tractable does not transfer.

The RL gap has a specific shape.
Testing whether a learned strategy generalizes across physics variations, or identifying which scene features a policy relies on, requires varying physics parameters while holding scene topology fixed.
Kinetix varies both jointly~\citep{matthews2025kinetix}; PHYRE and I-PHYRE expose no parametric control at all.
Without independent axes, a generalization result cannot be attributed to physics transfer rather than to a change in scene structure.


We introduce Interphyre, an extensible, intervention-aware 2D physics environment that addresses both gaps through three capabilities:
\begin{enumerate}[leftmargin=*,itemsep=2pt,parsep=0pt]

\item \textbf{An intervention API with snapshot/restore semantics.}
  Any running simulation can be snapshotted at an arbitrary point (on a physics event, a contact trigger, or a fixed timestep) and restored into two or more branches that diverge under controlled perturbations (impulses, forces, parameter changes, object additions or removals).
  Paired counterfactual evaluation follows: the same agent, on the same scene, under matched and perturbed physical conditions.
  No existing physical reasoning benchmark provides this programmatic primitive.

\item \textbf{Extensible level authoring as Python code.}
  Levels are Python classes built from a composable object API, with tunable physics
  parameters (gravity, friction, restitution, density) and user-defined success conditions.
  Like Toybox's reimplementation of Atari games~\citep{foley2018toybox}, Interphyre
  reimplements the 2D physics puzzle paradigm established by PHYRE, extending it with
  white-box access, parametric control, and intervention infrastructure.
  A curriculum of 25 named levels, each parameterized by a random seed, ships with
  250,000 pre-validated task instances; an oracle verification system certifies new seeds
  on demand, so the curriculum is editable and extensible rather than fixed.

\item \textbf{LLM-native interfaces and interpretability-ready data generation.}
  The simulator and intervention API are exposed as a tool-call surface that an LLM agent
  can invoke directly.
  The same interfaces support standalone generation of paired (factual, counterfactual)
  trajectory data for downstream interpretability and offline reinforcement learning
  pipelines.

\end{enumerate}

Figure \ref{fig:intervention-example} shows the key result: the same structural intervention has opposite causal relevance across seeds that use different solution strategies, a confound any probing study not conditioned on strategy will average over.

\subsection{Related Work}

\paragraph{Physical reasoning benchmarks.}
PHYRE~\citep{bakhtin2019phyre} introduced 2D physics puzzles for evaluating
sample-efficient learning, measuring agents by AUCCESS across within-template and
cross-template generalization splits.
Virtual Tools~\citep{allen2020virtualtools} developed a concurrent benchmark with
similar Box2D-style mechanics, grounded in a Bayesian ``sample, simulate, update''
cognitive model.
I-PHYRE~\citep{li2024iphyre} extended this to temporally sequenced interventions, where agents eliminate obstacles at precise moments as simulations unfold.
I-PHYRE's interventions are agent actions within a single trajectory; the environment produces one modified rollout, and paired counterfactual analysis requires two from a shared prefix.
Kinetix~\citep{matthews2025kinetix} takes a different direction: it is a JAX-accelerated physics engine supporting open-ended procedural generation for training general RL agents.
Interphyre provides a designed curriculum whose solution structure is known, which is necessary for controlled interpretability comparisons.
DeepPHY~\citep{xu2026deepphy} unifies several of these environments under a single
evaluation framework for agentic VLMs, but inherits the structural limitations of each
backend (no parametric control, no paired counterfactuals, no activation-level
instrumentation).

\paragraph{Counterfactual and causal evaluation environments.}
CoPhy~\citep{baradel2020cophy} introduced counterfactual learning of physical dynamics,
requiring models to predict outcomes after do-interventions on object positions in 3D
scenes.
CausalWorld~\citep{ahmed2021causalworld} exposed causal variables (masses, sizes,
friction, gravity) for do-interventions in robotic manipulation, demonstrating the
value of interventional evaluation for transfer learning.
CRAFT~\citep{ates2022craft} tested causal reasoning about forces and interactions
through video question answering over Box2D scenes, including counterfactual question
types.
ContPhy~\citep{zheng2024contphy} extended counterfactual physics questions to continuum
materials.
Each introduces counterfactual or interventional structure into physics evaluation; none implements the programmatic snapshot/restore primitive that generates paired rollouts from a bit-identical physical prefix, which Interphyre's paired-data generation requires.

\paragraph{LLM-based physical reasoning.}
LLMs underperform humans on physical attribute reasoning~\citep{wang2023newton} and struggle with physical dynamics even when strong on common-sense tasks~\citep{chow2025physbench}.
These systems evaluate behavior, not representations.
PhysGym~\citep{chen2025physgym} uses LLM agents to actively probe interactive physics simulations and formulate hypotheses, the closest existing paradigm to Interphyre, though PhysGym targets discovery of physical equations rather than intuitive task-solving.
LLMPhy~\citep{cherian2024llmphy} demonstrated a simulator-in-the-loop framework
where LLMs iteratively estimate physical parameters via code generation.
None provides the pairing mechanistic interpretability requires: model activations matched to counterfactual physical outcomes from a shared simulation state.

\paragraph{Interpretability on controlled reasoning tasks.}
Recent work has discovered world models in Othello~\citep{li2023othello}, chess~\citep{karvonen2024chess}, and mazes~\citep{spies2025mazes}, and circuit structure in board games~\citep{he2024othello_sae} and in-context RL~\citep{demircan2025td}.
All these results rely on discrete, fully observable ground-truth states.
Physics reasoning does not have this property.
Physical state is continuous and high-dimensional; the quantity a probe should predict depends on which contact event matters for the task, not on a fixed board position.
Counterfactual access requires rewinding a running simulation to a precise mid-trajectory event, not resetting a board to a saved position.
Thought Anchors~\citep{bogdan2025thought} identify important reasoning steps via black-box resampling of output traces; Interphyre supports a different analysis, pairing internal activations with causal physical ground truth obtained by intervening on the simulation directly rather than by resampling output traces.

\subsection{Design}
\label{sec:design}

\begin{figure}[t]
  \centering
  \includegraphics[width=\linewidth]{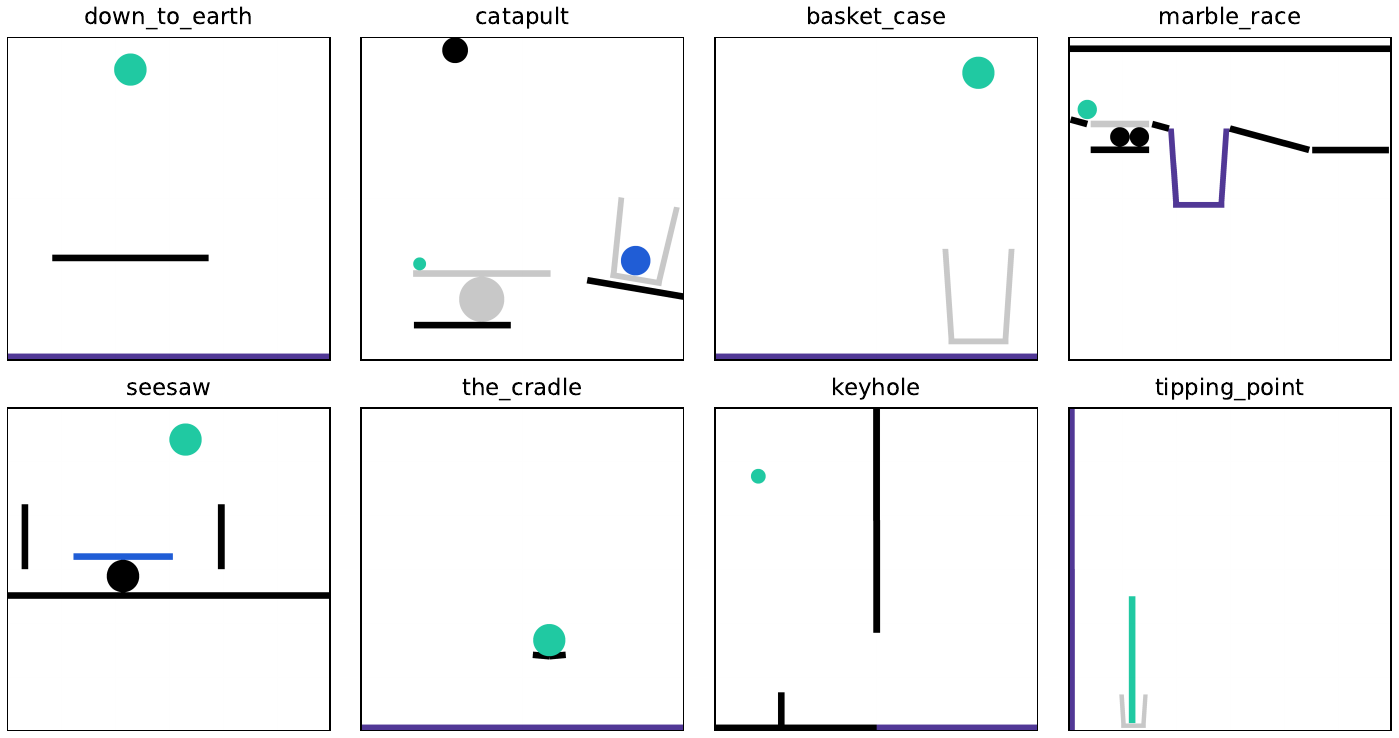}
  \caption{Eight of the 25 canonical levels in the Interphyre curriculum,
    shown at initial state.
    The curriculum is hand-authored to span a range of physical phenomena:
    \texttt{down\_to\_earth} (gravity-driven drop past an obstruction),
    \texttt{catapult} (impulse transfer), \texttt{basket\_case} (container
    avoidance), \texttt{marble\_race} (rolling dynamics along a track),
    \texttt{seesaw} (lever balance), \texttt{the\_cradle} (pendulum),
    \texttt{keyhole} (aperture navigation), and \texttt{tipping\_point}
    (unstable equilibrium).
    Each level is a Python class; each class is instantiable under a
    researcher-specified seed and physics configuration.}
  \label{fig:design-overview}
\end{figure}

Four design decisions give Interphyre its research capabilities: levels are Python code (\ref{sec:levels-as-code}), physics parameters are first-class inputs (\ref{sec:parameter-surface}), the curriculum is certified by an oracle (\ref{sec:oracle-verification}), and both RL and LLM interfaces share one simulator backend (\ref{sec:agent-interfaces}).
\ref{fig:design-overview} shows eight of the 25 canonical levels, spanning gravity-driven drops, collision chains, balancing tasks, and aperture navigation.

\subsubsection{Levels as Python Code}
\label{sec:levels-as-code}

Encoding levels as Python functions rather than data files means the geometry, action slots, and success conditions are ordinary code that any researcher can fork, modify, and register without rebuilding the package; Listing \ref{lst:custom-level} shows a minimal example.




\begin{listing}[t]
\centering
\begin{minipage}{0.98\linewidth}
\begin{lstlisting}[
  language=Python,
  basicstyle=\ttfamily\scriptsize,
  keywordstyle=\color{blue!55!black},
  commentstyle=\color{green!35!black},
  stringstyle=\color{red!60!black},
  breaklines=true,
  breakatwhitespace=false,
  columns=fullflexible,
  keepspaces=true,
  showstringspaces=false,
  frame=single,
  framesep=2mm,
  tabsize=2,
  aboveskip=0pt,
  belowskip=0pt
]
from interphyre import Level
from interphyre.levels import register_level
from interphyre.objects import Ball, Bar


def success_condition(engine):
    return engine.is_in_contact_for_duration(
        "green_ball",
        "purple_ground",
        3.0,
    )


@register_level
def build_level(seed=None, variant=0, scene=None) -> Level:
    ground = Bar(
        left=-5,
        right=5,
        y=-4.9,
        thickness=0.2,
        color="purple",
        dynamic=False,
    )

    green = Ball(
        x=0,
        y=4.0,
        radius=0.5,
        color="green",
        dynamic=True,
    )

    red = Ball(
        x=0,
        y=0.0,
        radius=0.4,
        color="red",
        dynamic=True,
    )

    return Level(
        name="falling",
        objects={
            "green_ball": green,
            "red_ball": red,
            "purple_ground": ground,
        },
        action_objects=["red_ball"],
        success_condition=success_condition,
    )
\end{lstlisting}
\end{minipage}

\caption{A minimal custom level.
  The geometry, the action slot, and the success condition are all ordinary
  Python; registering the builder makes it usable by the same evaluation
  pipeline as the bundled curriculum, with no package rebuild.}
\label{lst:custom-level}
\end{listing}

A researcher studying how behavior depends on scene topology writes a new subclass rather than waiting for a fixed benchmark to include the condition.

\begin{figure}[b]
  \centering
  \includegraphics[width=\linewidth]{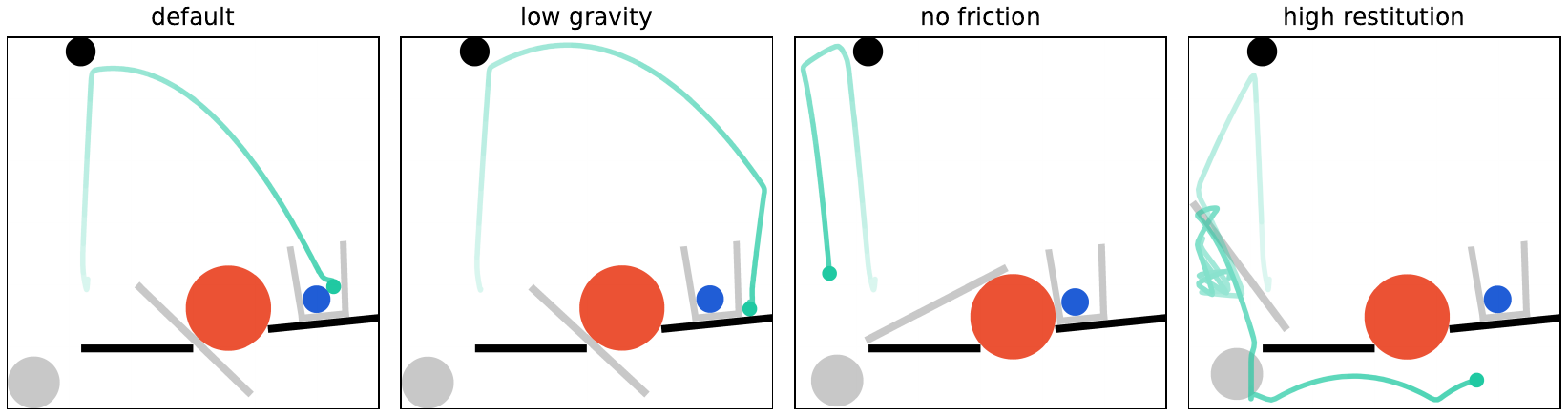}
  \caption{The \texttt{catapult} level (seed 8) under four physics
    configurations, each set by a single \texttt{SimulationConfig} field,
    with the same oracle action placed in every panel.
    Ghost trails show the green ball's in-flight trajectory.
    Default physics yields success.
    Reduced gravity preserves the qualitative mechanism but extends the arc,
    producing a near miss: the ball clears the basket opening rather than
    dropping in.
    Without friction the arm imparts less tangential force and the ball falls
    short; a higher-energy placement would compensate.
    With high restitution the elastic collision over-drives the arm and the
    ball overshoots; a softer placement would keep the launch on target.
    Holding scene and seed fixed, each parameter change shifts the required
    action.}
  \label{fig:parameter-sweep}
\end{figure}

\subsubsection{Physics Parameters as First-Class Inputs}
\label{sec:parameter-surface}

Physics parameters (gravity, friction, restitution, density, solver step size) are fields on a \texttt{SimulationConfig} object passed to the environment constructor, not constants compiled into the engine.
The same level, seed, and action can be evaluated under any physically plausible
parameter regime by swapping one configuration for another.
Figure \ref{fig:parameter-sweep} shows the consequence: the same oracle action succeeds
under default physics and fails in three distinct ways under reduced gravity, zero
friction, and high restitution, each failure requiring a different corrective action.

Unlike Kinetix, which varies physics parameters jointly with level topology~\citep{matthews2025kinetix}, Interphyre keeps scene structure and parameter regime independent; a generalization study holds the level and seed fixed while sweeping gravity or friction.

\subsubsection{Curriculum and Oracle Verification}
\label{sec:oracle-verification}

The curriculum ships with 25 canonical levels, each paired with 10{,}000
pre-validated seeds, for 250{,}000 pre-certified task instances.
Every seed in this distribution is guaranteed to admit at least one solving
action placement, verified by an exhaustive oracle search over a discretized
action grid with a bounded simulation budget per candidate rather than by
stochastic sampling.
The verification is decoupled from the environment runtime: a custom level or
an extended seed range can be certified offline and the resulting seed list
distributed alongside the environment code.
Solvability is therefore a guarantee of the curriculum rather than of any
particular bundled snapshot.

Oracle verification ensures that, up to the resolution of the action grid, every evaluation failure is attributable to the agent.
PHYRE's stochastic-validity heuristic~\citep{bakhtin2019phyre} admits unsolvable instances, a confound controlled interpretability studies cannot afford; Kinetix trades solvability guarantees for scale, reasonable for general RL pretraining but not for paired-counterfactual evaluation.

\subsubsection{Agent Interfaces and Observation Modes}
\label{sec:agent-interfaces}

Two interface surfaces share one simulator backend.
The gymnasium-compatible interface~\citep{towers2023gymnasium} exposes the
environment to standard RL training and evaluation code without modification:
action spaces, step semantics, and reset behavior follow the conventions of the
wider gymnasium ecosystem.
The tool-call interface exposes the same environment as a set of named
functions (\texttt{inspect\_scene}, \texttt{place\_action},
\texttt{advance\_simulation}, \texttt{query\_success}) an LLM can invoke
directly, so that an LLM reasoner operating under a tool-use harness can
drive the simulator in the same turn structure it would use for any other
external tool.

Three observation modes expose the same physics to different experimental designs: \texttt{physics\_state} returns a symbolic vector of positions, velocities, and contacts; \texttt{image} renders a frame in either RGB or a discrete-color palette (discrete colors remove the texture and shading confounds that bias VLM evaluation~\citep{chow2025physbench}); \texttt{both} returns both together for multimodal agents.
Because all three run the same simulation, observation modality is a controlled variable, not a design constraint.

\subsection{Intervention API}
\label{sec:interventions}

Paired counterfactual analysis requires two trajectories that share a bit-identical
physical prefix but diverge under exactly one controlled perturbation.
Forward rollout cannot produce these: without the ability to hold physical history
constant while changing a single variable mid-trajectory, interpretability studies that
depend on counterfactual contrast have no substrate.
Interphyre supplies this primitive via snapshot/restore, which makes the simulation
\emph{causally interrogable}.

\subsubsection{Snapshot/Restore and Counterfactual Branching}
\label{sec:snapshot-restore}

The environment is deterministic under a fixed seed: replaying the same action sequence on the same level, seed, and physics configuration yields a bit-identical trajectory, which is the property that makes a comparison genuinely paired.

The central primitive is the ability to capture an immutable snapshot of complete
physics state at any point during simulation and restore it into two or more independent
branches.
A snapshot captures the complete Box2D world state: positions, velocities, fixture
properties, solver configuration, active contacts, and contact-duration accumulators;
the accumulators are required because many levels define success as sustained contact,
and restoring position without restoring contact timing would break the pairing.

After \texttt{restore}, the simulation evolves identically to the original trajectory
unless the researcher introduces a perturbation.
This is what makes a comparison \emph{paired}: the two branches share a bit-identical
physical prefix, not merely an approximate distributional match.
The branching protocol is: snapshot at a physically meaningful event, restore into
branch~A with no modification, restore into branch~B with a controlled perturbation, and
compare the divergent outcomes.

In causal terms~\citep{pearl2009causality}, the branching protocol produces a matched counterfactual pair: the shared physical prefix is the conditioning history, the perturbation is the treatment whose effect is being measured, and the divergent suffix yields the factual and counterfactual outcomes.
What makes the comparison interpretable as evidence of causal effect rather than distributional difference is the bit-identical prefix: the only thing that differs between the two branches is the perturbation itself.

\subsubsection{Event-Driven Branch Points}
\label{sec:triggers}

The usefulness of snapshot/restore depends on \emph{where} to branch.
Raw step indices are fragile: the step at which a ball contacts a platform varies across seeds, coupling the experimental design to seed-specific timing.
Interphyre's declarative trigger language decouples branch points from timing: the researcher specifies a physical event, and the engine fires automatically when the condition is met.
Table \ref{tab:triggers} lists the core factories.


\begin{table}[t]
  \caption{Core trigger factories. Each factory returns a trigger that the engine
  evaluates per step; the researcher specifies \emph{what} physical event gates the
  intervention, not \emph{which step} it corresponds to.}
  \label{tab:triggers}
  \centering
  \small
  \setlength{\tabcolsep}{3.5pt}
  \renewcommand{\arraystretch}{1.08}
  \begin{tabularx}{\linewidth}{@{}
    >{\raggedright\arraybackslash}p{0.30\linewidth}
    >{\raggedright\arraybackslash}p{0.25\linewidth}
    >{\raggedright\arraybackslash}X
    @{}}
    \toprule
    \textbf{Factory} & \textbf{Fires when} & \textbf{Research use} \\
    \midrule
    \texttt{on\_contact(a, b)}
      & Objects \texttt{a}, \texttt{b} touch
      & Branch at collision to isolate post-contact dynamics \\

    \texttt{on\_contact\_duration(a, b, t)}
      & Sustained contact $\geq t$\,s
      & Branch near success to test perturbation sensitivity \\

    \texttt{on\_position\_threshold}
      & Object crosses position
      & Branch at spatial milestone, e.g., ball clears platform \\

    \texttt{on\_velocity\_threshold}
      & Speed crosses threshold
      & Test whether trajectory divergence depends on velocity at branch \\

    \texttt{at\_step(n)}
      & Fixed timestep
      & Uniform temporal sampling for trajectory datasets \\

    \texttt{when(predicate)}
      & Custom predicate on engine
      & Hypothesis-driven intervention on any state property \\
    \bottomrule
  \end{tabularx}
\end{table}

Triggers compose: \texttt{on\_sequence} fires an ordered list of triggers in sequence,
and \texttt{on\_any} fires when any child does.

\subsubsection{Worked Example: Strategy-Dependent Causal Structure}
\label{sec:worked-example}

Figure \ref{fig:intervention-example} runs this protocol on two seeds of the
\texttt{catapult} level whose oracle solutions use different strategies: in seed\,8, the
green projectile bounces off a black deflector ball to reach the basket; in seed\,5, it
flies directly in.

Both seeds succeed in the factual branch.
The two structural interventions, however, have opposite effects.
Removing the black deflector ball causes seed\,8 to fail (the deflector was necessary)
and leaves seed\,5 unaffected (the direct path does not require it).
Shrinking the red action ball to radius $r{=}0.4$ changes the launch dynamics for
seed\,5 (failure) while leaving seed\,8's deflector arc intact (success).
No single intervention explains both outcomes. Each identifies the causal role of one
object relative to one strategy.

Causal relevance, these results show, is not a property of an object in isolation.
The same \texttt{remove\_object} call is causally necessary for one strategy and
causally irrelevant for the other.
A probing study that pools across strategies averages over these opposite effects; if the two strategies occur at similar rates across seeds, the unconditional probe recovers neither.



\begin{figure}[t]
\centering
\begin{minipage}{0.98\linewidth}
\begin{lstlisting}[
  language=Python,
  basicstyle=\ttfamily\scriptsize,
  keywordstyle=\color{blue!55!black},
  commentstyle=\color{green!35!black},
  stringstyle=\color{red!60!black},
  breaklines=true,
  breakatwhitespace=false,
  columns=fullflexible,
  keepspaces=true,
  showstringspaces=false,
  frame=single,
  framesep=2mm,
  tabsize=2,
  aboveskip=0pt,
  belowskip=0pt
]
def run_branch(seed, action, intervene=None):
    env = InterphyreEnv("catapult", seed=seed, enable_interventions=True)
    env.reset(); env.place_action(action)

    snap, _ = env.run_until(
        on_contact("red_ball", "gray_platform")
    )

    env.restore(snap)
    env.step_physics(30)  # branch point: green ball airborne

    if intervene:
        intervene(env)

    return env.run_until(on_success(), max_steps=450)


for seed in [8, 5]:
    action = oracle_action(seed)

    # col 2: factual
    run_branch(seed, action)

    # col 3: remove deflector
    run_branch(
        seed, action,
        lambda e: e.remove_object("black_ball")
    )

    # col 4: shrink red ball
    run_branch(
        seed, action,
        lambda e: shrink_red_ball(e, r=0.4)
    )

    # seed 8: success / failure / success
    # seed 5: success / success / failure
\end{lstlisting}
\end{minipage}

\caption{Branching protocol for Figure~\ref{fig:intervention-example} 
  (catapult level, seeds~8 and~5).
  Each call snapshots at the moment the red ball contacts the catapult arm,
  advances 30~steps to the branch point, where the green ball is airborne,
  and then applies the structural intervention.
  Seed~8: removing the black deflector causes failure; shrinking the red ball does not.
  Seed~5: shrinking the red ball causes failure; removing the deflector does not.
  The object's causal role depends on which strategy the seed uses, not on the object itself.}
\label{fig:counterfactual-protocol}
\end{figure}

Figure \ref{fig:counterfactual-protocol} shows the protocol skeleton.
The snapshot is taken at the physical event that gates the experiment: the moment the
red action ball contacts the catapult arm.
The branch point is 30 steps later, once the green ball is airborne and the two seeds'
trajectories have diverged.
From that point, three branches diverge under different structural modifications.
Section \ref{sec:manipulation} formalizes the split between kinematic and structural
perturbations.

\subsubsection{Perturbation Primitives as Experimental Treatments}
\label{sec:manipulation}

Counterfactual branching requires not just rewinding but \emph{perturbing}, and the
space of perturbations determines the space of experimental questions.
Kinematic perturbations (impulse, continuous force, velocity override, position override,
and freeze) modify the state of existing objects.
These support experiments asking whether an outcome depends on a specific kinematic
variable (e.g., does a 5\% change in post-contact velocity flip the result?).
Structural perturbations (adding and removing objects) modify scene composition.
These support experiments asking whether an outcome depends on the presence of a
particular object (e.g., does removing a supporting surface change the agent's
prediction?). The kinematic/structural distinction is the experimental design axis that makes causal findings interpretable: the reversal in \ref{fig:intervention-example} is visible only when intervention type is held constant and seed varied.

\subsection{Summary}
\label{sec:ip_conclusion}

Conditioning on oracle solution strategy is necessary to recover probing signal in physics-reasoning models: pooling across strategies averages over opposite causal effects.
Linear probes on Qwen3-8B residual-stream activations predict counterfactual outcomes above chance across three levels, but only when indexed to the oracle strategy for each seed.
That result suggests probing without strategy conditioning conflates representations from different solution behaviors and understates what models encode about physical causality.
The same conditioning requirement applies to all three questions motivating the platform: probing how representations encode physical structure, whether reasoning steps commit to correct physics, and whether those commitments are causally related to task success all require probe targets conditioned on the oracle strategy for each seed.
Interphyre's snapshot/restore API makes this tractable: the two branches share a bit-identical physical prefix, so the probe's target is the perturbation's downstream effect rather than a distributional shift in scene statistics.

For RL research, the independent parameter axis is the primary contribution.
Fixing scene topology while varying physics parameters isolates physics transfer from structural variation, a confound Kinetix cannot separate.
The hand-authored curriculum provides concrete tasks with known solution structure, a substrate for curriculum learning and hypothesis-driven behavioral tests; a researcher can ask whether a strategy learned under one physics regime transfers to another, or whether it depends on properties specific to the training regime.

There are also limitations that motivate future work. The 25-level curriculum is hand-authored and does not scale indefinitely without additional authoring effort, though the oracle verification system applies to any new levels without modification to the runtime. We also see this in experiments where skill banks saturate and more test time compute does not help improve on the levels. So an interesting direction for future work is to integrate unsupervised environment design both to scale level complexity, experimentation and intervention tool design, and also to design unlimited newer levels. This would enable us to research building agents that continually discover and learn new knowledge and consolidate and store them in an ever-growing repertoire of skills. 

\newpage
\section{Other Benchmarks and Task Details}
\label{app:external_envs}

\textsc{InterPhyre} is our proposed environment and is documented in full in
Appendix~\ref{app:InterPhyre}. The other three are popular  task families existing in literature, so we
describe what a reader needs to interpret our numbers, what an instance
is, what the agent observes, what counts as success, and the configuration we
ran and defer to the originating papers for everything else.

\subsection{Interactive Games}
\label{app:external_envs:games}

All four games come from the suite of~\citep{lin2026orbit}. Mastermind, Wordle
and Minesweeper are \textsc{Gem} environments~\citep{liu2025gem} reached through
that work's environment adapter; Maze is its own implementation. 

\paragraph{Cross-episode protocol.}
Every game is played under the same multi-episode protocol. A task instance is
replayed for $T$ episodes against a fixed total step budget; each episode resets
the environment but leaves the hidden task unchanged, and the full interaction
history stays in the context window. The agent is told this explicitly in its
system prompt. Success is recorded per episode, so episode~1 measures
first-attempt competence and episode~3 measures what the agent carried across
resets. We use $T{=}3$ throughout, matching the published evaluation.

\paragraph{Mastermind (evaluation task).}
The agent must identify a hidden ordered code within a fixed number of attempts.
Each guess returns structured feedback: black pegs count digits that are correct
and in the correct position, white pegs count digits that are correct but
misplaced. We use the published evaluation parameters: code length $3$, digits drawn from $1$ to $6$, duplicates disallowed, and $3$ turns per episode against a total step cap of $9$. This makes the prompt character-identical to the one in the source paper's appendix. All three constructor arguments must be pinned together because the environment's \texttt{-random} variant otherwise redraws them at reset, and the outer loop terminates first, collapsing the run to a single episode.

\paragraph{Maze (evaluation task).}
A randomly initialised grid world with a start cell, walls, and a goal, generated
so that the shortest path falls in a pre-specified length range. The agent
observes only the four adjacent cells and emits one move per turn, with a cap of
$9$ steps per episode. An episode succeeds if the goal is reached within the cap.
Partial observability is what makes the task cross-episode: the map can only be
learned by traversing it.

\paragraph{Wordle and Minesweeper (transfer sources only).}
Neither is evaluated; both appear only as sources for the cross-task transfer
experiments of Section~\ref{sec:results:transfer}. In Wordle the agent guesses a
hidden five-letter word and receives per-letter feedback (correct and in place,
present but misplaced, absent). Our candidate list is the five-letter subset of
\texttt{en-basic}, $192$ words in total; train and evaluation pools are drawn from
one deduplicated stream so that disjointness is structural rather than incidental.
In Minesweeper the agent reveals or flags cells on a grid with hidden mines, and
revealing a safe cell displays the number of mines among its eight neighbours.
We pin the opening reveal to a fixed cell in every episode and every arm: the
generator excludes mines from the $3{\times}3$ region around the first reveal, so
without pinning the layout is chosen by where the agent happens to open and
episode~2 is a different board than episode~1 more than $90\%$ of the time ---
which would contradict the protocol's own premise that the task is identical
across episodes.

\subsection{Reasoning and Planning Tasks}
\label{app:external_envs:reasoning}

Four single-turn problem families, each with a generator and a programmatic
verifier. We construct every task through a single task factory, so prompt
templates, verifiers and per-task defaults have one source of truth, and we pin
the arguments in Table~\ref{tab:core_task_config} explicitly rather than relying
on the factory's defaults.

\begin{table}[h]
\centering
\small
\caption{The four reasoning and planning tasks and the arguments we pin. All four
use $100$ held-out items; our runs use $60$ training items.}
\label{tab:core_task_config}
\setlength{\tabcolsep}{5pt}
\begin{tabular}{lll}
\toprule
\textbf{Task} & \textbf{What an instance is} & \textbf{Pinned arguments} \\
\midrule
MathGAP~\citep{opedal2024mathgap}
  & arithmetic word problems with
  & none \\
  & controllable proof structure & \\[2pt]
ZebraLogic~\citep{lin2025zebralogic}
  & logic-grid puzzles with controllable
  & \texttt{size\_group="xlarge"} \\
  & search complexity & \\[2pt]
Tower of Hanoi~\citep{shojaee2025illusion}
  & valid action sequences for the
  & \texttt{variant="state"}, \\
  & classic recursive planning problem
  & train disks $[5]$, eval disks $[6]$ \\[2pt]
Matchstick \citep{nasvytis2026insight}
  & move one stick to repair an invalid
  & default dataset \\
  & Roman-numeral equation & \\
\bottomrule
\end{tabular}
\end{table}

\paragraph{Matchstick provenance.}
Matchstick arithmetic has no separate benchmark paper. It is a classic domain in
the study of human insight problem solving, its specific problem types are
adopted from~\citep{nasvytis2026insight}, and the generator and verifier we
run are those released with \textsc{Core}. Each puzzle presents an invalid arithmetic
equation in Roman numerals rendered as matchsticks, and the model must move a
single stick to make the equation valid.

\paragraph{Split behaviour.}
Three of the four evaluation sets are invariant to the training-set size: their
items come from a fixed file, a fixed shuffle slice, or a disjoint disk count.
MathGAP is the exception --- its train and evaluation sets are consecutive slices
\texttt{data[:n]} and \texttt{data[n:n+100]} of one shuffled $280$-row pool, so
raising the training size slides the evaluation window forward. The splits stay
disjoint either way, so this is a comparability constraint across regimes, not a
contamination risk, and we compare only within a regime.

\subsection{\textsc{DiscoverPhysics}}
\label{app:external_envs:discoverphysics}

\textsc{DiscoverPhysics}~\citep{wiemann2026discoverphysicsbenchmarkingllmsoutofthebox} asks an agent to
discover the laws of motion of a simulated world whose physics deliberately
deviates from our own, which is what separates discovery from recall of
established science. The benchmark contains $22$ such worlds ($11$ public, $11$
private); each is generated on demand by an $N$-body simulator, so there is no
static dataset to fit. We use three of the public worlds.

\paragraph{Experimentation through tools.}
\textsc{DiscoverPhysics} is the second environment in our evaluation, after
\textsc{InterPhyre}, in which the agent learns by running experiments through a
tool interface rather than by reading a fixed input. Each round the agent emits
exactly one action, as an XML-tagged tool call:
\begin{itemize}[leftmargin=*,itemsep=1pt,topsep=2pt]
  \item \texttt{<run\_experiment>} takes a JSON list of experiments. Each
    experiment sets the initial conditions of the particles the world lets the
    agent control and up to ten measurement times, and may set a total
    \texttt{duration}. The simulator integrates the hidden law and returns
    noisy positions and velocities of every particle at each requested time.
    A round may batch several experiments.
  \item \texttt{<final\_law>} submits a Python function
    \texttt{discovered\_law} that rolls the system forward under the inferred
    law. It may be accompanied by an optional \texttt{fit\_parameters()}
    declaring up to five free parameters with initial values and bounds, which
    the evaluator fits with \texttt{scipy.optimize} against the agent's own
    trajectories. The submission must also include an \texttt{<explanation>}
    of the world's physics in prose.
\end{itemize}
Which controls \texttt{<run\_experiment>} accepts depends on the world's
topology (listed per world below). The agent must run at least two rounds of
experiments before submitting and is forced to submit in the final round. The
benchmark also offers an optional mid-round tool, \texttt{<run\_mse\_fit>}, that
fits a candidate law to the data collected so far. It is not enabled in our
configuration, so the agent's only feedback before submission is the
trajectories it requests.

\paragraph{Scoring.}
Submissions are scored on two axes. The trajectory axis rolls out the fitted law
on noise-free held-out initial conditions and reports mean position error
normalised by the variance of the test trajectories (normalised MSE). The
explanation axis is an LLM judge comparing the explanation against an
expert-written per-world rubric on a $0$--$10$ scale, which we divide by $10$.
Following the benchmark, every explanation we report is scored by
\texttt{claude-opus-4-6}, a model different from the actor. An episode passes if
normalised MSE is below $0.1$ and the explanation score clears a gate. The
benchmark's gate is $0.90$; we also report $0.75$. Our reward, used to rank
episodes for distillation, is
$0.5\,s_{\text{expl}} + 0.5\,\mathrm{clip}(1 - \text{nMSE}/0.1)$, which lies in $[-0.5, 1]$.

\paragraph{Configuration.}
We run $10$ experimentation rounds per episode, against the benchmark's
$16$, with observation noise $\sigma{=}0.075$ against its $0.05$. A seed sets
only the observation noise; the hidden law of a world is fixed. So within-world
accumulation learns across noise realisations of one law, while cross-world
transfer is the setting in which the instance genuinely changes. Every arm is
evaluated on the same $15$ held-out seeds per world.

\paragraph{The three worlds we use.}
\texttt{oscillator} is a two-particle world. Particle~1 is a fixed source at
the origin and particle~2 is a mobile probe; the agent sets both scalar
properties $p_1, p_2$, the probe's initial position and velocity, the
measurement times and the duration. The spatial force is the 2D Poisson ($1/r$)
law, but its coupling varies in absolute time as
$G(t) = G_0 \cos(\omega t + \varphi)$, so the interaction periodically
reverses sign. Detecting this requires experiments run over a long enough
timescale.

\texttt{circle} has one centre particle and ten ring particles, all of which
source the field, under fractional gravitation derived from a fractional
Laplacian $-(-\nabla^2)^\alpha$. The agent controls only two scalars, the ring
radius and a tangential velocity shared by every ring particle, plus the
measurement times. It cannot place probes, address individual particles or set
masses.

\texttt{hubble} has $26$ particles: a heavy central anchor that is the only
field source, $20$ ring orbiters whose masses cycle through $\{1,2,4\}$, and
$5$ probes whose positions, velocities and masses the agent sets. A radial,
position-dependent Hubble flow $H(r)$ adds a mass-independent outward
acceleration on top of the central law. The agent must therefore disentangle
the central field-mediated force from a body force supplied by space itself.

\texttt{oscillator} and \texttt{circle} are within-world accumulation.
\texttt{hubble} is the cross-world evaluation in
Section~\ref{sec:results:transfer}: its bank is grown sequentially on nine
other public worlds, with \texttt{hubble} and \texttt{ether}, the two worlds
that share its central-law-plus-global-term structure, both held out.


\begin{figure}[H]
  \centering
  \includegraphics[width=0.85\linewidth]{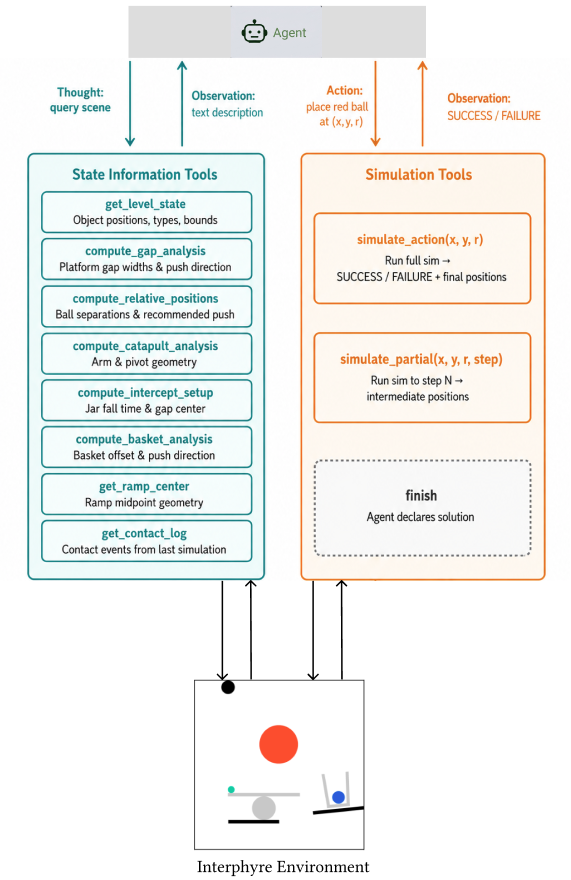}
  \caption{%
    \textbf{Tool interface between the \textsc{HExA} agent and the Interphyre environment.}
    The agent issues two classes of tool calls:
    \textcolor{teal}{\textbf{State Information Tools}} (left) query the scene
    without advancing the simulation---returning object positions, gap widths,
    relative ball positions, and level-specific geometry---and
    \textcolor{orange}{\textbf{Simulation Tools}} (right) place the red ball
    at $(x,y,r)$ and run the physics engine, returning
    \texttt{SUCCESS}/\texttt{FAILURE} with final object positions
    (\texttt{simulate\_action}) or intermediate state at a chosen timestep
    (\texttt{simulate\_partial}).
    The \texttt{finish} action terminates the episode.
    Level-specific state tools (\texttt{compute\_gap\_analysis},
    \texttt{compute\_relative\_positions}, \texttt{get\_ramp\_center}, etc.\@)
    are available only on their respective levels; shared tools are available
    on all levels.
    Selected tools presented here. Full tool signatures and usage instructions are provided in
    Appendix~\ref{app:tools}.
  }
  \label{fig:agent_tools}
\end{figure}

\begin{figure}[h]
  \centering
  \includegraphics[width=\linewidth]{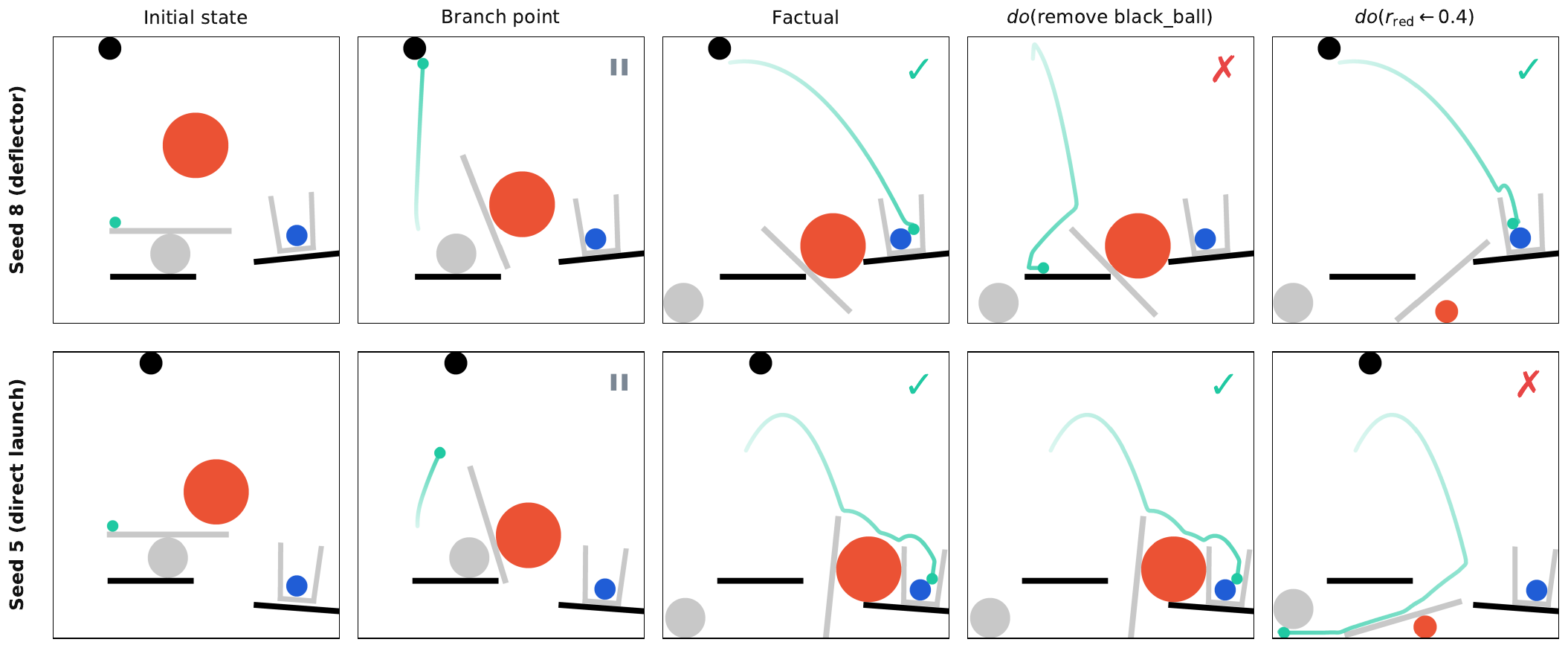}
  \caption{\textbf{\textsc{Interphyre}'s snapshot/restore API.} This branches a shared mid-trajectory state into a
    factual rollout and perturbed alternatives (Section \ref{sec:snapshot-restore}).
    Each row is one seed of the \texttt{catapult} level with a different oracle solution
    strategy.
    The leftmost column shows the initial scene for reference; the teal curve is the
    green ball's trajectory, shown up to the branch point in the \emph{Branch point}
    column and continuing through the outcome in each subsequent column.
    Top (seed\,8, deflector strategy): removing the deflector causes failure; shrinking
    the red action ball does not.
    Bottom (seed\,5, direct-launch strategy): shrinking the red ball causes failure;
    removing the deflector does not.
    The same interventions have opposite causal relevance across seeds, showing that causal
    relevance is a property of an object relative to the active strategy, not of the object
    in isolation.
    See Section \ref{sec:worked-example} for the branching protocol.}
  \label{fig:intervention-example_main_paper}
\end{figure}


\section{Tool Descriptions}
\label{app:tools}
This appendix provides complete descriptions of all tools available to the \textsc{ReAct} agent.
Tools are partitioned into \emph{shared tools} (available on every level) and \emph{level-specific tools} (exposed only on the corresponding puzzle).

\subsection{Shared Tools}
\label{app:tools:shared}

\paragraph{\texttt{get\_level\_state()}.}
Returns a structured text description of the current puzzle scene: the name, position, radius (for balls), dimensions and angle (for bars), and dynamic flag of every object; the world bounds ($x,y \in [-5, 5]$); and the level-specific success condition.

\paragraph{\texttt{simulate\_action(x, y, radius)}.}
Places the red ball at $(x, y)$ with the given \texttt{radius} $\in [0.1, 2.0]$ and runs the full physics simulation (up to 2{,}000 steps at 60\,Hz) to completion.
Returns the outcome (\texttt{SUCCESS}/\texttt{FAILURE}), total step count, and final positions and velocities of all objects.
Before launching the simulation the toolkit performs \emph{pre-simulation validation}: if the placement violates world bounds, overlaps an existing ball, or intersects a platform, the call returns a detailed error specifying the minimum adjustment required rather than consuming a simulation attempt.

\paragraph{\texttt{simulate\_partial(x, y, radius, stop\_step)}.}
Identical to \texttt{simulate\_action} but halts after \texttt{stop\_step} physics steps (or earlier if the success condition is satisfied).
Returns object positions and velocities at the chosen checkpoint, enabling mid-simulation diagnostics without committing to a full run.

\paragraph{\texttt{get\_contact\_log()}.}
Returns timestamped collision events from the most recent simulation: a list of \texttt{(step, object\_A, object\_B)} entries (capped at 20; a notice is appended if more exist), or \texttt{"No contact events recorded"} if no contacts occurred.
Must be called after \texttt{simulate\_action} or \texttt{simulate\_partial}.

\paragraph{\texttt{finish(x, y, radius)}.}
Submits the agent's final answer: places the red ball at $(x, y, r)$ and ends the episode.
The episode is scored as a success if the physics engine confirms the goal condition.

\subsection{Tools for Levels}
\label{app:tools:level}

\subsubsection{Down to Earth — \texttt{compute\_gap\_analysis()}}

Analyses the gaps on each side of the platform that separates the green ball from the floor.
Returns:
\begin{itemize}[nosep]
  \item platform span (left and right $x$-coordinates of the platform edges);
  \item green ball diameter;
  \item left gap width (distance from left wall to platform left edge) and right gap width (distance from platform right edge to right wall);
  \item \texttt{YES}/\texttt{NO} for whether the green ball can fit through each gap;
  \item a recommendation indicating which side has the larger viable gap, or a \texttt{WARNING} if neither gap is wide enough.
\end{itemize}

\subsubsection{Two Body Problem — \texttt{compute\_relative\_positions()}}

Analyses the spatial relationship between the green and blue balls.
Returns:
\begin{itemize}[nosep]
  \item position and radius of each ball;
  \item horizontal separation $\Delta x$, vertical separation $\Delta y$, and centre-to-centre distance;
  \item minimum contact distance (sum of radii);
  \item relative direction of the blue ball from the green ball (\texttt{LEFT}/\texttt{RIGHT}).
\end{itemize}
This information guides placement of the red ball to push the green ball into contact with the blue ball.

\subsubsection{Catapult — Four Analysis Tools}

The Catapult level exposes four complementary tools to handle its multi-body lever mechanics.

\paragraph{\texttt{describe\_scene\_geometry()}.}
Returns a strategy-neutral geometric inventory of the scene with no prescriptive advice:
all balls (position, radius, dynamic flag), all bars (centre, angle, length, dynamic flag), and all baskets (position, dynamic flag); plus the key pairwise distance (green--blue) and the success condition.
Intended as a first-pass survey before committing to a simulation strategy.
No arguments.

\paragraph{\texttt{simulate\_with\_trace(x, y, radius, object\_names, stop\_step)}.}
Places the red ball and runs a full (or truncated) simulation, then returns a per-object kinematic-extrema summary for every name in \texttt{object\_names} (e.g.\ \texttt{["green\_ball", "catapult\_arm"]}).
The summary per object includes: \texttt{peak\_y} (maximum height reached), \texttt{min\_y} (minimum height), \texttt{v\_max} (peak speed), $\Delta\mathrm{pos}$ (net displacement), and, for bodies that rotate appreciably, peak angular speed $\omega_{\max}$ and angle range $[\theta_{\min}, \theta_{\max}]$.
The most relevant contact events involving the red ball or any traced object are also returned (capped at 15).
Used to verify lever mechanics and confirm that the green ball reaches the basket.

\paragraph{\texttt{trace\_green\_ball(x, y, radius)}.}
A lighter trajectory probe than \texttt{simulate\_with\_trace}: only the green ball is sampled.
Places the red ball, runs the simulation up to a 600-step ceiling, and returns the green ball's $(x, y)$ waypoints at fixed 30-step intervals together with a start/end/peak summary (start position, end position, $\Delta$, peak~$y$, peak speed).
Stops early once the green ball settles (consecutive samples with $|\Delta\mathrm{pos}| < 0.10$) or once the success condition fires.
Use this when the only quantity of interest is \emph{where} the green ball travels rather than full per-object kinematics or contact events.

\paragraph{\texttt{predict\_first\_contact(x, y, radius)}.}
A cheap pre-simulation check (at most 90 physics steps, $\approx$1.5\,s of simulated time) that identifies the first object the red ball contacts after release.
Returns: placement validity, the name of the first object hit, the step at which contact occurs, approach speed, approximate contact point, and surface normal.
Used to verify that the red ball reaches the intended catapult arm before committing to a full simulation.

\subsubsection{Falling Into Place — \texttt{compute\_intercept\_setup()}}

Computes the intercept geometry for the timed-interception puzzle.
Returns:
\begin{itemize}[nosep]
  \item green ball position and which platform it rests on (\texttt{LEFT}/\texttt{RIGHT});
  \item blue jar position and travel direction the green ball must take;
  \item the platform edge the green ball must cross and the gap centre $x$-coordinate;
  \item estimated time until the jar falls to platform height (in seconds and in 60\,Hz physics steps).
\end{itemize}

\subsubsection{Basket Case — \texttt{compute\_basket\_analysis()}}

Analyses the spatial relationship between the green ball and the basket that the agent must avoid.
Returns:
\begin{itemize}[nosep]
  \item green ball position and radius;
  \item basket position and scale, including opening half-width;
  \item purple ground position (if present);
  \item horizontal distance from the green ball to the basket centre;
  \item recommended push direction (\texttt{LEFT}/\texttt{RIGHT}) to deflect the green ball away from the basket.
\end{itemize}.

\subsubsection{Pass the Parcel — \texttt{get\_ramp\_center()}}

Analyses the ramp-and-basket mechanism in the Pass the Parcel puzzle.
Returns the centre coordinates $(x, y)$ of the ramp, together with inferred geometry including the top-basket position, bottom-basket position, platform position, ramp angle, and ramp bounds.
This information guides placement of the red ball to roll it onto the ramp and interact with downstream mechanisms.

\subsubsection{Cliffhanger — \texttt{compute\_cliffhanger\_analysis()}}

Analyses the cliffhanger geometry: a vertical green bar standing on a black platform with a ceiling above and the purple ground below.
Returns:
\begin{itemize}[nosep]
  \item green bar centre, length, thickness, and angle, plus the $(x, y)$ coordinates of its bottom point (resting on the platform) and top point (opposite end);
  \item black platform extents (left and right $x$, centre $y$, top-surface $y$, thickness);
  \item ceiling $y$ and purple-ground $y$ where present;
  \item the bar's bottom-point distance to the platform's left and right edges;
  \item closer platform edge (\texttt{LEFT}/\texttt{RIGHT}) — i.e.\ the edge the bar must fall past — and the width of the falling gap on that side (between the platform edge and the world wall);
  \item the implied tip direction the bar's top end must rotate (\texttt{LEFT}/\texttt{RIGHT}) for the bar's centre of mass to cross the closer edge.
\end{itemize}

\subsubsection{Tipping Point — \texttt{compute\_tipping\_point\_analysis()}}

Analyses the tipping-point geometry: a vertical green bar pinned at its base by a small gray basket, with a static purple wall flush against either the left or right side of the box.
Returns:
\begin{itemize}[nosep]
  \item green bar centre, length, thickness, and angle, plus the $(x, y)$ coordinates of the free top end and the basket-pinned bottom end;
  \item gray basket centre (when present);
  \item purple wall $x$ and its top/bottom $y$;
  \item the wall's side relative to the bar (\texttt{LEFT}/\texttt{RIGHT});
  \item horizontal distance from the bar centre to the wall;
  \item approximate angle the bar must tip through, treated as a rigid stick pivoting at its base, to reach the wall ($\arcsin(\Delta x / L)$ when $\Delta x < L$, else \texttt{N/A});
  \item suggested tip direction (\texttt{LEFT}/\texttt{RIGHT}).
\end{itemize}

\lstdefinestyle{prompt}{
  basicstyle=\ttfamily\scriptsize,
  breaklines=true,
  breakatwhitespace=false,
  columns=fullflexible,
  keepspaces=true,
  showstringspaces=false,
  frame=single,
  xleftmargin=0pt,
  xrightmargin=0pt,
  aboveskip=0.5em,
  belowskip=0.5em,
  inputencoding=utf8,
  extendedchars=true,
  literate={—}{{---}}1
}
\section{Baseline Implementation Details}
\label{app:baselines}

This appendix documents every baseline used in our results for all environments : the \textsc{Zero-shot} no-interaction control, the per-step \textsc{ReAct} loop, the Reflexion
wrapper ($K{=}2$ trials), the two external
memory baselines (Episodic RAG and MemRL), and the GRPO fine-tune.  We run open weight models using vLLM, and Huggingface on upto 4 A100 GPUs with 80GB VRAM each and the Claude Sonnet model using the Anthropic API. Together with the per-level system
prompts (Appendix~\ref{app:systemprompts}) and tool descriptions
(Appendix~\ref{app:tools}), the contents below are sufficient to reproduce
every baseline number reported in Section~\ref{sec:results}.


\subsection{Zero-shot: One-Shot, No-Tool Baseline}
\label{app:baselines:direct}

\textsc{Zero-shot} removes the \textsc{ReAct} loop entirely. The agent emits a
single \texttt{Thought} and one \texttt{Action}, never iterates, never calls a
simulator-write tool of its own, and never observes the consequence of a
placement before committing. It is not, however, denied sight of the scene:
the harness performs one scripted \texttt{get\_level\_state} call on the
model's behalf and embeds the returned object table --- exact positions,
sizes, dynamic flags and any level-specific geometry --- verbatim in the user
message, so the model commits with the same numeric view of the initial state
that \textsc{ReAct} obtains on its first turn. What it lacks is not
information about the scene but the ability to act on it and observe the
result. It therefore serves as a lower-bound test for how much of each level
is solvable from a single static observation, with no interaction.


\paragraph{Per-seed protocol.}
The harness performs exactly two operations per seed:

\begin{enumerate}[leftmargin=*, nosep]
\item A single scripted call to \texttt{get\_level\_state} via the same
  \textsc{Interphyre} toolkit used by \textsc{ReAct}. The returned object
  table (positions, sizes, dynamic flags, and any level-specific
  geometry) is captured verbatim.
\item One non-streaming Claude Sonnet 4.6 call. The conversation has a
  level-specific system prompt and a single user message that embeds the
  scene table from step~1 and demands the final $(x, y, r)$ placement as
  JSON. The harness parses the JSON, runs \texttt{simulate\_action} once,
  and records success/failure.
\end{enumerate}

The model is given no read tools, no probe tools (\texttt{simulate\_partial},
\texttt{get\_contact\_log}), and no retries. There is one inspection of the
scene and one committed placement.

\paragraph{Actor model and decoding.}
Actor model: Claude Sonnet 4.6 invoked via the \texttt{claude -p} CLI with
\texttt{--max-turns 1} and \texttt{--no-session-persistence}, matching the
actor used by the \textsc{ReAct} and Reflexion baselines on the same level
so the comparison isolates the harness, not the model. Temperature $0.3$
and \texttt{max-new-tokens} $700$, identical to \textsc{ReAct}.

\paragraph{Answer prompt.}
The model reasons before answering: the answer block asks for a short
\texttt{Thought} followed by the committed placement, so every recorded
placement is preceded by explicit reasoning. The block appended to the user
message is:

\begin{lstlisting}[style=prompt]
Submit your FINAL red ball placement now using the finish action.

Output EXACTLY this format and NOTHING else (no preamble, no commentary
outside the Thought line):
Thought: <2-3 short sentences explaining your placement choice>
Action: finish
Action Input: {"x": <X>, "y": <Y>, "radius": <R>}

Replace <X>, <Y>, <R> with your chosen floating-point numbers (subject to
the placement constraints above).
\end{lstlisting}

The reported base rates are therefore not an artefact of suppressing
reasoning. On \texttt{catapult} --- the level where every model scores $0\%$
in Table~\ref{tab:direct_baseline_results_main} --- all 50 held-out seeds
produced a parseable placement and all 50 failed. Raising the reasoning
effort of a frontier reasoning model on the same seeds does not change the
outcome: $0/50$ again, this time spending $287$K reasoning tokens, on 13 of
which the model exhausted its token budget mid-reasoning and never committed
to a placement (scored, as always, as failures). One inspection and one
committed placement is the binding constraint on this level, not the amount
of deliberation before the answer.

\paragraph{System prompt.}
The system prompt is self-contained and does not reuse the \textsc{ReAct}
level prompts, so the baseline cannot accidentally inherit ReAct framing.
It states the world dimensions, gravity direction, placement constraints,
and a per-level \emph{elements} block enumerating the dynamic and static
objects with one-line descriptions. The full per-level blurbs are in the
released code (\texttt{react\_agent/run\_direct\_answer\_claude.py}).

\paragraph{Termination and scoring.}
A seed terminates after the single \texttt{simulate\_action} call. Success
is the same simulator predicate used by every other method (target pair in
contact for $\geq 3$\,s). If the model's response cannot be parsed into a
valid $(x, y, r)$ tuple, the seed is recorded as a failure with no
simulator call (these are rare: the answer block fixes a single syntactic
shape for the placement).
\subsection{ReAct: Inner Loop}
\label{app:baselines:react}

Every method we evaluate runs the same \textsc{ReAct}~\citep{yao2023reactsynergizingreasoningacting} loop on top of
the \textsc{Interphyre} tool API. \textsc{HExA} adds a skill block to the system
prompt; Reflexion adds a reflection block; both leave the inner loop
otherwise unchanged.

\paragraph{Per-step protocol.}
At each iteration the agent emits exactly one block of three lines:

\begin{lstlisting}[style=prompt]
Thought: <free-form reasoning over the conversation history>
Action: <tool name, drawn from the level's API>
Action Input: <JSON arguments, or empty for nullary tools>
\end{lstlisting}

The harness parses these three fields, dispatches the tool against the
simulator, and appends the returned text as

\begin{lstlisting}[style=prompt]
Observation: <tool output>
\end{lstlisting}

The agent then produces the next \texttt{Thought}/\texttt{Action}/
\texttt{Action Input} block, conditioned on the full conversation so far.

\paragraph{Termination, attempts, and scoring.}
A trial ends as soon as one of three things happens.
(1)~The agent emits \texttt{Action: finish} with arguments
$(x, y, r)$; the harness places the red ball and the simulator runs to
completion.
(2)~The agent emits \texttt{Action: simulate\_action} with arguments
$(x, y, r)$ and the simulator confirms the goal predicate (the named target
pair stays in contact for at least 3 seconds, i.e.\ 180 physics steps at
60\,Hz).
(3)~The iteration count reaches the cap of 25 without either of the above.
Each \texttt{simulate\_action} call counts as a placement attempt; an
agent may make up to roughly 24 attempts within a single trial.

\paragraph{Decoding.}
All baselines share decoding: temperature $0.3$, max-new-tokens $700$ per
agent turn, and a single tool call per turn (parsed greedily from the first
\texttt{Action} line in the model's reply).

\subsection{Reflexion}
\label{app:baselines:reflexion}

Reflexion~\citep{shinn2023reflexion} wraps the same single-trial
\textsc{ReAct} agent (above) in
$K{=}2$ trials per seed each trial runs for 12 iterations. Trial 1 runs as a standard \textsc{ReAct} loop. If
trial 1 fails, the harness calls a separate Claude Sonnet instance to write
a verbal reflection, prepends the reflection to the system prompt of trial
2, and runs trial 2 from a fresh 25-iteration budget. The episode is
scored a success if either trial succeeds; iteration count and wall-clock
time aggregate across the two trials.

\paragraph{Reflection model.}
The reflection step uses Claude Sonnet 4.6 (matching the actor for our
\texttt{catapult} run), invoked with a single non-streaming completion. The
reflection prompt is held fixed across seeds and levels.

\paragraph{Reflection prompt.}
The reflector receives the full \textsc{ReAct} trajectory of the failed
trial (\texttt{Thought}, \texttt{Action}, \texttt{Observation} per step) plus the
final observation. Long observations are truncated to the first 300 and
last 300 characters with an ellipsis to keep the prompt under context.
The prompt template is:

\begin{lstlisting}[style=prompt]
SYSTEM:
You are analyzing a failed attempt at a 2D physics puzzle.

You will receive the full Thought/Action/Observation trajectory from a single failed attempt by a ReAct-style agent, followed by the final outcome and any prior reflections accumulated on this same task.

Your job: produce a short reflection (<=5 sentences) that the agent will read before its next attempt on the SAME task. Cover:
(a) which strategy/approach the agent pursued in this attempt;
(b) the specific kinematic, geometric, or procedural reason it failed (cite concrete coordinates, distances, or contact events from the observations);
(c) one concrete different action or strategy to try next -- be specific (object names, approximate (x, y, radius), expected mechanism).

Hard rules:
- Do NOT repeat lessons that already appear in the prior reflections list.
- Do NOT re-state the goal or the puzzle rules.
- Output ONLY the reflection text. No preamble, no headers, no markdown.

USER:
Outcome: FAILURE.
Final observation:
{final_observation}

Trajectory:
--- Step 1 ---
Thought: {thought}
Action: {action}
Observation: {observation_truncated}
... (one block per step) ...

## Prior reflections (do not repeat these)
1. {prior_reflection_1}
...

Write the reflection now. Be specific and concise (<=5 sentences).
\end{lstlisting}

\paragraph{Trial 2 conditioning.}
The returned reflection text is appended to the level's system prompt under
a heading \texttt{\#\# Reflexion memory} before trial 2 starts. The actor
sees this block exactly once at trial 2's first turn, in addition to the
unchanged level system prompt and tool list. Trial 2 then proceeds as a
fresh \textsc{ReAct} loop with the standard 25-iteration cap.

\paragraph{Empty / failed reflections.}
If trial 1 already succeeds, no reflection is generated and trial 2 is
skipped (we still report the seed as a success but the reported iteration
count is just trial 1's). If the reflection model returns an empty string
(CLI error or empty output), trial 2 runs without a memory block, which
makes it equivalent to a second \textsc{ReAct} trial from scratch.

\subsection{Shared Harness for the External-Memory Baselines}
\label{app:baselines:memory_common}

Episodic RAG and MemRL differ only in what they store and how they select it.
Everything else below is held identical between them, so any gap between the two
is attributable to the memory representation rather than to the harness.

\paragraph{Online episode protocol.}
For each seed, the harness constructs and embeds the current intent, retrieves a
memory block once, and appends that block to the system prompt before the agent
begins its at-most-$25$-turn ReAct trajectory. After the environment returns the
episode outcome, the harness first applies the method's feedback update and then
writes the new episode to memory. The updated store and retrieval diagnostics
are flushed after every seed. Thus memory can change \emph{between} episodes,
but retrieval is not repeated inside an episode.

\paragraph{Store.}
Our port uses a JSON-backed store rather than MemOS or an external vector
database. Each record carries the level, seed, outcome class
(\textsc{success}/\textsc{failure}), intent text, intent embedding, injectable
experience text, utility $Q$, and update count. Retrieval computes cosine
similarity over the stored embeddings. Episodic RAG only appends records and
never changes $Q$; MemRL updates $Q$ on the same record schema.

\paragraph{Retrieval key (``intent'').}
Both baselines key memory on an \emph{intent}: the text that is embedded and
compared against past episodes. MemRL calls this $z$ (its ALFWorld runner uses
the goal string); Episodic RAG calls it the problem text. The obvious
construction --- the level description plus the toolkit's state dump --- does
not work on \textsc{InterPhyre}: within a level it is roughly $95\%$ identical
across seeds (same prose, same object names, same success condition, same world
bounds, only coordinates differ), and measured over $105$ seeds of
\texttt{catapult} and \texttt{pass\_the\_parcel} its pairwise cosine similarity
is $0.9994 \pm 0.0003$.
At that spread MemRL's first retrieval phase is all-or-nothing and its similarity
normalization is pure noise. We therefore use a \emph{geometry} intent: only the
instance-varying content, i.e.\ per-object coordinates and extents plus derived
relations between the objects the goal predicate names, with no prose and no
success condition. This is the honest analogue of MemRL's intent, since in
ALFWorld the goal text differs per task while in \textsc{InterPhyre} what differs
per instance is the geometry. Building it costs no LLM call and no rollout.

\paragraph{Embeddings.}
The source-faithful Episodic RAG configuration uses the local Jina v2 base
English encoder, as in \citet{nasvytis2026core}; the source-faithful MemRL
configuration uses OpenAI
\texttt{text-embedding-3-large}. Text embeddings are cached on disk
by (model, text hash), so the current intent is embedded once and the resulting
vector is reused when the episode is written back.

\paragraph{Injection point and context accounting.}
The retrieved block is appended to the agent's system prompt at exactly the
position where \textsc{HExA} appends its \texttt{LEARNED PHYSICS SKILLS} block,
so the base prompt is byte-identical across methods and only the contents of the
injected block differ. Episodic RAG injects the outcome label and
raw rollout content. MemRL injects, for each selected record, its outcome, source
seed, current $Q$, update count, and experience text. The intent itself is only
the retrieval key and is not repeated in the memory block.

\paragraph{Trajectory rendering.}
Both baselines render a stored trajectory as its last $12$ ReAct steps. A step is
a \texttt{Thought--Action--Observation} interaction; it is not a physics frame or
simulator time step. The observation is the textual return from a tool call
(e.g., scene state, placement error, or simulation outcome and final object
positions). Each observation is truncated to $600$ characters by retaining its
head and tail.
\subsection{Episodic RAG}
\label{app:baselines:rag}

Episodic RAG builds on retrieval-augmented generation
\citep{lewis2020rag}. Every completed episode is indexed by its problem intent.
For a new problem, the method supplies a semantically nearby successful attempt
and a nearby unsuccessful attempt, together with their verifier feedback. It has
no learned value function: improvement can only come from adding more examples
to the retrieval bank.

\paragraph{Retrieval.}
$k_{\mathrm{success}} = k_{\mathrm{failure}} = 1$. Candidates are ranked by cosine
similarity against the query intent, then split by outcome class and the top
entry of each class is taken. If either class is empty, the fallback in the quote
above applies and the two most similar rollouts are taken regardless of outcome;
the mode actually used is logged per episode

\paragraph{Prompt template.}
Episodic RAG makes no additional LLM call to summarize, rewrite, or otherwise
transform a retrieved episode. The following block is constructed
deterministically and appended to the unchanged \textsc{ReAct} system prompt.
Braced fields are filled at run time, and the rollout body repeats the step
template for at most the last $12$ steps. When two records are available, the
entire \texttt{Past rollout} section is repeated once per record.

\begin{lstlisting}[style=prompt]
## Retrieved Past Rollouts (from your own past attempts on this level)

### Past rollout {i} — {SUCCESS|FAILURE} (level {level}, seed {seed})
Level: {level}  |  Seed: {seed}
Verifier: {SUCCESS|FAILURE}
Iterations used: {iterations}
Simulator calls: {simulator_calls}
Shaped reward: {reward}
Final placement: x={x}, y={y}, radius={radius}
Final observation: {final_observation}

Rollout:
(showing last 12 of {total_steps} steps)   [only if truncated]
--- Step 1 ---
Thought: {thought}
Action: {action}
Observation: {observation}
...

These are complete past attempts with the environment's verdict. Use the successful rollout as a template where the geometry matches, and avoid repeating what the unsuccessful rollout did.
\end{lstlisting}

The footer above is reproduced from the implementation. In our port,
``complete'' means the complete stored representation: as noted below, that
representation is capped at the final $12$ steps rather than retaining the
unbounded original rollout.

\paragraph{Write-back.}
Every episode is stored with its verifier outcome, iterations, simulator-call
count, shaped trajectory reward, final placement, final observation, and raw
Thought--Action--Observation trajectory. There is no summarization, utility
update, or consolidation, so the bank grows monotonically.

\paragraph{Deviation from the published specification.}
We keep only the last $12$ steps of each rollout rather than all of them. The
rollouts are still raw --- no model summarizes them --- but they are no longer
complete. Two reasons: \texttt{gpt-oss-120b} must run with eager attention in our
setup and the untruncated block exhausts memory on a $4\times$A100 node during
the first episode; and holding trajectory length equal to MemRL's is what makes
the summarized-versus-raw contrast interpretable. Verbatim storage is recoverable
by setting the step cap to zero.

\subsection{MemRL}
\label{app:baselines:memrl}

MemRL~\citep{zhang2026memrl} is value-aware episodic memory over
(intent, experience, utility) triplets $(z, e, Q)$. We port the official release
to the \textsc{InterPhyre} episode interface using its \textbf{ALFWorld}
configuration, the closest analogue among its published settings: sequential,
exploration-heavy, and environment-verified. Table~\ref{tab:memrl_hparams} lists
the resolved values. In our store, $(z,e,Q)$ maps respectively to the record's
\texttt{intent}, \texttt{content}, and \texttt{q} fields; only $z$ is embedded.

\paragraph{Phase A: similarity recall.}
Candidates are scored by cosine similarity against the query intent; those above
a threshold $\delta$ are kept, capped at $k_1$. If the admitted set is empty, no
memory block is injected at all and the agent falls back on the frozen model, as
specified.

\paragraph{Phase B: value-aware selection.}
The admitted pool is re-ranked by
$\operatorname{score}_i=w_{\mathrm{sim}}\tilde{s}_i+w_Q\tilde{Q}_i$ and the top
$k_2$ are injected.

\paragraph{Utility update.}
Every memory injected during an episode receives a Monte-Carlo update
$Q \leftarrow (1-\alpha) Q + \alpha \, (r + \gamma \max Q')$ with $\gamma = 0$, so
the target is the terminal reward: $r = +1$ on success and $r = -1$ on failure. 

\paragraph{Actor prompt template.}
The selected records are ordered by the Phase-B score and rendered into the
following block, which is appended to the unchanged \textsc{ReAct} system
prompt. The \texttt{experience} field expands to either the successful or failed
record format shown immediately below.

\begin{lstlisting}[style=prompt]
## Archived Memories (from your own past experiences)

### Memory {i} — {SUCCESS|FAILURE} (seed {seed}, utility Q={Q}, updates={q_visits})
{experience_i}
...

These experiences were retrieved because they are similar to the current scene and have been useful in the past. Adapt them to this seed's geometry rather than copying coordinates blindly.
\end{lstlisting}

For a successful episode, \texttt{experience\_i} has this form:

\begin{lstlisting}[style=prompt]
Task: {level} (seed {seed})

SCRIPT:
{generated_success_script}

TRAJECTORY:
{rendered_last_12_steps}
\end{lstlisting}

For a failed episode, it has this form:

\begin{lstlisting}[style=prompt]
TASK REFLECTION:
Task: {level} (seed {seed})

What went wrong:
{generated_failure_reflection}

Failed approach:
{rendered_last_12_steps}
\end{lstlisting}

\paragraph{Memory-building prompts.}
MemRL makes one additional LLM call after each episode to construct the
experience text. A successful episode uses the following system prompt at
temperature $0.0$ on backends that expose temperature:

\begin{lstlisting}[style=prompt]
You are analyzing a SUCCESSFUL attempt at a 2D physics placement puzzle.

You will receive the task (level description and scene geometry) and the agent's Thought/Action/Observation trajectory.

Produce a concise, high-level script that captures the essential steps and decision points. The script must be:
1. Generic enough to apply to similar puzzle instances (different seeds, shifted geometry).
2. Specific enough to provide useful guidance (name objects, give approximate coordinates or offsets, state the mechanism).
3. At most 5 high-level steps.
4. Focused on strategy and key decisions, not on individual tool calls.

Output ONLY the numbered script. No preamble, no headers, no markdown fences.
\end{lstlisting}

A failed episode uses this system prompt at temperature $0.3$ when supported:

\begin{lstlisting}[style=prompt]
You are analyzing a FAILED attempt at a 2D physics placement puzzle.

You will receive the task (level description and scene geometry) and the agent's Thought/Action/Observation trajectory.

This attempt failed. Analyze what went wrong and suggest improvements for future similar tasks. Cover:
1. Incorrect assumptions the agent made about the physics or geometry.
2. Steps to improve — cite concrete coordinates, distances, or contact events from the observations.
3. What to avoid next time.

Keep it under 8 sentences. Output ONLY the reflection text. No preamble, no headers, no markdown fences.
\end{lstlisting}

Both calls receive the same user-message template:

\begin{lstlisting}[style=prompt]
Task:
{intent}

Trajectory:
{rendered_last_12_steps}
\end{lstlisting}

For the local/open-weight backend these are true system and user messages. The
Claude CLI backend serializes the same content with literal
\texttt{SYSTEM:}/\texttt{USER:}/\texttt{ASSISTANT:} role labels; as noted under
deviations, that backend does not expose the requested temperature setting.

\paragraph{Write-back.}
A successful episode is \emph{proceduralized} into a script of at most five
high-level steps, stored together with the trajectory; a failed episode is stored
as a reflection note plus the failed approach. The summarizer sees the intent and
the rendered last-$12$-step trajectory; it does not generate the retrieval key. An intent whose
nearest existing entry of the same outcome class exceeds the add-similarity
threshold refreshes that entry in place rather than growing the bank. The update
adopts the new content and episode provenance while retaining the old intent,
embedding, identifier, $Q$, and visit count. 

\begin{table}[h]
\centering
\small
\caption{MemRL hyperparameters. Values follow the official ALFWorld
configuration except where noted; $\delta$ and the similarity normalization
constants are dataset-dependent and calibrated per task.}
\label{tab:memrl_hparams}
\begin{tabular}{lll}
\toprule
\textbf{Parameter} & \textbf{Value} & \textbf{Role} \\
\midrule
$k_1$                        & 5      & Phase-A recall cap \\
$\delta$                     & calibrated & Phase-A similarity threshold \\
$k_2$                        & 3      & memories injected per episode \\
$w_{\mathrm{sim}}$ / $w_{Q}$ & 0.5 / 0.5 & Phase-B score weights \\
$\epsilon$                   & 0.0    & random-selection rate \\
$Q_{\min}$                   & $-10$  & utility floor \\
$\alpha$                     & 0.3    & utility learning rate \\
$\gamma$                     & 0.0    & discount (terminal reward) \\
$r$ (success / failure)      & $+1$ / $-1$ & episode reward \\
$Q_{\mathrm{init}}$          & 0.0    & initial utility \\
add-similarity threshold     & 0.99$^{\dagger}$ & update-in-place cutoff \\
novelty threshold            & 0.85   & reward-attribution cutoff \\
reward merge gain            & 0.1    & attribution scale \\
Trajectory steps kept        & 12     & shared with Episodic RAG \\
Observation truncation       & 600 chars & shared with Episodic RAG \\
\bottomrule
\end{tabular}
\\[3pt]
\end{table}

\paragraph{Calibration of $\delta$.}
MemRL's threshold is explicitly dataset-dependent: its appendix determines
$\delta$ from the pairwise cosine similarity distribution of task descriptions
within each benchmark, taking the top-$20\%$ quantile, and its similarity
normalization uses precomputed dataset statistics rather than per-pool ones. Both
therefore have to be \emph{measured} on our intents rather than copied from the
ALFWorld configuration. 

\paragraph{Numeric-geometry retrieval key.} We include a labeled numeric-key variant. It parses the fixed-format
geometry intent into coordinates, extents, angles, and pairwise offsets;
standardizes every feature over the calibration pool; and applies the same
cosine retrieval and two-phase MemRL selector. This changes only the retrieval
key, not the stored experience, utility update, actor, or interaction budget. It
is a documented departure from embedding a natural-language task description.

\paragraph{Deviations from the official release.}
First, \textsc{InterPhyre} uses
a geometry-derived intent because the goal sentence is constant within a level.
Second, merge-on-write uses $0.99$ and same-outcome replacement, rather than the
ALFWorld configuration's unrestricted $0.90$ match. Finally, with the Claude actor, memory-building calls use the
\texttt{claude -p} CLI, which exposes no temperature control; consequently the
release's $0.0$ proceduralization / $0.3$ reflection split cannot be reproduced
on that backend. Open-weight runs reuse the actor model and honor both
temperatures.

\subsection{GRPO: Post-Training Baseline with Gradient based Weight-Update}
\label{app:baselines:grpo}
We use Group Relative Policy Optimization~\citep{shao2024deepseekmath} to
fine-tune Qwen-2.5-3B-Instruct on the same two levels reported in Section~\ref{app:results:interphyre}
(\texttt{down\_to\_earth}, \texttt{two\_body\_problem}) using the VERL-TOOL framework~\citep{jiang2025verltoolholisticagenticreinforcement}. GRPO is included
purely as a sample-efficiency comparison for Section~\ref{app:results:grpo}; it is the only
method we evaluate that updates model weights.

\paragraph{Setup.}
Per-step batch size $1$, four rollouts per gradient step ($n{=}4$), binary
$\{0,1\}$ environment reward (the simulator's success predicate, identical to the
scoring rule used by every other method). One \emph{epoch} corresponds to a full
pass over the 50 training seeds, i.e.\ 50 gradient updates. We train for 10
epochs (500 total gradient steps).

\paragraph{Hyperparameters.}
All runs share the configuration in Table~\ref{tab:grpo_hparams}; the only
difference between the non-skilled and static-skilled variants is
\texttt{max\_prompt\_length} (2048 vs.\ 3072), which is increased to accommodate
the skill-bank prefix (${\approx}2200$ tokens vs.\ ${\approx}1400$ tokens without
skills).

\begin{table}[h]
\centering
\small
\caption{GRPO hyperparameters shared across all four runs.}
\label{tab:grpo_hparams}
\begin{tabular}{ll}
\toprule
\textbf{Parameter} & \textbf{Value} \\
\midrule
Model                   & \texttt{Qwen/Qwen2.5-3B-Instruct} \\
Algorithm               & GRPO \\
Training seeds          & 1--50 (val: 51--100) \\
Batch size              & 1 seed / gradient step \\
Group size $n$          & 4 rollouts per prompt \\
Total epochs            & 10 (500 gradient steps) \\
Learning rate           & $1\times10^{-6}$ \\
KL loss coefficient     & $1\times10^{-4}$ (\texttt{low\_var\_kl}) \\
Entropy coefficient     & $1\times10^{-3}$ \\
Temperature             & 1.0 \\
Max prompt length       & 2048 (non-skilled) / 3072 (static-skilled) \\
Max response length     & 4096 tokens \\
Max observation length  & 512 tokens \\
Max turns               & 25 \\
GPU memory utilization  & 0.3 \\
Observation masking     & True (observation tokens excluded from loss) \\
Hardware                & $1\times$A100 80\,GB \\
\bottomrule
\end{tabular}
\end{table}

\paragraph{Prompt variants.}
We evaluate two prompt variants. The \emph{non-skilled} variant uses the level's
full system prompt (Appendix~\ref{app:systemprompts}) without any skill block;
the \emph{static-skilled} variant pre-pends the same skill bank \textsc{HExA}
ends with after its final training pass, baked in at data-generation time so the
GRPO trainer sees a fixed augmented prompt across all rollouts. The
static-skilled variant therefore tests whether the skill bank is also useful as a
frozen prefix to gradient-based RL.

\paragraph{Results.}
Table~\ref{tab:grpo_val} reports greedy val success rate (\texttt{mean@1}, 50 val
seeds) at each epoch boundary. Figure~\ref{fig:grpo_comparison_main} visualises the
same data on a log-scaled $x$ axis to emphasise the low-data region.

\begin{table}[h]
\centering
\small
\caption{GRPO val success rate (greedy \texttt{mean@1}, 50 val seeds) per epoch.
Epoch 0 is the pre-training baseline (no gradient updates). \emph{ns} is
non-skilled and \emph{sk} static-skilled.}
\label{tab:grpo_val}
\begin{tabular}{lcccc}
\toprule
\textbf{Epoch} & \textbf{DtE (ns)} & \textbf{DtE (sk)} & \textbf{TBP (ns)} & \textbf{TBP (sk)} \\
\midrule
0  & \phantom{0}6\%  & 10\%  & \phantom{0}4\% & \phantom{00}4\% \\
1  & 20\%            & \phantom{0}6\%  & \phantom{0}6\% & \phantom{00}0\% \\
2  & 16\%            & 14\%  & 22\%           & \phantom{0}12\% \\
3  & 18\%            & \phantom{0}4\%  & 20\%           & \phantom{0}22\% \\
4  & 54\%            & 38\%  & 26\%           & 100\% \\
5  & 100\%           & 44\%  & 22\%           & 100\% \\
6  & 100\%           & 100\% & 50\%           & 100\% \\
7  & 100\%           & 100\% & 56\%           & 100\% \\
8  & 100\%           & 86\%  & 48\%           & 100\% \\
9  & 100\%           & 100\% & 50\%           & 100\% \\
10 & 100\%           & 100\% & 40\%           & 100\% \\
\midrule
\textsc{HExA}  & \multicolumn{2}{c}{23.3\%} & \multicolumn{2}{c}{12\%} \\
\bottomrule
\end{tabular}
\end{table}

\paragraph{Compute matching.}
The figure in Section~\ref{app:results:grpo} marks the matched-sample-budget point at
50 unique seeds (one epoch). At this point, GRPO has performed 50 gradient
updates and produced $50 \times 4 = 200$ rollouts; \textsc{HExA} has
generated $51$ trajectories total (5 warm-start + $17 \times 3$
training). We compare on \emph{unique seeds seen}, not on rollouts or
gradient steps, since counting rollouts or gradient steps would
artificially advantage the gradient-based method.

\subsection{Training and Evaluation Protocol}
\label{app:baselines:seeds}

This subsection states the training and evaluation budget behind every result in
Section~\ref{sec:results}. In every environment the training and evaluation
instances are disjoint, the memory-based baselines build their memory on the same
training instances as \textsc{HExA}, and every arm is scored on the same held-out
instances. At evaluation the bank (or store) is frozen: nothing is distilled,
evolved or written, and no state carries over between held-out instances.
Table~\ref{tab:seed_ranges} summarises the budgets.

\begin{table}[h]
\centering
\small
\caption{%
  Training and evaluation budget per environment. Warm start is the number of
  instances played by the base actor with an empty bank and distilled into the
  initial bank (round~$0$ in the games and the reasoning tasks); training is the
  number of instances seen by the remaining rounds of the evolution loop, as
  rounds $\times$ instances per round; evaluation is the number of held-out
  instances every arm is scored on. Maze counts are for \texttt{gpt-oss-120b}.
  Episodic RAG and MemRL build their memory on the same training instances as
  \textsc{HExA}.
}
\label{tab:seed_ranges}
\setlength{\tabcolsep}{4.5pt}
\begin{tabular}{llll}
\toprule
\textbf{Environment} & \textbf{Warm start} & \textbf{Training} & \textbf{Held-out eval} \\
\midrule
\textsc{InterPhyre} (each level)  & $5$  & $51$ ($17\times3$) & $50$ \\
Mastermind                        & $10$ & $70$ ($7\times10$) & $40$ \\
Maze                              & $10$ & $190$ ($19\times10$) & $243$ \\
Reasoning and planning (each task) & $2$ & $58$ ($29\times2$, $10$ attempts) & $100$ \\
\textsc{DiscoverPhysics} \texttt{oscillator}, \texttt{circle} & $5$ & $15$ ($5\times3$) & $15$ \\
\midrule
\multicolumn{4}{l}{\emph{Transfer (no target-task trajectories)}}\\
\textsc{InterPhyre} cross-level     & \multicolumn{2}{l}{source-level banks} & $50$ \\
Games cross-task                   & \multicolumn{2}{l}{Wordle $10{+}120$, Minesweeper $10{+}190$} & $40$ / $243$ \\
\textsc{DiscoverPhysics} \texttt{hubble} & $18$ & $24$ ($8$ worlds $\times3$) & $15$ \\
\bottomrule
\end{tabular}
\end{table}

\subsubsection{\textsc{InterPhyre}}
\label{app:baselines:seeds:interphyre}

Every level's seeds are split once, before any run, into three disjoint blocks
that are never mixed:

\begin{enumerate}[leftmargin=*, nosep]
\item \textbf{Warm start, $5$ seeds.} Base-actor trajectories distilled into
  the initial bank $\mathcal{K}_0$ under \emph{Off2On} initialisation. These
  seeds are never scored.
\item \textbf{Training, $51$ seeds.} The evolution loop runs here: $17$ rounds
  of $3$ fresh seeds each. Episodic RAG and MemRL build their stores on the same
  $51$ seeds, so every method is given the same budget of experience on the same
  instances.
\item \textbf{Held-out evaluation, $50$ seeds.} These seeds,
  produce the reported solve rates for \textsc{Zero-shot}, \textsc{ReAct},
  Reflexion ($K{=}2$), Episodic RAG, MemRL and \textsc{HExA}. No method has seen
  any of them during training.
\end{enumerate}

Seeds are passed directly to the simulator's level generator and determine the
randomised positions, masses and dimensions of the level's objects
(Appendix~\ref{app:InterPhyre}), so the three blocks are disjoint sets of
\emph{instances}, not merely disjoint indices.

\paragraph{The round-by-round loop is training, not evaluation.}
During training, round $n$ attempts its $3$ seeds with the bank
$\mathcal{K}_{n-1}$ fixed, and only afterwards is the evolver allowed to update
the bank from those trajectories; no attempt is ever made with a bank that has
been distilled on the seed being attempted. Those per-round solve rates are a
\emph{training} diagnostic --- they are what Figure~\ref{fig:build_then_freeze}
plots are computed from them. At
evaluation the bank is frozen at the end of round $17$.

\paragraph{Two protocols, kept separate.}
Table~\ref{tab:variants_compact} (initialisation and update rule) and the
reward-free rows of Table~\ref{tab:reward_ablation_main} report the
\emph{training-time} round-by-round rate on the training seeds rather than a
frozen held-out rate, because their purpose is to compare how fast different
update rules accumulate skill. Their numbers are paired across configurations
but are \emph{not} comparable to the held-out rates in
Table~\ref{tab:catapult_cost} --- which is why, for instance, Off2On Evolving reads
$84\%$ there and $82.7\%$ as a frozen bank on the held-out pool. Every other
\textsc{InterPhyre} number in the paper is a frozen bank on the $50$ held-out
seeds.

\paragraph{Cross-level transfer.}
The target bank is synthesised from the \textsc{HExA} banks of the source levels,
each evolved on its own level's warm-start and training seeds, plus a textual
description of the target level. It is evaluated on the target level's $50$
held-out seeds, the same seeds as that level's \textsc{ReAct} baseline. The
synthesis never sees a target-level trajectory, so the comparison is zero-shot
at the trajectory level.

\paragraph{GRPO.}
GRPO updates weights and has no bank to freeze, so it uses its own split of
$50$ training and $50$ validation seeds per level
(Appendix~\ref{app:baselines:grpo}).

\subsubsection{Interactive Games}
\label{app:baselines:seeds:games}

An instance is one hidden task played for $3$ episodes with a cumulative chat
(Appendix~\ref{app:external_envs:games}). One memory record is one three-episode
instance, retrieved once at instance start.

\paragraph{Mastermind.}
The $120$ possible codes are partitioned into $80$ training and $40$ held-out
instances. \textsc{HExA} warm-starts on $10$ of the training instances
(round~$0$, played by the base actor with an empty bank) and trains on the
remaining $70$ in $7$ rounds of $10$; Episodic RAG and MemRL build on the same
$80$ instances in batches of $10$. Every arm, including the no-memory arm and
Reflexion, is scored on the same $40$ held-out instances with the bank or store
frozen.

\paragraph{Maze.}
Training and evaluation layouts are drawn independently. \textsc{HExA} uses
$200$ training instances. For \texttt{gpt-oss-120b} it warm-starts on $10$
(round~$0$) and trains on the remaining $190$ in $19$ rounds of $10$; for
Qwen3-8B it warm-starts on $20$ and trains on $180$ in $9$ rounds of $20$.
Episodic RAG and MemRL build on the same pool after removing duplicate layouts
($193$ instances). Every
arm is scored on the same $243$ held-out layouts.

\paragraph{Cross-task transfer.}
The source banks are \textsc{HExA} banks trained on the source games: Wordle on
$130$ instances ($13$ rounds of $10$) and Minesweeper on $200$ instances ($20$
rounds of $10$). A target bank is synthesised from one or both source banks plus
a textual description of the target game, with no target-game trajectories, and
is scored on the target's held-out pool: the same $40$ Mastermind instances or
$243$ Maze layouts as the within-task arms.

\subsubsection{Reasoning and Planning Tasks}
\label{app:baselines:seeds:reasoning}

Each of MathGAP, ZebraLogic, Tower of Hanoi and Matchstick uses $60$ training
items and $100$ held-out items (Appendix~\ref{app:external_envs:reasoning}).
\textsc{HExA} runs $30$ rounds of $2$ problems with $10$ attempts each, i.e.\
$600$ training rollouts in which every training item is visited once. Round~$0$
is the warm start: its $2$ problems are attempted by the base actor with an
empty bank, and the remaining $29$ rounds train on the other $58$ items; a further $300$ no-memory calls ($5$ per training item) estimate the
per-item baseline, for a total budget of $900$ rollouts. Episodic RAG and MemRL
build on the same $60$ items with the same $900$-rollout budget. Every arm is
scored on the same $100$ held-out items with the bank or store frozen. We run no
transfer experiments on these tasks.

\subsubsection{\textsc{DiscoverPhysics}}
\label{app:baselines:seeds:discoverphysics}

A seed sets only the observation noise; the hidden law of a world is fixed
(Appendix~\ref{app:external_envs:discoverphysics}). Within a world the held-out
seeds are therefore new noise realisations of the law the bank was trained on,
while \texttt{hubble} is a world the bank has never seen.

\paragraph{Within-world (\texttt{oscillator}, \texttt{circle}).}
The initial bank is distilled from $5$ warm-start episodes of the base actor.
\textsc{HExA} then trains for $5$ rounds of $3$ episodes ($15$ episodes), and the
bank is frozen after round $5$. Episodic RAG and MemRL build on the same $5$
warm-start and $15$ training episodes. Every arm is scored on the same $15$
held-out seeds per world.

\paragraph{Cross-world (\texttt{hubble}).}
\texttt{hubble} is held out. The bank is initialised
from $18$ warm-start episodes pooled across five other worlds and grown online
on $8$ other public worlds with $3$ episodes each ($24$ episodes). Episodic RAG
and MemRL build on the same $18$ warm-start and $24$ training episodes. Every
arm is scored on the same $15$ held-out \texttt{hubble} seeds, and no
\texttt{hubble} trajectory is seen during training.

\section{Details of Experiments and Results}
\label{app:results}
\label{app:exp_res_final}

This appendix holds the full results behind Section~\ref{sec:results}: the
per-level and per-task tables that the main-text figures summarise, together
with per-episode metrics, iteration counts, injected-token budgets and
significance tests.

\begin{table}[!htbp]
\centering
\small
\caption{%
  \textbf{A distilled bank beats storing the experience itself, at a third of
  the context.} We report Claude Sonnet 4.6 results on the hardest levels of \textsc{Interphyre}. We build the frozen bank using HExA training loop and evaluate on  same 50 held-out seeds. \emph{Success}: solve rate (\%), mean
  $\pm$ one standard deviation over runs where more than one exists.
  \emph{Avg iters}: mean turns per seed. \emph{Tokens/ep.}: mean memory block
  injected into the actor's prompt per held-out episode (\texttt{cl100k\_base}),
  excluding the task prompt and transcript. HExA improves across first two metrics and uses substantially less tokens compared to other baselines.
}
\label{tab:catapult_cost}
\setlength{\tabcolsep}{6pt}
\begin{tabular}{lccr}
\toprule
\textbf{Method} & \textbf{Success (\%)} & \textbf{Avg iters} & \textbf{Tokens/ep.} \\
\midrule
\multicolumn{4}{l}{\emph{\texttt{catapult}} --- 50 held-out seeds}\\
\midrule
\multicolumn{4}{l}{\emph{No persistent memory}}\\
\textsc{Zero-Shot}                & \phantom{0}0.0 & \phantom{0}1.0 & 0 \\
\textsc{ReAct}                    & $11.0{\pm}3.0$ & $23.1{\pm}0.2$ & 0 \\
Reflexion                         & $21.3{\pm}2.5$ & $20.4{\pm}0.8$ & \phantom{0,}287 \\
\multicolumn{4}{l}{\emph{External memory}}\\
MemRL                             & $49.0{\pm}1.0$ & $13.3{\pm}0.1$ & 10{,}149 \\
Episodic RAG                      & $62.9{\pm}3.7$ & $12.8{\pm}0.2$ & 6{,}513 \\
\textbf{\textsc{HExA}}            & $\mathbf{82.7{\pm}9.3}$ & $\mathbf{12.5{\pm}0.9}$ & \textbf{2{,}392} \\
\midrule
\multicolumn{4}{l}{\emph{\texttt{pass\_the\_parcel}} --- 50 held-out seeds }\\
\midrule
\multicolumn{4}{l}{\emph{No persistent memory}}\\
\textsc{Zero-Shot}                & \phantom{0}0.0 & 1 & 0 \\
\textsc{ReAct}                    & $20.0{\pm}5.7$ & $18.7{\pm}1.3$ & 0 \\
Reflexion                         & $32.0{\pm}6.6$ & $20.1{\pm}0.8$ & \phantom{0,}263 \\
\multicolumn{4}{l}{\emph{External memory}}\\
MemRL                             & $46.0{\pm}7.0$ & $16.8{\pm}1.3$ & 10{,}023 \\
Episodic RAG                      & $60.0{\pm}6.9$ & $15.0{\pm}1.4$ & \phantom{0,}5{,}735 \\
\textbf{\textsc{HExA}}            & $\mathbf{96.0{\pm}2.8}$ & $\mathbf{5.8{\pm}0.8}$ & \textbf{\phantom{0,}2{,}631} \\
\bottomrule
\end{tabular}
\\[3pt]
\end{table}

\begin{table}[t]
\centering
\small
\caption{%
  \textbf{A distilled skill bank is the only memory that helps a frozen
  open-weight actor on both \textsc{InterPhyre} levels, at both model sizes.}
  Actor and evolver are the same model in every row. \textsc{DtE} is
  \texttt{down\_to\_earth} and \textsc{TBP} is \texttt{two\_body\_problem}. Every
  arm builds its memory on 50 training seeds, freezes it, and is scored on a
  disjoint pool of 50 held-out seeds it never saw; no distillation happens during
  evaluation. \textsc{ReAct} is the identical harness with an empty bank, so the
  only variable is what the retriever places in the prompt. Reflexion keeps no
  persistent store, carrying only a verbal lesson across the attempts on a seed;
  Episodic RAG retrieves past rollouts verbatim; MemRL maintains
  (intent, experience, utility) triplets. \emph{Success} is the percentage of
  held-out seeds solved, $\pm$ one standard deviation over $n=3$ runs.
  Bold marks the best value in each column.
}
\label{tab:interphyre_qwen_full}
\setlength{\tabcolsep}{6pt}
\begin{tabular}{lcccc}
\toprule
& \multicolumn{2}{c}{\textbf{Qwen2.5-7B-Instruct}} & \multicolumn{2}{c}{\textbf{Qwen2.5-3B-Instruct}} \\
\cmidrule(lr){2-3}\cmidrule(lr){4-5}
\textbf{Method} & \textbf{DtE} & \textbf{TBP} & \textbf{DtE} & \textbf{TBP} \\
\midrule
\multicolumn{5}{l}{\emph{No persistent memory}}\\
\textsc{ReAct}                   & 63.3 {\scriptsize$\pm$3.4}  & 12.0 {\scriptsize$\pm$0.0} & \phantom{0}8.3 {\scriptsize$\pm$0.9}                        & \phantom{0}7.3 {\scriptsize$\pm$0.9} \\
Reflexion ($K{=}2$)              & 76.0 {\scriptsize$\pm$0.0}  & 27.9 {\scriptsize$\pm$0.1} & \phantom{0}2.0 {\scriptsize$\pm$0.0} & \phantom{0}8.0 {\scriptsize$\pm$0.0} \\
\midrule
\multicolumn{5}{l}{\emph{External memory}}\\
Episodic RAG                     & 42.7 {\scriptsize$\pm$11.4} & 18.7 {\scriptsize$\pm$7.6} & \phantom{0}4.0 {\scriptsize$\pm$4.0} & \phantom{0}5.3 {\scriptsize$\pm$3.1} \\
MemRL (numeric key)              & 36.0 {\scriptsize$\pm$6.0}  & 23.3 {\scriptsize$\pm$4.2} & \phantom{0}3.3 {\scriptsize$\pm$3.1} & \phantom{0}6.0 {\scriptsize$\pm$2.0} \\
\midrule
\textbf{\textsc{HExA}}           & \textbf{82.7} {\scriptsize$\pm$5.2} & \textbf{50.0} {\scriptsize$\pm$3.3} & \textbf{\phantom{0}23.3} {\scriptsize$\pm$0.9} & \textbf{12.0} {\scriptsize$\pm$0.0} \\
\bottomrule
\end{tabular}
\end{table}

\subsection{\textsc{InterPhyre}}
\label{app:results:interphyre}

Table~\ref{tab:catapult_cost} gives the held-out solve rate and cost accounting
(iterations and injected tokens) of every method on the two hardest levels with
Claude Sonnet, and Table~\ref{tab:interphyre_qwen_full} gives the open-weight
head-to-head solve rate on two levels with each open-weight model, with
per-replicate standard deviations. The \textsc{Zero-shot} base rate on every
level is in the main body (Table~\ref{tab:direct_baseline_results_main}).

\paragraph{Does active experimentation and skill creation help agents solve complex tasks?}
We test whether distilling interaction into an evolving skill bank helps on
levels where a single \textsc{ReAct} trajectory fails. We compare against two
kinds of baseline: methods that accumulate no reusable skills
(\textsc{Zero-shot}, \textsc{ReAct}, Reflexion), and external-memory methods
that retain past episodes rather than distilling them (Episodic RAG, MemRL).
On the two hardest levels with Claude Sonnet, where the base model solves
nothing, \textsc{HExA} is the strongest method by a wide margin, reaching
$82.7\%$ on \texttt{catapult} and $96.0\%$ on \texttt{pass\_the\_parcel}
(Table~\ref{tab:catapult_cost}, Figure~\ref{fig:build_then_freeze}). The
memory-free baselines stay low, and the memory baselines recover only part of
the gain. Figure~\ref{fig:catapult45_simframes} shows a representative
\texttt{catapult} seed where the retrieved skills guide \textsc{HExA} to a
solution in a handful of interactions, while \textsc{ReAct} exhausts its budget.

The same pattern holds for smaller open-weight actors
(Table~\ref{tab:interphyre_qwen_full}). With Qwen-2.5-7B and Qwen-2.5-3B,
\textsc{HExA} is the best method on both levels at both sizes, and on
\texttt{two\_body\_problem} it lifts the 7B model from $12.0\%$ under
\textsc{ReAct} to $50.0\%$. Storing experience alone is not enough: Episodic
RAG and MemRL often fall below memoryless \textsc{ReAct}. Their retrieved
rollouts come from seeds with different geometry, so a small actor tends to
copy placements that do not transfer, and the long verbatim transcripts crowd
out its reasoning about the current scene. Reflexion comes
closest only on \texttt{down\_to\_earth} with the 7B model, where it needs a
second attempt on every seed. The gain therefore comes from distilling prior
interactions into reusable strategies, not from extra retries or from placing
past experience in the prompt. A run with GPT-OSS-120B on \texttt{catapult}
shows the same effect, with \textsc{HExA} lifting it from $0\%$ under
\textsc{ReAct} to $54\%$.

\begin{figure}[t]
  \centering
  \includegraphics[width=\linewidth]{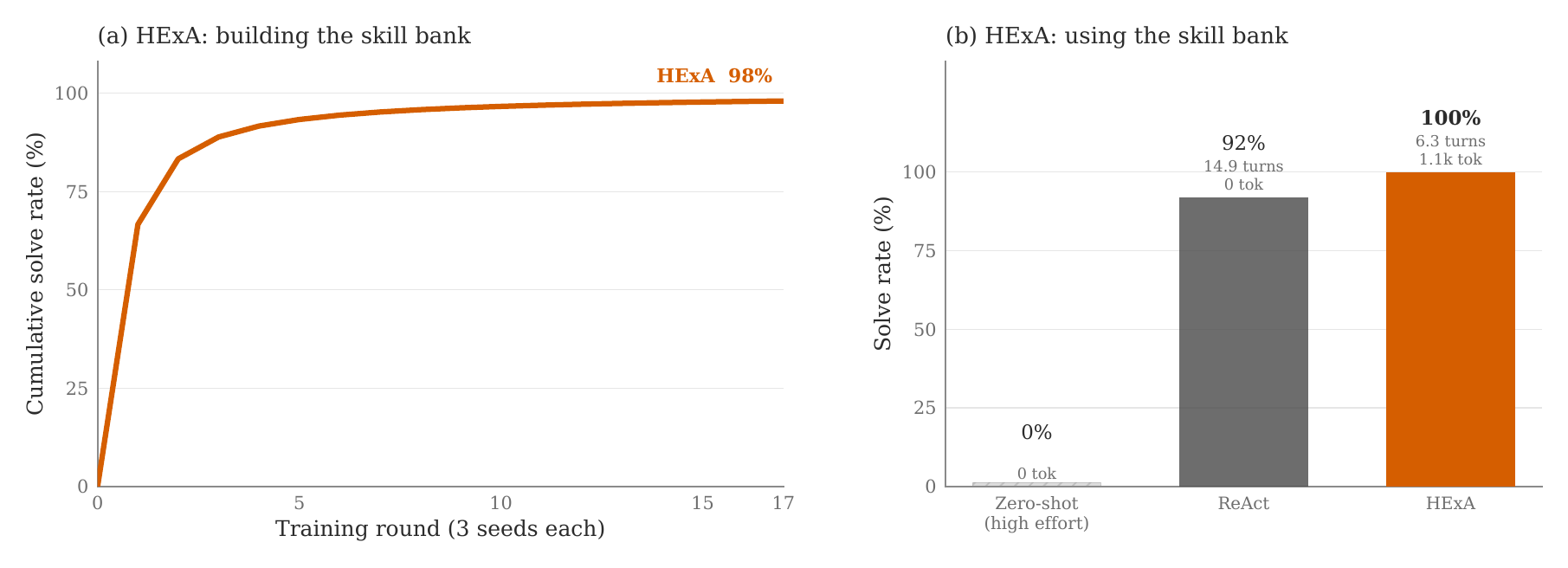}
  \caption{%
    \textbf{GPT-6-Astra on \texttt{catapult}.} \emph{(a)} Cumulative solve rate
    on the training seeds as the bank is built ($17$ rounds of $3$ seeds).
    \emph{(b)} Solve rate on the $50$ held-out seeds with the bank frozen. Mean
    turns per seed and the injected memory per episode are printed above each
    bar. \textsc{Zero-shot} is run at high reasoning effort; the default-effort
    run also scores $0\%$. \textsc{HExA} solves every held-out seed in fewer than
    half the turns of \textsc{ReAct}.}
  \label{fig:app_astra_catapult}
\end{figure}

\begin{figure}[t]
  \centering
  \includegraphics[width=\linewidth]{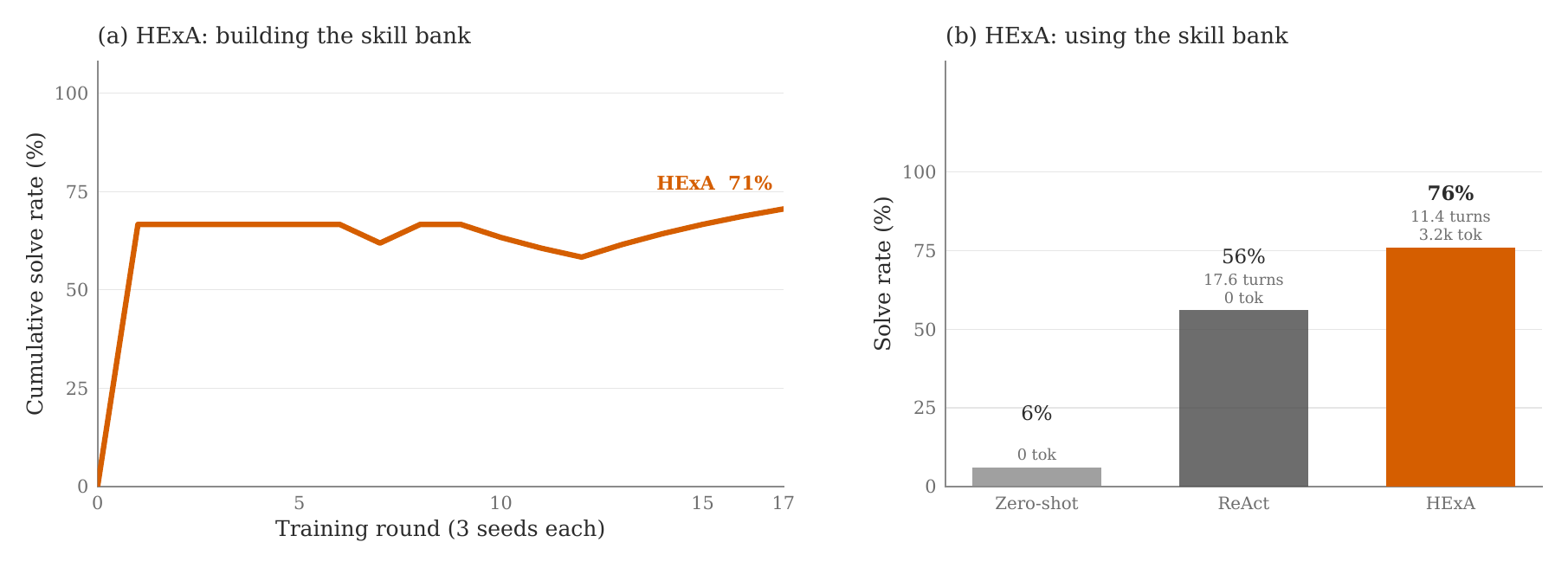}
  \caption{%
    \textbf{GPT-6-Astra on \texttt{pass\_the\_parcel}.} Panels as in
    Figure~\ref{fig:app_astra_catapult}; \textsc{Zero-shot} is run at default
    reasoning effort. \textsc{HExA} raises the held-out solve rate from $56\%$
    under \textsc{ReAct} to $76\%$ while cutting turns per seed from $17.6$ to
    $11.4$.}
  \label{fig:app_astra_ptp}
\end{figure}

\paragraph{A frontier actor.} We repeat the build-then-freeze protocol with
GPT-6-Astra on the two hardest levels
(Figures~\ref{fig:app_astra_catapult} and~\ref{fig:app_astra_ptp}): the bank is
built over $17$ rounds of $3$ training seeds and then frozen for $50$ held-out
seeds. Scale alone does not solve these levels without interaction. On
\texttt{catapult}, \textsc{Zero-shot} Astra scores $0\%$ at both default and
high reasoning effort, and at high effort it runs out of its token budget
mid-reasoning without committing an action on $13$ seeds. On
\texttt{pass\_the\_parcel} it reaches $6\%$. The stronger actor does make better
use of interaction: \textsc{ReAct} reaches $92\%$ and $56\%$, far above Claude
Sonnet's $11.0\%$ and $20.0\%$. \textsc{HExA} still improves on it on both axes.
On \texttt{catapult} it solves every held-out seed ($100\%$) in $6.3$ turns per
seed against $14.9$ for \textsc{ReAct}, injecting $1{,}113$ tokens. On
\texttt{pass\_the\_parcel} it reaches $76\%$ against $56\%$, in $11.4$ turns
against $17.6$, injecting $3{,}190$ tokens. The bank builds faster with the
stronger actor. Its cumulative training solve rate passes $80\%$ by the second
round on \texttt{catapult} and ends at $98\%$. On \texttt{pass\_the\_parcel} it
plateaus near $67\%$ after the first round and ends at $71\%$. So the value of
the distilled bank is not specific to a mid-sized actor. As the actor gets
stronger, the gain shifts from solve rate alone toward interaction cost: at
saturation on \texttt{catapult}, \textsc{HExA} still halves the turns per seed.


\paragraph{Can \textsc{HExA} leverage learned skills for more efficient
exploration and experimentation?}
Spending more compute on interaction does not by itself raise task success, so we
record the average number of turns per seed as an efficiency proxy alongside
solve rate, Table~\ref{tab:catapult_cost}.
Accuracy and cost move together: \textsc{HExA} lowers the average from $23.1 \pm 0.2$
turns under \textsc{ReAct} to $12.5 \pm 0.9$ on \texttt{catapult}, roughly $46\%$
fewer, and from $18.7$ to $5.8$ on \texttt{pass\_the\_parcel}, so the extra
successes come with less interaction per instance rather than more. It is also
the cheapest memory to carry, at $2{,}392$ injected tokens per episode on
\texttt{catapult} against $6{,}513$ for Episodic RAG and $10{,}149$ for MemRL
(Table~\ref{tab:catapult_cost}); the bank is a fixed size, so its cost is
constant in the number of episodes played, whereas storing rollouts scales with
the rollout cap rather than with anything learned. Converting past experience
into a reusable skill bank therefore amortizes the cost of future
experimentation instead of re-paying it each episode.

\paragraph{What does it cost to construct and use the memory?}
\label{app}
We report (Table~\ref{tab:total_cost}) the total token cost used by the memory
baselines, Episodic RAG and MemRL, and \textsc{HExA} for constructing their
memory during the training phase, which includes actor calls and evolver calls
(per round; for MemRL, its summariser per episode), and for evaluation on
held-out seeds using the final memories or skill banks. \textsc{HExA}
constructs its memory at less than half the cost of the memory baselines
($4.9$M tokens on \texttt{catapult} against $10.5$M and $12.3$M), and its $17$
evolver calls are only about $5\%$ of that. Evaluation is also cheaper, so in
total \textsc{HExA} uses $2$--$3.5\times$ fewer tokens than either memory
baseline. \textsc{ReAct} and Reflexion construct no memory, so the fair
comparison with them is the cost of a solved seed. Even with its entire memory
construction included, \textsc{HExA} needs $0.21$M tokens per solved seed on
\texttt{catapult} against $0.32$M for Reflexion and $1.03$M for
\textsc{ReAct} ($0.14$M against $0.19$M and $0.43$M on
\texttt{pass\_the\_parcel}), making it the cheapest method per solved seed on
both levels. \\

\begin{table}[!htbp]
\centering
\small
\caption{%
  \textbf{Total LLM token cost, memory construction included.} Claude Sonnet 4.6 on the
  two hardest \textsc{InterPhyre} levels, in \texttt{cl100k\_base} tokens.
  \emph{Memory construction}: building the skill bank or memory store on the $51$ training seeds.
  \emph{Actor}: prompts and outputs of every actor turn. \emph{Memory calls}:
  LLM calls outside the actor (\textsc{HExA}'s evolver, one per round;
  MemRL's summariser, one per episode). \emph{Eval}: actor tokens over the
  $50$ held-out episodes, including the injected memory (for Reflexion, both
  trials and its reflections). \emph{Per solve}: total divided by the number of
  held-out seeds solved. \textsc{ReAct} and Reflexion construct no memory.}
\label{tab:total_cost}
\setlength{\tabcolsep}{4.5pt}
\begin{tabular}{lrrrrrr}
\toprule
 & \multicolumn{2}{c}{\textbf{Memory construction}} & \textbf{Eval} & & & \\
\cmidrule(lr){2-3}\cmidrule(lr){4-4}
\textbf{Method} & \textbf{Actor} & \textbf{Memory calls} & \textbf{Actor} &
\textbf{Total} & \textbf{Success (\%)} & \textbf{Per solve} \\
\midrule
\multicolumn{7}{l}{\emph{\texttt{catapult}}}\\
\textsc{ReAct}       & $0$       & $0$                       & $5.66$M  & $5.66$M  & $11.0$ & $1.03$M \\
Reflexion            & $0$       & $0$                       & $3.43$M  & $3.43$M  & $21.3$ & $0.32$M \\
MemRL                & $12.14$M  & $0.20$M {\scriptsize(51)} & $10.11$M & $22.45$M & $49.0$ & $0.92$M \\
Episodic RAG         & $10.49$M  & $0$                       & $7.29$M  & $17.78$M & $62.9$ & $0.57$M \\
\textbf{\textsc{HExA}} & $4.61$M & $0.27$M {\scriptsize(17)} & $3.71$M  & $8.58$M  & $\mathbf{82.7}$ & $\mathbf{0.21}$M \\
\midrule
\multicolumn{7}{l}{\emph{\texttt{pass\_the\_parcel}}}\\
\textsc{ReAct}       & $0$       & $0$                       & $4.26$M  & $4.26$M  & $20.0$ & $0.43$M \\
Reflexion            & $0$       & $0$                       & $3.02$M  & $3.02$M  & $32.0$ & $0.19$M \\
MemRL                & $10.72$M  & $0.17$M {\scriptsize(51)} & $12.49$M & $23.38$M & $46.0$ & $1.02$M \\
Episodic RAG         & $8.34$M   & $0$                       & $7.74$M  & $16.08$M & $60.0$ & $0.54$M \\
\textbf{\textsc{HExA}} & $4.52$M & $0.25$M {\scriptsize(17)} & $\mathbf{1.79}$M & $6.56$M & $\mathbf{96.0}$ & $\mathbf{0.14}$M \\
\bottomrule
\end{tabular}
\end{table}

\begin{figure}[t]
\centering
\setlength{\tabcolsep}{1.2pt}
\renewcommand{\arraystretch}{0.85}
{\footnotesize\itshape \textsc{ReAct} (25 iterations, FAILURE)}\\[1pt]
\begin{tabular}{@{}ccccccc@{}}
  \includegraphics[width=0.135\linewidth]{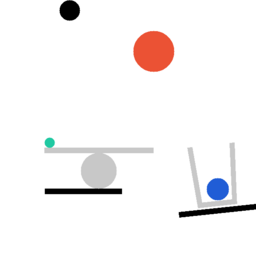} &
  \includegraphics[width=0.135\linewidth]{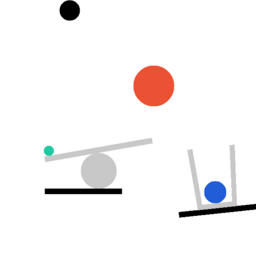} &
  \includegraphics[width=0.135\linewidth]{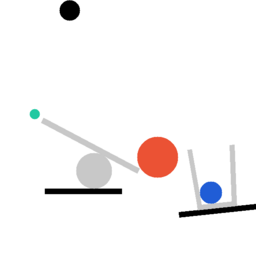} &
  \includegraphics[width=0.135\linewidth]{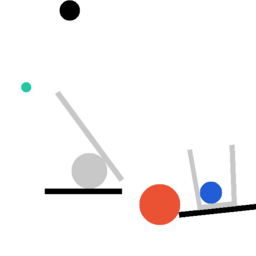} &
  \includegraphics[width=0.135\linewidth]{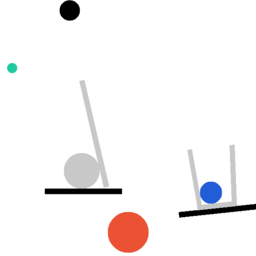} &
  \includegraphics[width=0.135\linewidth]{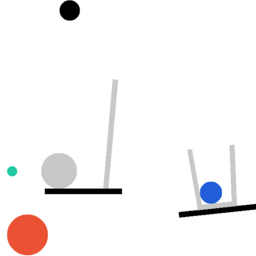} &
  \includegraphics[width=0.135\linewidth]{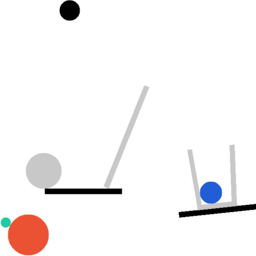} \\
\end{tabular}\\[2pt]
{\footnotesize\itshape \textsc{HExA} (6 iterations, SUCCESS)}\\[1pt]
\begin{tabular}{@{}ccccccc@{}}
  \includegraphics[width=0.135\linewidth]{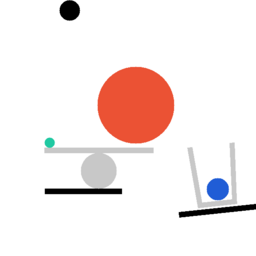} &
  \includegraphics[width=0.135\linewidth]{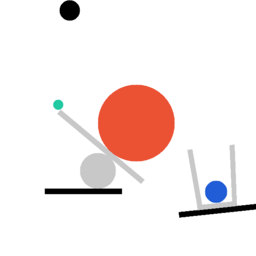} &
  \includegraphics[width=0.135\linewidth]{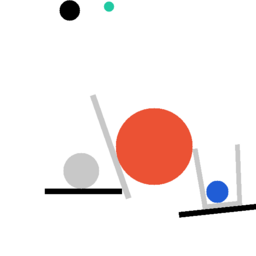} &
  \includegraphics[width=0.135\linewidth]{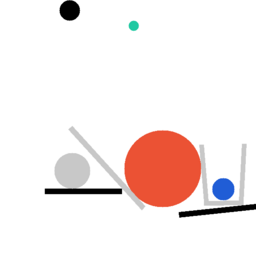} &
  \includegraphics[width=0.135\linewidth]{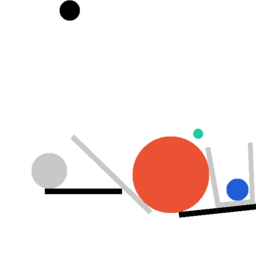} &
  \includegraphics[width=0.135\linewidth]{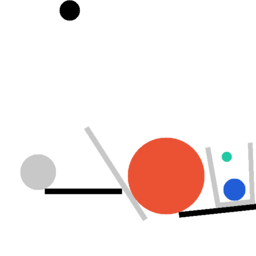} &
  \includegraphics[width=0.135\linewidth]{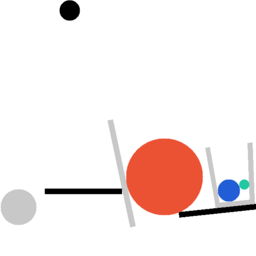} \\
\end{tabular}
\caption{%
\textbf{How the evolved bank guides the agent on a held-out \texttt{catapult}
seed.} Top: \textsc{ReAct}, with no cross-episode memory, exhausts its
25-iteration budget on ineffective placements. Bottom: \textsc{HExA} opens with
the bank's learned default, reads the resulting failure through the bank's
diagnosis rather than as a reason to keep tuning the same parameter, applies the
corrective rule, and succeeds in six interaction iterations. Frames are sampled
from the final simulations; full prompts, bank entries and trajectories are in
Appendix~\ref{app:trajectories:bank}.}
\label{fig:catapult45_simframes}
\end{figure}

\subsection{Instantiation and ablations}
\label{app:results:instantiation}
\label{app:exp_res_final_instantiation}
Section~\ref{sec:regimes} compares the five instantiations of the bank
(initialization $\times$ update rule); Table~\ref{tab:variants_compact} gives
their training solve rate and mean turns per seed on both levels.

\begin{table}[!htbp]
\centering
\small
\setlength{\tabcolsep}{6pt}
\caption{%
  \textbf{Initialisation and update rule.} Claude Sonnet on the two hard
  \textsc{InterPhyre} levels under the round-by-round protocol
  (Appendix~\ref{app:baselines:seeds}),on $50$ training seeds of \texttt{catapult} and \texttt{pass\_the\_parcel}.
  \emph{Solve}: solve rate (\%); \emph{Iters.}: mean turns per seed, with a
  failure counted at the $25$-turn cap. \emph{Evolving} passes the evolver both
  the current bank and the new round's trajectories; \emph{Iterative} rebuilds
  from the current rollouts alone. \emph{Off2On} warm-starts from base-actor
  trajectories; \emph{Online} starts cold; \emph{Offline} keeps its warm-start
  bank frozen. Off2On Evolving, written simply as \textsc{HExA} elsewhere, is
  best on both levels on both metrics. (Numbers are on single runs)
}
\label{tab:variants_compact}
\begin{tabular}{lcccc}
\toprule
& \multicolumn{2}{c}{\textbf{Catapult}} & \multicolumn{2}{c}{\textbf{Pass the Parcel}} \\
\cmidrule(lr){2-3}\cmidrule(lr){4-5}
\textbf{Configuration} & \textbf{Solve} & \textbf{Iters.} & \textbf{Solve} & \textbf{Iters.} \\
\midrule
Off2On Evolving (\textsc{HExA}) & \textbf{84.0} & \textbf{11.8} & \textbf{68.0} & \textbf{12.7} \\
Off2On Iterative                & 56.0          & 15.7          & 48.0          & 17.7          \\
Online Evolving                 & 44.0          & 18.2          & 58.0          & 16.4          \\
Online Iterative                & 36.0          & 20.4          & 56.0          & 18.9          \\
Offline Static                  & 60.0          & 17.0          & 48.0          & 18.5          \\
\bottomrule
\end{tabular}
\end{table}

\paragraph{Reward labels.}
%
\begin{table}[!htbp]
\centering
\scriptsize
\setlength{\tabcolsep}{5pt}
\renewcommand{\arraystretch}{1.0}
\caption{%
  \textbf{Reward labels on the skill bank.} Solve rate (\%) and mean iterations
  per seed, both arms scored on 50 \textbf{training} seeds . The \emph{no
  reward} variant treats every success and failure identically, so the evolver
  decides what to retain and prune from its own prior rather than from
  environment feedback; everything else is held fixed
  The accuracy shows mean $\pm$ one standard deviation over $n{=}3$ independently built banks per arm.
}
\label{tab:reward_ablation_main}
\label{tab:reward_ablation}
\begin{tabular}{llcc}
\toprule
\textbf{Actor / Level} & \textbf{Method} & \textbf{Solve rate (\%)} & \textbf{Iters} \\
\midrule
\multirow{2}{*}{\shortstack[l]{Claude Sonnet\\\texttt{catapult}}}
  & \textsc{HExA}, no reward        & 51.0 {\tiny$\pm$10.1}          & 15.8 \\
  & \textbf{\textsc{HExA}, reward}  & \textbf{78.0} {\tiny$\pm$10.7} & \textbf{13.3} \\
\midrule
\multirow{2}{*}{\shortstack[l]{Qwen-2.5-7B\\\texttt{down\_to\_earth}}}
  & \textsc{HExA}, no reward        & 64.0 {\tiny$\pm$1.7}          & 14.6 \\
  & \textbf{\textsc{HExA}, reward}  & \textbf{72.0} {\tiny$\pm$5.2} & \textbf{12.6} \\
\midrule
\multirow{2}{*}{\shortstack[l]{Qwen-2.5-7B\\\texttt{two\_body\_problem}}}
  & \textsc{HExA}, no reward        & 20.0 {\tiny$\pm$2.7}         & 22.2 \\
  & \textbf{\textsc{HExA}, reward}  & \textbf{42.0} {\tiny$\pm$3.3} & \textbf{17.7} \\
\bottomrule
\end{tabular}
\end{table}

\textsc{HExA} can ingest additional experimental evidence such as trajectory
rewards, which are a function of outcome reward and efficiency
(Appendix~\ref{sec:reward}), to further guide skill evolution and retrieval. The
skill reward label $r_k$ is in turn a function of the trajectory rewards of the
trajectories that invoked the skill,
\begin{equation}
\label{eq:weight}
r_k = \operatorname{clamp}\!\left(\frac{\bar{r}_{\mathrm{src}}+1}{2},\,0.1,\,1.0\right),
\end{equation}
where $\bar{r}_{\mathrm{src}}$ is the mean reward of those source trajectories.
Users can define any trajectory and skill reward; \textsc{HExA} as a framework
is agnostic of these choices. We also run skill evolution without rewards, where
every success and failure trajectory, and every skill distilled out of them, is
treated equally, so the LLM evolver updates the bank from its prior knowledge
rather than from environment feedback. Removing the labels costs accuracy on
every level we test (Figure~\ref{fig:reward_ablation},
Table~\ref{tab:reward_ablation_main}): on \texttt{catapult} the reward-free
variant drops Claude Sonnet from $78.0 \pm 10.7\%$ to $51.0 \pm 10.1\%$, and on
Qwen-2.5-7B it reaches $64.0 \pm 1.7\%$ versus $72.0 \pm 5.2\%$ with rewards on
\texttt{down\_to\_earth} and $20.0 \pm 2.7\%$ versus $42.0 \pm 3.3\%$ on
\texttt{two\_body\_problem}. It also costs efficiency: the reward-free bank
needs $15.8$ iterations per seed on \texttt{catapult} against $13.3$ with
labels, $14.6$ against $12.6$ on \texttt{down\_to\_earth} and $22.2$ against
$17.7$ on \texttt{two\_body\_problem}, so the labelled bank solves more seeds
with fewer interactions. The reward labels thus provide environment-grounded
feedback from the actor's own exploration that helps assign credit for effective
and efficient skill acquisition, and better guides the evolver in composing,
retaining or pruning skills. Our default setting for \textsc{HExA} is therefore
Off2On~+~Evolution with reward-labelled skills.

\subsubsection{Frozen-bank content and ranking ablations}
\label{app:bank-content}

\paragraph{Setup.} Because HExA's skill bank is injected as a textual prefix,
any improvement could in principle be attributed to prompt length, in-context
formatting, or generic priming toward physics reasoning rather than to the
specific knowledge the evolver distilled. We isolate this with frozen banks---no
evolution during evaluation---under four content conditions and four ranking
conditions (Figure~\ref{fig:bank_ablation_full}; the numbers are those of
Table~\ref{tab:ablations} below).

\begin{figure}[H]
  \centering
  \includegraphics[width=1\linewidth]{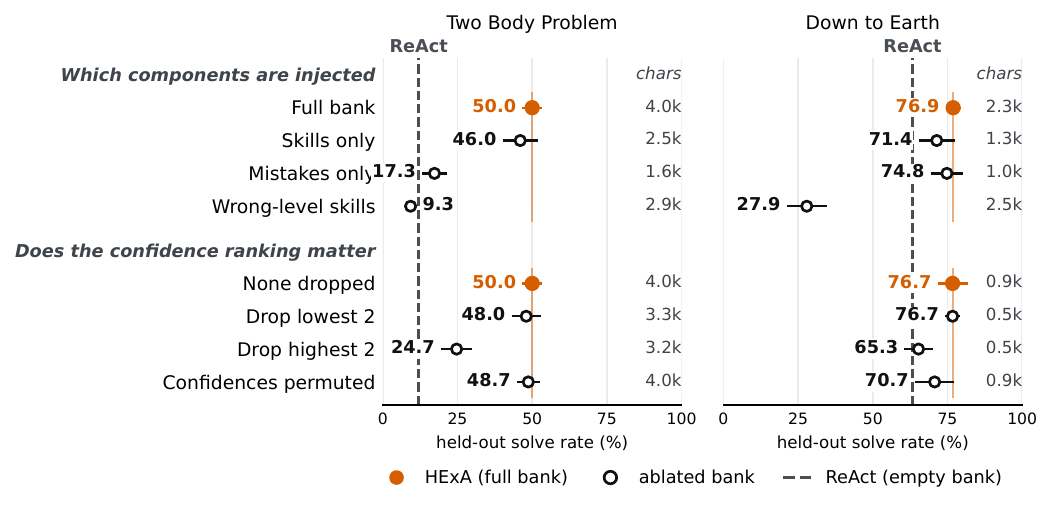}
  \caption{%
    \textbf{Frozen-bank content and ranking ablations, full results.}
    Held-out solve rate (\%) on 50 unseen seeds of \texttt{two\_body\_problem}
    and \texttt{down\_to\_earth} with Qwen-2.5-7B and a frozen skill bank;
    points are means over 3 independent runs and bars denote $\pm 1$ s.d.
    The dashed line marks \textsc{ReAct} with an empty bank
    ($\mathcal{K} = \varnothing$), the shaded orange region marks performance
    worse than injecting nothing, and the blue band marks $\pm 1$ s.d.\ around
    the full-bank reference. The \emph{injected context} column reports the
    character count of the retrieved prefix, so each condition's accuracy can
    be read against the amount of text it adds.
    \emph{Top:} which components of the bank are injected---full bank,
    strategy skills only, mistake records only, or a length-matched bank
    evolved on a different level.
    \emph{Bottom:} whether the evolver's reward-derived ranking
    carries information---all skills retained, the two lowest-rewards
    skills removed, the two highest removed, or rewards randomly permuted
    while retaining all content.}
  \label{fig:bank_ablation_full}
\end{figure}
%
\begin{table}[t]
\centering
\small
\caption{%
  \textbf{Ablating the bank on \textsc{InterPhyre}}, Qwen-2.5-7B throughout.
  \emph{Success} is the percentage of 50 held-out seeds solved, $\pm$ one
  standard deviation over $n{=}3$ evaluations; \emph{Chars} counts the injected
  block. Within a block every arm retrieves from a \emph{single} frozen bank, so
  no difference can come from a different training run.
  \textbf{Top:} which components are injected --- the wrong-level arm swaps in
  fluent skills distilled from the \emph{other} level and keeps this level's own
  mistakes, at comparable injected length, so only content changes.
  \textbf{Bottom:} whether the reward ranking matters --- the drop arms
  remove disjoint pairs of skills, on \textsc{DtE} at lengths within $0.9\%$ of
  each other, and the permuted arm keeps every skill and only shuffles the
  rewards.The two blocks retrieve from different
  \textsc{DtE} banks, hence its two \emph{None} rows; \textsc{TBP} uses one bank
  throughout. Bold marks the best value per block and level.
}
\label{tab:ablations}
\label{tab:interphyre_ablation}
\setlength{\tabcolsep}{5pt}
\begin{tabular}{@{}lcccc@{}}
\toprule
& \multicolumn{2}{c}{\textbf{Two Body Problem}}
& \multicolumn{2}{c}{\textbf{Down to Earth}} \\
\cmidrule(lr){2-3}\cmidrule(lr){4-5}
\textbf{Injected bank} & \textbf{Chars} & \textbf{Succ.}
                       & \textbf{Chars} & \textbf{Succ.} \\
\midrule
\multicolumn{5}{@{}l}{\emph{Which components are injected}}\\
Full bank             & 4004 & \textbf{50.0} {\tiny$\pm$3.3} & 2335 & \textbf{76.9} {\tiny$\pm$2.5} \\
Skills only           & 2490 & 46.0 {\tiny$\pm$5.9}          & 1345 & 71.4 {\tiny$\pm$6.0} \\
Mistakes only         & 1564 & 17.3 {\tiny$\pm$4.1}          & 1040 & 74.8 {\tiny$\pm$5.4} \\
\emph{Wrong-level} skills & 2862 & \phantom{0}9.3 {\tiny$\pm$1.9} & 2476 & 27.9 {\tiny$\pm$6.7} \\
\midrule
\multicolumn{5}{@{}l}{\emph{Whether the reward ranking matters}}\\
None dropped          & 4004 & \textbf{50.0} {\tiny$\pm$3.3} & \phantom{0}892 & \textbf{76.7} {\tiny$\pm$5.0} \\
Drop \emph{lowest} 2  & 3293 & 48.0 {\tiny$\pm$4.9}          & \phantom{0}543 & \textbf{76.7} {\tiny$\pm$2.5} \\
Drop \emph{highest} 2 & 3193 & 24.7 {\tiny$\pm$5.2}          & \phantom{0}548 & 65.3 {\tiny$\pm$4.7} \\
rewards permuted  & 4004 & 48.7 {\tiny$\pm$3.8}          & \phantom{0}892 & 70.7 {\tiny$\pm$6.6} \\
\midrule
\textsc{ReAct} (empty) & \phantom{000}0 & 12.0 {\tiny$\pm$0.0} & \phantom{000}0 & 63.3 {\tiny$\pm$3.4} \\
\bottomrule
\end{tabular}
\end{table}

\paragraph{Strategy skills and mistake records are not interchangeable.}
Removing either component costs accuracy, but the two behave differently across
levels. Strategy skills alone retain most of the gain on both levels
($46.0\%$ and $71.4\%$). Mistake records alone are nearly sufficient on
\texttt{down\_to\_earth} ($74.8\%$ of a $76.9\%$ ceiling) yet collapse on
\texttt{two\_body\_problem} ($17.3\%$). We read this as a difference in failure
structure: on the easier level most failures are recurring, locally correctable
errors that a record of past mistakes prevents directly, whereas the harder
level requires a positive multi-step strategy---bisection over a contact
parameter, in the worked example of Appendix~\ref{sec:worked-example}---that
descriptions of what went wrong do not supply. This supports maintaining both
skill types rather than a single undifferentiated memory.

\paragraph{The evolver's rewards are informative.}
On \texttt{down\_to\_earth}, removing the two highest-reward skills costs
$11.4$ points ($76.7\% \rightarrow 65.3\%$), whereas removing the two lowest
leaves performance unchanged ($76.7\%$) despite halving the injected context
from $0.9$k to $0.5$k characters: the ranking identifies which entries carry the
task, and the low-ranked tail is close to free to discard. Permuting rewards
while retaining all content costs $6.0$ points on \texttt{down\_to\_earth}
($76.7\% \rightarrow 70.7\%$), indicating that retrieval order matters beyond
mere inclusion, though on \texttt{two\_body\_problem} the same manipulation lies
within run-to-run variance ($50.0\% \rightarrow 48.7\%$). We therefore treat
ranking quality as a secondary rather than primary driver of performance.

\paragraph{Relevance, not volume: the wrong-level control.}
We replace the bank with one evolved on a different \textsc{Interphyre} level,
matched in length and indistinguishable in surface form: fluent, well-structured
physics guidance with the same schema of titles, principles and applicability
conditions. On \texttt{two\_body\_problem} this reduces solve rate from $50.0\%$
to $9.3\%$, below the $12.0\%$ empty-bank \textsc{ReAct} baseline, while
injecting $2.9$k characters against the full bank's $4.0$k; the same ordering
holds on \texttt{down\_to\_earth} ($76.9\% \rightarrow 27.9\%$). Injecting
plausible but misapplied procedural knowledge is therefore actively harmful
relative to injecting nothing, which rules out length and priming explanations:
a longer prompt that is wrong about the task is worse than no prompt at all. We
attribute this to misdirected exploration---the actor commits early to a
strategy whose applicability condition does not in fact hold, and spends its
$25$-turn budget refining the wrong hypothesis rather than forming a new one.
This is also why cross-level transfer (Section~\ref{sec:transfer}) requires the
evolver's de-binding and re-grounding step rather than direct reuse of a source
bank.

\paragraph{Summary.} These ablations locate HExA's gain in task-relevant
distilled content and its reward-derived prioritization, not in context length.
Both skill types are load-bearing, their relative importance tracks the failure
structure of the level, and the bank's value concentrates in a small number of
high-reward strategies.

\subsubsection{Is experimentation needed once the skill bank is built?}
\label{app:skills-no-tools}

\paragraph{Motivation.} \textsc{HExA} interacts with the environment in two
places: while the bank is evolved and built, and again at evaluation, where the actor calls the tools with the frozen bank in context. One possibility is that, once built, the bank is
sufficient on its own, so experimentation matters only for building it. The
other is that the bank guides how the actor experiments, so the actor still
needs experimentation at inference. To ablate this, we separate the two by giving the final skill
bank to an agent for which inference time experimentation is disallowed.

\paragraph{Setup.} We take the final skill bank of Claude Sonnet on catapult and pass\_the\_parcel levels and inject it exactly as in \textsc{HExA}'s
frozen evaluation. The agent receives
the \textsc{Zero-shot} prompt: the task and a single description of the
scene layout (object positions and sizes), with no experimentation tools. Only the skill bank is added. We evaluate it on the same $50$ held-out seeds under three
protocols:
\begin{itemize}
  \item \textbf{Zero-Shot.} One call and one final placement.
  \item \textbf{Extended reasoning.} $25$ turns with zero environment rewards or feedback.
  Turns $1$--$24$ are a free scratchpad for forming hypotheses, working
  through the skills and eliminating options, and the agent commits one
  placement on turn $25$. This matches
  \textsc{HExA}'s $25$-turn budget, but the turns are extended chain of thought
  rather than active experiments or interactions.
  \item \textbf{Binary feedback.} Up to $25$ placements. After each one
  the agent is told only \emph{success}, \emph{failure} or \emph{invalid}
  (overlap at $t{=}0$, which follows from the scene layout). There is no
  trajectory, no probe or trace tool, and no other simulator feedback. This tests
  whether a binary experiment decision is enough, or whether \textsc{HExA}'s richer
  intervention tools are needed.
\end{itemize}

\begin{table}[!htbp]
\centering
\small
\caption{%
  \textbf{The frozen bank without the tool interface.} Claude Sonnet 4.6 on the
  $50$ held-out seeds of the two hardest \textsc{InterPhyre} levels, with the
  final frozen \textsc{HExA} bank of one build per level injected exactly as in
  \textsc{HExA}'s frozen evaluation. In the three middle rows the agent has no simulator
  tools; it is given the scene layout once and the prompt of the
  \textsc{Zero-shot} baseline. \emph{Signal}: what the environment returns to
  the agent before its answer is scored. \emph{Turns}: mean model calls per
  seed (placements for outcome-only feedback, including failed ones). The three
  middle rows are single runs
  over the $50$ seeds. \textsc{Zero-shot} and \textsc{ReAct}
  are the rows of Table~\ref{tab:catapult_cost}; \textsc{HExA} is the frozen
  evaluation of the same bank. (All numbers are on single run)}
\label{tab:skills_no_tools}
\setlength{\tabcolsep}{4pt}
\begin{tabular}{llcccc}
\toprule
 & & & & \multicolumn{2}{c}{\textbf{Success (\%) / Turns}} \\
\cmidrule(lr){5-6}
\textbf{Arm} & \textbf{Signal} & \textbf{Skills} & \textbf{Tools} & \texttt{catapult} & \texttt{pass\_the\_parcel} \\
\midrule
\textsc{Zero-shot}                & none              & --         & --         & \phantom{0}$0.0$ / $1$  & \phantom{0}$0.0$ / $1$ \\
\textsc{ReAct}                    & full simulator    & --         & \checkmark & $11.0$ / $23.1$ & $20.0$ / $18.7$ \\
\midrule
Zero-Shot (Skill Bank)                       & none              & \checkmark & --         & $16.0$ / $1$    & $22.0$ / $1$ \\
Extended reasoning (25 turns)     & none              & \checkmark & --         & $42.0$ / $25$   & $48.0$ / $25$ \\
Binary feedback (25 turns)  & success / failure & \checkmark & --         & $90.0$ / $8.0$  & $90.0$ / $5.1$ \\
\midrule
\textsc{HExA} (frozen, same bank) & full simulator    & \checkmark & \checkmark & $86.0$ / $12.7$ & $96.0$ / $5.8$ \\
\bottomrule
\end{tabular}
\end{table}

\begin{figure}[t]
  \centering
  \includegraphics[width=\linewidth]{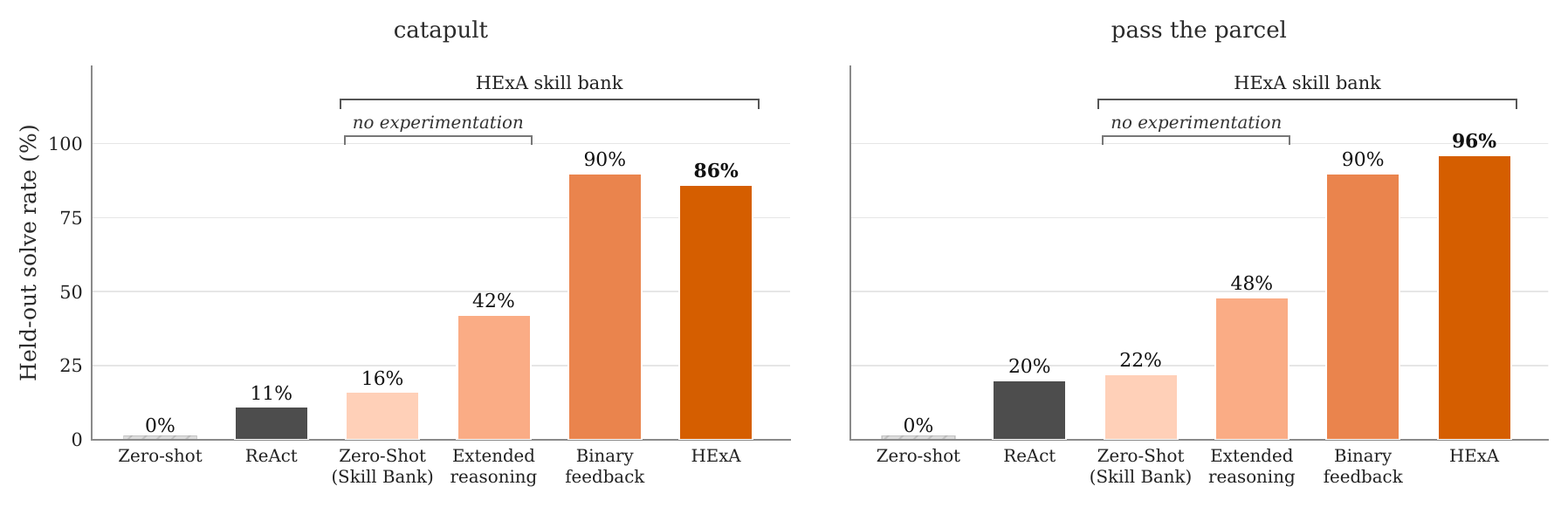}
  \caption{%
    \textbf{The \textsc{HExA} skill bank with and without
    experimentation tools.} Claude Sonnet 4.6, solve rate on the $50$ held-out
    seeds of each level. Greys use no skills: \textsc{Zero-shot} (no
    interaction) and \textsc{ReAct}. Oranges all use the same
    frozen \textsc{HExA} skill bank, with lighter shades receiving less
    environment feedback at evaluation. \emph{Zero-Shot (Skill Bank)}: one call,
    one placement. \emph{Extended reasoning}: $24$ scratchpad turns without
    feedback, then one placement. \emph{Binary feedback}: up to $25$
    placements, each answered only with success or failure. \textsc{HExA}: the
    full simulator. Numbers as in Table~\ref{tab:skills_no_tools}.}
  \label{fig:skills_no_tools}
\end{figure}

\paragraph{Distilled skills improve the actor's physical reasoning.} Without any
environment feedback the skill bank lifts the actor well above its baselines
(Table~\ref{tab:skills_no_tools}). Zero-Shot reaches $16.0\%$ on
\texttt{catapult} and $22.0\%$ on \texttt{pass\_the\_parcel}, against $0\%$
for \textsc{Zero-shot}. With the $25$-turn reasoning budget this rises to
$42.0\%$ and $48.0\%$, above \textsc{ReAct} with the full simulator ($11.0\%$
and $20.0\%$). Because every seed is held out and has a new layout, the gain
cannot come from recalling past placements. The skills supply transferable
physical understanding of the level: how the mechanism behaves (a drop right of
the catapult pivot launches the green ball rightward), which parameters matter,
and which errors to avoid. The actor applies this to a new scene and reasons its
way to a correct placement on roughly half the seeds without touching the
simulator. The gain also grows with the reasoning budget ($16\%\rightarrow42\%$
and $22\%\rightarrow48\%$), which is what one expects if the skills act as a
scaffold for reasoning rather than as a lookup.

\paragraph{Experimentation at inference remains necessary.} Reasoning alone
stays $44$ and $48$ points below \textsc{HExA} with the same skill bank ($86\%$ and
$96\%$). The only difference between extended reasoning and \textsc{HExA} is experimentation and environment
feedback. Adding even binary success or failure feedback of the experiments after each placement, the actor reaches $90.0\%$ on both levels, in
$8.0$ and $5.1$ placements per seed. With extended reasoning, the actor typically settles on its answer within the first half
of the budget and restates it for the remaining turns, because no new
information arrives.

This ablation thus clearly shows that Experimentation matters in both phases. While the bank is
built, it produces transferable skills that sharpen the actor's reasoning about
the level: these alone take the actor from $0\%$ to $42$--$48\%$. At inference,
it lets the actor carry out the search the skills describe, and even
outcome-only feedback is enough for this (about $90\%$). Neither ingredient suffices alone. Interaction without skills
(\textsc{ReAct}) reaches $11$--$20\%$, and skills without interaction reach
$42$--$48\%$. Together they reach $86$--$96\%$.
\subsection{Other benchmarks and tasks}
\label{app:results:external}

We repeat the same frozen-bank protocol on three external benchmark families:
interactive games, reasoning and planning, and open-ended scientific discovery,
against the same non-experimental and external-memory baselines.
Figure~\ref{fig:external_bench} summarizes each family as one score per actor,
averaged over that family's environments; Table~\ref{tab:external} reports the
per-task values it averages, and the three subsections below give token budgets,
per-episode breakdowns and exact paired tests for each family.

Across all four actor groups, \textsc{HExA} is the best-performing method in 9 of 10 task-actor settings; on Maze with gpt-oss-120b, MemRL is marginally higher (0.666 vs. 0.649) at about 8× the injected context.
\textsc{Episodic RAG} falls below \textsc{Zero-shot} on the interactive games
with \texttt{gpt-oss-120b}, while \textsc{MemRL} falls below it on
\textsc{DiscoverPhysics}. Thus, each episodic-memory method imposes a cost in at
least one family, whereas the distilled bank improves every reported actor group.

\begin{table}[!b]
\centering
\scriptsize
\captionsetup{skip=3pt}
\caption{%
  \textbf{The bank lifts a frozen actor on all three external families.} One row
  per task/actor cell, on the metric each block states. Every bank is frozen
  before evaluation and scored on instances it never saw, under each family's own
  split; values are the mean over $n{=}3$ replicates. On the games and \textsc{DiscoverPhysics} all arms are paired on
  identical instances. Bold marks the best value per row. Significance
  tests, token budgets and per-episode breakdowns are in
  Appendices~\ref{app:results:games}--\ref{app:results:discovery}.
}
\label{tab:external}
\vspace{-2pt}
\setlength{\tabcolsep}{4pt}
\renewcommand{\arraystretch}{0.88}
\begin{tabular}{@{}llcccc@{}}
\toprule
\textbf{Task} & \textbf{Actor} & \textbf{Zero-Shot} & \textbf{Episodic RAG} & \textbf{MemRL} & \textbf{\textsc{HExA}} \\
\midrule
\multicolumn{6}{@{}l}{\emph{Interactive games}--- success rate on the metric each block reports}\\
Mastermind (\texttt{any}) & \texttt{gpt-oss-120b} & .683 & .650 & .690 & \textbf{.883} \\
Mastermind (\texttt{any}) & Qwen3-8B              & .367 & .492 & .392 & \textbf{.640} \\
Maze (\texttt{any})       & \texttt{gpt-oss-120b} & .607 & .601 & \textbf{.667} & .649 \\
Maze (\texttt{any})       & Qwen3-8B              & .479 & .483 & .513 & \textbf{.564} \\
\midrule
\multicolumn{6}{@{}l}{\emph{Reasoning \& planning} --- accuracy on 100 held-out items}\\
MathGAP         & \texttt{gpt-oss-120b} & 0.510 & 0.560 & 0.600 & \textbf{0.863} \\
ZebraLogic      & \texttt{gpt-oss-120b} & 0.560 & 0.720 & 0.650 & \textbf{0.807} \\
Tower of Hanoi  & \texttt{gpt-oss-120b} & 0.263 & 0.377 & 0.273 & \textbf{0.520} \\
Matchstick      & \texttt{gpt-oss-120b} & 0.737 & 0.767 & 0.703 & \textbf{0.787} \\
\midrule
\multicolumn{6}{@{}l}{\emph{\textsc{DiscoverPhysics}} --- explanation score over 15 test seeds}\\
\texttt{oscillator} & \texttt{claude-sonnet-5} & 0.516 & 0.822 & 0.780 & \textbf{0.853} \\
\texttt{circle}     & \texttt{claude-sonnet-5} & 0.580 & 0.580 & 0.569 & \textbf{0.916} \\
\bottomrule
\end{tabular}
\vspace{-4pt}
\end{table}

\subsubsection{Interactive games}
\label{app:results:games}

Table~\ref{tab:orbit_memory} is the full head-to-head on Mastermind and Maze:
four task/actor blocks, each arm playing an identical held-out pool, with
first-episode, last-episode and \texttt{any}-episode success, the injected token
budget, and a two-sided paired sign test on episode-3 outcomes against that
block's \textsc{Zero-shot} arm.

\paragraph{Do distilled skills help when each instance is replayed?}
For the interactive games, each instance is played for three episodes with the
hidden task held fixed. Figure~\ref{fig:external_bench} reports \texttt{any},
the fraction of instances solved in at least one of the three episodes;
Table~\ref{tab:orbit_memory} additionally reports \emph{ep1} and \emph{ep3},
success on the first and final episodes, which measure how the agent's
behavior evolves across repeated attempts on the same instance.

The bank improves both actors and is also the cheapest persistent memory.
On Mastermind with GPT-OSS-120B, \textsc{HExA} injects $971$ tokens per episode
compared with $1{,}957$ for Episodic RAG and $3{,}416$ for \textsc{MemRL}, while
reaching $88.3\%$ \texttt{any}-episode success versus $68.3\%$ for no memory.
With Qwen3-8B the token budgets differ: \textsc{HExA} uses $359$ tokens
versus $1{,}965$ for Episodic RAG and $3{,}192$ for \textsc{MemRL}, and achieves
$64.0\%$ \texttt{any}-episode success against $36.7\%$ baseline.
On Maze, token costs again vary by model: GPT-OSS-120B requires $945$, $4{,}699$,
and $7{,}572$ tokens respectively, with \textsc{HExA} at $65\%$ \texttt{any}-episode
success; Qwen3-8B requires $429$, $5{,}457$, and $7{,}027$ tokens, with
\textsc{HExA} at $56.4\%$ \texttt{any}-episode success against $47.9\%$ baseline.
Thus, the episodic baselines inject between $3.5\times$ and $16\times$ as much
context as the skill bank across all settings, yet consistently underperform.

Storing episodes verbatim does not consistently transfer across instances:
a retrieved transcript from a previous instance can displace the actor's own
cross-episode observations rather than supplement them with a reusable abstraction.
\textsc{Reflexion}, which keeps only verbal lessons with minimal injected context,
shows the lowest persistent memory cost but also the lowest \texttt{any}-episode
success across most settings. In contrast, \textsc{HExA} compresses prior
experience into a much smaller set of strategies that can be reused across
instances while leaving more context available for the current interaction.

\begin{table}[t]
\centering
\scriptsize
\caption{%
  \textbf{\textsc{HExA} gives the largest gains in three of four settings at the lowest memory cost among memory baselines.} Four task/actor settings on
  the two games, each a separate block. Each instance is played for three episodes:
  \emph{ep1} and \emph{ep3} are first- and last-episode success rates and
  \texttt{any} is success in at least one of the three. Every value is the mean
  $\pm$ one standard deviation over three independently built memories per arm.
  \emph{Tokens} is the memory block injected into the actor's prompt each
  episode. Episodic RAG retrieves past rollouts verbatim; MemRL maintains
  (intent, experience, utility) triplets; Reflexion keeps no persistent store,
  carrying only a verbal lesson across episodes, and is the one arm that
  \emph{removes} context by truncating the cross-episode transcript.
  A dash marks an arm that was not run at this block's budget. Bold marks \textsc{HExA} and the best value
  per block.
}
\label{tab:orbit_memory_headtohead}
\label{tab:orbit_memory}
\label{tab:mastermind_qwen}
\label{tab:maze_qwen}
\setlength{\tabcolsep}{3.5pt}
\begin{tabular}{lccccr}
\toprule
& \textbf{Method} & \textbf{ep1} & \textbf{ep3} & \textbf{\texttt{any}} & \textbf{Tokens} \\
\midrule
\multirow{5}{*}{\shortstack{Mastermind\\\texttt{gpt-oss-120b}\\$n{=}40$}}
  & Zero-Shot              & $.100\pm.035$ & $.650\pm.020$ & $.683\pm.031$ & 0 \\
  & \textbf{\textsc{HExA}} & $.058\pm.024$ & $\mathbf{.875\pm.020}$ & $\mathbf{.883\pm.012}$ & \textbf{971} \\
  & Episodic RAG           & $.083\pm.031$ & $.633\pm.085$ & $.650\pm.094$ & 1{,}957 \\
   & MemRL                  & $.070\pm.021$ & $.685\pm.250$ & $.690\pm.244$ & 3{,}416 \\
  & Reflexion  & $\mathbf{.117\pm.047}$ & $.133\pm.051$ & $.283\pm.031$ & 483 \\
\addlinespace
\midrule
\multirow{6}{*}{\shortstack{Mastermind\\\texttt{Qwen3-8B}\\$n{=}40$}}
  & Zero-Shot                          & $.092\pm.012$ & $.283\pm.083$ & $.367\pm.072$ & 0 \\
  & \textbf{\textsc{HExA}}    & $.079\pm.047$ & $\mathbf{.471\pm.080}$ & $\mathbf{.640\pm.082}$ & \textbf{359} \\
  & Episodic RAG                       & $.092\pm.012$ & $.458\pm.043$ & $.492\pm.031$ & 1{,}965 \\
  & MemRL                              & $.017\pm.024$ & $.300\pm.054$ & $.392\pm.043$ & 3{,}192 \\
  & Reflexion  & $\mathbf{.158\pm.012}$ & $.325\pm.041$ & $.433\pm.024$ & 269 \\
\addlinespace
\midrule
\multirow{5}{*}{\shortstack{Maze\\\texttt{gpt-oss-120b}\\$n{=}243$}}
  & Zero-Shot                          & $.244\pm.005$ & $.610\pm.014$ & $.607\pm.016$ & 0 \\
  & \textbf{\textsc{HExA}}              & $.198\pm.040$ & $.604\pm.033$ & $.649\pm.045$ & \textbf{945} \\
  & Episodic RAG                       & $\mathbf{.410\pm.037}$ & $.549\pm.024$ & $.601\pm.020$ & 4{,}699 \\
  & MemRL                              & $.341\pm.022$ & $\mathbf{.632\pm.026}$ & $\mathbf{.667\pm.024}$ & 7{,}572 \\
  & Reflexion  & $.229\pm.028$ & $.308\pm.018$ & $.519\pm.043$ & 358 \\
\addlinespace
\midrule
\multirow{6}{*}{\shortstack{Maze\\\texttt{Qwen3-8B}\\$n{=}243$}}
  & Zero-Shot                          & $.243\pm.038$ & $.472\pm.025$ & $.479\pm.019$ & 0 \\
  & \textbf{\textsc{HExA}}    & $.278\pm.010$ & $\mathbf{.542\pm.011}$ & $\mathbf{.564\pm.010}$ & \textbf{429} \\
  & Episodic RAG                       & $\mathbf{.333\pm.037}$ & $.379\pm.060$ & $.483\pm.067$ & 5{,}457 \\
  & MemRL                              & $.296\pm.033$ & $.425\pm.020$ & $.513\pm.032$ & 7{,}027 \\
  & Reflexion  & $.263\pm.018$ & $.299\pm.032$ & $.465\pm.022$ & 194 \\
\bottomrule
\end{tabular}
\\[3pt]
\end{table}

\subsubsection{Reasoning and planning}
\label{app:results:reasoning}

Table~\ref{tab:reasoning} gives the per-task head-to-head on the four reasoning
and planning tasks. Every arm is our own run on our provider at the same rollout
budget as \textsc{HExA}, so all rows are paired on identical items.

\paragraph{Do distilled skills help on reasoning and planning tasks?}
Every method receives the same rollout budget, so the comparison differs only in
what information is carried between items. \textsc{HExA} improves the frozen actor
on all four tasks, raising mean accuracy from $0.518$ with \textsc{Zero-shot} to
$0.744$, and is the only memory method that improves every task rather than
trading gains on one task against losses on another. Episodic RAG and
\textsc{MemRL} reach mean accuracies of $0.606$ and $0.556$, respectively.
Storing previous rollouts therefore recovers part of the benefit, but not most
of it, despite the equal rollout budget. The additional gain from \textsc{HExA}
comes from making distilled strategies immediately available to later items
through context rather than carrying the original trajectories themselves.

\begin{table}[t]
\centering
\small
\caption{%
  \textbf{\textsc{HExA} improves a frozen model on all four reasoning and
  planning tasks.} Actor and evolver are both \texttt{gpt-oss-120b}. All cells
  are accuracy, the fraction of held-out items answered correctly; where more
  than one run exists the value is the mean over three runs with $\pm$ one
  standard deviation. Every arm is our own run on our provider at the same
  rollout budget, so all rows are paired on identical items and compared by
  exact McNemar on per-item outcomes. Bold marks the best value in each column.
}
\label{tab:reasoning}
\label{tab:core}
\setlength{\tabcolsep}{5pt}
\begin{tabular}{lcccc}
\toprule
\textbf{Method} & \textbf{MathGAP} & \textbf{ZebraLogic} & \textbf{Hanoi} & \textbf{Matchstick} \\
\midrule
Zero-Shot      & 0.510 {\scriptsize$\pm$.006} & 0.560 {\scriptsize$\pm$.026} & 0.263 {\scriptsize$\pm$.007} & 0.737 {\scriptsize$\pm$.023} \\
Episodic RAG   & 0.560{\scriptsize$\pm$.010} & 0.720{\scriptsize$\pm$.030} & 0.377{\scriptsize$\pm$.037} & 0.767{\scriptsize$\pm$.009} \\
MemRL          & 0.600{\scriptsize$\pm$.015}   & 0.650{\scriptsize$\pm$.002}   & 0.273{\scriptsize$\pm$.027}  & 0.703{\scriptsize$\pm$.023}  \\
\textbf{\textsc{HExA}} & \textbf{0.863} {\scriptsize$\pm$.009} & \textbf{0.807} {\scriptsize$\pm$.018} & \textbf{0.520} {\scriptsize$\pm$.010} & \textbf{0.787} {\scriptsize$\pm$.050} \\
\midrule
\emph{McNemar $p$, \textsc{HExA} vs.\ no memory} & \emph{4.7e-23} & \emph{2.8e-14} & \emph{7.4e-12} & \emph{0.0081} \\
\bottomrule
\end{tabular}
\end{table}

\subsubsection{\textsc{DiscoverPhysics}}
\label{app:results:discovery}

Table~\ref{tab:discovery} gives explanation score, benchmark reward and pass-rate
at the 0.90 gate for all three worlds, comparing a frozen skill bank against
\textsc{Episodic RAG}, \textsc{MemRL}, and no memory, all paired on the same 15 seeds.

\paragraph{Do distilled skills help when the task is open-ended scientific discovery?}
On \textsc{DiscoverPhysics}, the skill bank contains experimentation strategies
rather than task solutions. \textsc{HExA} is the only method that improves explanation
on all three worlds. On \texttt{oscillator}, both \textsc{Episodic RAG} ($0.773$)
and \textsc{HExA} ($0.860$) improve over the baseline ($0.493$), but \textsc{HExA}
reaches $11/15$ pass-rate at the strict 0.90 explanation gate versus Episodic RAG's
$6/15$ and MemRL's $7/15$. On \texttt{circle}, neither episodic baseline improves
explanation (Episodic RAG $0.553$, MemRL $0.540$), leaving \textsc{HExA} ($0.887$)
as the sole improvement on the metric. On \texttt{hubble}, cross-world transfer,
\textsc{HExA} achieves $0.840$ explanation while the episodic baselines reach
$0.740$ (Episodic RAG) and $0.753$ (MemRL).

The improvement also appears under the benchmark's strict success criterion.
On \texttt{oscillator}, pass rate at the $0.90$ gate rises from $0/15$ to $11/15$
with \textsc{HExA}, whereas Episodic RAG reaches $6/15$ and MemRL $7/15$.
On \texttt{circle}, the benchmark reward does not improve for any method,
yet the explanation-score gain is the most pronounced across all three worlds.
This is not inconsistent: the evaluator re-optimizes the submitted parameters against
the agent's own trajectories, so an incorrect functional form can still fit the observed
data well. The explanation metric therefore exposes improvements in discovering the
underlying law that the fit-based reward may not fully reflect, and distilled strategies
succeed where verbatim-retrieval and learned-summarisation baselines do not.

\begin{table}[t]
\centering
\scriptsize
\caption{%
  \textbf{\textsc{DiscoverPhysics}: distilled skills vs.\ episodic memory baselines.}
  The agent designs experiments in a world governed by an unknown field equation,
  observes noisy particle trajectories, and submits the governing law as
  executable Python with a prose explanation. Actor and teacher are both
  \texttt{claude-sonnet-5}; every number is scored by \texttt{claude-opus-4-6}, a
  judge independent of the actor and the one the benchmark itself uses. All arms
  are paired on the same 15 held-out seeds per world, evaluated under a protocol
  registered before evaluation. Entries are \textbf{mean $\pm$ s.d.\ over three
  independent evaluations} of that arm (pass counts are means out of 15).
  \emph{Expl.}\ is the judge's explanation score and \emph{Reward} the
  benchmark's fit score, both higher-is-better; \emph{pass@1}$_{.90}$ counts
  episodes with explanation $\geq 0.90$ and normalised MSE $<0.1$.
  \textsc{Episodic RAG} retrieves past rollouts verbatim; \textsc{MemRL}
  maintains (intent, experience, utility) triplets with LLM summarisation;
  \textsc{HExA} uses a frozen distilled bank. \texttt{oscillator} and
  \texttt{circle} are within-world accumulation where the bank encodes that
  world's own physics; \texttt{hubble} is cross-world transfer, grown on nine
  other worlds with \texttt{hubble} unseen --- the two episodic baselines store
  raw target-world rollouts and so have no cross-world form, and are reported
  on the within-world designs only. Bold marks \emph{Expl.}\ and \emph{Reward}
  entries that improve significantly over the same world's baseline.
}
\label{tab:discovery}
\setlength{\tabcolsep}{3pt}
\begin{tabular}{lllcccc}
\toprule
\textbf{World} & \textbf{Design} & \textbf{Method} & \textbf{Expl.} & \textbf{Reward} & \textbf{pass@1}$_{.75}$ & \textbf{pass@1}$_{.90}$ \\
\midrule
\multirow{4}{*}{\texttt{oscillator}} & \multirow{4}{*}{within-world}
  & Zero-Shot & 0.516 $\pm$ 0.044 & $-$0.013 $\pm$ 0.111 & 2.0/15 & 1.7/15 \\
& & \textsc{Episodic RAG} & 0.822 $\pm$ 0.008 & $+$0.811 $\pm$ 0.038 & 11.3/15 & 11.0/15 \\
& & \textsc{MemRL} & 0.780 $\pm$ 0.007 & $+$0.761 $\pm$ 0.08 & 12.0/15 & 11.3/15 \\
& & \textbf{\textsc{HExA}} & \textbf{0.853 $\pm$ 0.018} & \textbf{$+$0.832 $\pm$ 0.048} & 12.3/15 & 11.3/15 \\
\midrule
\multirow{4}{*}{\texttt{circle}} & \multirow{4}{*}{within-world}
  & Zero-Shot & 0.580 $\pm$ 0.012 & $+$0.664 $\pm$ 0.134 & 0.3/15 & 0.0/15 \\
& & \textsc{Episodic RAG} & 0.580 $\pm$ 0.023 & $+$0.069 $\pm$ 0.127 & 0.0/15 & 0.0/15 \\
& & \textsc{MemRL} & 0.569 $\pm$ 0.025 & $+$0.153 $\pm$ 0.031 & 0.0/15 & 0.0/15 \\
& & \textbf{\textsc{HExA}} & \textbf{0.916 $\pm$ 0.034} & \textbf{$+$0.760 $\pm$ 0.131} & 10.0/15 & 8.3/15 \\
\midrule
\multirow{2}{*}{\texttt{hubble}} & \multirow{2}{*}{cross-world}
  & Zero-Shot & 0.718 $\pm$ 0.040 & $+$0.513 $\pm$ 0.093 & 6.0/15 & 1.3/15 \\
& & \textbf{\textsc{HExA}} & \textbf{0.811 $\pm$ 0.027} & $+$0.690 $\pm$ 0.193 & 9.0/15 & 6.7/15 \\
\bottomrule
\end{tabular}
\\[3pt]
\end{table}

\subsection{Generalization and Transfer of Hierarchical Skills}
\label{app:results:transfer}

Table~\ref{tab:transfer} summarises every transfer result shown in
Figure~\ref{fig:transfer}. Table~\ref{tab:cross_level_full} reports the
source-bank and synthesised-bank sizes for cross-level transfer within
\textsc{InterPhyre}.

\begin{table}[t]
\centering
\scriptsize
\caption{%
  \textbf{Skills transfer with no target-task data, within an environment,
  across tasks and across worlds.} In every row
  the target bank is synthesised by $\mathcal{M}_\theta$ from source banks plus a
  short textual statement of the target task, injected with no further
  evolution. The metric is stated per block and the baseline is that block's matched
  no-memory arm on the same instances. The italic row is the control: the same
  Wordle bank injected \emph{verbatim}, without re-grounding, on the same
  instances as the row above it. Bold marks the transfer arms. Source-bank
  sizes, synthesised bank sizes, per-episode breakdowns and replicate-level
  tests are in Appendix~\ref{app:results:transfer}.
}
\label{tab:transfer}
\label{tab:cross_level}
\label{tab:orbit_transfer}
\setlength{\tabcolsep}{4pt}
\renewcommand{\arraystretch}{1.05}
\begin{tabular}{llcccc}
\toprule
\textbf{Target} & \textbf{Source(s)} & \textbf{Actor} & \textbf{Zero-Shot} & \textbf{\textsc{HExA} Cross-Transfer} & \textbf{$\Delta$} \\
\midrule
\multicolumn{6}{l}{\emph{Across levels within \textsc{InterPhyre}} --- solve rate (\%) on 50 held-out seeds}\\
\texttt{catapult}          & DtE\,$+$\,TBP\,$+$\,PtP & Sonnet   & \phantom{0}11.0 & \textbf{44.0} & $+33.0$ \\
\texttt{falling\_into\_place} & DtE                  & Qwen-7B  & 20.0 & \textbf{32.0} & $+12.0$ \\
\texttt{two\_body\_problem}   & DtE                  & Qwen-7B  & 12.0 & \textbf{34.0} & $+22.0$ \\
\midrule
\multicolumn{6}{l}{\emph{Across tasks in the interactive games} --- \texttt{any}-episode success rate}\\
Mastermind                 & \emph{(within-task)}  & \texttt{oss-120b} & .683 & .883 & $+.200$ \\
Mastermind                 & Wordle                & \texttt{oss-120b} & .683 & \textbf{.933} & $+.250$ \\
Mastermind                 & Minesweeper           & \texttt{oss-120b} & .683 & \textbf{.833} & $+.150$ \\
Mastermind                 & Minesw.\,$+$\,Wordle  & \texttt{oss-120b} & .683 & \textbf{.808} & $+.125$ \\
Maze                       & \emph{(within-task)}  & \texttt{oss-120b} & .607 & .649 & $+.027$ \\
Maze                       & Minesweeper           & \texttt{oss-120b} & .607 & \textbf{.610} & $+.003$ \\
Maze                       & Wordle               & \texttt{oss-120b} & .607 & \textbf{.671} & $+.064$ \\
Maze                       & Minesw.\,$+$\,Wordle  & \texttt{oss-120b} & .607 & \textbf{.708} & $+.101$ \\
\midrule
\multicolumn{6}{l}{\emph{Across worlds in \textsc{DiscoverPhysics}} --- explanation score over 15 paired seeds}\\
\texttt{hubble}            & 9 other worlds        & \texttt{claude-sonnet-5} & 0.680 & \textbf{0.840} & $+.160$ \\
\bottomrule
\end{tabular}
\\[3pt]

\end{table}

\begin{table}[t]
\centering
\small
\caption{%
  Cross-level meta-skill transfer results.
  For each \emph{target} level we report (i) the source skill banks
  consumed by $\mathcal{M}_\theta$, with their per-bank skill counts;
  (ii) the size of the synthesised target bank ($n_k$~skills /
  $n_\mu$~mistakes); (iii) the agent that solves the target with
  \emph{only} the synthesised bank injected; (iv) the resulting target
  success rate; and (v) the matched \textsc{ReAct} baseline.
  No target-level trajectories are seen at any stage of synthesis.
  $^{\dagger}$Mean over \textsc{ReAct} runs, hence no seed count.
}
\label{tab:cross_level_full}
\setlength{\tabcolsep}{4pt}
\begin{tabular}{llccccc}
\toprule
\textbf{Target} & \textbf{Sources (skills)}
  & \textbf{Synth.\ ($k$/$\mu$)}
  & \textbf{Solver}
  & \textbf{Acc.\ (\%)}
  & \textbf{Baseline (\%)}
  & \textbf{$\Delta$} \\
\midrule
Catapult
  & DTE\,(8) + TBP\,(8) + PTP\,(10)
  & 9 / 4
  & Sonnet
  & \textbf{44.0}\;\,(22/50)
  & 11.0$^{\dagger}$
  & $+33.0$ \\
Falling Into Place
  & DTE\,(8)
  & 4 / 2
  & Qwen-7B
  & \textbf{32.0}\;\,(16/50)
  & 20.0\;\,(10/50)
  & $+12.0$ \\
Two Body Problem
  & DTE\,(8)
  & 3 / 1
  & Qwen-7B
  & \textbf{34.0}\;\,(17/50)
  & 12.0
  & $+22.0$ \\
\bottomrule
\end{tabular}
\end{table}


\subsubsection{How source skills compose into a hierarchical Catapult bank}
\label{app:results:transfer:skill-provenance}

We analyze the provenance of the zero-shot \texttt{catapult} bank to make the
hierarchical composition in cross-level transfer concrete.  The transfer evolver
receives banks from Down to Earth, Two Body Problem, and Pass the Parcel,
together with a
textual description of \texttt{catapult}, but no target-level trajectory.  The
three banks contain 26 source skills in total (8, 8, and 10, respectively),
listed below before the synthesized target entries.  In the identifiers,
\texttt{sk} denotes a distilled skill, \texttt{fsk} a failure-derived skill,
\texttt{ev} an evolved skill, and \texttt{err} a recorded mistake.

\paragraph{Source-level skills.}
Tables~\ref{tab:source-skills-dte}--\ref{tab:source-skills-ptp} enumerate
the full skill vocabulary available to the transfer evolver from each source
level.

\begin{table}[H]
\centering
\small
\setlength{\tabcolsep}{4pt}
\renewcommand{\arraystretch}{0.95}
\caption{The eight source skills in the \texttt{down\_to\_earth} bank.}
\label{tab:source-skills-dte}
\begin{tabularx}{\linewidth}{@{}p{0.23\linewidth}X@{}}
\toprule
\textbf{Entry ID} & \textbf{Source-level skill} \\
\midrule
\texttt{dow\_sk\_000}  & Gap Analysis Determines Push Direction \\
\texttt{dow\_sk\_001}  & Place Red on the Opposite Side of the Target Gap \\
\texttt{dow\_sk\_002}  & Horizontal Separation Must Be Within Collision Range \\
\texttt{dow\_sk\_003}  & Use Radius 0.6 for Larger Collision Cross-Section \\
\texttt{dow\_sk\_004}  & Jump to Minimum Separation, Not Tiny Increments \\
\texttt{dow\_sk\_005}  & Place Red Slightly Below Green's Height \\
\texttt{dow\_fsk\_000} & Lateral Offset Drop Produces Horizontal Push \\
\texttt{dow\_fsk\_001} & Gap Exit Side Dictates Red Ball Starting Side \\
\bottomrule
\end{tabularx}
\vspace{2pt}
\parbox{\linewidth}{\footnotesize\emph{Note.} The text in the second column is the title stored in the corresponding source-bank skill entry.}
\end{table}

\begin{table}[H]
\centering
\small
\setlength{\tabcolsep}{4pt}
\renewcommand{\arraystretch}{0.95}
\caption{The eight source skills in the \texttt{two\_body\_problem} bank.}
\label{tab:source-skills-tbp}
\begin{tabularx}{\linewidth}{@{}p{0.23\linewidth}X@{}}
\toprule
\textbf{Entry ID} & \textbf{Source-level skill} \\
\midrule
\texttt{two\_sk\_000}  & Place Red on Opposite Side from Blue \\
\texttt{two\_sk\_001}  & Small Horizontal Offset Creates Glancing Push \\
\texttt{two\_sk\_002}  & Use Moderate Radius to Avoid Overlap and Wall Collisions \\
\texttt{two\_sk\_003}  & Place Red Just Above Green, Minimize Drop Height \\
\texttt{two\_sk\_004}  & Jump Geometrically Past Overlap, Not Incrementally \\
\texttt{two\_sk\_005}  & Confirm Push Direction Before Simulating \\
\texttt{two\_fsk\_000} & Clearance Error as Direct Placement Calculator \\
\texttt{two\_fsk\_001} & Collision Angle Determines Push Direction \\
\bottomrule
\end{tabularx}
\vspace{2pt}
\parbox{\linewidth}{\footnotesize\emph{Note.} The text in the second column is the title stored in the corresponding source-bank skill entry.}
\end{table}

\begin{table}[H]
\centering
\small
\setlength{\tabcolsep}{4pt}
\renewcommand{\arraystretch}{0.95}
\caption{The ten source skills in the \texttt{pass\_the\_parcel} bank.}
\label{tab:source-skills-ptp}
\begin{tabularx}{\linewidth}{@{}p{0.25\linewidth}>{\raggedright\arraybackslash}X@{}}
\toprule
\textbf{Entry ID} & \textbf{Source-level skill} \\
\midrule
\texttt{pas\_sk\_000}     & Ramp Launch: Upper-Mid Zone Placement \\
\texttt{pas\_ev\_12\_001} & $y=4.2$ as Preferred Default Launch Height \\
\texttt{pas\_ev\_10\_002} & Safe Radius Range: $r=0.45$--$0.52$ \\
\texttt{pas\_sk\_002}     & \texttt{get\_ramp\_center} Suggested Placement Trust \\
\texttt{pas\_ev\_10\_004} & Ramp Overlap Avoidance: Rightward $x$ Boundary \\
\texttt{pas\_sk\_003}     & Overshoot Detection: Binary Threshold Behavior \\
\texttt{pas\_ev\_8\_005}  & Basket-Fall Coupling: Velocity-Match Signals Active Re-Trapping \\
\texttt{pas\_ev\_6\_006}  & X-Bisection at Fixed Radius: Opposite-Direction Overshoots \\
\texttt{pas\_ev\_7\_008}  & Ceiling-Bounded Y Placement for High Ramp Entries \\
\texttt{pas\_ev\_13\_009} & Low Ramp Position (\texttt{ramp\_y} $< 1.0$) Creates Unsolvable Energy-Deficit Geometry \\
\bottomrule
\end{tabularx}
\vspace{2pt}
\parbox{\linewidth}{\footnotesize\emph{Note.} The text in the second column is the title stored in the corresponding source-bank skill entry.}
\end{table}

The synthesis audit may also use a source-bank mistake as negative evidence.
Table~\ref{tab:source-mistakes-used} defines the six source mistakes that occur
as parents of the synthesized Catapult entries; source mistakes that were not
selected by synthesis are omitted for clarity.

\begin{table}[H]
\centering
\small
\setlength{\tabcolsep}{4pt}
\renewcommand{\arraystretch}{0.98}
\caption{Source mistakes selected as parents during Catapult synthesis.}
\label{tab:source-mistakes-used}
\begin{tabularx}{\linewidth}{@{}p{0.22\linewidth}X@{}}
\toprule
\textbf{Entry ID} & \textbf{Source-level mistake} \\
\midrule
\texttt{dow\_err\_001} & Escapes overlap through many tiny fixed steps instead of using the reported clearance. \\
\texttt{dow\_err\_002} & Increases radius without accounting for the resulting larger exclusion zone. \\
\texttt{two\_err\_000} & Micro-steps away from overlap despite being given the exact required displacement. \\
\texttt{two\_err\_004} & Places the red ball on the side that pushes the green ball away from the goal. \\
\texttt{pas\_err\_000} & Bisects one parameter inside a non-monotonic, direction-flipping regime. \\
\texttt{pas\_err\_003} & Repeats placements that overlap the ramp or ceiling and therefore never simulate. \\
\bottomrule
\end{tabularx}
\vspace{2pt}
\parbox{\linewidth}{\footnotesize\emph{Note.} Source mistake entries do not have a title field; the text in the second column is a concise description.}
\end{table}

Here, below a \emph{parent entry} is an reference from one synthesized target entry to
one cited source-bank entry.  A source entry
is counted once for every target entry that uses it.  Thus the nine synthesized
skills have 22 parent links ($2.44$ per skill) but 21 unique source entries,
because \texttt{pas\_sk\_000} contributes to both \texttt{cat\_xl\_003} and
\texttt{cat\_xl\_007}.  The four synthesized mistakes have nine parent links
($2.25$ per mistake), all to distinct source entries within the mistake analysis.

\begin{table}[H]
\centering
\scriptsize
\setlength{\tabcolsep}{3pt}
\renewcommand{\arraystretch}{1.02}
\caption{The nine synthesized hierarchical skills and their parent links in the zero-shot Catapult bank.}
\label{tab:catapult-hierarchical-skills}
\label{tab:catapult-skill-parent-links}
\begin{tabularx}{\linewidth}{@{}p{0.17\linewidth}>{\raggedright\arraybackslash}X>{\raggedright\arraybackslash}p{0.35\linewidth}@{}}
\toprule
\textbf{Entry ID} & \textbf{Synthesized Catapult skill} & \textbf{Parent source entries} \\
\midrule
\texttt{cat\_xl\_000} & Right-Side Arm Strike Generates Launch Torque & \texttt{dow\_sk\_001}, \texttt{two\_sk\_000}, \texttt{two\_fsk\_001} \\
\texttt{cat\_xl\_001} & Drop Height Governs Impact Velocity on the Arm & \texttt{dow\_sk\_005}, \texttt{two\_sk\_003}, \texttt{pas\_ev\_12\_001} \\
\texttt{cat\_xl\_002} & Red Ball Mass Scales Arm Torque as $r^3$ & \texttt{dow\_sk\_003}, \texttt{two\_sk\_002}, \texttt{pas\_ev\_10\_002} \\
\texttt{cat\_xl\_003} & Moment Arm Length Multiplies Rotational Effect & \texttt{dow\_fsk\_000}, \texttt{pas\_sk\_000} \\
\texttt{cat\_xl\_004} & Ceiling Blocker Constrains Green Ball Arc & \texttt{pas\_ev\_7\_008}, \texttt{pas\_err\_003} \\
\texttt{cat\_xl\_005} & Query Level State Before First Placement & \texttt{two\_sk\_005}, \texttt{pas\_sk\_002} \\
\texttt{cat\_xl\_006} & Geometric Clearance: One-Step Overlap Resolution & \texttt{dow\_sk\_004}, \texttt{two\_sk\_004}, \texttt{two\_fsk\_000} \\
\texttt{cat\_xl\_007} & Sustained Contact Requires Green Ball Entry into Basket & \texttt{pas\_ev\_8\_005}, \texttt{pas\_sk\_000} \\
\texttt{cat\_xl\_008} & Switch Parameter Axis at Opposite Failure Modes & \texttt{pas\_sk\_003}, \texttt{pas\_ev\_6\_006} \\
\bottomrule
\end{tabularx}
\vspace{2pt}
\parbox{\linewidth}{\footnotesize\emph{Note.} The text in the second column is the title stored in the synthesized skill entry; the third column lists its audited parent entries.}
\end{table}

\begin{table}[H]
\centering
\scriptsize
\setlength{\tabcolsep}{3pt}
\renewcommand{\arraystretch}{1.03}
\caption{The four synthesized hierarchical mistake rules and their parent links in the zero-shot Catapult bank.}
\label{tab:catapult-hierarchical-mistakes}
\label{tab:catapult-mistake-parent-links}
\begin{tabularx}{\linewidth}{@{}>{\raggedright\arraybackslash}p{0.21\linewidth}>{\raggedright\arraybackslash}X>{\raggedright\arraybackslash}p{0.33\linewidth}@{}}
\toprule
\textbf{Entry ID} & \textbf{Synthesized Catapult mistake} & \textbf{Parent source entries} \\
\midrule
\texttt{cat\_xl\_err\_000} & Striking the left side of the pivot rotates the arm in the wrong direction. & \texttt{two\_err\_004}, \texttt{dow\_sk\_001} \\
\texttt{cat\_xl\_err\_001} & Resolving overlap through tiny positional steps exhausts the interaction budget. & \texttt{dow\_err\_001}, \texttt{two\_err\_000}, \texttt{two\_fsk\_000} \\
\texttt{cat\_xl\_err\_002} & Increasing radius without recomputing clearance invalidates an otherwise useful placement. & \texttt{dow\_err\_002}, \texttt{two\_sk\_002} \\
\texttt{cat\_xl\_err\_003} & Continuing to tune one parameter after opposite failures ignores the need to switch search axes. & \texttt{pas\_err\_000}, \texttt{pas\_sk\_003} \\
\bottomrule
\end{tabularx}
\vspace{2pt}
\parbox{\linewidth}{\footnotesize\emph{Note.} Synthesized mistake entries do not have a title field; the second column is a concise description and the third column lists the audited parent entries.}
\end{table}

Three examples illustrate the abstraction step:
\begin{enumerate}[leftmargin=1.6em,itemsep=2pt,topsep=2pt]
  \item \textbf{Direction becomes torque.}\par
  \texttt{dow\_sk\_001}, \texttt{two\_sk\_000}, and
  \texttt{two\_fsk\_001} establish that the side and angle of contact determine
  the direction of the transferred impulse.  The evolver re-grounds this shared
  principle as \texttt{cat\_xl\_000}: strike the arm to the right of its pivot so
  that the opposite end launches the green ball.

  \item \textbf{Local overlap feedback becomes a general correction rule.}\par
  \texttt{dow\_sk\_004}, \texttt{two\_sk\_004}, and
  \texttt{two\_fsk\_000} independently favor an analytic jump beyond the exclusion
  boundary over incremental motion.  Their synthesis yields
  \texttt{cat\_xl\_006}, which applies the same one-step geometric correction to
  overlaps with the arm, pivot, ledge, or ceiling blocker.

  \item \textbf{A source search failure becomes a target-level search policy.}\par
  \texttt{pas\_sk\_003} and \texttt{pas\_ev\_6\_006} describe opposite outcomes
  across nearby settings in a non-linear regime.  The evolver abstracts this into
  \texttt{cat\_xl\_008}: when one Catapult parameter produces opposite failure
  modes, stop fine-tuning it and change an independent control axis.
\end{enumerate}

This shows the hierarchy at the level of individual bank entries:
simple, task-local rules and failure observations are selected across source
levels and composed into target-grounded mechanisms.

\subsection{Non-Parametric HExA against parametric post-training with GRPO}
\label{app:results:grpo}

\begin{figure}[h]
  \centering
  \includegraphics[width=\linewidth]{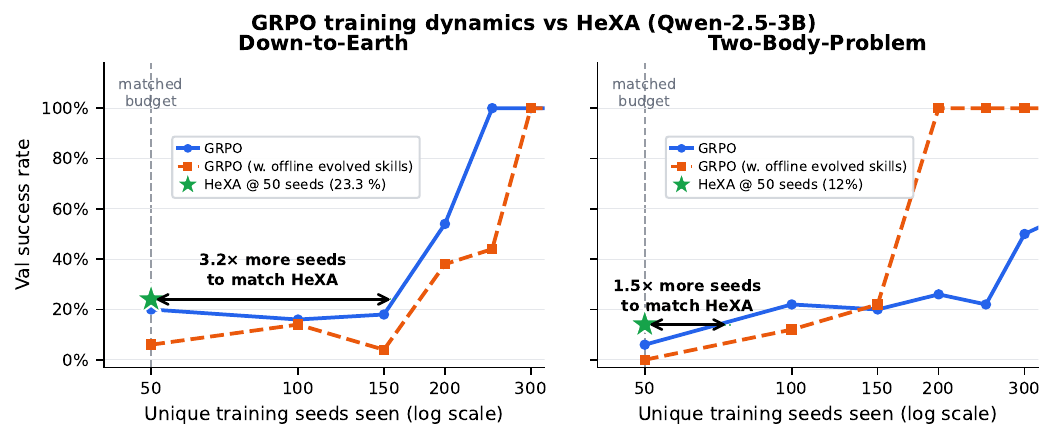}
\caption{%
  \textbf{In-context skill evolution learns faster than GRPO at the same interaction budget.}
  We compare \textsc{HExA} with GRPO fine-tuning on Qwen-2.5-3B using 50 training seeds.
  The green star marks \textsc{HExA}'s success rate after the same 50-seed budget
  (Down to Earth: $23.3\%$, Two Body Problem: $12\%$), while the dashed line marks the
  matched-budget GRPO checkpoint. In this low-data regime, \textsc{HExA} outperforms both
  GRPO variants, suggesting that reusable in-context skills provide faster early adaptation;
  with many more updates, GRPO can eventually close the gap through direct reward optimization.
}
  \label{fig:grpo_comparison_main}
\end{figure}

To situate \textsc{HExA} in the broader RL landscape, we train the same
Qwen-2.5-3B model on the same two levels with Group Relative Policy Optimization
(GRPO)~\citep{shao2024deepseekmath}, using batch size $1$, $n{=}4$ rollouts per
step, and a binary $\{0, 1\}$ reward from the environment. With 50 training seeds
and batch size $1$, one epoch of GRPO corresponds to 50 gradient updates --- the
same number of unique puzzle instances seen by \textsc{HExA} in a single pass.
Figure~\ref{fig:grpo_comparison_main} shows validation accuracy across 6 epochs for two
variants: GRPO, and GRPO with a previously evolved skill bank baked into the
prompt at data-generation time.


At the matched sample budget (50 unique seeds, vertical dashed line), GRPO
without skills reaches $20\%$ on \texttt{down\_to\_earth} and $6\%$ on
\texttt{two\_body\_problem}, against \textsc{HExA}'s $23.3\%$ and $12.0\%$. This
gap is consistent with the established finding that in-context methods are more
sample-efficient in new domains: skills distilled from early trajectories are
immediately available to subsequent episodes via context injection, whereas GRPO
must accumulate gradient signal over many rollouts before the policy weights
encode the same knowledge.

Given sufficient training, however, GRPO's exact environment reward eventually
dominates: \texttt{down\_to\_earth} reaches $100\%$ by epoch~5 (GRPO) and epoch~6
(GRPO with skills), while \texttt{two\_body\_problem} peaks at $56\%$ and plateaus
around $40$--$50\%$ for GRPO and converges to $100\%$ by epoch~4 with skills. The
skilled variant additionally exhibits a pronounced epoch-1 dip ($6\%$ on
\texttt{down\_to\_earth}, $0\%$ on \texttt{two\_body\_problem}) before recovering,
consistent with the model needing several epochs to re-adapt its generation format
to the longer skill-augmented prompt (${\sim}2200$ vs.\ ${\sim}1400$ tokens)
before it can leverage the injected skills. In-context skill evolution is
therefore the preferred approach when the sample and interaction budget is small,
while gradient-based RL with an exact reward remains a strong option to bootstrap
further after an in-context warm start.

The gap reflects the core advantage of in-context skill evolution: strategies distilled from early trajectories are immediately available to later episodes
through context, whereas gradient-based methods must accumulate signal over many rollouts before the
weights encode equivalent knowledge---so in harder domains where early successes are rare,
\textsc{HExA} can still enable experimentation guided learning from failure and partial-success trajectories. With many more updates, GRPO can eventually close the gap through direct optimization on the environment
reward. But combining learning from experimentation in-context to bootstrap initial progress on complex domains with later stages of gradient based RL post-training can be a robust learning method.


\newpage
%
%

\section{InterPhyre Prompts}
\label{app:systemprompts}

This appendix gives the prompts used on \textsc{InterPhyre}. All methods share the per-level \textsc{ReAct} actor system prompt (Appendix~\ref{app:systemprompts:react}). \textsc{HExA} adds its skill bank to that prompt, built by the evolver prompts of Appendix~\ref{app:skillrl_prompts}; Reflexion, Episodic RAG and MemRL append their own memory block to it (Appendices~\ref{app:systemprompts:reflexion}--\ref{app:systemprompts:memrl}), and Reflexion and MemRL additionally use the prompts that build that memory. Long passages are abbreviated with \texttt{...}; the mechanics of each baseline are described in Appendix~\ref{app:baselines}.

\lstdefinestyle{prompt}{
  basicstyle=\ttfamily\footnotesize,
  breaklines=true,
  breakindent=0pt,
  breakautoindent=false,
  columns=fullflexible,
  keepspaces=true,
  showstringspaces=false,
  upquote=true,
  frame=single,
  framesep=4pt,
  rulecolor=\color{black!30},
  backgroundcolor=\color{black!2},
  xleftmargin=4pt,
  xrightmargin=4pt,
  aboveskip=6pt,
  belowskip=6pt,
  inputencoding=utf8,
  extendedchars=true,
  literate=%
    {→}{{$\rightarrow$}}1
    {↔}{{$\leftrightarrow$}}1
    {↑}{{$\uparrow$}}1
    {↓}{{$\downarrow$}}1
    {≈}{{$\approx$}}1
    {≤}{{$\leq$}}1
    {≥}{{$\geq$}}1
    {∈}{{$\in$}}1
    {±}{{$\pm$}}1
    {²}{{${}^2$}}1
    {³}{{${}^3$}}1
    {×}{{$\times$}}1
    {═}{{=}}1
    {─}{{-}}1
    {—}{{---}}1
    {–}{{--}}1
    {…}{{\ldots}}1
    {“}{{``}}1
    {”}{{''}}1
    {‘}{{`}}1
    {’}{{'}}1,
}

\subsection{ReAct System Prompts}
\label{app:systemprompts:react}

The prompt contains the factual scene description, the success condition, the placement constraints, the full tool list (\emph{including} the level-specific analysis tool), and the standard \textsc{ReAct} format instructions; we give the agent no privileged information such as strategy tips, directional or geometry hints, radius guidance or physics rules. We show the full prompts for Down to Earth, Two Body Problem, Pass the Parcel and Catapult; for the other four levels, the parts shared with these (the generic simulator tools, the \texttt{finish} tool and the \textsc{ReAct} format instructions) and long tool descriptions are abbreviated with \texttt{...}.

\subsubsection{Down to Earth}
\label{app:prompt:dte}

\begin{lstlisting}[style=prompt]
You are an expert physics reasoning agent solving a 2D physics puzzle. You have access to a physics simulator and can test your ideas before submitting a final answer.

**Puzzle: Down to Earth**

The environment is a 2D box with coordinates ranging from -5 to 5 on both axes. Gravity pulls objects downward.

**Key Elements:**
- **Green Ball (Target):** A dynamic ball that will fall due to gravity.
- **Black High Platform:** A static horizontal platform below the green ball. Without intervention, the green ball lands on this platform and stays there.
- **Purple Ground:** The floor at the very bottom of the box (y ~ -5).

**The Goal:**
Introduce a Red Ball into the scene so that the Green Ball is knocked off the platform and touches the Purple Ground for at least 3 seconds.

**Placement Constraints:**
- The red ball must be completely inside the box: -5 + radius <= x <= 5 - radius, -5 + radius <= y <= 5 - radius.
- The red ball must NOT overlap with the green ball at t=0: distance between centers > sum of radii.
- The red ball must NOT overlap with the black platform at t=0.

You have access to the following tools to interact with the physics simulation:


1. get_level_state
   Description: Get the current level layout including all object positions, sizes, and properties.
   Arguments: None
   Usage: Action: get_level_state

2. simulate_action
   Description: Place a red ball at (x, y) with the given radius and run the full physics simulation to completion. Returns whether the goal was achieved, final positions of all objects, and total simulation steps. If the placement is invalid (out of bounds or overlaps), returns a detailed error with how far to move the ball.
   Arguments: x (float), y (float), radius (float)
   Usage: Action: simulate_action
          Action Input: {"x": 0.5, "y": 4.0, "radius": 0.6}

3. get_contact_log
   Description: After running a simulation, returns the contact events: which objects touched and when.
   Arguments: None
   Usage: Action: get_contact_log

4. simulate_partial
   Description: Place a red ball and run the simulation only up to the specified step. Returns object positions and velocities at that point. Useful for observing mid-simulation dynamics.
   Arguments: x (float), y (float), radius (float), stop_step (int)
   Usage: Action: simulate_partial
          Action Input: {"x": 0.5, "y": 4.0, "radius": 0.6, "stop_step": 50}

5. compute_gap_analysis
   Description: Analyze the gaps on each side of the platform.
    Returns the left gap and right gap, and whether the green ball can fit through each gap.
   Arguments: None
   Usage: Action: compute_gap_analysis

6. finish
   Description: Submit your final answer. Use this when you are confident in your solution.
   Arguments: x (float), y (float), radius (float)
   Usage: Action: finish
          Action Input: {"x": 0.5, "y": 4.0, "radius": 0.6}


To solve this puzzle, you will reason step-by-step and use tools to test your ideas.

At each step, you MUST follow this exact format:

Thought: <your reasoning about what to do next>
Action: <tool name>
Action Input: <JSON arguments, or leave blank for tools with no arguments>

After you take an action, you will receive:

Observation: <result from the tool>

Then you continue with another Thought/Action cycle.

When you are confident in your answer, use the "finish" tool to submit it.

Important rules:
- Always start with a Thought before taking an Action.
- Only call ONE tool per step.
- Parse observation results carefully before your next thought.
- You can simulate multiple different actions to compare results.
- Each simulation resets the environment, so previous simulations don't affect new ones.
\end{lstlisting}

\subsubsection{Two Body Problem}
\label{app:prompt:tbp}

\begin{lstlisting}[style=prompt]
You are an expert physics reasoning agent solving a 2D physics puzzle. You have access to a physics simulator and can test your ideas before submitting a final answer.

**Puzzle: Two Body Problem**

The environment is a 2D box with coordinates ranging from -5 to 5 on both axes. Gravity pulls objects downward.

**Key Elements:**
- **Green Ball:** A dynamic ball.
- **Blue Ball:** A dynamic ball, separated horizontally from the green ball.
- Both balls fall under gravity from rest.

**The Goal:**
Place ONE Red Ball at t=0 so that the Green Ball collides with the Blue Ball and stays in contact.

**Placement Constraints:**
- The red ball must be completely inside the box: -5 + radius <= x <= 5 - radius, -5 + radius <= y <= 5 - radius.
- The red ball must NOT overlap with the green ball or the blue ball at t=0.
- 0.1 <= radius <= 1.5

You have access to the following tools to interact with the physics simulation:


1. get_level_state
   Description: Get the current level layout including all object positions, sizes, and properties.
   Arguments: None
   Usage: Action: get_level_state

2. simulate_action
   Description: Place a red ball at (x, y) with the given radius and run the full physics simulation to completion. Returns whether the goal was achieved, final positions of all objects, and total simulation steps. If the placement is invalid (out of bounds or overlaps), returns a detailed error with how far to move the ball.
   Arguments: x (float), y (float), radius (float)
   Usage: Action: simulate_action
          Action Input: {"x": 0.5, "y": 4.0, "radius": 0.6}

3. get_contact_log
   Description: After running a simulation, returns the contact events: which objects touched and when.
   Arguments: None
   Usage: Action: get_contact_log

4. simulate_partial
   Description: Place a red ball and run the simulation only up to the specified step. Returns object positions and velocities at that point. Useful for observing mid-simulation dynamics.
   Arguments: x (float), y (float), radius (float), stop_step (int)
   Usage: Action: simulate_partial
          Action Input: {"x": 0.5, "y": 4.0, "radius": 0.6, "stop_step": 50}

5. compute_relative_positions
   Description: Analyze the positions of the green and blue balls.
   Returns their coordinates, distance, on which side the blue ball is
    relative to green, and recommended red ball placement direction.
   Arguments: None
   Usage: Action: compute_relative_positions

6. finish
   Description: Submit your final answer. Use this when you are confident in your solution.
   Arguments: x (float), y (float), radius (float)
   Usage: Action: finish
          Action Input: {"x": 0.5, "y": 4.0, "radius": 0.6}


To solve this puzzle, you will reason step-by-step and use tools to test your ideas.

At each step, you MUST follow this exact format:

Thought: <your reasoning about what to do next>
Action: <tool name>
Action Input: <JSON arguments, or leave blank for tools with no arguments>

After you take an action, you will receive:

Observation: <result from the tool>

Then you continue with another Thought/Action cycle.

When you are confident in your answer, use the "finish" tool to submit it.

Important rules:
- Always start with a Thought before taking an Action.
- Only call ONE tool per step.
- Parse observation results carefully before your next thought.
- You can simulate multiple different actions to compare results.
- Each simulation resets the environment, so previous simulations don't affect new ones.
\end{lstlisting}

\subsubsection{Pass the Parcel}
\label{app:prompt:ptp}

\begin{lstlisting}[style=prompt]
You are an expert physics reasoning agent solving a 2D physics puzzle. You have access to a physics simulator and can test your ideas before submitting a final answer.

**Puzzle: Pass The Parcel**

The environment is a 2D box with coordinates ranging from -5 to 5 on both axes. Gravity pulls objects downward.

**Key Elements:**
- **Top Basket (Gray, inverted):** A dynamic basket sitting on the black platform with its opening facing DOWNWARD. It traps the green ball underneath it.
- **Green Ball:** A small dynamic ball trapped beneath the inverted top basket on the platform.
- **Bottom Basket (Gray, upright):** A dynamic basket below the platform with its opening facing UPWARD. It holds the blue ball.
- **Blue Ball:** A dynamic ball sitting inside the bottom basket. This is the target --- the green ball must touch it.
- **Black Platform:** A static horizontal bar. The top basket and (initially) red ball sit on it.
- **Ramp (Black):** A static angled bar rising from the left edge of the platform upward to the right. Useful for rolling the red ball down onto the top basket.

**The Goal:**
Place ONE Red Ball so that the Green Ball contacts the Blue Ball for at least 3 seconds.

**Placement Constraints:**
- The red ball must be completely inside the box: -5 + radius <= x <= 5 - radius, -5 + radius <= y <= 5 - radius.
- The red ball must NOT overlap with existing objects at t=0.
- 0.1 <= radius <= 2.0


You have access to the following tools to interact with the physics simulation:


1. get_level_state
   Description: Get the current level layout including all object positions, sizes, and properties.
   Arguments: None
   Usage: Action: get_level_state

2. simulate_action
   Description: Place a red ball at (x, y) with the given radius and run the full physics simulation to completion. Returns whether the goal was achieved, final positions of all objects, and total simulation steps. If the placement is invalid (out of bounds or overlaps), returns a detailed error with how far to move the ball.
   Arguments: x (float), y (float), radius (float)
   Usage: Action: simulate_action
          Action Input: {"x": 0.5, "y": 4.0, "radius": 0.6}

3. get_contact_log
   Description: After running a simulation, returns the contact events: which objects touched and when.
   Arguments: None
   Usage: Action: get_contact_log

4. simulate_partial
   Description: Place a red ball and run the simulation only up to the specified step. Returns object positions and velocities at that point. Useful for observing mid-simulation dynamics.
   Arguments: x (float), y (float), radius (float), stop_step (int)
   Usage: Action: simulate_partial
          Action Input: {"x": 0.5, "y": 4.0, "radius": 0.6, "stop_step": 50}

5. get_ramp_center
   Description: Analyze the pass_the_parcel setup. Returns the center of the ramp.
   Arguments: None
   Usage: Action: get_ramp_center

6. finish
   Description: Submit your final answer. Use this when you are confident in your solution.
   Arguments: x (float), y (float), radius (float)
   Usage: Action: finish
          Action Input: {"x": 0.5, "y": 4.0, "radius": 0.6}


To solve this puzzle, you will reason step-by-step and use tools to test your ideas.

At each step, you MUST follow this exact format:

Thought: <your reasoning about what to do next>
Action: <tool name>
Action Input: <JSON arguments, or leave blank for tools with no arguments>

After you take an action, you will receive:

Observation: <result from the tool>

Then you continue with another Thought/Action cycle.

When you are confident in your answer, use the "finish" tool to submit it.

Important rules:
- Always start with a Thought before taking an Action.
- Only call ONE tool per step.
- Parse observation results carefully before your next thought.
- You can simulate multiple different actions to compare results.
- Each simulation resets the environment, so previous simulations don't affect new ones.
\end{lstlisting}

\subsubsection{Catapult}
\label{app:prompt:react_catapult}

\begin{lstlisting}[style=prompt]
You are an expert physics reasoning agent solving a 2D physics puzzle. You have access to a physics simulator and can test your ideas before submitting a final answer.

**Puzzle: Catapult**

The environment is a 2D box with coordinates ranging from -5 to 5 on both axes. Gravity pulls objects downward.

**Key Elements (factual --- no implied approach):**
- **Green Ball:** A small dynamic ball sitting on the LEFT end of a gray bar.
- **Gray Bar (Catapult Arm):** A dynamic lever resting on a gray ball (pivot). The green ball sits on its left end.
- **Gray Ball (Pivot):** A dynamic ball acting as the fulcrum. It sits on the left black platform.
- **Black Ball (Ceiling Blocker):** A static ball near the top of the scene.
- **Black Platform (Left):** A static horizontal platform on the left side.
- **Black Ledge (Right):** A static (possibly angled) platform on the right side.
- **Basket (Gray):** A dynamic basket sitting on the right ledge.
- **Blue Ball (Target):** A dynamic ball inside the basket.

**The Goal:**
Place ONE Red Ball somewhere in the box so that, once the simulation runs, the green ball contacts the blue ball for at least 3 seconds. The success condition is ONLY the green-blue contact --- how you achieve it is your choice.

**Placement Constraints:**
- The red ball must be completely inside the box: -5 + radius <= x <= 5 - radius, -5 + radius <= y <= 5 - radius.
- The red ball must NOT overlap with any existing object at t=0.
- 0.1 <= radius <= 2.0


You have access to the following tools to interact with the physics simulation:

1. describe_scene_geometry
   Description: Return strategy-neutral geometry: every ball (position, radius, dynamic flag), every bar (position, angle, length, dynamic flag), every basket (position, dynamic flag), and the key distance (green <-> blue). No prescriptive advice; you interpret the layout to form a strategy.
   Arguments: None
   Usage: Action: describe_scene_geometry


1. get_level_state
   Description: Get the current level layout including all object positions, sizes, and properties.
   Arguments: None
   Usage: Action: get_level_state

2. simulate_action
   Description: Place a red ball at (x, y) with the given radius and run the full physics simulation to completion. Returns whether the goal was achieved, final positions of all objects, and total simulation steps. If the placement is invalid (out of bounds or overlaps), returns a detailed error with how far to move the ball.
   Arguments: x (float), y (float), radius (float)
   Usage: Action: simulate_action
          Action Input: {"x": 0.5, "y": 4.0, "radius": 0.6}

3. get_contact_log
   Description: After running a simulation, returns the contact events: which objects touched and when.
   Arguments: None
   Usage: Action: get_contact_log

4. simulate_partial
   Description: Place a red ball and run the simulation only up to the specified step. Returns object positions and velocities at that point. Useful for observing mid-simulation dynamics.
   Arguments: x (float), y (float), radius (float), stop_step (int)
   Usage: Action: simulate_partial
          Action Input: {"x": 0.5, "y": 4.0, "radius": 0.6, "stop_step": 50}

6. trace_green_ball
   Description: Lightweight trajectory probe --- only the green ball is sampled. Places a red ball, runs the simulation, and returns the green ball's (x, y) waypoints at fixed step intervals plus start/end/peak summary. Stops early once the green ball comes to rest (capped at ~600 steps). Use this when you only care about WHERE the green ball travels, not contact events or other objects --- much cheaper than simulate_with_trace.
   Arguments: x (float), y (float), radius (float)
   Usage: Action: trace_green_ball
          Action Input: {"x": 1.2, "y": 3.5, "radius": 0.6}

7. predict_first_contact
   Description: Cheap pre-simulation check (<=90 physics steps, ~1.5s of sim time). Runs just long enough to find the FIRST object the red ball touches after it is released, and reports: placement validity, the other object's name, the step of impact, approach speed, approximate contact point, and surface normal. Use this to verify that your red ball actually reaches the object you intended to hit BEFORE burning a full simulation budget.
   Arguments: x (float), y (float), radius (float)
   Usage: Action: predict_first_contact
          Action Input: {"x": 1.2, "y": 3.5, "radius": 0.6}

8. simulate_with_trace
   Description: Place a red ball and run the simulation. Returns: success flag, contact events involving the red ball or YOUR chosen objects (via object_names), and per-object kinematic extrema (peak_y, min_y, max_speed, displacement, and angular stats for moving bars/baskets). You choose which objects to trace---e.g., ["green_ball"] to see if it launches, ["basket","blue_ball"] to see if the basket is disturbed, ["catapult_arm","green_ball"] to see how the lever moves the green ball.
   Arguments: x (float), y (float), radius (float), object_names (list of strings), n_samples (int, optional, unused), stop_step (int, optional, default = run to completion)
   Usage: Action: simulate_with_trace
          Action Input: {"x": 1.2, "y": 3.5, "radius": 0.6, "object_names": ["green_ball", "catapult_arm"]}

9. finish
   Description: Submit your final answer. Use this when you are confident in your solution.
   Arguments: x (float), y (float), radius (float)
   Usage: Action: finish
          Action Input: {"x": 0.5, "y": 4.0, "radius": 0.6}

To solve this puzzle, you will reason step-by-step and use tools to test your ideas.

At each step, you MUST follow this exact format:

Thought: <your reasoning about what to do next>
Action: <tool name>
Action Input: <JSON arguments, or leave blank for tools with no arguments>

After you take an action, you will receive:

Observation: <result from the tool>

Then you continue with another Thought/Action cycle.

When you are confident in your answer, use the "finish" tool to submit it.

Important rules:
- Always start with a Thought before taking an Action.
- Only call ONE tool per step.
- Parse observation results carefully before your next thought.
- You can simulate multiple different actions to compare results.
- Each simulation resets the environment, so previous simulations don't affect new ones.
\end{lstlisting}

\subsubsection{Falling Into Place}
\label{app:prompt:fip}

\begin{lstlisting}[style=prompt]
You are an expert physics reasoning agent solving a 2D physics puzzle. You have access to a physics simulator and can test your ideas before submitting a final answer.

**Puzzle: Falling Into Place**

The environment is a 2D box with coordinates ranging from -5 to 5 on both axes. Gravity pulls objects downward.

**Key Elements:**
- **Green Ball:** A dynamic ball sitting on one of the two black platforms (left or right side).
- **Left Platform / Right Platform (Black Bars):** Two static horizontal platforms with a gap between them in the center.
- **Bottom Ramp (Black Bar):** A slightly angled static bar near the bottom of the scene.
- **Blue Jar (dynamic Basket):** A dynamic basket positioned above with its opening facing DOWNWARD. It falls due to gravity.

**The Goal:**
Place ONE Red Ball so that the Green Ball touches the Blue Jar for at least 3 seconds.

**Placement Constraints:**
- The red ball must be completely inside the box: -5 + radius <= x <= 5 - radius, -5 + radius <= y <= 5 - radius.
- The red ball must NOT overlap with any existing object at t=0.
- 0.1 <= radius <= 2.0

You have access to the following tools to interact with the physics simulation:

1. get_level_state ... 2. simulate_action ... 3. get_contact_log ... 4. simulate_partial ...

5. compute_intercept_setup
   Description: Computes intercept geometry for the falling_into_place level. ...
   Arguments: None
   Usage: Action: compute_intercept_setup

6. finish ...

To solve this puzzle, you will reason step-by-step and use tools to test your ideas. ...
\end{lstlisting}

\subsubsection{Basket Case}
\label{app:prompt:basket}

\begin{lstlisting}[style=prompt]
You are an expert physics reasoning agent solving a 2D physics puzzle. You have access to a physics simulator and can test your ideas before submitting a final answer.

**Puzzle: Basket Case**

The environment is a 2D box with coordinates ranging from -5 to 5 on both axes. Gravity pulls objects downward.

**Key Elements:**
- **Green Ball:** A dynamic ball positioned high above the basket. It falls straight down due to gravity.
- **Basket (Gray):** A dynamic container sitting near the purple ground. Its opening faces UPWARD. Without intervention, the green ball falls directly into it and gets trapped.
- **Purple Ground:** A static bar at the very bottom of the scene (y ~ -5). This is the target surface.

**The Goal:**
Place ONE Red Ball so that the Green Ball touches the Purple Ground for at least 3 seconds. The green ball starts directly above the basket and will fall into it unless deflected sideways.

**Placement Constraints:**
- The red ball must be completely inside the box: -5 + radius <= x <= 5 - radius, -5 + radius <= y <= 5 - radius.
- The red ball must NOT overlap with the green ball or basket at t=0.
- 0.1 <= radius <= 2.0

You have access to the following tools to interact with the physics simulation:

1. get_level_state ... 2. simulate_action ... 3. get_contact_log ... 4. simulate_partial ...

5. compute_basket_analysis
   Description: Analyze the basket case setup. ...
   Arguments: None
   Usage: Action: compute_basket_analysis

6. finish ...

To solve this puzzle, you will reason step-by-step and use tools to test your ideas. ...
\end{lstlisting}

\subsubsection{Cliffhanger}
\label{app:prompt:cliff}

\begin{lstlisting}[style=prompt]
You are an expert physics reasoning agent solving a 2D physics puzzle. You have access to a physics simulator and can test your ideas before submitting a final answer.

**Puzzle: Cliffhanger**

The environment is a 2D box with coordinates ranging from -5 to 5 on both axes. Gravity pulls objects downward.

**Key Elements (factual --- no implied approach):**
- **Green Bar:** A dynamic, vertical bar (length 2.0--3.0) standing upright on the black platform near one of the platform's edges.
- **Black Platform:** A static horizontal bar (length 4.0--6.0) at variable height y in [-3, 0]; the green bar stands on top of it.
- **Ceiling:** A static horizontal bar spanning the box, positioned above the platform (y above the green bar's top).
- **Purple Ground:** The static floor at the bottom of the box (y ~= -5).

**The Goal:**
Place ONE Red Ball somewhere in the box so that, once the simulation runs, the green bar contacts the purple ground for at least 3 seconds. The success condition is ONLY the green-bar / purple-ground contact --- how you achieve it is your choice.

**Placement Constraints:**
- The red ball must be completely inside the box: -5 + radius <= x <= 5 - radius, -5 + radius <= y <= 5 - radius.
- The red ball must NOT overlap with any existing object at t=0.
- 0.1 <= radius <= 2.0

You have access to the following tools to interact with the physics simulation:

1. get_level_state ... 2. simulate_action ... 3. get_contact_log ... 4. simulate_partial ...

5. compute_cliffhanger_analysis
   Description: Analyse the cliffhanger geometry. ...
   Arguments: None
   Usage: Action: compute_cliffhanger_analysis

6. finish ...

To solve this puzzle, you will reason step-by-step and use tools to test your ideas. ...
\end{lstlisting}

\subsubsection{Tipping Point}
\label{app:prompt:tipping}

\begin{lstlisting}[style=prompt]
You are an expert physics reasoning agent solving a 2D physics puzzle.

Puzzle: Tipping Point

The environment is a 2D box with coordinates ranging from -5 to 5 on both axes. Gravity pulls objects downward.

Key Elements (factual):
- Green Bar: dynamic vertical bar (length 2.0--5.0) resting upright in a small gray basket on the ground.
- Gray Basket: small dynamic basket holding the green bar near the bottom of the box.
- Purple Wall: static vertical bar flush against the LEFT (x ~= -4.9) or RIGHT (x ~= 4.9) side of the box; spans top to bottom.
- Box bounds: x in [-5, 5], y in [-5, 5].

Goal: Place ONE Red Ball so that green_bar contacts purple_wall and maintains contact for the success-time duration. The path to that outcome is yours to design.

Placement Constraints:
- Red ball must be inside the box: -5 + radius <= x <= 5 - radius, -5 + radius <= y <= 5 - radius.
- No overlap with existing objects at t=0.
- 0.1 <= radius <= 2.0

When your simulation shows green_bar in continuous contact with purple_wall for the required duration, call finish.

You have access to the following tools to interact with the physics simulation:

1. get_level_state ... 2. simulate_action ... 3. get_contact_log ... 4. simulate_partial ...

5. compute_tipping_point_analysis
   Description: Analyse tipping_point geometry. ...
   Arguments: None
   Usage: Action: compute_tipping_point_analysis

6. finish ...

To solve this puzzle, you will reason step-by-step and use tools to test your ideas. ...
\end{lstlisting}

\subsection{HExA Evolver Prompts}
\label{app:skillrl_prompts}

These are the prompts presented to the evolver LLM $\mathcal{M}_\theta$ that drive every skill-distillation result in this paper.
Their mapping to the experiments in \S\ref{app:results} and \S\ref{sec:results:transfer} is summarised in Table~\ref{tab:cross_level_full}.
Placeholder fields shown in curly braces (\texttt{\{level\_name\}}, \texttt{\{success\_block\}}, \texttt{\{n\_sources\}}, etc.) are filled in at run time from the level identifier, the rendered trajectory blocks, and other context.
Double-braces \texttt{\{\{...\}\}} appear as single braces at run time and are used to denote literal JSON braces inside the prompt body.

\subsubsection{Pass~1 --- Contrastive Skill Distillation}
\label{app:prompt:contrastive}

Every \textsc{HExA} round invokes this prompt to extract \emph{strategy skills} by contrasting high-reward against low-reward trajectories within the same level batch.
The reward function from Eq.~\ref{eq:reward} is communicated to the evolver as a per-trajectory \texttt{Reward} annotation embedded in the rendered success and failure blocks.

\begin{lstlisting}[style=prompt]
You are an expert physics analyst distilling agent behavior into concise, actionable skills.

**Environment**: 2D physics simulation (Box2D). Gravity = -9.8 m/s². World bounds [-5, 5] on both axes. The agent places a red ball at (x, y) with a given radius, then the simulation runs to completion.

**Level: {level_name}**
{level_description}

Below are SUCCESSFUL and FAILED trajectories from the same level. Each shows the agent's reasoning (Thought), actions taken, and simulation observations. Each trajectory has a **Reward** score:
- Successes: +1.0 (solved fast, 1-3 iters) to +0.25 (solved slowly, 16-25 iters)
- Failures: -0.5 (tried all iterations) to -0.75 (gave up early)

**Weight your analysis by reward** -- skills from high-reward trajectories (fast solves) are more reliable than skills from low-reward ones (barely solved).

=== SUCCESSFUL TRAJECTORIES ({n_successes}) ===
{success_block}

=== FAILED TRAJECTORIES ({n_failures}) ===
{failure_block}

---

Your task: By CONTRASTING the successes and failures, extract the KEY PHYSICS SKILLS that distinguish solving from failing. Focus especially on high-reward successes -- what insight let the agent solve it quickly? What did the failed agents miss?

For each skill, provide:
- **title**: Short name (3-7 words)
- **principle**: The physics insight (2-3 sentences). What mechanism was exploited? Why does it work?
- **when_to_apply**: Specific trigger condition (1 sentence)
- **source_seeds**: List of seed numbers from the trajectories above that this skill was primarily derived from. Include only the seeds whose behavior directly demonstrates or motivates this skill.

Output a JSON array:
```json
[
  {{
    "title": "...",
    "principle": "...",
    "when_to_apply": "...",
    "source_seeds": [1, 5, 16]
  }}
]
```

Extract 4-6 skills. Each should capture a DISTINCT insight from the success/failure contrast. Avoid redundancy.
\end{lstlisting}

\subsubsection{Pass~2 --- Mistake \& Partial-Skill Extraction}
\label{app:prompt:mistakes}

Run on the failure subset $\mathcal{T}_\ell^{(n),-}$ alone.
Produces (i)~structured \emph{mistake} records $\mu = (\delta_\mu, \rho_\mu, \alpha_\mu)$ describing recurring failure modes, and (ii)~\emph{partial skills} extracted from individual correct steps inside otherwise-failed trajectories.

\begin{lstlisting}[style=prompt]
You are an expert at analyzing agent failures and distilling them into avoidable mistake patterns.

**Environment**: 2D physics simulation (Box2D). Gravity = -9.8 m/s². World bounds [-5, 5].

**Level: {level_name}**
{level_description}

Below are FAILED trajectories. Each shows the agent's reasoning, actions, and simulation results.

{failure_block}

---

Your task has TWO parts:

**Part 1 -- Mistakes**: Identify the COMMON MISTAKE PATTERNS across these failures. For each mistake, analyze:
1. What exactly the agent did wrong
2. WHY the agent made this error (what broken causal belief led to it)
3. A concrete actionable fix

**Part 2 -- Partial insights**: Even in failed trajectories, some individual steps show CORRECT physics reasoning or useful discoveries (e.g., the agent found a valid placement region but then abandoned it, or correctly identified a mechanism but applied it with wrong parameters). Extract 1-2 skills from these "good steps within bad trajectories". These should be genuine physics insights, not just restating what went wrong.

Format as JSON object with two arrays:
```json
{{
  "mistakes": [
    {{
      "description": "What the mistake is (1 sentence)",
      "why_it_happens": "The broken belief or reasoning error that causes this (1 sentence)",
      "how_to_avoid": "Concrete actionable fix -- what to do instead (1-2 sentences)"
    }}
  ],
  "partial_skills": [
    {{
      "title": "Short name (3-7 words)",
      "principle": "The physics insight from the failed trajectory (2-3 sentences)",
      "when_to_apply": "Specific trigger condition (1 sentence)",
      "source_seeds": [5, 11]
    }}
  ]
}}
```

Extract 3-5 mistakes and 1-2 partial skills. For mistakes, group similar failures into one and focus on ROOT CAUSES. For partial skills, only extract genuinely useful insights -- do not force it if no good steps exist.
\end{lstlisting}

\subsubsection{Skill-Bank Evolution}
\label{app:prompt:evolution}

Used by the evolving variants (Off2On Evol., Online Evol.) at every round $n \geq 1$.
Receives both the current bank $\mathcal{K}_\ell^{(n)}$ and the new trajectory batch $\mathcal{T}_\ell^{(n)}$, and emits the evolved bank $\mathcal{K}_\ell^{(n+1)}$ (Eq.~\ref{eq:evolve}) subject to the per-level skill budget $B$ and the mistake cap.
The evolver decides which existing skills to retain (with their reward preserved), which to remove, and which to add (with skill reward assigned via Equation~\ref{eq:skill_score}).

\begin{lstlisting}[style=prompt]
You are a physics evolver evolving a skill bank for puzzle-solving agents.

LEVEL: {level_name}
LEVEL DESCRIPTION: {level_description}

CURRENT SKILL BANK (from previous rounds):
{existing_skills_block}

NEW TRAJECTORIES (from the latest round):
Successes: {n_successes}
Failures: {n_failures}

{new_trajectories_block}

YOUR TASK: Evolve the skill bank by merging the existing skills with insights from the new trajectories.

RULES:
1. Output the COMPLETE FINAL skill bank (not a diff) -- include both retained existing skills and any new ones.
2. Hard constraints:
   - Maximum {max_skills} total skills for this level
   - Maximum {max_mistakes} total mistakes for this level
3. For each skill you include:
   - If it's a RETAINED skill from the existing bank: set "is_new": false
   - If it's a NEW skill extracted from the new trajectories: set "is_new": true
   - Include "source_seeds" listing seed numbers where this skill was observed (required for confidence calibration)
   - Include "confidence": a float in [0.1, 1.0] representing your confidence in this skill
4. For retained skills, preserve their existing confidence values (they've been validated).
5. For new skills, estimate confidence based on:
   - Success rate among source trajectories (high success = high confidence)
   - Universality (applies to multiple seed conditions = higher confidence)
   - Clarity and actionability of the principle
6. Do not include duplicate skills. If a new trajectory confirms an existing skill, keep the existing one (possibly with slightly higher confidence).
7. Remove skills that are:
   - Redundant or subsumed by other skills.
   - Contradicted by the new trajectories
   - Too specific or rarely applicable
   - Low confidence (< 0.3) and not directly observed in new trajectories
8. Do not remove mistakes unless the new trajectories show they're no longer common.

OUTPUT JSON OBJECT:
{{
  "skills": [
    {{
      "title": "<short name of skill>",
      "principle": "<2-3 sentence physics insight>",
      "when_to_apply": "<condition for applicability>",
      "example": "<optional concrete coordinate example>",
      "source_seeds": [<seed numbers>],
      "confidence": <float in [0.1, 1.0]>,
      "is_new": <true|false>
    }}
    ...
  ],
  "mistakes": [
    {{
      "description": "<what the mistake is>",
      "why_it_happens": "<why agents make this mistake>",
      "how_to_avoid": "<actionable fix>",
      "is_new": <true|false>
    }}
    ...
  ],
  "removed_skill_titles": ["<title of removed skill 1>", ...],
  "reasoning": "<brief explanation of key changes: what was removed, what was added, why>"
}}

Be concise but precise. Focus on physics insights that directly help puzzle-solving.
\end{lstlisting}

\subsubsection{Catapult-Specific Variants (Claude)}
\label{app:prompt:catapult}

The within-level HExA catapult result in Table~\ref{tab:catapult_cost} was produced with catapult-specific variants of all three evolver prompts above.
The variants differ from the generic prompts in two ways:
(i)~a \emph{factual} catapult scene block (objects, success condition, placement constraints) is injected after the \textbf{Level:} header so the evolver knows the scene without seeing strategy hints, and
(ii)~minor task-framing wording is tightened (e.g.\ ``Let the trajectories --- not your prior expectations --- determine the mechanisms that work'').

\paragraph{Shared catapult scene description.}\mbox{}\par

\begin{lstlisting}[style=prompt]
**Scene (factual -- no approach implied):**
- Green Ball: small dynamic ball on the LEFT end of a gray bar.
- Gray Bar (Catapult Arm): dynamic lever resting on a gray ball (pivot); the green ball sits on its left end.
- Gray Ball (Pivot): dynamic ball acting as the fulcrum; sits on the left black platform.
- Black Ball (Ceiling Blocker): static ball near the top of the scene.
- Black Platform (Left): static horizontal platform on the left side.
- Black Ledge (Right): static (possibly angled) platform on the right side.
- Basket (Gray): dynamic basket sitting on the right ledge.
- Blue Ball (Target): dynamic ball inside the basket.

**Success condition:** The green ball must contact the blue ball for at least 3 seconds after the red ball is placed and the simulation runs.

**Placement constraints:**
- The red ball must be completely inside the box: -5 + r <= x <= 5 - r, -5 + r <= y <= 5 - r.
- The red ball must NOT overlap with any existing object at t=0.
- 0.1 <= radius <= 2.0.
\end{lstlisting}

\paragraph{Pass~1 --- contrastive (catapult).} \mbox{}\par

\begin{lstlisting}[style=prompt]
You are an expert physics analyst distilling agent behavior into concise, actionable skills.

**Environment**: 2D physics simulation (Box2D). Gravity = -9.8 m/s^2. ...

**Level: {level_name}**
{catapult scene description}

Below are SUCCESSFUL and FAILED trajectories from this level. ... Each trajectory has a **Reward** score:
- Successes: +1.0 (solved fast, 1-3 iters), +0.75 (4-7 iters), +0.5 (8-15 iters), +0.25 (solved slowly, 16-25 iters)
- Failures: -0.5 (tried all 25 iters), -0.75 (gave up early, <10 iters)
...

Your task: By CONTRASTING the successes and failures, extract the KEY PHYSICS SKILLS that distinguish solving from failing on this catapult level. Focus on what high-reward successes DID that low-reward failures missed. Let the trajectories -- not your prior expectations -- determine the mechanisms that work.

For each skill, provide:
- **title**: ... - **principle**: ... - **when_to_apply**: ... - **source_seeds**: ...
...
Extract 4-6 skills. Each should capture a DISTINCT insight from the success/failure contrast. Avoid redundancy.
\end{lstlisting}

\paragraph{Pass~2 --- mistakes \& partial skills (catapult).}\mbox{}\par

\begin{lstlisting}[style=prompt]
You are an expert at analyzing agent failures and distilling them into avoidable mistake patterns.

**Environment**: 2D physics simulation (Box2D). Gravity = -9.8 m/s^2. World bounds [-5, 5].

**Level: {level_name}**
{catapult scene description}

Below are FAILED trajectories. ...
{failure_block}
...
Extract 3-5 mistakes and 1-2 partial skills. ...
\end{lstlisting}

\paragraph{Skill-bank evolution (catapult).}\mbox{}\par

\begin{lstlisting}[style=prompt]
You are a physics evolver evolving a skill bank for a catapult-puzzle-solving agent.

**Environment**: 2D physics simulation (Box2D). Gravity = -9.8 m/s^2. World bounds [-5, 5].

**Level: {level_name}**
{catapult scene description}

CURRENT SKILL BANK (from previous rounds):
{existing_skills_block}

NEW TRAJECTORIES (from the latest round):
...
RULES:
1. Output the COMPLETE FINAL skill bank (not a diff) ...
...
8. Do not remove mistakes unless the new trajectories show they're no longer common.

OUTPUT JSON OBJECT:
{{ "skills": [...], "mistakes": [...], "removed_skill_titles": [...], "reasoning": "..." }}

Be concise but precise. Focus on physics insights grounded in the trajectories.
\end{lstlisting}

\subsubsection{Cross-Level Synthesis --- Full (Claude)}
\label{app:prompt:cross_level_full}

Used for the catapult cross-level result in Table~\ref{tab:cross_level} where the evolver is Claude Sonnet~4.6 and three source banks are consumed (\texttt{down\_to\_earth}, \texttt{two\_body\_problem}, \texttt{pass\_the\_parcel}).
The eight hard constraints --- citation requirement, no invented coordinates, target-side entities only, no platitudes, mandatory transfer rationale, confidence calibration, no redundancy, mistake constraints --- are enumerated explicitly so that every synthesised skill remains traceable to the source skill IDs that motivated it.

\begin{lstlisting}[style=prompt]
You are an expert physics analyst performing CROSS-LEVEL OFFLINE SKILL TRANSFER.

You will be given expert skill banks from {n_sources} SOURCE physics levels.
Each source bank was distilled and evolved over many rounds of an agent solving
that level. You will also be given a factual description of a TARGET level
that the agent has NEVER attempted yet -- no target-level trajectories,
no target-level skills, no target-level success/failure data exist.

Your task: synthesise a TARGET skill bank by extracting transferable physics
PRINCIPLES from the source banks, re-grounding them in the target scene, and
predicting which principles the target level will reward.

==========================================================================
**Environment (shared across all levels)**: 2D physics simulation (Box2D).
Gravity = -9.8 m/s^2. World bounds [-5, 5] on both axes. The agent places ONE
red ball at (x, y) with a chosen radius (0.1 <= r <= 2.0) at t=0; the simulation
then runs to completion. Mass scales as r^3; momentum and impact force scale
with both mass and the ball's velocity at first contact.
==========================================================================

=== SOURCE LEVELS ===

{source_blocks}

=== TARGET LEVEL: {target_level} ===

{target_block}

=== TASK ===

Produce a SKILL BANK for the target level. The bank must contain:
- 6 to 10 SKILLS (physics principles you predict will help on the target).
- 2 to 4 MISTAKES (anti-patterns transferable from source failures).

=== HARD CONSTRAINTS ===

1. **Every skill MUST cite source skills.** For each output skill, include
   `source_skills`: a non-empty list of objects of the form
   `{{"source_level": "<level>", "skill_id": "<id>"}}`, listing the source-bank
   entries that motivated this skill. If you cannot cite at least one source
   skill, do NOT emit the skill -- synthesise something else instead.

2. **No invented coordinates.** You have NEVER seen the target level solved.
   Do NOT predict specific (x, y, r) values that "work" for the target. The
   `example` field, if used, must describe a *qualitative* placement never specific numbers.

3. **Use target-level entities.** Phrase each skill in terms of the entities
   that exist in the TARGET scene description above. Do NOT mention source-
   level-only entities (e.g. "ramp", "inverted basket") in the principle or
   when_to_apply, even if those entities motivated the transfer. The
   `transfer_rationale` field is where you may name source entities to
   explain the analogy.

4. **No platitudes.** Skills like "use physics intuition", "consider gravity",
   "be careful" are forbidden. Each skill must name a SPECIFIC mechanism,.

5. **Each skill must include a `transfer_rationale`** (1-2 sentences) explaining
   what physics primitive bridges the source observation to the target
   prediction. This is the audit trail for your synthesis.

6. **Confidence calibration.** Weight a skill's confidence by:
   - How many source banks corroborate the underlying primitive (more = higher).
   - How directly the primitive applies to the target scene (direct = higher).
   - How load-bearing the source skill was (high source confidence = higher).
   Confidence is a float in [0.1, 1.0]. Assign 0.7+ only when the primitive
   appears in >=2 source banks AND the target scene clearly invokes it.

7. **Avoid redundancy.** Each skill must capture a DISTINCT mechanism. Do not
   emit two skills that are paraphrases of the same idea.

8. **Mistakes**: derive each from source-bank mistakes or from the failure
   modes those mistakes imply. Apply the same hard constraints (cite sources,
   no invented coordinates, target-side entities, specific mechanism).

=== OUTPUT FORMAT (single JSON object) ===

```json
{{
  "skills": [
    {{
      "title": "<3-7 word name>",
      "principle": "<2-3 sentence physics insight, in target-level terms>",
      "when_to_apply": "<specific condition, in target-level terms>",
      "example": "<qualitative placement description, NO numbers -- or omit>",
      "confidence": <float in [0.1, 1.0]>,
      "source_skills": [
        {{"source_level": "<level>", "skill_id": "<id>"}}
      ],
      "transfer_rationale": "<1-2 sentences naming the bridging physics primitive>"
    }}
  ],
  "mistakes": [
    {{
      "description": "<1 sentence in target-level terms>",
      "why_it_happens": "<1 sentence root cause>",
      "how_to_avoid": "<1-2 sentence concrete fix>",
      "source_skills": [
        {{"source_level": "<level>", "skill_id": "<id>"}}
      ],
      "transfer_rationale": "<1-2 sentences>"
    }}
  ],
  "synthesis_reasoning": "<2-4 sentences: which source levels contributed most heavily, which primitives you treated as load-bearing, which source skills you deliberately did NOT transfer and why>"
}}
```

Be precise and specific. Do not pad. Do not produce extra fields. Output ONLY
the JSON object inside a single fenced ```json block.
\end{lstlisting}

\subsubsection{Cross-Level Synthesis --- Slim (Qwen-7B)}
\label{app:prompt:cross_level_qwen}

Used for the Falling Into Place and Two Body Problem cross-level rows of Table~\ref{tab:cross_level}, where the evolver is Qwen-2.5-7B.
The slim variant drops audit fields (\texttt{source\_skills}, \texttt{transfer\_rationale}), reduces to three rules, and tightens form constraints on \texttt{principle} / \texttt{when\_to\_apply} / \texttt{example} to prevent the platitude-and-hallucination failure mode the long prompt produces on smaller models (e.g.\ inventing a lever on a level with no pivot).
Output budget is reduced to 4--6 skills and 1--3 mistakes.

\begin{lstlisting}[style=prompt]
You are extracting transferable physics PRINCIPLES from one or more source level skill banks for a NEW target level.

ENVIRONMENT: 2D physics (Box2D), gravity = -9.8 m/s^2, bounds [-5, 5]. Agent places ONE red ball (x, y, radius 0.1-2.0) at t=0; simulation runs to completion.

=== SOURCE LEVELS ===

{source_blocks}

=== TARGET LEVEL: {target_level} ===

{target_block}

=== TASK ===

Output a skill bank for the target level. Produce 4-6 skills and 1-3 mistakes.

RULES (FOLLOW STRICTLY):

1. SKILLS ARE PRINCIPLES, NOT COMMANDS. Each skill states a physics observation, mechanism, or relationship the agent can REASON FROM. Skills MUST NOT tell the agent what to do. Each field has a strict FORM:
   - `principle`: starts with a condition or relationship ("When X holds, Y occurs", "X scales with Y", "A and B differ in Z"). NEVER an imperative verb. FORBIDDEN openings: place / position / put / drop / move / set / use / adjust.
   - `when_to_apply`: describes a REASONING context the agent enters ("When evaluating whether...", "When comparing alternatives that..."). NEVER a target outcome to achieve ("when you need to tip...", "when knocking the bar off...").
   - `example`: contrasts two physical scenarios or illustrates the principle abstractly. NEVER a placement, never a specific scene action.

2. Each skill names a SPECIFIC mechanism. FORBIDDEN platitudes: "use physics", "place strategically", "consider gravity", "adjust as needed", "be careful".

3. Use ONLY entities present in the TARGET scene above. Do not mention source-only entities anywhere. If the target scene has no pivot, do NOT write skills about levers or torque-around-a-pivot.

4. NO coordinates and NO placements anywhere in any field.

5. Mistakes describe MISCONCEPTIONS the agent might form (wrong mental models), NOT actions to avoid. `how_to_avoid` is a CORRECTED UNDERSTANDING, not a placement instruction.

OUTPUT FORMAT (single JSON object, fenced ```json block, no extra fields):
```json
{{
  "skills": [
    {{"title": "...", "principle": "...", "when_to_apply": "...", "example": "...", "confidence": 0.X}}
  ],
  "mistakes": [
    {{"description": "<misconception, NOT an action>", "why_it_happens": "...", "how_to_avoid": "<corrected understanding, NOT a placement command>"}}
  ],
  "synthesis_reasoning": "<1-2 sentences: which source primitive transferred>"
}}
```

Be concise. Output ONLY the JSON object.
\end{lstlisting}

\subsection{Reflexion}
\label{app:systemprompts:reflexion}

After a failed trial, the reflection model receives the following prompt, built from the failed \textsc{ReAct} trajectory (observations truncated to 600 characters by head and tail) and any earlier reflections on the same seed.

\begin{lstlisting}[style=prompt]
SYSTEM:
You are analyzing a failed attempt at a 2D physics puzzle.

You will receive the full Thought/Action/Observation trajectory from a single failed attempt by a ReAct-style agent, followed by the final outcome and any prior reflections accumulated on this same task.

Your job: produce a short reflection (≤5 sentences) that the agent will read before its next attempt on the SAME task. Cover:
(a) which strategy/approach the agent pursued in this attempt;
(b) the specific kinematic, geometric, or procedural reason it failed (cite concrete coordinates, distances, or contact events from the observations);
(c) one concrete different action or strategy to try next -- be specific (object names, approximate (x, y, radius), expected mechanism).

Hard rules:
- Do NOT repeat lessons that already appear in the prior reflections list.
- Do NOT re-state the goal or the puzzle rules. ...

USER:
Outcome: FAILURE.
Final observation:
{final_observation}

Trajectory:
--- Step 1 ---
Thought: {thought}
Action: {action}
Observation: {observation}
...

## Prior reflections (do not repeat these)
1. {prior_reflection_1}
...

Write the reflection now. Be specific and concise (≤5 sentences).
\end{lstlisting}

The returned reflection is appended to the level's \textsc{ReAct} system prompt for the next trial:

\begin{lstlisting}[style=prompt]
{ReAct system prompt of the level}

## Past Lessons (from prior failed attempts on this seed)
1. {reflection_1}
...
Use these lessons to choose a different approach this trial. Do not repeat the same failed strategy.
\end{lstlisting}

\subsection{Episodic RAG}
\label{app:systemprompts:rag}

Episodic RAG makes no LLM call to build its memory. The retrieved successful and unsuccessful rollouts are rendered deterministically and appended to the level's \textsc{ReAct} system prompt; the \texttt{Past rollout} section repeats once per retrieved record.

\begin{lstlisting}[style=prompt]
{ReAct system prompt of the level}

## Retrieved Past Rollouts (from your own past attempts on this level)

### Past rollout {i} — {SUCCESS|FAILURE} (level {level}, seed {seed})
Level: {level}  |  Seed: {seed}
Verifier: {SUCCESS|FAILURE}
Iterations used: {iterations}
Simulator calls: {simulator_calls}
Shaped reward: {reward}
Final placement: x={x}, y={y}, radius={radius}
Final observation: {final_observation}

Rollout:
(showing last 12 of {total_steps} steps)
--- Step 1 ---
Thought: {thought}
Action: {action}
Observation: {observation}
...

These are complete past attempts with the environment's verdict. Use the successful rollout as a template where the geometry matches, and avoid repeating what the unsuccessful rollout did.
\end{lstlisting}

\subsection{MemRL}
\label{app:systemprompts:memrl}

\paragraph{Retrieval key.} Both Episodic RAG and MemRL embed a geometry intent: one line per scene object (position, extents, angle, dynamic flag) followed by the offsets between the objects named by the goal. For Catapult, seed 1:

\begin{lstlisting}[style=prompt]
basket +3.50 -3.33 a+8.97 dyn
black_ball -2.46 +4.60 r+0.40 static
black_platform -1.52 -3.71 l-3.02 r-0.02 t-3.71 b-3.71 a+0.00 static
blue_ball +3.50 -2.75 r+0.38 dyn
...
green_ball->blue_ball dx+6.35 dy-0.94 d+6.42
green_ball->gray_ball dx+1.92 dy-1.10 d+2.22
...
\end{lstlisting}

\paragraph{Memory-building prompts.} After each episode MemRL makes one LLM call to turn the trajectory into experience text: a script for a successful episode and a reflection for a failed one.

\begin{lstlisting}[style=prompt]
SYSTEM (successful episode):
You are analyzing a SUCCESSFUL attempt at a 2D physics placement puzzle.

You will receive the task (level description and scene geometry) and the agent's Thought/Action/Observation trajectory.

Produce a concise, high-level script that captures the essential steps and decision points. The script must be:
1. Generic enough to apply to similar puzzle instances (different seeds, shifted geometry).
2. Specific enough to provide useful guidance (name objects, give approximate coordinates or offsets, state the mechanism).
3. At most 5 high-level steps.
4. Focused on strategy and key decisions, not on individual tool calls.

Output ONLY the numbered script. ...

SYSTEM (failed episode):
You are analyzing a FAILED attempt at a 2D physics placement puzzle.

You will receive the task (level description and scene geometry) and the agent's Thought/Action/Observation trajectory.

This attempt failed. Analyze what went wrong and suggest improvements for future similar tasks. Cover:
1. Incorrect assumptions the agent made about the physics or geometry.
2. Steps to improve — cite concrete coordinates, distances, or contact events from the observations.
3. What to avoid next time.

Keep it under 8 sentences. Output ONLY the reflection text. ...

USER (both):
Task:
{intent}

Trajectory:
{rendered_last_12_steps}
\end{lstlisting}

\paragraph{Actor prompt.} The selected memories are ordered by their Phase-B score and appended to the level's \textsc{ReAct} system prompt. A successful record's experience is \texttt{Task: \{level\} (seed \{seed\})}, then \texttt{SCRIPT:} \texttt{\{script\}}, then \texttt{TRAJECTORY:} \texttt{\{rendered\_last\_12\_steps\}}; a failed record's is shown below.

\begin{lstlisting}[style=prompt]
{ReAct system prompt of the level}

## Archived Memories (from your own past experiences)

### Memory {i} — {SUCCESS|FAILURE} (seed {seed}, utility Q={Q}, updates={q_visits})
TASK REFLECTION:
Task: {level} (seed {seed})

What went wrong:
{reflection}

Failed approach:
{rendered_last_12_steps}
...

These experiences were retrieved because they are similar to the current scene and have been useful in the past. Adapt them to this seed's geometry rather than copying coordinates blindly.
\end{lstlisting}


\newpage
\section{Prompts for the Other Benchmarks}
\label{app:ext_prompts}

\lstdefinestyle{promptwrap}{%
  style=prompt,%
  basicstyle=\ttfamily\scriptsize,%
  columns=fullflexible,%
  keepspaces=true,%
  breaklines=true,%
  breakatwhitespace=false,%
  postbreak=\mbox{\ensuremath{\hookrightarrow}\space},%
  showstringspaces=false%
}

This appendix reproduces the actor, reflector, summariser and evolver
prompts of every arm --- \textsc{Zero-shot}, Reflexion, Episodic RAG, MemRL and
\textsc{HExA} --- on the interactive games
(\S\ref{app:ext_prompts:games}), the reasoning and planning tasks
(\S\ref{app:ext_prompts:reasoning}) and \textsc{DiscoverPhysics}
(\S\ref{app:ext_prompts:discovery}); the environments themselves are described in
Appendix~\ref{app:external_envs}. Long passages are abbreviated with \texttt{...}.
Fields in braces, such as \texttt{\{problem\}}, are filled at run time; a listing
with several parts separates them with banners, which are not part of the prompt.

\subsection{Interactive games: Mastermind and Maze}
\label{app:ext_prompts:games}
{\sloppy All five arms share one actor. It receives the multi-episode system prompt
of~\citep{lin2026orbit} and, as user messages, the raw text the environment returns; every per-turn user
message is identical across arms. Each memory arm differs from \textsc{Zero-shot}
only in one block appended to the system message as
\texttt{SYSTEM + "\textbackslash n\textbackslash n" + BLOCK}, so the system prompt
below is the \textsc{Zero-shot} prompt when \texttt{\{BLOCK\}} is empty. The chat is
cumulative over the three episodes of an instance, except that Reflexion clears it
back to the system message at each episode boundary. The reflector, the MemRL
summariser and the \textsc{HExA} evolver are the same model as the actor in each
block (\texttt{gpt-oss-120b} or Qwen3-8B); the \textsc{HExA} evolver is sent a
single user message with no system prompt.\par}

\subsubsection{Actor system prompt (all arms)}
\label{app:ext_prompts:games:actor_system}
{\sloppy \texttt{\{BLOCK\}} is empty for \textsc{Zero-shot} and is the injected block of
\S\ref{app:ext_prompts:games:reflexion_block} (Reflexion),
\S\ref{app:ext_prompts:games:rag} (Episodic RAG),
\S\ref{app:ext_prompts:games:memrl_block} (MemRL) or
\S\ref{app:ext_prompts:games:hexa_block} (\textsc{HExA}) otherwise.\par}

\begin{lstlisting}[style=promptwrap]
============================================================
  SYSTEM PROMPT
============================================================
You are solving the same task across multiple episodes with a fixed ...

{BLOCK}
\end{lstlisting}

\subsubsection{Mastermind environment messages (user turns, all arms)}
\label{app:ext_prompts:games:mastermind_env}
{\sloppy Configuration: code length 3, digits 1--6, no duplicates, 3 guesses per
episode. \texttt{\{EXAMPLE\}} is a random distinct-digit guess such as
\texttt{\textbackslash boxed\{4 3 2\}}.\par}

\begin{lstlisting}[style=promptwrap]
============================================================
  USER MESSAGE: OPENING OF EPISODE {k}
============================================================
[Episode {k}] New episode begins.
You are playing Mastermind.
You have to guess the secret 3 digit code within 3 turns.
The code consists of digits from 1 to 6 (inclusive). Duplicate numbers are not allowed.
After you enter your guess, I will say mark your guess with black and white pegs, where
  a black peg indicates a correct digit in the correct position, while a white peg
  indicates a correct digit in the wrong position.
After thinking, format your final answer inside \boxed{...}, for example, {EXAMPLE}.
As you play, the history of your guesses will be appended below. Use the information to
  complete the game before you run out of guesses.
Enter your first guess to start the game.

============================================================
  USER MESSAGE: FEEDBACK AFTER EACH GUESS
============================================================
At turn {t}, you guessed [{d1}, {d2}, {d3}]. This guess receives {B} black peg(s) and
  {W} white peg(s).Enter your guess.

============================================================
  USER MESSAGE: OTHER PER-TURN REPLIES
============================================================
At turn {t}, you did not provide a valid guess.Enter your guess.
At turn {t}, you guessed [{..}] which has {n} entries but the code has length 3.Enter
  your guess.
At turn {t}, you guessed [{..}], which has numbers outside the range 1 to 6.Enter your
  guess.
At turn {t}, you guessed [{..}], which has been already guessed before.Enter your guess.

============================================================
  USER MESSAGE: END OF EPISODE
============================================================
Congratulations! You guessed the correct code [{d1}, {d2}, {d3}] in {t} turns.Enter
  your guess.
You have reached the maximum number of turns.Enter your guess.

============================================================
  USER MESSAGE: EPISODE BOUNDARY (ONE MESSAGE)
============================================================
{TERMINAL_REPLY_OF_EPISODE_k}

[Episode {k+1}] New episode begins.
You are playing Mastermind.
... (opening text as above) ...
\end{lstlisting}

\subsubsection{Maze environment messages (user turns, all arms)}
\label{app:ext_prompts:games:maze_env}
{\sloppy Configuration: $6{\times}6$ to $7{\times}7$ mazes, 9 moves per episode;
\texttt{\{up\}}, \texttt{\{down\}}, \texttt{\{left\}} and \texttt{\{right\}} are each
one of \texttt{wall}, \texttt{path} or \texttt{goal}.\par}

\begin{lstlisting}[style=promptwrap]
============================================================
  USER MESSAGE: OPENING OF EPISODE {k}
============================================================
[Episode {k}] New episode begins.
You are a maze-solving agent. Your goal is to navigate from the START position to the
  GOAL position in the fewest turns possible. You are at the START position. Now you
  are at position ({x}, {y}) in the maze. Around you, up leads to {up}, down leads to
  {down}, left leads to {left}, and right leads to {right}.
Output your next move from up/down/left/right within \boxed{}.

============================================================
  USER MESSAGE: AFTER A LEGAL MOVE
============================================================
You move to {action}. Now you are at position ({x}, {y}) in the maze. Around you, up
  leads to {up}, down leads to {down}, left leads to {left}, and right leads to
  {right}.
Output your next move from up/down/left/right within \boxed{}.

============================================================
  USER MESSAGE: OTHER PER-TURN REPLIES
============================================================
You hit the wall. Please choose another direction.
Invalid action. Please try again.

============================================================
  USER MESSAGE: END OF EPISODE
============================================================
Congratulations! You arrived at the goal! Let's play again.

You have reached the maximum number of turns (9). Let's play again.

============================================================
  USER MESSAGE: EPISODE BOUNDARY (ONE MESSAGE)
============================================================
{TERMINAL_REPLY_OF_EPISODE_k}

[Episode {k+1}] New episode begins.
You are a maze-solving agent. ... (opening text as above) ...
\end{lstlisting}

\subsubsection{Reflexion: reflector prompts}
\label{app:ext_prompts:games:reflexion}
{\sloppy The reflector runs only after a failed episode. \texttt{\{PAST\_BLOCK\}} is
empty when there are no earlier reflections; \texttt{\{TRIAL\}} renders the finished
episode, and the \texttt{(episode ended)} marker appears only on its last turn.
Observations longer than 400 characters are cut to their first and last 200.\par}

\begin{lstlisting}[style=promptwrap]
============================================================
  SYSTEM PROMPT (MASTERMIND)
============================================================
You are analysing a failed attempt at Mastermind.

The secret is a 3-digit code using digits 1-6 with no repeats. ...

Write a short reflection (at most 5 sentences) that the agent will ...
(a) which guesses it made and what the feedback ruled in or out;
(b) where its deduction went wrong -- a guess inconsistent with earlier ...
(c) a concrete opening guess and follow-up policy for the next attempt.

Hard rules:
- The agent will NOT have the failed transcript. ...
- Do NOT state or guess the secret code as though you ...
- Do NOT repeat lessons already listed as prior reflections.
- Output ONLY the reflection text. No preamble, no headings, no markdown.

============================================================
  SYSTEM PROMPT (MAZE)
============================================================
You are analysing a failed attempt at a grid maze navigation task.

The agent sees only its own (x, y) position and whether each ...

Write a short reflection (at most 5 sentences) that the agent will ...
(a) what search or movement strategy it actually followed;
(b) the specific reason that failed -- cite concrete coordinates, the cells ...
(c) one concrete different strategy for the next attempt, naming coordinates or ...

Hard rules:
- The agent will NOT have the failed transcript. ...
- Do NOT invent map knowledge the observations did not show.
- Do NOT repeat lessons already listed as prior reflections.
- Output ONLY the reflection text. No preamble, no headings, no markdown.

============================================================
  USER PROMPT (BOTH GAMES)
============================================================
Outcome: FAILURE.

What happened:
{TRIAL}{PAST_BLOCK}

Write the reflection now. At most 5 sentences.

============================================================
  {PAST_BLOCK} (WHEN PRIOR REFLECTIONS EXIST)
============================================================

Prior reflections (do not repeat these):
1. {REFLECTION_1}
2. {REFLECTION_2}

============================================================
  {TRIAL}
============================================================

=== Episode {k} ===
turn {t}: {OBSERVATION} ...

Verdict: episode {k} FAILED after {n} turns.
\end{lstlisting}

\subsubsection{Reflexion: lessons block}
\label{app:ext_prompts:games:reflexion_block}
{\sloppy Appended to the actor system prompt from episode~2 on; the list holds one
lesson per failed episode so far.\par}

\begin{lstlisting}[style=promptwrap]
============================================================
  INJECTED BLOCK
============================================================
## Lessons from your previous attempts at THIS task You have attempted ...

1. {REFLECTION_1}
2. {REFLECTION_2}

Use these lessons to try a different approach. ...
\end{lstlisting}

\subsubsection{Episodic RAG: injected block and stored record}
\label{app:ext_prompts:games:rag}
{\sloppy Episodic RAG makes no LLM call besides the actor. It retrieves one success
and one failure (the top two by similarity if either class is empty), and each
stored record is the verbatim rollout, at most 12 turns per episode with observations
cut to 600 characters. The \texttt{Best guess placed} line appears on Mastermind only,
the position suffix on \texttt{Result} on Maze only, and the \texttt{NOTE} line only
when every episode repeated the same action sequence. \texttt{\{outcome label\}} is
\texttt{GOAL}, \texttt{wall}, \texttt{invalid}, \texttt{budget exhausted} or
\texttt{moved} on Maze, and \texttt{SOLVED}, \texttt{rejected},
\texttt{out of guesses} or the black/white peg counts on Mastermind.\par}

\begin{lstlisting}[style=promptwrap]
============================================================
  INJECTED BLOCK
============================================================
## Retrieved Past Attempts (your own transcripts on other instances of this task)

### Past attempt 1 --- {SUCCESS|FAILURE} (task {task}, instance {instance_id})
{RECORD_CONTENT_1}

### Past attempt 2 --- {SUCCESS|FAILURE} (task {task}, instance {instance_id})
{RECORD_CONTENT_2}

Each block above is a complete past attempt at a different instance ...

============================================================
  STORED RECORD ({RECORD_CONTENT_i})
============================================================
Task: {task}  |  Instance: {instance_id}
Verifier: {solved first at episode N | never solved} (R={success_count}/3) ...

Rollout:
== Episode 1 --- {SOLVED|FAILED} ({n} turns) == ...
\end{lstlisting}

\subsubsection{MemRL: memory-building prompts}
\label{app:ext_prompts:games:memrl_build}
{\sloppy After each instance the summariser writes a \emph{script} (success, temperature
0.0) or a \emph{reflection} (failure, temperature 0.3). \texttt{\{INTENT\}} is the
opening observation of episode~1 without the episode header (with the example guess
masked on Mastermind, and only the position clause kept on Maze).\par}

\begin{lstlisting}[style=promptwrap]
============================================================
  SYSTEM PROMPT, SUCCESS (MASTERMIND)
============================================================
You are analyzing a SUCCESSFUL attempt at Mastermind (a hidden 3-digit code, ...

The agent played ONE instance of this task for THREE episodes. ...

The hidden part of the task is different on every instance. ...

After each guess the agent is told how many digits are correct ...

Produce a concise, high-level script capturing the guessing strategy that worked. It
  must be:
1. Generic across codes: about how to choose an opening guess and ...
2. Specific enough to act on: state the inference rule, not a vague intention.
3. At most 5 high-level steps.
4. Explicit about what the agent carried from earlier episodes, if anything.

Do NOT state the secret code or the winning guess.

Output ONLY the numbered script. No preamble, no headers, no markdown fences.

============================================================
  SYSTEM PROMPT, FAILURE (MASTERMIND)
============================================================
You are analyzing a FAILED attempt at Mastermind (a hidden 3-digit code, ...

The agent played ONE instance of this task for THREE episodes. ...

The hidden part of the task is different on every instance. ...

After each guess the agent is told how many digits are correct ...

This attempt failed. Analyze what went wrong and what a future attempt ...
1. Whether any guess was inconsistent with the peg feedback the agent ...
2. Whether episodes 2 and 3 repeated guesses from episode 1 instead ...
3. What to avoid next time, as a rule that applies to any code.

Do NOT state this instance's secret code.

Keep it under 8 sentences. ...

============================================================
  SYSTEM PROMPT, SUCCESS (MAZE)
============================================================
You are analyzing a SUCCESSFUL attempt at a hidden-maze navigation task.

The agent played ONE instance of this task for THREE episodes. ...

The hidden part of the task is different on every instance. ...

The agent sees only its own coordinates and whether each of the ...

Produce a concise, high-level script capturing the search behaviour that worked. It
  must be:
1. Generic across mazes: about how to explore, backtrack and use what ...
2. Specific enough to act on: name the decision rule and the ...
3. At most 5 high-level steps.
4. Explicit about what changed between episodes, if anything did.

Do NOT state the route, the goal coordinates, or any specific sequence of moves.

Output ONLY the numbered script. No preamble, no headers, no markdown fences.

============================================================
  SYSTEM PROMPT, FAILURE (MAZE)
============================================================
You are analyzing a FAILED attempt at a hidden-maze navigation task.

The agent played ONE instance of this task for THREE episodes. ...

The hidden part of the task is different on every instance. ...

The agent sees only its own coordinates and whether each of the ...

This attempt failed. Analyze what went wrong and what a future attempt ...
1. The search behaviour that wasted turns --- revisiting cells, walking into ...
2. Whether episodes 2 and 3 changed anything after episode 1 failed, ...
3. What to avoid next time, as a rule that applies to any maze.

Do NOT state this maze's route or goal position.

Keep it under 8 sentences. ...

============================================================
  USER PROMPT (BOTH GAMES)
============================================================
Task:
{INTENT}

Trajectory:
{TRAJECTORY_TEXT}
\end{lstlisting}

\subsubsection{MemRL: injected block and stored records}
\label{app:ext_prompts:games:memrl_block}
{\sloppy The top three memories are injected, labelled with their learned utility
$Q$. \texttt{\{TRAJECTORY\}} uses the Episodic RAG rollout renderer of
\S\ref{app:ext_prompts:games:rag}, from \texttt{== Episode 1} down.\par}

\begin{lstlisting}[style=promptwrap]
============================================================
  INJECTED BLOCK
============================================================
## Archived Memories (from your own past experiences on this task)

### Memory 1 --- {SUCCESS|FAILURE} (instance {instance_id}, utility Q={q:+.3f},
  updates={q_visits})
{RECORD_CONTENT_1}

### Memory 2 --- ...
{RECORD_CONTENT_2}

### Memory 3 --- ...
{RECORD_CONTENT_3}

These were retrieved because they have been useful before on instances like this one.
  ...

============================================================
  STORED RECORD (SUCCESS)
============================================================
Task: {task}  |  Instance: {instance_id}

SCRIPT:
{SCRIPT}

TRAJECTORY:
{TRAJECTORY}

============================================================
  STORED RECORD (FAILURE)
============================================================
TASK REFLECTION:
Task: {task}  |  Instance: {instance_id}

What went wrong:
{REFLECTION}

Failed approach:
{TRAJECTORY}
\end{lstlisting}

\subsubsection{\textsc{HExA} evolver: contrastive skill distillation (Mastermind)}
\label{app:ext_prompts:games:mastermind_distill}
{\sloppy Round~0, pass~1, on up to five successful and five failed instances.\par}

\begin{lstlisting}[style=promptwrap]
============================================================
  USER PROMPT
============================================================
You are analysing an agent playing Mastermind and distilling what separates its
successes from its failures into reusable skills.

**Task: mastermind**
The agent must find a hidden code of distinct digits. ...

**Instance reward** grades *when* the code was first cracked, not how many
episodes succeeded:
- near +1.00 -- solved on episode 1, in few guesses. The cold-start case.
- +0.60 -- first solved on episode 2. Cross-episode deduction worked.
- +0.30 -- first solved on episode 3. It worked, slowly.
- negative -- never solved. ...
- -1.00 -- never solved, and every episode repeated the same guesses. ...

=== SOLVED ({n_successes} instances) ===
{success_block}

=== NOT SOLVED ({n_failures} instances) ===
{failure_block}

---

Compare them. Each block shows every guess and the pegs it drew, ...

Write the most general skill the comparison supports. ...

**What the agent is already told.** Every agent receives this system prompt ...

    "You are solving the same task across multiple episodes with a fixed ...

and the environment restates the rules of the game -- code length, ...

**The test.** Ask of each candidate skill: ...

Bank only method the prompt does NOT supply: ...

Write every skill as guidance for a DIFFERENT hidden code. ...

Concretely, NEVER write a skill that contains:
- a specific code or digit sequence to try ("start with 1 ...
- a claim about which digits are in the code,
- an opening guess justified by what worked on these instances rather ...

A skill that would stop working if the hidden code changed is ...

OUTPUT JSON OBJECT:
{
  "skills": [
    ...
  ]
}
\end{lstlisting}

\subsubsection{\textsc{HExA} evolver: mistake extraction (Mastermind)}
\label{app:ext_prompts:games:mastermind_mistakes}
{\sloppy Round~0, pass~2, on up to eight failed instances. Mastermind has no partial-skill field.\par}

\begin{lstlisting}[style=promptwrap]
============================================================
  USER PROMPT
============================================================
You are analysing failed Mastermind attempts to name the mistakes behind them.

**Task: mastermind**
The agent must find a hidden code of distinct digits. ...

=== FAILED ATTEMPTS ===
{failure_block}

---

Name what these agents did wrong, in terms of the peg feedback ...

**What the agent is already told.** Every agent receives this system prompt ...

    "You are solving the same task across multiple episodes with a fixed ...

and the environment restates the rules of the game -- code length, ...

**The test.** Ask of each candidate skill: ...

Bank only method the prompt does NOT supply: ...

OUTPUT JSON OBJECT:
{
  "mistakes": [
    ...
  ]
}
\end{lstlisting}

\subsubsection{\textsc{HExA} evolver: bank evolution (Mastermind)}
\label{app:ext_prompts:games:mastermind_evolve}
{\sloppy Rounds~1 onward, once per round, on the round's selected instances
(12 on \texttt{gpt-oss-120b}, 6 on Maze with Qwen3-8B).\par}

\begin{lstlisting}[style=promptwrap]
============================================================
  USER PROMPT
============================================================
You are curating a skill bank for an agent playing Mastermind.

**Task: mastermind**
The agent must find a hidden code of distinct digits. ...

**Instance reward** grades *when* the code was first cracked, not how many
episodes succeeded:
- near +1.00 -- solved on episode 1, in few guesses. The cold-start case.
- +0.60 -- first solved on episode 2. Cross-episode deduction worked.
- +0.30 -- first solved on episode 3. It worked, slowly.
- negative -- never solved. ...
- -1.00 -- never solved, and every episode repeated the same guesses. ...

=== CURRENT SKILL BANK ===
{existing_skills_block}

=== NEW EVIDENCE: {n_successes} solved, {n_failures} unsolved instances ===
{new_trajectories_block}

---

**First: what is invariant here?** Read the block above for regularities that ...

**Then generalise before you write anything down.** Look at the invariants you just
  listed. ...

Concretely, before adding an entry:
  - If it fires only on one specific feedback value, one count, ...
  - If two or more entries you are considering differ only in ...
  - Prefer an entry that applies at every turn to several that ...

Then revise the bank against this new evidence. Rules:

1. **Keep** skills the new instances still support, preserving their confidence.
2. **Update** a skill whose wording is right but whose trigger the evidence sharpens.
3. **Add** a skill for a pattern the new instances support and ...
   - **Below the cap, add it.** A free slot is not a ...
   - **At the cap**, an addition must displace something: ...
4. **Remove** skills that are redundant, contradicted by the new instances, tied ...

**What the agent is already told.** Every agent receives this system prompt ...

    "You are solving the same task across multiple episodes with a fixed ...

and the environment restates the rules of the game -- code length, ...

**The test.** Ask of each candidate skill: ...

Bank only method the prompt does NOT supply: ...

Write every skill as guidance for a DIFFERENT hidden code. ...

Concretely, NEVER write a skill that contains:
- a specific code or digit sequence to try ("start with 1 ...
- a claim about which digits are in the code,
- an opening guess justified by what worked on these instances rather ...

A skill that would stop working if the hidden code changed is ...

OUTPUT JSON OBJECT:
{
  "invariants": ["<a regularity that held, with the count supporting it>"],
  "skills": [
    ...
  ],
  "mistakes": [
    ...
  ],
  "removed_skill_titles": ["<title>"],
  "reasoning": "<what was removed, what was added, why>"
}
\end{lstlisting}

\subsubsection{\textsc{HExA} evolver: contrastive skill distillation (Maze)}
\label{app:ext_prompts:games:maze_distill}
{\sloppy Round~0, pass~1, on up to five successful and five failed instances.\par}

\begin{lstlisting}[style=promptwrap]
============================================================
  USER PROMPT
============================================================
You are an expert analyst of agent behaviour in interactive environments, distilling ...

**Setting**: an interactive, partially observed navigation task played over repeated
  attempts. Each ...

This means there are two distinct things a trajectory can teach:
- how to explore an unknown maze efficiently on a cold start, and
- how to convert what episode 1 revealed into a better episode ...

The second is only visible because you are shown all three episodes at once. ...

**Reward** is printed in each instance header and grades *when* the maze ...
- near +1.00 -- solved on episode 1, at or near the ...
- +0.70 to +1.00 -- solved on episode 1 with some wasted moves.
- +0.60 -- first solved on episode 2; +0.30 -- first solved ...
- -0.50 -- never solved, but later episodes tried something genuinely different.
- -1.00 -- never solved, and every episode repeated the same move sequence. ...

So instances are **not** equally informative. ...

**Task: maze**
Navigate a grid maze from a fixed start cell to a fixed goal cell. ...

Below are SOLVED and UNSOLVED instances. ...

=== SOLVED INSTANCES ({n_successes}) ===
{success_block}

=== UNSOLVED INSTANCES ({n_failures}) ===
{failure_block}

---

Your task: by CONTRASTING the solved and unsolved instances, extract the KEY ...

**What the agent is already told.** Every agent in this task receives ...

    "You are solving the same task across multiple episodes with a fixed ...

**The test, which is not a list of banned phrases.** Ask of ...

Applying the test honestly matters more than avoiding particular wording. ...

So a skill about reusing earlier episodes is admissible, but ONLY if ...

Bank only what the prompt does NOT say -- concrete method the ...

Write every skill as guidance for a FRESH, unseen maze: ...

Concretely, NEVER write a skill that contains:
- a sequence of specific moves ("go left, left, then up"),
- specific coordinates ("the goal is near (5, 2)", "avoid (3, 3)"),
- any claim about where the goal is in a particular maze.

A skill that would stop working if the walls were rearranged is ...

For each skill, provide:
- **title**: Short name (3-7 words)
- **principle**: The insight (2-3 sentences). What was done, and why does it work?
- **when_to_apply**: Specific trigger condition (1 sentence)
- **source_seeds**: the instance identifiers this skill was primarily derived from.
  Copy ...

Output a JSON array:
[
  {
    ...
  }
]
\end{lstlisting}

\subsubsection{\textsc{HExA} evolver: mistake extraction (Maze)}
\label{app:ext_prompts:games:maze_mistakes}
{\sloppy Round~0, pass~2, on up to eight failed instances.\par}

\begin{lstlisting}[style=promptwrap]
============================================================
  USER PROMPT
============================================================
You are an expert analyst of agent failures in interactive environments.

**Setting**: an interactive, partially observed navigation task played over repeated
  attempts. Each ...

This means there are two distinct things a trajectory can teach:
- how to explore an unknown maze efficiently on a cold start, and
- how to convert what episode 1 revealed into a better episode ...

The second is only visible because you are shown all three episodes at once. ...

**Reward** is printed in each instance header and grades *when* the maze ...
- near +1.00 -- solved on episode 1, at or near the ...
- +0.70 to +1.00 -- solved on episode 1 with some wasted moves.
- +0.60 -- first solved on episode 2; +0.30 -- first solved ...
- -0.50 -- never solved, but later episodes tried something genuinely different.
- -1.00 -- never solved, and every episode repeated the same move sequence. ...

So instances are **not** equally informative. ...

**Task: maze**
Navigate a grid maze from a fixed start cell to a fixed goal cell. ...

Below are UNSOLVED instances -- three episodes each, none of which reached the goal.

{failure_block}

---

Your task, in two parts.

1. **Mistakes.** Identify the recurring errors that cost these agents the maze. ...

2. **Partial skills.** Some unsolved instances still contain good behaviour -- a ...

**What the agent is already told.** Every agent in this task receives ...

    "You are solving the same task across multiple episodes with a fixed ...

**The test, which is not a list of banned phrases.** Ask of ...

Applying the test honestly matters more than avoiding particular wording. ...

So a skill about reusing earlier episodes is admissible, but ONLY if ...

Bank only what the prompt does NOT say -- concrete method the ...

Write every skill as guidance for a FRESH, unseen maze: ...

Concretely, NEVER write a skill that contains:
- a sequence of specific moves ("go left, left, then up"),
- specific coordinates ("the goal is near (5, 2)", "avoid (3, 3)"),
- any claim about where the goal is in a particular maze.

A skill that would stop working if the walls were rearranged is ...

Output a JSON object:
{
  "mistakes": [
    ...
  ],
  "partial_skills": [
    ...
  ]
}
\end{lstlisting}

\subsubsection{\textsc{HExA} evolver: bank evolution (Maze)}
\label{app:ext_prompts:games:maze_evolve}
{\sloppy Rounds~1 onward, once per round, on the round's selected instances
(12 on \texttt{gpt-oss-120b}, 6 on Maze with Qwen3-8B).\par}

\begin{lstlisting}[style=promptwrap]
============================================================
  USER PROMPT
============================================================
You are curating a skill bank for an agent solving interactive maze navigation.

**Setting**: an interactive, partially observed navigation task played over repeated
  attempts. Each ...

This means there are two distinct things a trajectory can teach:
- how to explore an unknown maze efficiently on a cold start, and
- how to convert what episode 1 revealed into a better episode ...

The second is only visible because you are shown all three episodes at once. ...

**Reward** is printed in each instance header and grades *when* the maze ...
- near +1.00 -- solved on episode 1, at or near the ...
- +0.70 to +1.00 -- solved on episode 1 with some wasted moves.
- +0.60 -- first solved on episode 2; +0.30 -- first solved ...
- -0.50 -- never solved, but later episodes tried something genuinely different.
- -1.00 -- never solved, and every episode repeated the same move sequence. ...

So instances are **not** equally informative. ...

**Task: maze**
Navigate a grid maze from a fixed start cell to a fixed goal cell. ...

=== CURRENT SKILL BANK ===
{existing_skills_block}

=== NEW EVIDENCE: {n_successes} solved, {n_failures} unsolved instances ===
{new_trajectories_block}

---

**First, state what is invariant.** From the pooled counts above, list regularities ...

Then revise the bank against the new evidence:

1. **Keep** skills the new instances still support.
2. **Update** a skill whose trigger condition or reasoning the evidence sharpens.
3. **Add** a skill for a supported, transferable pattern no entry covers ...
4. **Remove** skills that are redundant, contradicted, route-specific, or low
  confidence (< ...
5. Keep mistakes unless the new instances show they are no longer common.
6. **Cite a turn**: for every skill, name an instance and the ...

**Bank only what the actor's system prompt does not already say.** Every ...

OUTPUT JSON OBJECT:
{
  "invariants": [
    ...
  ],
  "skills": [
    ...
  ],
  "mistakes": [
    ...
  ],
  "removed_skill_titles": ["<title of removed skill 1>", ...],
  "confidence_changes": [
    ...
  ],
  "reasoning": "<what was removed, what was added, why -- and for any invariant you
  stated but banked no skill from, name it and say why it is unactionable. If a
  quantity inside a retained skill changed, name the instance that justified it>"
}

Be concise but precise. Focus on transferable exploration strategy grounded in the
  trajectories.
\end{lstlisting}

\subsubsection{\textsc{HExA} evolver inputs: instance and bank rendering}
\label{app:ext_prompts:games:hexa_inputs}
{\sloppy How each instance fills \texttt{\{success\_block\}}, \texttt{\{failure\_block\}}
and \texttt{\{new\_trajectories\_block\}}, and how the current bank fills
\texttt{\{existing\_skills\_block\}}. The \texttt{note:} lines appear only when they
apply. On Maze the evolution prompt (not the round-0 passes) prefixes the individual
transcripts with a cross-instance aggregate of what the agent observed. The bank has
one entry per skill and per mistake, and reads
\texttt{(No existing skills or mistakes yet)} when empty.\par}

\begin{lstlisting}[style=promptwrap]
============================================================
  INSTANCE RENDER (MASTERMIND)
============================================================
=== Instance {instance_id} | reward {reward:+.2f} ({reward_bin}) | first solved on
  episode ...
  note: every episode repeated the same guesses -- this agent did not ...

============================================================
  INSTANCE RENDER (MAZE)
============================================================
=== Instance {instance_id} | reward {reward:+.2f} ({reward_bin}) | solved first on
  episode ...
  note: {n} turn(s) produced no parseable action and were spent doing nothing ...

Episodes solved: {solved}/{total}
Turns used, all episodes:        {histogram} ...

=== INDIVIDUAL TRANSCRIPTS ===

{per-instance renders as above}

============================================================
  EXISTING BANK
============================================================
EXISTING LEVEL-SPECIFIC SKILLS:
  - [{CONFIDENCE_PCT}%] {TITLE} ...

EXISTING MISTAKES:
  - {MISTAKE_DESCRIPTION}
\end{lstlisting}

\subsubsection{\textsc{HExA}: injected skill block}
\label{app:ext_prompts:games:hexa_block}
{\sloppy Skills are ordered by confidence and capped at 10 skills and 5 mistakes, then
truncated to 1{,}000 tokens; entries distilled from failures are tagged
\texttt{Lesson} instead of \texttt{Strategy}. The Maze block is identical with
\texttt{(maze)} in the section header.\par}

\begin{lstlisting}[style=promptwrap]
============================================================
  INJECTED BLOCK
============================================================
=== LEARNED PROBLEM-SOLVING STRATEGIES ===

## Task-Specific Strategies (mastermind)
1. [Strategy: {TITLE_1}] {PRINCIPLE_1} ...

2. [Strategy: {TITLE_2}] {PRINCIPLE_2}
   Apply when: {WHEN_TO_APPLY_2} ...

3. [Lesson: {TITLE_3}] {PRINCIPLE_3}
   Apply when: {WHEN_TO_APPLY_3} ...

## Common Mistakes to Avoid
1. **{DESCRIPTION_1}** ...

2. **{DESCRIPTION_2}**
   Why it happens: {WHY_IT_HAPPENS_2} ...

=== END SKILLS ===
\end{lstlisting}

\subsection{Reasoning and planning tasks}
\label{app:ext_prompts:reasoning}
{\sloppy All arms build the actor prompt with the same task factory's prompt
builder (\texttt{build\_solve\_prompt}; Appendix~\ref{app:external_envs:reasoning}),
sent as a single user message. \textsc{Zero-shot} passes no
guidelines. Reflexion, Episodic RAG, MemRL and \textsc{HExA} place their block in the
\texttt{\{guidelines\}} slot, so each task's prompt is shown once with that slot; the
\textsc{Zero-shot} prompt is the same text without the three lines that start at
\texttt{Before you begin} (for MathGAP, it is \texttt{\{problem\}} alone). Sampling
for every actor call is temperature 0.6, top-$p$ 0.95, reasoning effort high and a
60{,}000-token budget.\par}

\subsubsection{Actor prompt: MathGAP}
\label{app:ext_prompts:reasoning:mathgap}

\begin{lstlisting}[style=promptwrap]
============================================================
  USER PROMPT (ZERO-SHOT)
============================================================
{problem}

============================================================
  USER PROMPT (WITH MEMORY BLOCK)
============================================================
Before you begin, here are your past reflections on the most helpful ...
{guidelines}

Now here is the task:
{problem}
\end{lstlisting}

\subsubsection{Actor prompt: ZebraLogic}
\label{app:ext_prompts:reasoning:zebra}
{\sloppy \texttt{\{output\_schema\_json\}} is the puzzle's JSON answer schema.\par}

\begin{lstlisting}[style=promptwrap]
============================================================
  USER PROMPT
============================================================
Solve the ZebraLogic puzzle below.

# Example Puzzle
There are 3 houses, numbered 1 to 3 from left to right, ...
- Each person has a unique name: Peter, Eric, Arnold
- Each person has a unique favorite drink: tea, water, milk

## Clues for the Example Puzzle
1. Peter is in the second house.
2. Arnold is directly left of the one who only drinks water.
3. The one who only drinks water is directly left of the ...

## Answer to the Example Puzzle
{
  "reasoning": "Given Clue 1, Peter is in House 2. According to Clue 2, Arnold is
  directly left of the one who only drinks water, so Arnold must be in House 1 and
  Peter must drink water. Then Clue 3 implies the person in House 3 drinks milk, so
  Eric is in House 3 and Arnold drinks tea.",
  "solution": {
    ...
  }
}

Before you begin, here are your past reflections on the most helpful ...
{guidelines}

# Puzzle to Solve
{problem}

# Instruction
Now please solve the above puzzle. ...
\end{lstlisting}

\subsubsection{Actor prompt: Tower of Hanoi}
\label{app:ext_prompts:reasoning:hanoi}
{\sloppy \texttt{\{problem\}} is filled from the template in the second block.\par}

\begin{lstlisting}[style=promptwrap]
============================================================
  USER PROMPT
============================================================
You are a helpful assistant. ...

1. Only one disk can be moved at a time.

2. Each move consists of taking the upper disk from one stack ...

3. A larger disk may not be placed on top of a smaller disk.

The goal is to transform the initial configuration into the goal configuration. ...

moves = [[1, 0, 2], [2, 0, 1], [1, 2, 1], [3, 0, 2],
         [1, 1, 0], [2, 1, 2], [1, 0, 2]]

Requirements:

- When exploring potential solutions in your thinking process, always include the ...

- The positions are 0-indexed (the leftmost peg is 0).

- Ensure your final answer includes the complete list of moves in the format:
  moves = [[disk_id, from_peg, to_peg], ...]

Before you begin, here are your past reflections on the most helpful ...
{guidelines}

{problem}

Ensure your final answer includes the complete list of moves in the format: ...

============================================================
  {problem} TEMPLATE
============================================================
I have a puzzle with {num_disks} disks of different sizes with
Initial configuration:

- Peg 0: {initial_peg_0}

- Peg 1: {initial_peg_1}

- Peg 2: {initial_peg_2}

Goal configuration:

- Peg 0: {goal_peg_0}

- Peg 1: {goal_peg_1}

- Peg 2: {goal_peg_2}

Rules:

- Only one disk can be moved at a time.

- Only the top disk from any stack can be moved.

- A larger disk may not be placed on top of a smaller disk.

Remember that the peg contents are listed from bottom to top.

Find the sequence of moves to transform the initial configuration into the goal
  configuration.
\end{lstlisting}

\subsubsection{Actor prompt: Matchstick}
\label{app:ext_prompts:reasoning:matchstick}

\begin{lstlisting}[style=promptwrap]
============================================================
  USER PROMPT
============================================================
You are an expert problem solver. ...

Imagine that each number and operator in the equation that you will ...
- 'I' is composed of one vertical matchstick.
- 'II' is composed of two vertical matchsticks, stacked side by side.
- 'X' is composed of two matchsticks, which are crossed diagonally.
- '=' is composed of two horizontal matchsticks stacked on top of each other.

The equation you will see below will be incorrect. ...

Important constraints: You cannot simply remove a matchstick; you must relocate it ...

Here is an example problem: I = VI - III
The solution is: I = IV - III

Before you begin, here are your past reflections on the most helpful ...
{guidelines}

Now here is the new equation that you have to correct, following ...
{problem}

Please think step by step. ...
\end{lstlisting}

\subsubsection{Reflexion: reflector prompts and lessons block}
\label{app:ext_prompts:reasoning:reflexion}
{\sloppy Three trials per item. After a failed trial the reflector, the only call in any
arm that sends a system message, writes one lesson; the list of prior reflections
grows by one per failed trial. The lessons block fills \texttt{\{guidelines\}} on the
next trial.\par}

\begin{lstlisting}[style=promptwrap]
============================================================
  SYSTEM PROMPT
============================================================
You are analysing a FAILED attempt at a reasoning problem with a ...

You will be shown the problem and the solver's own reasoning-and-answer text. ...

Write a short reflection (at most 5 sentences) that a fresh attempt ...
(a) the specific step where the reasoning went wrong -- an arithmetic ...
(b) what to verify before committing to a final answer next time;
(c) one concrete different approach to try.

Hard rules:
- Do NOT state or imply what the correct answer is; you ...
- Do NOT repeat lessons already listed as prior reflections.
- Output ONLY the reflection text. No preamble, no headings, no markdown.

============================================================
  USER PROMPT (AFTER TRIAL 1)
============================================================
Problem:
{problem_text}

Solver's reasoning and answer:
{raw_text}

Parsed final answer: {parsed_answer}
Verdict: INCORRECT

Write the reflection now. At most 5 sentences.

============================================================
  USER PROMPT (AFTER TRIAL 2; PRIOR LIST GROWS)
============================================================
Problem:
{problem_text}

Solver's reasoning and answer:
{raw_text}

Parsed final answer: {parsed_answer}
Verdict: INCORRECT

Prior reflections (do not repeat these):
1. {reflection_1}

Write the reflection now. At most 5 sentences.

============================================================
  INJECTED BLOCK (TRIAL 2)
============================================================
## Lessons from your previous attempts at THIS problem
You have attempted this exact problem before and failed. You cannot see
those attempts, but you wrote down what went wrong:

1. {reflection_1}

Use these lessons to try a different approach. ...

============================================================
  INJECTED BLOCK (TRIAL 3)
============================================================
## Lessons from your previous attempts at THIS problem
You have attempted this exact problem before and failed. You cannot see
those attempts, but you wrote down what went wrong:

1. {reflection_1}
2. {reflection_2}

Use these lessons to try a different approach. ...
\end{lstlisting}

\subsubsection{Episodic RAG: injected block}
\label{app:ext_prompts:reasoning:rag}
{\sloppy Stored records are the raw rollouts, cut to 3{,}000 characters (head and
tail); no LLM call builds them. One success and one failure are retrieved.\par}

\begin{lstlisting}[style=promptwrap]
============================================================
  INJECTED BLOCK
============================================================
## Retrieved Past Rollouts (from your own past attempts on this level)

### Past rollout 1 --- SUCCESS (level {task_name}, seed {problem_id})
Task: {task_name}  |  Instance: {task_name}:{problem_id} ...

Rollout:
--- Step 1 --- ...

### Past rollout 2 --- FAILURE (level {task_name}, seed {problem_id})
Task: {task_name}  |  Instance: {task_name}:{problem_id} ...

Rollout:
--- Step 1 --- ...

These are complete past attempts with the environment's verdict. ...
\end{lstlisting}

\subsubsection{MemRL: memory-building prompts and injected block}
\label{app:ext_prompts:reasoning:memrl}
{\sloppy Each build call is one user message. Up to three memories are injected.\par}

\begin{lstlisting}[style=promptwrap]
============================================================
  USER PROMPT, SUCCESS (SCRIPT)
============================================================
You are analyzing a SUCCESSFUL solution to a reasoning problem with a ...

You will receive the problem statement and the solver's full reasoning-plus-answer text.

Produce a concise, high-level script that captures the essential solution strategy. ...
1. Generic enough to apply to similar problem instances (different numbers, different
  ...
2. Specific enough to provide useful guidance (name the technique, the key ...
3. At most 5 high-level steps.
4. Focused on strategy and the decisive insight, not a transcription of the arithmetic.

Output ONLY the numbered script. No preamble, no headers, no markdown fences.

Problem:
{problem_text}

Solver's reasoning and answer:
--- Step 1 --- ...

============================================================
  USER PROMPT, FAILURE (REFLECTION)
============================================================
You are analyzing a FAILED attempt at a reasoning problem with a ...

You will receive the problem statement and the solver's full reasoning-plus-answer
  text, ...

This attempt failed. Analyze what went wrong and suggest improvements for future ...
1. The incorrect assumption or arithmetic/logic slip that produced the wrong answer.
2. What to check or verify before committing to a final answer next time.
3. What to avoid repeating.

Keep it under 8 sentences. ...

Problem:
{problem_text}

Solver's reasoning and answer:
--- Step 1 --- ...

============================================================
  INJECTED BLOCK
============================================================
## Archived Memories (from your own past experiences)

### Memory 1 --- SUCCESS (seed {problem_id}, utility Q={q:+.3f}, updates={q_visits})
Task: {task_name}  |  Instance: {task_name}:{problem_id}

SCRIPT:
{script}

TRAJECTORY:
--- Step 1 --- ...

### Memory 2 --- FAILURE (seed {problem_id}, utility Q={q:+.3f}, updates={q_visits})
TASK REFLECTION: ...

What went wrong:
{reflection}

Failed approach:
--- Step 1 --- ...

### Memory 3 --- SUCCESS (seed {problem_id}, utility Q={q:+.3f}, updates={q_visits})
{memory_3_content}

These experiences were retrieved because they are similar to the current scene ...
\end{lstlisting}

\subsubsection{\textsc{HExA} evolver: bank evolution}
\label{app:ext_prompts:reasoning:hexa_evolve}
{\sloppy The only evolver prompt called in the reported runs, once per round, as a
single user message. \texttt{\{level\_name\}} is the task and
\texttt{\{level\_description\}} its description from
\S\ref{app:ext_prompts:reasoning:hexa_inputs}.\par}

\begin{lstlisting}[style=promptwrap]
============================================================
  USER PROMPT
============================================================
You are a teacher curating a strategy bank for a model solving reasoning problems.

**Setting**: a single-turn reasoning benchmark. For each problem the model receives one
  ...

**Reward** is printed in each attempt's header and is **baseline-relative**: ...

- A success on a problem the model rarely solves unaided scores near +1.00. ...
- A success on a problem it almost always solves unaided scores near 0.00. ...
- A failure scores negatively, most negatively when the problem was easy.
- A reward of exactly +1.00 or -1.00 means no baseline was ...

So attempts are **not** equally informative, and a skill drawn from high-reward ...

**Task: {level_name}**
{level_description}

CURRENT SKILL BANK (from previous rounds):
{existing_skills_block}

NEW ATTEMPTS (from the latest round):
Accepted: {n_successes} ...

{new_trajectories_block}

YOUR TASK: evolve the bank using the evidence in the new attempts.

FIRST, before writing any skill, work out the CONTRAST:
- What did the higher-reward attempts do that the lower-reward ones did not?
- Look at the reasoning itself, not just the verdicts: ...
- **If a behaviour appears in both accepted and rejected attempts, it ...

A skill earns its place only if it names something the attempts revealed. ...

RULES:
1. Output the COMPLETE FINAL bank (not a diff) --- include both ...
2. Hard constraints:
   - Maximum 10 total skills for this task.
   - Maximum 5 total mistakes for this task.
3. For each skill you include:
   - If it is a RETAINED skill from the existing bank: ...

Write every skill as guidance for a FRESH, unseen problem of the same task: ...

OUTPUT JSON OBJECT:
{
  "skills": [
    ...
  ],
  "mistakes": [
    ...
  ],
  "removed_skill_titles": ["<title of removed skill 1>", ...],
  "reasoning": "<brief explanation of key changes: what was removed, what was added,
  why>"
}

Be concise but precise. Focus on transferable reasoning strategies grounded in the
  attempts.
\end{lstlisting}

\subsubsection{\textsc{HExA} evolver inputs: task descriptions and rendering}
\label{app:ext_prompts:reasoning:hexa_inputs}
{\sloppy The task description filling \texttt{\{level\_description\}}, how each attempt
fills \texttt{\{new\_trajectories\_block\}}, and how the bank fills
\texttt{\{existing\_skills\_block\}}.\par}

\begin{lstlisting}[style=promptwrap]
============================================================
  TASK DESCRIPTION (MATHGAP)
============================================================
Each problem is an arithmetic word problem about how many of one ...

============================================================
  TASK DESCRIPTION (ZEBRALOGIC)
============================================================
Each problem is a constraint-satisfaction grid puzzle: ...

============================================================
  TASK DESCRIPTION (TOWER OF HANOI)
============================================================
Each problem gives a three-peg Tower of Hanoi position: ...

============================================================
  TASK DESCRIPTION (MATCHSTICK)
============================================================
Each problem shows one arithmetic equation written in Roman numerals with the ...

============================================================
  ATTEMPT RENDER (ACCEPTED)
============================================================
--- Problem {problem_id} | Task: {task_name} | Verdict: CORRECT | Reward: {reward:+.2f}
  ---
  Step 1:
    Thought: {raw_text: reasoning+final, clipped to 6000 chars + '... [reasoning
  truncated]', then middle-elided to 2400 chars}
    Action: {parsed_answer}
    Observation: CORRECT --- the verifier accepted the answer

  Final answer: {parsed_answer}

============================================================
  ATTEMPT RENDER (REJECTED, TRUNCATED)
============================================================
--- Problem {problem_id} | Task: {task_name} | Verdict: INCORRECT | Reward:
  {reward:+.2f} | truncated at the token limit ---
  Step 1:
    Thought: {raw_text: reasoning+final, clipped to 6000 chars + '... [reasoning
  truncated]', then middle-elided to 2400 chars}
    Action: {parsed_answer}
    Observation: INCORRECT --- the verifier rejected the answer (generation hit the
  token limit; the answer may be truncated)

  Final answer: {parsed_answer}

============================================================
  EXISTING BANK (ROUND 0)
============================================================
(No existing skills or mistakes yet)

============================================================
  EXISTING BANK (LATER ROUNDS)
============================================================
EXISTING LEVEL-SPECIFIC SKILLS:
  - [{confidence_pct}%] {title}
      principle: {principle}
      apply when: {when_to_apply}
      example: {example}

EXISTING MISTAKES:
  - {description}
\end{lstlisting}

\subsubsection{\textsc{HExA}: injected skill block}
\label{app:ext_prompts:reasoning:hexa_block}
{\sloppy Fills \texttt{\{guidelines\}}, capped at 1{,}000 tokens; the section header
names the task.\par}

\begin{lstlisting}[style=promptwrap]
============================================================
  INJECTED BLOCK
============================================================
=== LEARNED PROBLEM-SOLVING STRATEGIES ===

## Task-Specific Strategies (mathgap)
1. [Strategy: {title_1}] {principle_1} ...

2. [Strategy: {title_2}] {principle_2}
   Apply when: {when_to_apply_2} ...

## Common Mistakes to Avoid
1. **{mistake_description_1}** ...

=== END SKILLS ===
\end{lstlisting}

\subsection{\textsc{DiscoverPhysics}}
\label{app:ext_prompts:discovery}
{\sloppy All arms share one actor (Claude Sonnet, at most 10 rounds, at least 2 before a
final law, no critic). The arms differ only in a block appended to the end of the
system prompt, and \textsc{Zero-shot} is the system prompt alone. The system prompt
is identical across worlds except for its world-specific block. We compare
\textsc{Zero-shot}, Episodic RAG, MemRL and \textsc{HExA} on this benchmark.\par}

\subsubsection{Actor system prompt (oscillator, all arms)}
\label{app:ext_prompts:discovery:actor_system}
{\sloppy For circle and hubble, the block under \texttt{World-Specific Instructions} is
replaced by the one in \S\ref{app:ext_prompts:discovery:worlds}.\par}

\begin{lstlisting}[style=promptwrap]
============================================================
  SYSTEM PROMPT
============================================================
You are an expert physics and AI research scientist tasked with discovering ...

**Workflow:**
1. Analyse the mission description provided.
2. Design a set of experiments to test your hypotheses.
3. Use the `<run_experiment>` tag to submit your experimental inputs.
4. The system will return the results in an `<experiment_output>` tag.
   - If a returned value is `nan`, ignore it --- it indicates ...
5. Only one action is allowed per round: either `<run_experiment>` or `<final_law>`.
...

(Round budget --- total rounds available and minimum rounds before final submission ...

## World-Specific Instructions

The block below describes this world's topology: ...

### Topology: 2-particle

Two particles in a 2D universe.

Your task: discover the law of motion governing particle 2.

### Control Parameters

| Field | Meaning | Typical range |
|---|---|---| ...

(Some worlds in this topology accept additional optional fields --- e.g. ...

### Input Format (interactive mode only)

```
<run_experiment>
...
```

You may submit several experiments per round in the JSON array. ...

### Output Format

Each experiment returns:

```
<experiment_output>
...
```

`pos1` and `velocity1` are reported for completeness even though particle 1 is held
  fixed.

### `discovered_law` Signature

```python
def discovered_law(pos1, pos2, p1, p2, velocity2, duration, **params):
...
```

The positional arguments must appear in exactly this order. ...

**Noise:** Reported particle **positions may contain Gaussian observation noise of
  unknown scale** ...

## Final Submission

When confident, submit your findings as a single Python function enclosed in
  `<final_law>` tags.

**Submission Requirements:**
1. The function must be named `discovered_law`.
2. The positional signature must match the stub given in the world-specific ...
3. Return what the world-specific stub says to return (typically the final ...
4. If you conclude that one of the input arguments does not ...
5. Constants you are CERTAIN about (e.g. a dimensional prefactor you have ...
6. Constants that remain UNCERTAIN --- exponents, screening lengths, diffusion
  coefficients, wave ...

**Implementation note --- time integration:**
The evaluator calls your `discovered_law` many times --- once per (case x ...

Skeleton:

```python
from scipy.integrate import solve_ivp
...
```

If your law genuinely cannot be expressed as an ODE, vectorise with ...

**Fittable Parameters (optional, recommended when you have uncertain constants):**
Alongside `discovered_law`, you may define a second function `fit_parameters()` that
  returns a ...

```python
def fit_parameters():
...
```

Inside `discovered_law`, read fitted values from `**params` (e.g. ...

**Critical Boundaries:**
- Do NOT include any explanation or commentary inside the `<final_law>` block ...
- In your final-submission round, output ONLY the `<final_law>` block followed by ...

**Explanation Tag (required in the final submission round):**
Alongside your `<final_law>`, you MUST also include a separate `<explanation>` tag
  containing ...

**Reminder:**
The laws of physics in this universe may differ from those in ...

## RUN-SPECIFIC CONSTRAINTS (these override any conflicting numbers above)
- For this session you have EXACTLY 10 round(s) of experiments. ...
- You must run at least 2 round(s) of experiments before submitting a <final_law>.
- If your `discovered_law` integrates the trajectory with a for-loop over some ...
- ALWAYS set `dt > 0.005` in EVERY test simulation AND in ...
\end{lstlisting}

\subsubsection{World-specific instructions (circle, hubble)}
\label{app:ext_prompts:discovery:worlds}

\begin{lstlisting}[style=promptwrap]
============================================================
  CIRCLE
============================================================
### Topology: Centre + ring (constrained input)

A 2D universe containing 11 particles:
- **Particle 0**: starts at the centre `[0, 0]`.
- **Particles 1--10**: arranged equally spaced on a ring of radius `ring_radius`, ...

All 11 particles interact through an unknown field. ...

### Control Parameters

| Field | Meaning | Typical range |
|---|---|---| ...

### Input Format (interactive mode only)

```
<run_experiment>
...
```

You may submit several experiments per round in the JSON array. ...

### Output Format

```
<experiment_output>
...
```

Particle ordering is fixed: ...

### `discovered_law` Signature

```python
def discovered_law(positions, velocities, duration, **params):
...
```

The signature takes raw per-particle initial conditions (not `ring_radius` /
  `initial_tangential_velocity`) --- ...

============================================================
  HUBBLE
============================================================
### Topology: Anchor + ring orbiters + 5 probes (with masses)

A 2D universe containing 26 particles:
- **Particle 0**: a single "anchor" near the centre with very large ...
- **Particles 1--20**: 20 background "orbiter" particles arranged on a ring around ...
- **Particles 21--25**: 5 probe particles you fully control --- position, velocity, AND
  mass.

You are scored on how accurately your law predicts the trajectories of ...

### Control Parameters

| Field | Meaning | Typical range |
|---|---|---| ...

Avoid placing probes exactly on the anchor (origin) or on top of ...

### Input Format (interactive mode only)

```
<run_experiment>
...
```

You may submit several experiments per round in the JSON array. ...

### Output Format

```
<experiment_output>
...
```

Particle ordering: index 0 = anchor, indices 1--20 = ring orbiters, indices ...

### `discovered_law` Signature

```python
def discovered_law(positions, velocities, masses, duration):
...
```

Your function must simulate **all 26** particles forward (the orbiters and anchor ...
\end{lstlisting}

\subsubsection{Actor user messages (all arms)}
\label{app:ext_prompts:discovery:turns}
{\sloppy The first user message, identical for all three worlds, followed by every
message the harness can send on later turns. \texttt{\{law\_stub\}} is the world's
\texttt{discovered\_law} signature and \texttt{\{experiment\_format\}} a one-line
\texttt{<run\_experiment>} example.\par}

\begin{lstlisting}[style=promptwrap]
============================================================
  USER PROMPT (FIRST TURN)
============================================================
You are an expert physics and AI research scientist tasked with discovering ...

============================================================
  USER MESSAGE: ROUND 9 OF 10 WARNING
============================================================
Warning: this is round {round_num} of {max_rounds}. You have one round remaining. Your
  next response MUST be a <final_law> submission --- you will not be able to run
  further experiments after this.

============================================================
  USER MESSAGE: FINAL ROUND (ROUND 10)
============================================================
This is your final round. You MUST now submit your best guess as a <final_law>
  regardless of confidence. Do not run more experiments.

<final_law>
{law_stub}</final_law>

============================================================
  USER MESSAGE: MISSING EXPLANATION
============================================================
Your <final_law> submission is accepted, but you did not include the required
  <explanation> tag. Reply NOW with ONLY a single <explanation>...</explanation> block
  containing a 2--3 sentence plain-English description of the physical system you
  discovered. Do not include any other text or tags.

============================================================
  USER MESSAGE: FINAL LAW BEFORE ROUND 2
============================================================
You have only run {round_num} round(s) of experiments. You must run at least
  {min_rounds} rounds before submitting a final law. Please design and run at least one
  more experiment.

============================================================
  USER MESSAGE: NO RECOGNISED TAG
============================================================
ERROR: No <run_experiment>, <run_mse_fit>, or <final_law> tag found in your response.
  You must respond with one of these XML tags --- no code fences, no markdown, just the
  raw tag.

Option 1 --- run an experiment:
{experiment_format}

Option 2 --- submit your final law:
<final_law> ...

Option 3 --- fit your candidate law's free parameters against the data ...
<run_mse_fit> ...

Respond with the XML tag NOW. No explanation before or after.

============================================================
  USER MESSAGE: EXPERIMENT RESULT
============================================================
<experiment_output>
{results_json}
</experiment_output>

============================================================
  USER MESSAGE: EXPERIMENT ERROR
============================================================
<experiment_output>
Error running experiment: {error}
</experiment_output>

============================================================
  USER MESSAGE: REPLY TO <run_mse_fit>
============================================================
<mse_fit_output>
{"error":"trajectory CSV logging is disabled, so no data is available for MSE fitting
  in this run.","loss_before":null,"loss_after":null,"fitted_params":{},
  "declared_params":{},"n_training":0}
</mse_fit_output>
\end{lstlisting}

\subsubsection{\textsc{HExA}: injected skill block}
\label{app:ext_prompts:discovery:hexa_block}
{\sloppy Skills are sorted by confidence (12 skills and 6 mistakes within a world, 8
and 4 for the cross-world hubble bank).\par}

\begin{lstlisting}[style=promptwrap]
============================================================
  INJECTED BLOCK
============================================================
The following experimentation and reasoning skills were learned from discovery episodes
  on ...

=== LEARNED PHYSICS SKILLS ===

## Cross-World Experimentation Skills (learned on other worlds --- verify here)
{i}. [{Strategy|Lesson}: {skill.title}] {skill.principle} ...

... (repeated per skill)
## Common Mistakes to Avoid ...

... (repeated per mistake)
=== END SKILLS ===
\end{lstlisting}

\subsubsection{\textsc{HExA} evolver: contrastive skill distillation}
\label{app:ext_prompts:discovery:contrastive}
{\sloppy Round~0, pass~1. Sent as a single user message to \texttt{claude-sonnet-5}.\par}

\begin{lstlisting}[style=promptwrap]
============================================================
  USER PROMPT
============================================================
You are an expert experimental physicist distilling agent behavior into concise,
transferable discovery skills.

**Environment**: an LLM agent does open-ended physics discovery. Each episode it is ...

**Reward** per trajectory is a blend of explanation quality and normalized trajectory
  ...

**These skills are CROSS-WORLD**: ...

**Current world (setup only --- the governing law is hidden even from you):**
{level_description}

Below are SUCCESSFUL and FAILED discovery episodes on this world. ...

=== SUCCESSFUL EPISODES ({n_successes}) ===
{success_block}

=== FAILED EPISODES ({n_failures}) ===
{failure_block}

---

By CONTRASTING successes and failures, extract the KEY EXPERIMENTATION SKILLS that
  distinguished ...

For each skill provide:
- **title**: short name (3-7 words)
- **principle**: the transferable method (2-3 sentences). What to do and why ...
- **when_to_apply**: the trigger condition (1 sentence)
- **source_seeds**: seed numbers from the episodes above this skill came from

Output a JSON array:

```json
[
...
```

Extract 4-6 DISTINCT skills. Avoid redundancy. No fitted constant values.
\end{lstlisting}

\subsubsection{\textsc{HExA} evolver: mistake and partial-skill extraction}
\label{app:ext_prompts:discovery:common}
{\sloppy Round~0, pass~2. Sent as a single user message to \texttt{claude-sonnet-5}.\par}

\begin{lstlisting}[style=promptwrap]
============================================================
  USER PROMPT
============================================================
You are an expert at analyzing physics-discovery failures and distilling avoidable
mistake patterns.

**Environment**: an LLM agent does open-ended physics discovery. Each episode it is ...

**Reward** per trajectory is a blend of explanation quality and normalized trajectory
  ...

**These skills are CROSS-WORLD**: ...

**Current world (setup only):**
{level_description}

Below are FAILED discovery episodes.

{failure_block}

---

**Part 1 --- Mistakes**: identify COMMON MISTAKE PATTERNS across these failures. For
  each:
1. what the agent did wrong (experiment design, data interpretation, fitting, or ...
2. WHY (the broken belief or shortcut behind it),
3. a concrete, transferable fix.

**Part 2 --- Partial insights**: ...

Output a JSON object:

```json
{
...
```

Extract 3-5 mistakes and 1-2 partial skills. Group similar failures; focus on root
causes. World-agnostic methods only --- no fitted constant values.
\end{lstlisting}

\subsubsection{\textsc{HExA} evolver: bank evolution}
\label{app:ext_prompts:discovery:bank}
{\sloppy Rounds~1 onward, once per round. Sent as a single user message to \texttt{claude-sonnet-5}.\par}

\begin{lstlisting}[style=promptwrap]
============================================================
  USER PROMPT
============================================================
You are a physics-discovery mentor evolving a CROSS-WORLD skill bank for a discovery
  agent. ...

**Environment**: an LLM agent does open-ended physics discovery. Each episode it is ...

**Reward** per trajectory is a blend of explanation quality and normalized trajectory
  ...

**These skills are CROSS-WORLD**: ...

**World that produced the NEW episodes below (setup only):**
{level_description}

CURRENT SKILL BANK (from previous worlds/rounds):
{existing_skills_block}

EPISODES:
Successes: {n_successes}   Failures: {n_failures}

Episodes are marked by provenance. ...

{new_trajectories_block}

YOUR TASK: merge the existing bank with insights from the new episodes into the
COMPLETE FINAL bank (not a diff).

RULES:
1. Output the full final bank --- retained skills AND new ones.
2. Hard caps: at most {max_skills} skills and {max_mistakes} mistakes total.
3. Per skill: set "is_new" (true = extracted from the new episodes, ...
4. Retained skills keep their existing confidence unless the new episodes clearly ...
5. New-skill confidence reflects: reward of source episodes, and how broadly the ...
6. REMOVE skills that are: redundant/subsumed, contradicted by the new episodes,
  world-specific ...

OUTPUT JSON OBJECT:
{
  "skills": [
    ...
  ],
  "mistakes": [
    ...
  ],
  "removed_skill_titles": ["..."],
  "reasoning": "<brief: what changed and why, esp. anything demoted/removed>"
}

Be concise and precise. Ground every skill in the episodes.
\end{lstlisting}

\subsubsection{\textsc{HExA} evolver input: world description (oscillator)}
\label{app:ext_prompts:discovery:hexa_level}
{\sloppy \texttt{\{level\_description\}} for the oscillator evolver prompts: the world's
facts and controls, then the first 6{,}000 characters of the actor's own instructions,
so that the evolver does not distil skills the actor is already told.\par}

\begin{lstlisting}[style=promptwrap]
A central two-/N-body configuration. ...

Controls you can set in each experiment (and no others): ...
**WHAT THE AGENT IS ALREADY TOLD.** Every episode the agent receives the ...
\end{lstlisting}

\subsubsection{Episodic RAG: injected block}
\label{app:ext_prompts:discovery:rag}
{\sloppy Numeric literals and law forms in the stored steps are masked, and each rollout
keeps its last 12 steps with observations cut to 600 characters.\par}

\begin{lstlisting}[style=promptwrap]
============================================================
  INJECTED BLOCK
============================================================
## Retrieved Past Attempts (your own transcripts from earlier discovery episodes)

### Past attempt {i} --- {SUCCESS|FAILURE} (world {W#}, seed {seed})
Level: {scope_level}  |  Seed: {seed} ...

Rollout:
--- Step {k} --- ...

Each block above is one complete past attempt at discovering a hidden ...
\end{lstlisting}

\subsubsection{MemRL: memory-building prompts}
\label{app:ext_prompts:discovery:memrl_build}
{\sloppy Each build call is one prompt, with the role labels written as plain text.\par}

\begin{lstlisting}[style=promptwrap]
============================================================
  PROMPT, SUCCESS (SCRIPT)
============================================================
SYSTEM:
You are analyzing a SUCCESSFUL attempt to discover the hidden law of ...

You will receive the world's setup (what is observable and which controls ...

Produce a concise, high-level script capturing the experimental strategy. It must be:
1. Generic enough to apply to a different world with a different hidden law.
2. Specific about EXPERIMENT DESIGN: which quantity was varied, what was held ...
3. At most 5 high-level steps.
4. Restricted to controls named in the world setup you are given. ...
5. Free of any claim about what the law turned out to ...

Output ONLY the numbered script. No preamble, no headers, no markdown fences.

USER:
World setup: ...

Trajectory:
{trajectory_text}

ASSISTANT:

============================================================
  PROMPT, FAILURE (REFLECTION)
============================================================
SYSTEM:
You are analyzing a FAILED attempt to discover the hidden law of ...

You will receive the world's setup (what is observable and which controls ...

This attempt failed. Analyze what went wrong at the level of EXPERIMENT ...
1. What the agent failed to measure, or measured in a way ...
2. Concrete design fixes -- which variable to sweep, over what range, ...
3. What to avoid next time.

Constraints:
- Restrict every suggestion to controls named in the world setup you are given. ...
- Do not state or guess what the hidden law is. ...

Keep it under 8 sentences. ...

USER:
World setup: ...

Trajectory:
{trajectory_text}

ASSISTANT:
\end{lstlisting}

\subsubsection{MemRL: injected block}
\label{app:ext_prompts:discovery:memrl_block}

\begin{lstlisting}[style=promptwrap]
============================================================
  INJECTED BLOCK
============================================================
## Archived Memories (from your own past discovery episodes)

### Memory {i} --- SUCCESS (world {W#}, seed {seed}, utility Q={q:+.3f},
  updates={q_visits})
Level: {scope_level}  |  Seed: {seed}

SCRIPT:
{script}

TRAJECTORY:
--- Step {k} --- ...

### Memory {i} --- FAILURE (world {W#}, seed {seed}, utility Q={q:+.3f},
  updates={q_visits})
TASK REFLECTION: ...

What went wrong:
{reflection}

Failed approach:
--- Step {k} --- ...

These were retrieved because they have been useful before on tasks like this one. ...
\end{lstlisting}

\newpage

%
%

\section{Working Example: ReAct vs.\ \textsc{HExA} on Level Catapult of InterPhyre}
\label{app:trajectories}

\lstdefinestyle{promptwrap}{%
  style=prompt,%
  basicstyle=\ttfamily\scriptsize,%
  columns=fullflexible,%
  keepspaces=true,%
  breaklines=true,%
  breakatwhitespace=false,%
  postbreak=\mbox{\ensuremath{\hookrightarrow}\space},%
  showstringspaces=false%
}
\lstdefinestyle{promptwrapHL}{%
  style=promptwrap,%
  moredelim=[is][\color{skillblue}\bfseries]{@@}{@@},%
  moredelim=[is][\color{scalarDark}\bfseries]{\#\#}{\#\#},%
  moredelim=[is][\color{black!55}\itshape]{[[}{]]}%
}

This appendix walks through a single Interphyre seed --- \texttt{catapult}
seed~64 --- on which the \textsc{ReAct} baseline fails and \textsc{HExA}
succeeds, with both agents using Claude Sonnet~4.6 as the actor. Seed~64 is a
held-out evaluation seed: the bank was frozen after round~17 and was never
distilled on this seed. We reproduce, in order: the system prompt and full
Thought/Action/Observation transcript for \textsc{ReAct} (failure,
25~iterations); the round-17 evolving skill bank that the \textsc{HExA} agent
has access to; the system prompt and full transcript for \textsc{HExA} (success
in 6~iterations, final placement $(x{=}0.3, y{=}-0.25, r{=}2.0)$).
The two prompts differ only in the \textsc{LEARNED PHYSICS SKILLS} block
injected by the retriever ahead of the user prompt; the tool list,
format instructions, and kickoff message are identical, isolating the
contribution of the skill bank.

\subsection{Sparsified simulation frames (catapult seed 64)}
\label{app:trajectories:simframes}

\begin{figure}[!htbp]
\centering
\setlength{\tabcolsep}{1.2pt}
\renewcommand{\arraystretch}{0.85}
{\footnotesize\itshape \textsc{ReAct} (25 iterations, FAILURE):}\\[1pt]
\begin{tabular}{@{}ccccccc@{}}
  \includegraphics[width=0.135\linewidth]{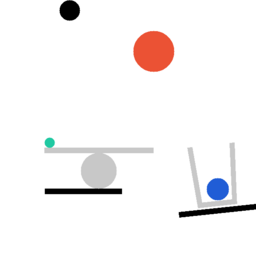} &
  \includegraphics[width=0.135\linewidth]{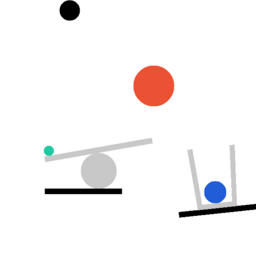} &
  \includegraphics[width=0.135\linewidth]{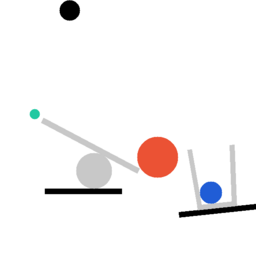} &
  \includegraphics[width=0.135\linewidth]{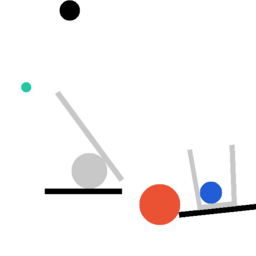} &
  \includegraphics[width=0.135\linewidth]{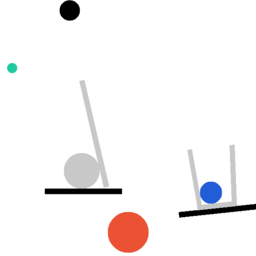} &
  \includegraphics[width=0.135\linewidth]{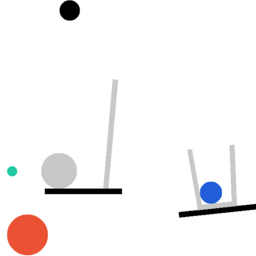} &
  \includegraphics[width=0.135\linewidth]{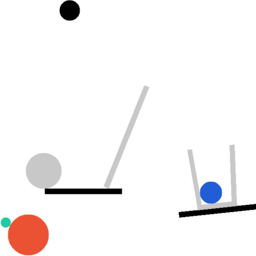} \\[-1pt]
  {\scriptsize step 0} & {\scriptsize step 30} & {\scriptsize step 57} &
  {\scriptsize step 70} & {\scriptsize step 91} &
  {\scriptsize step 151} & {\scriptsize step 193} \\
\end{tabular}\\[6pt]
{\footnotesize\itshape \textsc{HExA} (6 iterations, SUCCESS):}\\[1pt]
\begin{tabular}{@{}ccccccc@{}}
  \includegraphics[width=0.135\linewidth]{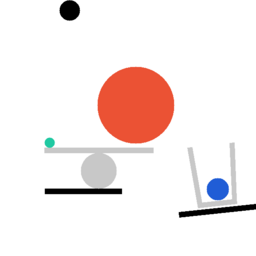} &
  \includegraphics[width=0.135\linewidth]{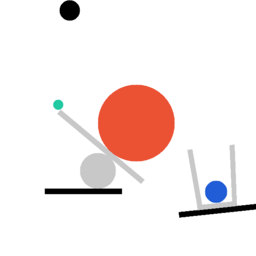} &
  \includegraphics[width=0.135\linewidth]{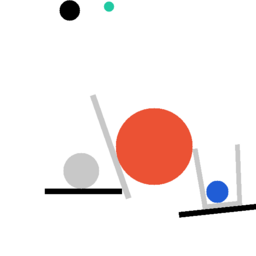} &
  \includegraphics[width=0.135\linewidth]{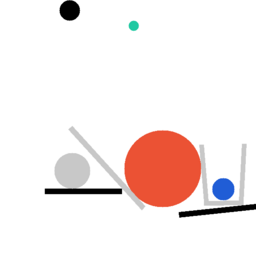} &
  \includegraphics[width=0.135\linewidth]{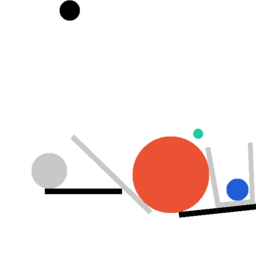} &
  \includegraphics[width=0.135\linewidth]{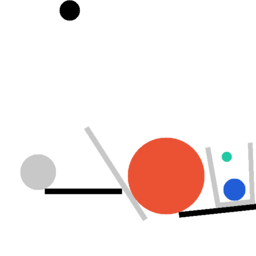} &
  \includegraphics[width=0.135\linewidth]{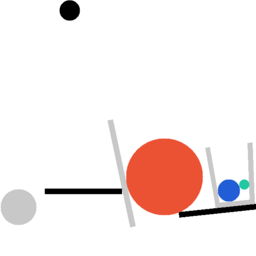} \\[-1pt]
  {\scriptsize step 0} & {\scriptsize step 25} & {\scriptsize step 57} &
  {\scriptsize step 74} & {\scriptsize step 119} &
  {\scriptsize step 140} & {\scriptsize step 172} \\
\end{tabular}
\caption{%
  Sparsified simulation frames for catapult seed 64, time increasing
  left to right. Top row: \textsc{ReAct} exhausts its 25-iteration
  budget without a green--blue contact. Bottom row: \textsc{HExA}'s
  successful $(x{=}0.3,\,y{=}{-}0.25,\,r{=}2.0)$ placement; the lever
  launches the green ball into the basket, where it makes the required
  3-second contact with the blue ball.
}
\label{fig:catapult45_simframes_app}
\end{figure}

Figure~\ref{fig:catapult45_simframes_app} shows seven sparsely-sampled frames
from each agent's final simulation, in left-to-right time order.
The top row is \textsc{ReAct}: after 25 iterations the agent never produces
a placement that brings the green ball into contact with the blue ball.
The bottom row is \textsc{HExA}'s 6-iteration solve at
$(x{=}0.3,\,y{=}{-}0.25,\,r{=}2.0)$: the large red ball strikes the catapult
arm just right of the pivot, the lever launches the green ball in a high
arc across the scene, and the green ball comes to rest in the basket
against the blue ball.


\subsection{\textsc{ReAct} system prompt (catapult seed 64)}
\label{app:trajectories:react_prompt}
The full \textsc{ReAct} catapult system prompt is given in
Appendix~\ref{app:prompt:react_catapult} and the tool descriptions in
Appendix~\ref{app:tools}; we omit them here and show only the opening lines and the
user prompt used for this run.

\begin{lstlisting}[style=promptwrap]
============================================================
  SYSTEM PROMPT
============================================================
You are an expert physics reasoning agent solving a 2D physics puzzle. You have access 
  to a physics simulator and can test your ideas before submitting a final answer.

**Puzzle: Catapult**

[... puzzle description, placement constraints, tool list, Thought/Action format and
  rules omitted; see text ...]

============================================================
  USER PROMPT
============================================================
Solve the Catapult puzzle. You have 25 iterations to solve it. The success condition is:
   green_ball must contact blue_ball. Use tools effectively and think about alternate 
  approaches if one does not work.
- **ALWAYS call predict_first_contact first** --- it is cheap and tells you if your red 
  ball hits the intended object.
- Only if predict confirms the right contact: call trace_green_ball to track green ball.
\end{lstlisting}

\subsection{\textsc{ReAct} trajectory (catapult seed 64, 25 iterations, FAILURE)}
\label{app:trajectories:react_run}

\begin{lstlisting}[style=promptwrap]
--- Step 1 ---
Thought:
Let me start by understanding the scene geometry to plan my approach.

Action:
describe_scene_geometry({})

Observation:
=== Scene Geometry ===
World bounds: x in [-5, 5], y in [-5, 5]

Balls:
  green_ball: pos=(-3.06, -1.97), r=0.20, dynamic=True
  blue_ball: pos=(3.50, -1.99), r=0.32, dynamic=True
  black_ball: pos=(-1.17, 4.60), r=0.40, dynamic=False
  gray_ball: pos=(-1.14, -3.07), r=0.70, dynamic=True

Bars:
  ledge: pos=(3.50, -2.71), angle=-3.6 deg, len=3.01, dynamic=False
  black_platform: pos=(-1.74, -3.87), angle=0.0 deg, len=3.00, dynamic=False
  gray_platform: pos=(-1.14, -2.27), angle=0.0 deg, len=4.25, dynamic=True

Baskets:
  basket: pos=(3.50, -2.51), dynamic=True

Key distances:
  green<->blue: 6.56

Success condition: green_ball <-> blue_ball for 3.0s
==================================================

--- Step 2 ---
Thought:
The green ball is at (-3.06, -1.97) sitting on the left end of the gray bar (catapult
  arm). The gray ball pivot is at (-1.14, -3.07). The blue ball is at (3.50, -1.99)
  inside the basket. The catapult mechanism: if I drop something heavy on the right end
  of the gray bar, it will launch the green ball upward and to the right. The gray bar
  is at (-1.14, -2.27) with length 4.25. The right end of the bar would be at
  approximately (-1.14 + 4.25/2, -2.27) = (0.99, -2.27). Let me drop the red ball on the
  right side of the catapult arm to launch the green ball toward the blue ball.

Action:
predict_first_contact({'x': 1.0, 'y': 0.0, 'radius': 0.5})

Observation:
=== Predicted First Contact ===
placement_valid: True
red_ball drop: x=1.000, y=0.000, radius=0.500
first_contact_object: gray_platform
contact_step: 31
approach_speed: 2.945
contact_point (approx midpoint): (-0.079, -1.792)
contact_normal (red->other inverted): (0.918, 0.396)

--- Step 3 ---
Thought:
The red ball hits the gray platform (catapult arm) at step 31. Now let me trace the
  green ball to see if it gets launched toward the blue ball.

Action:
trace_green_ball({'x': 1.0, 'y': 0.0, 'radius': 0.5})

Observation:
=== Green Ball Trajectory (steps run: 330) ===
red_ball drop: x=1.000, y=0.000, radius=0.500
Success: False
green_ball: start=(-3.06, -1.97) end=(-4.78, -4.78) Delta=(-1.72, -2.81) peak_y=-0.54 v_max=7.32

Waypoints (step: x, y):
  step    1: (-3.06, -1.97)
  step   30: (-3.09, -2.26)
  step   60: (-3.24, -0.54)
  step   90: (-3.24, -0.57)
  step  120: (-3.25, -3.04)
  step  150: (-3.58, -3.76)
  step  180: (-3.96, -4.78)
  step  210: (-4.33, -4.78)
  step  240: (-4.70, -4.78)
  step  270: (-4.78, -4.78)
  step  300: (-4.78, -4.78)
  step  330: (-4.78, -4.78)

--- Step 4 ---
Thought:
The green ball went left and down, not toward the blue ball. I need to hit the right end
  of the catapult arm more precisely to launch the green ball to the right. The gray bar
  extends from about (-1.14 - 4.25/2) = (-3.26) to (-1.14 + 4.25/2) = (0.99) in x. I
  need to hit further right on the bar. Let me try dropping further to the right end.

Action:
predict_first_contact({'x': 2.0, 'y': -1.0, 'radius': 0.5})

Observation:
=== Predicted First Contact ===
placement_valid: True
red_ball drop: x=2.000, y=-1.000, radius=0.500
first_contact_object: ledge
contact_step: 27
approach_speed: 0.883
contact_point (approx midpoint): (2.750, -2.362)
contact_normal (red->other inverted): (-0.906, 0.423)

--- Step 5 ---
Thought:
The red ball hits the ledge, not the catapult arm. I need to hit the right end of the
  gray bar more precisely. The bar extends to about x=0.99 on the right. Let me try
  dropping closer to the right end of the bar.

Action:
predict_first_contact({'x': 0.5, 'y': -1.0, 'radius': 0.5})

Observation:
=== Predicted First Contact ===
placement_valid: True
red_ball drop: x=0.500, y=-1.000, radius=0.500
first_contact_object: gray_platform
contact_step: 20
approach_speed: 1.569
contact_point (approx midpoint): (-0.322, -1.905)
contact_normal (red->other inverted): (0.916, 0.401)

--- Step 6 ---
Thought:
The red ball hits the gray platform (catapult arm). Let me trace the green ball with
  this placement.

Action:
trace_green_ball({'x': 0.5, 'y': -1.0, 'radius': 0.5})

Observation:
=== Green Ball Trajectory (steps run: 300) ===
red_ball drop: x=0.500, y=-1.000, radius=0.500
Success: False
green_ball: start=(-3.06, -1.97) end=(-4.66, -4.78) Delta=(-1.60, -2.81) peak_y=-0.21 v_max=4.62

Waypoints (step: x, y):
  step    1: (-3.06, -1.97)
  step   30: (-3.09, -1.70)
  step   60: (-2.65, -0.21)
  step   90: (-2.11, -0.92)
  step  120: (-2.38, -3.03)
  step  150: (-3.86, -3.88)
  step  180: (-4.67, -4.67)
  step  210: (-4.62, -4.78)
  step  240: (-4.63, -4.78)
  step  270: (-4.65, -4.78)
  step  300: (-4.66, -4.78)

--- Step 7 ---
Thought:
The green ball is going up but then falling to the left. I need it to go right toward
  the blue ball at (3.50, -1.99). The catapult needs to be hit on the far right end so
  the green ball launches to the right. Let me try hitting closer to x=0.9 which should
  be near the right end of the bar.

Action:
predict_first_contact({'x': 0.8, 'y': 0.5, 'radius': 0.3})

Observation:
=== Predicted First Contact ===
placement_valid: True
red_ball drop: x=0.800, y=0.500, radius=0.300
first_contact_object: gray_platform
contact_step: 37
approach_speed: 1.724
contact_point (approx midpoint): (-0.189, -1.812)
contact_normal (red->other inverted): (0.914, 0.406)

--- Step 8 ---
Thought:
The red ball is hitting the bar but not at the right end. The contact point is at
  x=-0.189, which means the ball is landing near the center-left of the bar. I need to
  position the ball so it hits the far right end of the bar (near x=0.99). Let me try a
  smaller radius and position more to the right.

Action:
trace_green_ball({'x': 0.8, 'y': 0.5, 'radius': 0.3})

Observation:
=== Green Ball Trajectory (steps run: 240) ===
red_ball drop: x=0.800, y=0.500, radius=0.300
Success: False
green_ball: start=(-3.06, -1.97) end=(-4.71, -4.78) Delta=(-1.65, -2.81) peak_y=-1.58 v_max=8.08

Waypoints (step: x, y):
  step    1: (-3.06, -1.97)
  step   30: (-3.09, -2.26)
  step   60: (-3.47, -1.58)
  step   90: (-3.91, -1.96)
  step  120: (-4.36, -4.79)
  step  150: (-4.77, -4.78)
  step  180: (-4.75, -4.78)
  step  210: (-4.73, -4.78)
  step  240: (-4.71, -4.78)

--- Step 9 ---
Thought:
The green ball keeps going left. The catapult lever needs to be hit on the right side of
  the pivot to launch the green ball (on left end) upward and to the right. The pivot is
  at x=-1.14, so the right end of the bar is around x=0.99. The ball hitting the bar
  seems to be tilting it the wrong way. Let me try hitting very close to the right tip
  of the bar.

Action:
predict_first_contact({'x': 0.9, 'y': -1.5, 'radius': 0.2})

Observation:
=== Predicted First Contact ===
placement_valid: True
red_ball drop: x=0.900, y=-1.500, radius=0.200
first_contact_object: gray_platform
contact_step: 16
approach_speed: 0.270
contact_point (approx midpoint): (-0.121, -2.054)
contact_normal (red->other inverted): (0.980, 0.198)

--- Step 10 ---
Thought:
The contact point is still at x=-0.121, not at the right end. The red ball seems to be
  rolling/sliding along the bar to where contact is detected. Let me try placing the red
  ball directly above the far right end of the bar at x=0.99.

Action:
trace_green_ball({'x': 0.9, 'y': -1.5, 'radius': 0.2})

Observation:
=== Green Ball Trajectory (steps run: 360) ===
red_ball drop: x=0.900, y=-1.500, radius=0.200
Success: False
green_ball: start=(-3.06, -1.97) end=(-4.78, -4.78) Delta=(-1.72, -2.81) peak_y=-1.90 v_max=2.74

Waypoints (step: x, y):
  step    1: (-3.06, -1.97)
  step   30: (-3.09, -2.02)
  step   60: (-3.17, -1.93)
  step   90: (-3.24, -1.90)
  step  120: (-3.27, -2.01)
  step  150: (-3.36, -2.22)
  step  180: (-3.70, -2.72)
  step  210: (-4.16, -4.71)
  step  240: (-4.63, -4.78)
  step  270: (-4.78, -4.78)
  step  300: (-4.78, -4.78)
  step  330: (-4.78, -4.78)
  step  360: (-4.78, -4.78)

--- Step 11 ---
Thought:
The green ball keeps going left after launch. Let me try a large, heavy drop from high
  up on the far right end of the catapult arm to create more rightward force. I'll use a
  bigger radius ball dropped from high above the right end of the bar.

Action:
predict_first_contact({'x': 0.9, 'y': 3.0, 'radius': 1.0})

Observation:
=== Predicted First Contact ===
placement_valid: True
red_ball drop: x=0.900, y=3.000, radius=1.000
first_contact_object: gray_platform
contact_step: 50
approach_speed: 6.945
contact_point (approx midpoint): (-0.176, -1.352)
contact_normal (red->other inverted): (0.769, 0.639)

--- Step 12 ---
Thought:
The red ball hits the gray platform with high speed (6.945). Let me trace the green ball
  to see if this launches it toward the blue ball.

Action:
trace_green_ball({'x': 0.9, 'y': 3.0, 'radius': 1.0})

Observation:
=== Green Ball Trajectory (steps run: 420) ===
red_ball drop: x=0.900, y=3.000, radius=1.000
Success: False
green_ball: start=(-3.06, -1.97) end=(-4.78, -4.78) Delta=(-1.71, -2.81) peak_y=+0.89 v_max=8.35

Waypoints (step: x, y):
  step    1: (-3.06, -1.97)
  step   30: (-3.09, -2.26)
  step   60: (-3.83, -1.46)
  step   90: (-4.63, +0.89)
  step  120: (-4.30, +0.52)
  step  150: (-3.96, -2.31)
  step  180: (-4.45, -3.26)
  step  210: (-4.78, -3.54)
  step  240: (-4.78, -3.61)
  step  270: (-4.78, -3.75)
  step  300: (-4.78, -4.16)
  step  330: (-4.78, -4.78)
  step  360: (-4.78, -4.78)
  step  390: (-4.78, -4.78)
  step  420: (-4.78, -4.78)

--- Step 13 ---
Thought:
The green ball keeps going left instead of right. Let me get the full level state to
  better understand the bar geometry and try a different approach.

Action:
get_level_state({})

Observation:
=== Level State: catapult ===

* green_ball (green Ball): position=(-3.0633, -1.9686), radius=0.2000, dynamic=True
* blue_ball (blue Ball): position=(3.5000, -1.9920), radius=0.3200, dynamic=True
* ledge (black Bar): position=(3.5000, -2.7124), left=2.0000, right=5.0000, top=-2.6172, bottom=-2.8075, thickness=0.2000, dynamic=False
* basket: {'x': 3.5, 'y': np.float64(-2.5119720543860606), 'color': 'gray', 'dynamic': True, 'type': 'Basket', 'angle': -3.6288244368770233, 'scale': 0.7880514159250508}
* black_ball (black Ball): position=(-1.1695, 4.6000), radius=0.4000, dynamic=False
* black_platform (black Bar): position=(-1.7383, -3.8686), left=-3.2383, right=-0.2383, top=-3.8686, bottom=-3.8686, thickness=0.2000, dynamic=False
* gray_ball (gray Ball): position=(-1.1383, -3.0686), radius=0.7000, dynamic=True
* gray_platform (gray Bar): position=(-1.1383, -2.2686), left=-3.2633, right=0.9867, top=-2.2686, bottom=-2.2686, thickness=0.2000, dynamic=True

World bounds: x in [-5, 5], y in [-5, 5]
Action objects: ['red_ball'] -- you control the red ball placement
Success condition: green_ball must touch blue_ball for 3 seconds
============================================================

--- Step 14 ---
Thought:
The green ball keeps launching left because the catapult rotation sends the left end
  upward-left tangentially. I need a different approach - let me try placing the red
  ball to the LEFT of the green ball to directly push it rightward off the arm.

Action:
predict_first_contact({'x': -3.6, 'y': 0.0, 'radius': 0.5})

Observation:
=== Predicted First Contact ===
placement_valid: True
red_ball drop: x=-3.600, y=0.000, radius=0.500
first_contact_object: green_ball
contact_step: 38
approach_speed: 4.580
contact_point (approx midpoint): (-3.359, -2.227)
contact_normal (red->other inverted): (-0.733, 0.680)

--- Step 15 ---
Thought:
The red ball hits the green ball! Let me trace the green ball to see if this pushes it
  toward the blue ball.

Action:
trace_green_ball({'x': -3.6, 'y': 0.0, 'radius': 0.5})

Observation:
=== Green Ball Trajectory (steps run: 240) ===
red_ball drop: x=-3.600, y=0.000, radius=0.500
Success: False
green_ball: start=(-3.06, -1.97) end=(-3.80, -4.78) Delta=(-0.74, -2.81) peak_y=-1.97 v_max=3.94

Waypoints (step: x, y):
  step    1: (-3.06, -1.97)
  step   30: (-3.09, -2.26)
  step   60: (-3.02, -3.52)
  step   90: (-3.27, -3.56)
  step  120: (-3.63, -4.39)
  step  150: (-3.80, -4.78)
  step  180: (-3.80, -4.78)
  step  210: (-3.80, -4.78)
  step  240: (-3.80, -4.78)

--- Step 16 ---
Thought:
The green ball goes down instead of right. Let me try dropping from directly above the
  green ball to push it rightward, or try hitting it from the left side more precisely.
  Let me try placing the red ball to the left of the green ball at the same height to
  push it horizontally.

Action:
predict_first_contact({'x': -4.0, 'y': -1.97, 'radius': 0.5})

Observation:
=== Predicted First Contact ===
placement_valid: True
red_ball drop: x=-4.000, y=-1.970, radius=0.500
first_contact_object: bottom_wall
contact_step: 43
approach_speed: 1.405
contact_point (approx midpoint): (-2.000, -4.732)
contact_normal (red->other inverted): (-0.991, 0.130)

--- Step 17 ---
Thought:
The red ball falls through to the bottom wall without hitting the green ball. Let me try
  a different approach - place the red ball just to the left of the green ball at the
  same height so it rolls into it.

Action:
predict_first_contact({'x': -4.5, 'y': -1.97, 'radius': 0.3})

Observation:
=== Predicted First Contact ===
placement_valid: True
red_ball drop: x=-4.500, y=-1.970, radius=0.300
first_contact_object: bottom_wall
contact_step: 45
approach_speed: 1.470
contact_point (approx midpoint): (-2.250, -4.828)
contact_normal (red->other inverted): (-0.997, 0.074)

--- Step 18 ---
Thought:
The red ball isn't hitting the green ball when placed at the same height. Let me try a
  different approach - place the red ball directly above the blue ball/basket area to
  push the blue ball down, or better, try to have the green ball land in the basket. Let
  me try placing the red ball on the right side of the catapult arm near the pivot to
  tilt the arm differently.

Action:
predict_first_contact({'x': -0.5, 'y': 2.0, 'radius': 0.8})

Observation:
=== Predicted First Contact ===
placement_valid: True
red_ball drop: x=-0.500, y=2.000, radius=0.800
first_contact_object: gray_platform
contact_step: 47
approach_speed: 5.033
contact_point (approx midpoint): (-0.860, -1.645)
contact_normal (red->other inverted): (0.510, 0.860)

--- Step 19 ---
Thought:
The red ball hits the gray platform near the center. Let me trace the green ball
  trajectory.

Action:
trace_green_ball({'x': -0.5, 'y': 2.0, 'radius': 0.8})

Observation:
=== Green Ball Trajectory (steps run: 300) ===
red_ball drop: x=-0.500, y=2.000, radius=0.800
Success: False
green_ball: start=(-3.06, -1.97) end=(-3.44, -3.94) Delta=(-0.38, -1.97) peak_y=+2.86 v_max=9.56

Waypoints (step: x, y):
  step    1: (-3.06, -1.97)
  step   30: (-3.09, -2.26)
  step   60: (-3.84, -0.61)
  step   90: (-4.68, +2.50)
  step  120: (-4.38, +2.86)
  step  150: (-4.09, +0.77)
  step  180: (-3.74, -3.53)
  step  210: (-3.44, -3.83)
  step  240: (-3.44, -3.92)
  step  270: (-3.44, -3.94)
  step  300: (-3.44, -3.94)

--- Step 20 ---
Thought:
The green ball keeps going left. The catapult mechanism launches it to the left due to
  the bar angle. Let me try a completely different approach - place the red ball to
  knock the blue ball out of the basket toward the green ball, or drop it to push the
  blue ball left toward the green ball.

Action:
predict_first_contact({'x': 3.5, 'y': 2.0, 'radius': 0.5})

Observation:
=== Predicted First Contact ===
placement_valid: True
red_ball drop: x=3.500, y=2.000, radius=0.500
first_contact_object: basket
contact_step: 43
approach_speed: 4.784
contact_point (approx midpoint): (3.504, -1.580)
contact_normal (red->other inverted): (0.003, 1.000)

--- Step 21 ---
Thought:
The red ball hits the basket. Let me simulate this to see if the blue ball gets knocked
  toward the green ball.

Action:
simulate_action({'x': 3.5, 'y': 2.0, 'radius': 0.5})

Observation:
FAILURE. The green ball did NOT contact the blue ball in the basket.
Total simulation steps: 2000
Reward: -0.1

Final object positions:
  green_ball: pos=(-4.777, -4.780), vel=(0.000, 0.000)
  red_ball: pos=(3.552, -1.247), vel=(0.000, 0.000)
  blue_ball: pos=(3.621, -2.103), vel=(0.000, 0.000)
  ledge: pos=(3.500, -2.712), vel=(0.000, 0.000)
  basket: pos=(3.470, -2.593), vel=(0.000, 0.000)
  black_ball: pos=(-1.170, 4.600), vel=(0.000, 0.000)
  black_platform: pos=(-1.738, -3.869), vel=(0.000, 0.000)
  gray_ball: pos=(-2.160, -3.064), vel=(0.000, 0.000)
  gray_platform: pos=(-3.222, -3.040), vel=(0.000, 0.000)

--- Step 22 ---
Thought:
The green ball keeps launching left. Let me try a completely different approach - drop
  the red ball to push the blue ball out of the basket to the left, so the blue ball
  travels toward the green ball's final position at the bottom left.

Action:
predict_first_contact({'x': 4.5, 'y': 0.0, 'radius': 0.5})

Observation:
=== Predicted First Contact ===
placement_valid: True
red_ball drop: x=4.500, y=0.000, radius=0.500
first_contact_object: right_wall
contact_step: 1
approach_speed: 0.163
contact_point (approx midpoint): (4.738, -0.001)
contact_normal (red->other inverted): (-1.000, -0.005)

--- Step 23 ---
Thought:
The green ball keeps launching to the upper-left. Let me try hitting the blue
  ball/basket from the right side to knock the blue ball leftward toward where the green
  ball ends up (around -4.78, -4.78). I'll drop the red ball to the right of the basket.

Action:
predict_first_contact({'x': 4.8, 'y': -1.5, 'radius': 0.2})

Observation:
=== Predicted First Contact ===
placement_valid: True
red_ball drop: x=4.800, y=-1.500, radius=0.200
first_contact_object: right_wall
contact_step: 1
approach_speed: 0.163
contact_point (approx midpoint): (4.888, -0.751)
contact_normal (red->other inverted): (-0.141, -0.990)

--- Step 24 ---
Thought:
Let me try to knock the blue ball out of the basket by dropping from above and to the
  right of it, so it moves left toward where the green ball will be.

Action:
predict_first_contact({'x': 3.8, 'y': -1.5, 'radius': 0.3})

Observation:
INVALID placement:
  - Overlaps with blue_ball: dist=0.5762, sum_of_radii=0.6200. Move at least 0.0938 units further away.
placement_valid=False; no contact check performed.

--- Step 25 ---
Thought:
Let me try to drop the red ball to hit the blue ball from above to knock it out of the
  basket toward the left.

Action:
predict_first_contact({'x': 3.8, 'y': -1.0, 'radius': 0.3})

Observation:
INVALID ACTION: Action object 0 at (3.80, -1.00) with radius 0.30 is invalid
\end{lstlisting}

\subsection{Round-17 evolving skill bank used by \textsc{HExA}}
\label{app:trajectories:bank}

\begin{lstlisting}[style=promptwrap]
# CATAPULT SKILLS

[cat_ev_9_000]
  Title:
    Right-of-Pivot Arm Strike Zone (x~=0.2--0.5 Default)
  generation: 11    confidence: 0.93
  source seeds: [43, 48, 49, 51, 52, 53]
  Principle:
    Placing the red ball at x~=0.2--0.5 on the right half of the catapult arm creates a
      downward force that lifts the green ball in a rightward arc toward the basket.
      x=0.3 is the default starting point; x=0.2 at standard y reduces range ~1.3 units.
      Shifting x rightward (>=0.5) at standard-y positions risks diminishing returns.
  When to apply:
    Default for all catapult strikes; start with x=0.3. At moderate elevated y (~=0.5),
      decreasing x below 0.3 increases range (see leftward-x skill). Do not exceed x~=0.5
      without checking for diminishing returns.
  Example:
    Seed 43: x=0.3 y=-0.5 -> green at x=3.10, success. Seed 48: x=0.3 y=0.3 -> success.
      Seed 51: x=0.3 y=0.0 -> success.

[cat_ev_6_001]
  Title:
    Low Drop (y~=-0.5 to 0.0) Maximizes Range; Binary-Search y First
  generation: 11    confidence: 0.93
  source seeds: [48, 49, 51, 53]
  Principle:
    Drop height y is the primary range control: y~=-0.5 produces maximum range, y~=0.0
      produces moderate range, y~=0.3--0.5 produces reduced range. A +/-0.25 change in y at
      standard heights shifts landing x by 1.5--2.0 units. When standard y (<=0.0) is
      blocked, the operative range shifts to y=0.25--0.5, where lower y still produces
      more range.
  When to apply:
    Default for all seeds; start at y=-0.5, then binary-search toward y=0.0 if
      overshooting. If gray_ball blocks all y<=0.0, shift to y=0.5 as first valid
      placement, then binary-search DOWN to y=0.3 or y=0.25 to increase range. Use
      y=-0.25 as first midpoint in standard regime.
  Example:
    Seed 49: y=-0.5 blocked, y=0.0->x=4.26 (overshoot), y=0.25->x=3.71, success. Seed 53:
      y=0.0->x=4.45, y=0.25->x=4.19, y=0.5->x=4.17 (y-dominant regime detected).

[cat_ev_11_002]
  Title:
    Overlap Fallback Sequence: y=-0.5 -> y=0.0 -> y=0.5 -> Binary-Search y=0.3--0.5
  generation: 11    confidence: 0.91
  source seeds: [48, 50, 51, 52, 53]
  Principle:
    When y=-0.5 is blocked, try y=0.0, then y=0.5 with r=2.0. Once a valid placement at
      y=0.5 is confirmed, binary-search DOWNWARD within y=0.3--0.5 to fine-tune range,
      since lower y in this zone still increases range. Only escalate to y=1.0 if y=0.3
      is also blocked or all y=0.3--0.5 attempts miss by more than 1 unit.
  When to apply:
    When predict_first_contact returns invalid for y=-0.5; follow sequence y=0.0 -> y=0.5
      r=2.0 -> binary-search y=0.3--0.5. Skip x=1.0 at y=1.0 always. For elevated green
      ball scenes (y>=-0.5), standard y slots may all be blocked and sequence starts at
      y=1.0.
  Example:
    Seed 48: y=-0.5/0.0 blocked -> y=0.5 (x=2.37 undershoot) -> y=0.3 (x=3.14, success).
      Seed 51: y=-0.5 blocked -> y=0.0 valid (gap 0.21, success). Seed 52: y=-0.5/0.0
      blocked -> y=0.5 undershoot -> bisect down.

[cat_ev_11_003]
  Title:
    Trace-then-Simulate: Gap Threshold, Bisect Chaotic Cliffs, Simulate Narrow Boundary
  generation: 11    confidence: 0.87
  source seeds: [49, 50, 51, 52]
  Principle:
    At standard drop heights (y<=0.0), commit to simulate when gap is under 0.8 units. At
      elevated heights (y=0.25--0.5), commit at gap <0.5. When a chaotic cliff is
      detected in y=0.3--0.5, bisect to the midpoint rather than simulating immediately.
      After bisection has narrowed the cliff to Deltay<=0.01 (minimum resolution), simulating
      the overshoot boundary IS valid---the green ball can contact the blue ball during
      its arc even if the final resting position overshoots.
  When to apply:
    At y<=0.0: gap <0.8 -> simulate. At y=0.25--0.5: gap <0.5 -> simulate. If chaotic cliff:
      bisect first. Once Deltay<=0.01 and no gap reduction is possible, simulate the
      overshoot side---arc contact is achievable at that boundary.
  Example:
    Seed 51: gap ~0.21 at y=0.0 -> success. Seed 52: cliff at y=0.34/0.35 after full
      bisection -> simulated y=0.34 (overshoot) -> success via arc contact.

[cat_ev_1_005]
  Title:
    Seed-Dependent Scene Geometry Requires Per-Seed Calibration
  generation: 11    confidence: 0.85
  source seeds: [42, 43, 51, 52, 53]
  Principle:
    Arm height, pivot position, blue ball position, green ball starting position, and
      arm tilt vary meaningfully across seeds. If green ball starts at y>=-0.5
      (elevated), standard low-y drops will be blocked, requiring elevated-y placements
      starting at y~=1.0--1.1. If blue ball is at y<-3.0, standard r=2.0 placements may
      produce too flat a trajectory.
  When to apply:
    Always call describe_scene_geometry first; if green ball y>=-0.5, treat as
      elevated-arm seed requiring y>=1.0 placements. If first 2--3 standard placements
      show green going left, treat as non-standard geometry seed requiring extra
      calibration.
  Example:
    Seed 43: standard geometry -> x=0.3 y=-0.5 -> success in 4 steps. Seed 42: green at
      y=-0.38 (elevated), standard y blocked, needed y=1.0 x-sweep. Seed 53: blue ball
      at y=-3.02, needed x=0.1 at y=0.5.

[cat_ev_9_006]
  Title:
    Chaotic Trajectory Zones: Avoid Fine-Tuning in Sensitive Ranges
  generation: 11    confidence: 0.91
  source seeds: [44, 45, 50, 52]
  Principle:
    Chaotic zones exist in both y and x dimensions where tiny parameter changes cause
      large trajectory swings: (1) y~=-0.15 to -0.25 at x=0.3; (2) y~=0.3--0.4 at r=2.0 in
      some seeds (Deltay=0.1 -> >1.9 range jump); (3) y~=1.05--1.1 at elevated arms; (4)
      x~=0.35--0.40 at y=0.75. When a chaotic cliff is detected, bisect methodically; the
      cliff boundary can be as narrow as Deltay=0.01.
  When to apply:
    When two consecutive traces with small Delta in x or y produce landing positions >1.5
      units apart: for y-chaotic cliffs bisect (e.g., try y=0.35 between y=0.3 and
      y=0.4); for x-chaotic cliffs simulate overshoot side. Do not iterate further
      within the chaotic range without a clear bisection strategy.
  Example:
    Seed 44: y=0.5->x=4.50, y=0.7->x=-0.33 (y-cliff). Seed 52: y=0.3->x=4.23, y=0.4->x=2.71
      (cliff); bisected to y=0.34/y=0.35 boundary; simulated y=0.34 -> success.

[cat_ev_6_009]
  Title:
    Identical Range at Different x/r Signals y-Dominant Regime
  generation: 11    confidence: 0.93
  source seeds: [44, 47, 53]
  Principle:
    When trace_green_ball returns nearly identical green ball landing positions despite
      changing x or radius, the scene is in a y-dominant regime where drop height
      controls range. Stop iterating x or r and switch to y-binary-search or x-sweep at
      a different y level. If lower y is blocked, try x<0.3 at moderate elevated y
      (~=0.5) to escape the regime.
  When to apply:
    When two consecutive traces with different x or r values produce landing positions
      within 0.2 units of each other; immediately switch to adjusting y. If y is blocked
      and regime persists at x=0.3, try x=0.1--0.2 at y=0.5 to break out.
  Example:
    Seed 44: x=0.3 and x=0.2 at r=2.0 both ->x=4.50. Seed 47: x=0.3 y=0.5 and various r
      all gave x~=2.69. Seed 53: y=0.0->x=4.45, y=0.25->x=4.19, y=0.5->x=4.17 (plateau) ->
      switched to x=0.1 -> success.

[cat_ev_10_007]
  Title:
    x=0.2 Escape for Standard-y Chaotic Cliff (y<=0.0 Only)
  generation: 11    confidence: 0.85
  source seeds: [41]
  Principle:
    When y binary-search at standard heights (y~=-0.5 to 0.0) reveals that y=-0.5 and
      y=-0.25 produce identical overshooting landings but y=0.0 flips the green ball far
      left, shifting x from 0.3 to 0.2 at y=-0.5 reduces range by ~1.3 units. This
      escape is strictly for standard y (<=0.0): at elevated y (~=0.5), x=0.2 increases
      range (opposite effect).
  When to apply:
    Only when standard-y (y~=-0.5 to 0.0) binary search hits chaotic cliff with both
      y=-0.5 and y=-0.25 overshooting identically but y=0.0 going far left. Do NOT apply
      at elevated y (>=0.5)---use leftward x-sweep there for the opposite
      (range-increasing) effect.
  Example:
    Seed 41: x=0.3 y=-0.5->x=4.33, y=-0.25->x=4.33 (identical), y=0.0->x=-4.78; x=0.2
      y=-0.5->x=3.05, success.

[cat_ev_9_009]
  Title:
    Elevated-y Fallback: x=2.0, r=1.5 at y=1.0--1.5 as Last Resort
  generation: 11    confidence: 0.8
  source seeds: [36, 40]
  Principle:
    When the overlap fallback sequence reaches y>=1.0 and x-sweep at y=1.0 (0.3->0.5)
      still fails, escalate to x=2.0, r=1.5---not x=1.0 at y=1.0 (INVALID ACTION). At
      x=2.0, y=1.0, r=1.5, the red ball contacts the arm at a rightward strike point,
      sending green ball with full range. An alternative for elevated scenes is x=0.3,
      y=1.1, r=2.0 when x=0.3 y=1.0 is blocked by gray_ball.
  When to apply:
    After exhausting x-sweep (0.3--0.5) at y=1.0 in elevated-arm seeds; use x=2.0 r=1.5
      at y=1.0, then y=1.5 if still blocked. Skip x=1.0 at y=1.0 always.
  Example:
    Seed 36: x=2.0, y=1.0, r=1.5 -> success. Seed 40: x=0.3, y=1.1, r=2.0 -> green at
      x=4.32, success.

[cat_ev_11_009]
  Title:
    Leftward x Shift (<0.3) at Moderate Elevated y (~=0.5) Dramatically Increases Range
  generation: 11    confidence: 0.72
  source seeds: [47, 53]
  Principle:
    At moderate elevated drop heights (y~=0.5), decreasing x below 0.3 dramatically
      increases green ball range---opposite to standard-y behavior. This occurs because at
      y=0.5 the arm contact geometry changes such that a more central strike delivers
      more angular momentum to the arm. When stuck in a y-dominant regime with
      insufficient or excessive range at x=0.3, try x=0.1--0.2 at y=0.5 before escalating
      to y=1.0.
  When to apply:
    When y=0.5 with x=0.3 produces insufficient range AND lower y is blocked or chaotic;
      try x=0.2 (binary mid) then x=0.1 at y=0.5. Binary-search x between 0.1 and 0.3 to
      target blue ball position. Do NOT apply at y>=1.0.
  Example:
    Seed 47: x=0.3 y=0.5->x=2.72, x=0.1 y=0.5->x=4.78; x=0.2 y=0.5 simulated -> success.
      Seed 53: x=0.3 y=0.5->x=4.17 (regime), x=0.1 y=0.5->x=3.63 (gap 0.13) -> success.

# CATAPULT COMMON MISTAKES

[cat_err_000]
  Description:
    Fine-tuning x or y in a chaotic zone when trajectory response is nonlinear.
  generation: 11
  Why it happens:
    Agents assume monotonic response and keep incrementing when the system is actually
      sensitive/chaotic.
  How to avoid:
    When two consecutive traces produce landing positions >1.5 units apart with small Delta,
      switch to bisection rather than linear search.

[cat_err_001]
  Description:
    Using x=1.0 at y=1.0 in elevated-arm scenes.
  generation: 11
  Why it happens:
    Agents choose an intermediate x value between 0.3 and 2.0 without checking validity.
  How to avoid:
    Skip x=1.0 at y=1.0 always; go directly to x=2.0 r=1.5 as the elevated fallback.

[cat_err_002]
  Description:
    Simulating the overshoot side of a y-chaotic cliff (y=0.3--0.5) without bisecting
      first.
  generation: 11
  Why it happens:
    Agents see a near-target landing on overshoot trace and try to simulate without
      confirming the cliff boundary.
  How to avoid:
    Always bisect the chaotic cliff first. Only simulate the overshoot boundary after
      bisection has narrowed to Deltay<=0.01 and no further resolution is possible.

[cat_err_003]
  Description:
    Placing the red ball at x>3 (far right of scene).
  generation: 11
  Why it happens:
    Agents try to increase rightward strike force by moving the contact point further
      right.
  How to avoid:
    Keep x within 0.1--0.5 for standard strikes; x=2.0 is maximum for elevated fallback.

[cat_err_004]
  Description:
    Treating x=0.2 as a universal range-reducer regardless of drop height.
  generation: 11
  Why it happens:
    Agents apply the x=0.2 escape learned at standard y to elevated y situations where x
      behavior is reversed.
  How to avoid:
    x=0.2 reduces range only at y<=0.0. At elevated y (~=0.5), x=0.2 and x=0.1 INCREASE
      range---use this deliberately when in a low-range y-dominant regime.
\end{lstlisting}

\subsection{\textsc{HExA} system prompt (catapult seed 64)}
\label{app:trajectories:hexa_prompt}
The \textsc{HExA} prompt is the \textsc{ReAct} prompt of
\S\ref{app:trajectories:react_prompt} plus the retrieved skills and mistakes,
and a four-line \texttt{IMPORTANT} reminder to consult them after a failed
simulation. The reminder is emitted only when the retrieved block is non-empty,
since it refers to a section that does not otherwise exist; on an empty bank the
prompt is byte-for-byte \textsc{ReAct}. It carries no physics, geometry or
placement guidance of its own.

\begin{lstlisting}[style=promptwrap]
============================================================
  SYSTEM PROMPT
============================================================
[... ReAct system prompt (puzzle description, placement constraints, tool list,
  Thought/Action format and rules) omitted; identical to the ReAct prompt above ...]

=== LEARNED PHYSICS SKILLS ===

## Catapult-Specific Skills
1. [Strategy: Right-of-Pivot Arm Strike Zone (x~=0.2--0.5 Default)] Placing the red ball
  at x~=0.2--0.5 on the right half of the catapult arm creates a downward force that lifts
  the green ball in a rightward arc toward the basket. x=0.3 is the default starting
  point; x=0.2 at standard y reduces range ~1.3 units. Shifting x rightward (>=0.5) at
  standard-y positions risks diminishing returns.
   Apply when: Default for all catapult strikes; start with x=0.3. At moderate elevated
   y (~=0.5), decreasing x below 0.3 increases range (see leftward-x skill). Do not
   exceed x~=0.5 without checking for diminishing returns.
   Example: Seed 43: x=0.3 y=-0.5 -> green at x=3.10, success. Seed 48: x=0.3 y=0.3 ->
   success. Seed 51: x=0.3 y=0.0 -> success.

2. [Strategy: Low Drop (y~=-0.5 to 0.0) Maximizes Range; Binary-Search y First] Drop
  height y is the primary range control: y~=-0.5 produces maximum range, y~=0.0 produces
  moderate range, y~=0.3--0.5 produces reduced range. A +/-0.25 change in y at standard
  heights shifts landing x by 1.5--2.0 units. When standard y (<=0.0) is blocked, the
  operative range shifts to y=0.25--0.5, where lower y still produces more range.
   Apply when: Default for all seeds; start at y=-0.5, then binary-search toward y=0.0
   if overshooting. If gray_ball blocks all y<=0.0, shift to y=0.5 as first valid
   placement, then binary-search DOWN to y=0.3 or y=0.25 to increase range. Use y=-0.25
   as first midpoint in standard regime.
   Example: Seed 49: y=-0.5 blocked, y=0.0->x=4.26 (overshoot), y=0.25->x=3.71, success.
   Seed 53: y=0.0->x=4.45, y=0.25->x=4.19, y=0.5->x=4.17 (y-dominant regime detected).

3. [Strategy: Identical Range at Different x/r Signals y-Dominant Regime] When
  trace_green_ball returns nearly identical green ball landing positions despite
  changing x or radius, the scene is in a y-dominant regime where drop height controls
  range. Stop iterating x or r and switch to y-binary-search or x-sweep at a different y
  level. If lower y is blocked, try x<0.3 at moderate elevated y (~=0.5) to escape the
  regime.
   Apply when: When two consecutive traces with different x or r values produce landing
   positions within 0.2 units of each other; immediately switch to adjusting y. If y is
   blocked and regime persists at x=0.3, try x=0.1--0.2 at y=0.5 to break out.
   Example: Seed 44: x=0.3 and x=0.2 at r=2.0 both ->x=4.50. Seed 47: x=0.3 y=0.5 and
   various r all gave x~=2.69. Seed 53: y=0.0->x=4.45, y=0.25->x=4.19, y=0.5->x=4.17
   (plateau) -> switched to x=0.1 -> success.

4. [Strategy: Overlap Fallback Sequence: y=-0.5 -> y=0.0 -> y=0.5 -> Binary-Search
  y=0.3--0.5] When y=-0.5 is blocked, try y=0.0, then y=0.5 with r=2.0. Once a valid
  placement at y=0.5 is confirmed, binary-search DOWNWARD within y=0.3--0.5 to fine-tune
  range, since lower y in this zone still increases range. Only escalate to y=1.0 if
  y=0.3 is also blocked or all y=0.3--0.5 attempts miss by more than 1 unit.
   Apply when: When predict_first_contact returns invalid for y=-0.5; follow sequence
   y=0.0 -> y=0.5 r=2.0 -> binary-search y=0.3--0.5. Skip x=1.0 at y=1.0 always. For
   elevated green ball scenes (y>=-0.5), standard y slots may all be blocked and sequence
   starts at y=1.0.
   Example: Seed 48: y=-0.5/0.0 blocked -> y=0.5 (x=2.37 undershoot) -> y=0.3 (x=3.14,
   success). Seed 51: y=-0.5 blocked -> y=0.0 valid (gap 0.21, success). Seed 52:
   y=-0.5/0.0 blocked -> y=0.5 undershoot -> bisect down.

5. [Strategy: Chaotic Trajectory Zones: Avoid Fine-Tuning in Sensitive Ranges] Chaotic
  zones exist in both y and x dimensions where tiny parameter changes cause large
  trajectory swings: (1) y~=-0.15 to -0.25 at x=0.3; (2) y~=0.3--0.4 at r=2.0 in some seeds
  (Deltay=0.1 -> >1.9 range jump); (3) y~=1.05--1.1 at elevated arms; (4) x~=0.35--0.40 at
  y=0.75. When a chaotic cliff is detected, bisect methodically; the cliff boundary can
  be as narrow as Deltay=0.01.
   Apply when: When two consecutive traces with small Delta in x or y produce landing
   positions >1.5 units apart: for y-chaotic cliffs bisect (e.g., try y=0.35 between
   y=0.3 and y=0.4); for x-chaotic cliffs simulate overshoot side. Do not iterate
   further within the chaotic range without a clear bisection strategy.
   Example: Seed 44: y=0.5->x=4.50, y=0.7->x=-0.33 (y-cliff). Seed 52: y=0.3->x=4.23,
   y=0.4->x=2.71 (cliff); bisected to y=0.34/y=0.35 boundary; simulated y=0.34 -> success.

6. [Strategy: Trace-then-Simulate: Gap Threshold, Bisect Chaotic Cliffs, Simulate Narrow
  Boundary] At standard drop heights (y<=0.0), commit to simulate when gap is under 0.8
  units. At elevated heights (y=0.25--0.5), commit at gap <0.5. When a chaotic cliff is
  detected in y=0.3--0.5, bisect to the midpoint rather than simulating immediately.
  After bisection has narrowed the cliff to Deltay<=0.01 (minimum resolution), simulating the
  overshoot boundary IS valid---the green ball can contact the blue ball during its arc
  even if the final resting position overshoots.
   Apply when: At y<=0.0: gap <0.8 -> simulate. At y=0.25--0.5: gap <0.5 -> simulate. If
   chaotic cliff: bisect first. Once Deltay<=0.01 and no gap reduction is possible, simulate
   the overshoot side---arc contact is achievable at that boundary.
   Example: Seed 51: gap ~0.21 at y=0.0 -> success. Seed 52: cliff at y=0.34/0.35 after
   full bisection -> simulated y=0.34 (overshoot) -> success via arc contact.

## Common Mistakes to Avoid
1. **Fine-tuning x or y in a chaotic zone when trajectory response is nonlinear.**
   Why it happens: Agents assume monotonic response and keep incrementing when the
   system is actually sensitive/chaotic.
   How to avoid: When two consecutive traces produce landing positions >1.5 units apart
   with small Delta, switch to bisection rather than linear search.

2. **Using x=1.0 at y=1.0 in elevated-arm scenes.**
   Why it happens: Agents choose an intermediate x value between 0.3 and 2.0 without
   checking validity.
   How to avoid: Skip x=1.0 at y=1.0 always; go directly to x=2.0 r=1.5 as the elevated
   fallback.

3. **Simulating the overshoot side of a y-chaotic cliff (y=0.3--0.5) without bisecting
  first.**
   Why it happens: Agents see a near-target landing on overshoot trace and try to
   simulate without confirming the cliff boundary.
   How to avoid: Always bisect the chaotic cliff first. Only simulate the overshoot
   boundary after bisection has narrowed to Deltay<=0.01 and no further resolution is
   possible.

4. **Placing the red ball at x>3 (far right of scene).**
   Why it happens: Agents try to increase rightward strike force by moving the contact
   point further right.
   How to avoid: Keep x within 0.1--0.5 for standard strikes; x=2.0 is maximum for
   elevated fallback.

=== END SKILLS ===

============================================================
  USER PROMPT
============================================================
Solve the Catapult puzzle. You have 25 iterations to solve it. The success condition is:
   green_ball must contact blue_ball. Use tools effectively and think about alternate 
  approaches if one does not work.
- **ALWAYS call predict_first_contact first** --- it is cheap and tells you if your red 
  ball hits the intended object.
- Only if predict confirms the right contact: call trace_green_ball to track green ball.


IMPORTANT --- After each FAILED simulation, before trying a new placement, you MUST:
1. Review the LEARNED PHYSICS SKILLS listed in the system prompt above.
2. Identify which skill applies to the failure you just observed.
3. Explicitly state which skill you are applying and why in your next Thought.
\end{lstlisting}

\subsection{\textsc{HExA} trajectory (catapult seed 64)}
\label{app:trajectories:hexa_run}

Highlighting is ours and is not part of the model's output:
\textcolor{skillblue}{blue} marks reasoning that invokes the injected skill bank,
\textcolor{scalarDark}{red} marks the observations that drive each decision, and
\textcolor{black!55}{\itshape grey} annotations give the bank entry being applied.
Skill numbers (\#1--\#6) refer to the retrieved block in
\S\ref{app:trajectories:hexa_prompt}.

\begin{lstlisting}[style=promptwrapHL]
--- Step 1 ---
Thought:
Let me start by describing the scene geometry to understand the layout.

Action:
describe_scene_geometry({})

Observation:
=== Scene Geometry ===
World bounds: x in [-5, 5], y in [-5, 5]

Balls:
  green_ball: pos=(-3.06, -1.97), r=0.20, dynamic=True
  blue_ball: pos=(3.50, -1.99), r=0.32, dynamic=True
  black_ball: pos=(-1.17, 4.60), r=0.40, dynamic=False
  gray_ball: pos=(-1.14, -3.07), r=0.70, dynamic=True

Bars:
  ledge: pos=(3.50, -2.71), angle=-3.6 deg, len=3.01, dynamic=False
  black_platform: pos=(-1.74, -3.87), angle=0.0 deg, len=3.00, dynamic=False
  gray_platform: pos=(-1.14, -2.27), angle=0.0 deg, len=4.25, dynamic=True

Baskets:
  basket: pos=(3.50, -2.51), dynamic=True

Key distances:
  green<->blue: 6.56

Success condition: green_ball <-> blue_ball for 3.0s
==================================================

--- Step 2 ---
Thought:
The scene shows a catapult setup. The green ball is at (-3.06, -1.97) on the left end of
  the gray platform (catapult arm). The gray ball pivot is at (-1.14, -3.07). The blue
  ball is at (3.50, -1.99) in a basket. I need to drop a red ball on the right side of
  the catapult arm to launch the green ball toward the blue ball. @@Following skills, I'll@@
  @@start with x=0.3, y=-0.5, r=2.0.@@ First, let me predict first contact.
  [[<- Skill #1 cat_ev_9_000 (x~=0.3 strike zone) + Skill #2 cat_ev_6_001 (low drop y=-0.5, search y first)]]

Action:
predict_first_contact({'x': 0.3, 'y': -0.5, 'radius': 2.0})

Observation:
=== Predicted First Contact ===
placement_valid: True
red_ball drop: x=0.300, y=-0.500, radius=2.000
##first_contact_object: gray_platform##
contact_step: 10
approach_speed: 1.464
contact_point (approx midpoint): (-0.420, -1.454)
contact_normal (red->other inverted): (0.664, 0.748)

--- Step 3 ---
Thought:
The red ball contacts the gray_platform (catapult arm) at step 10. This is correct - it
  hits the right side of the arm which should launch the green ball. Let me trace the
  green ball trajectory.

Action:
trace_green_ball({'x': 0.3, 'y': -0.5, 'radius': 2.0})

Observation:
=== Green Ball Trajectory (steps run: 360) ===
red_ball drop: x=0.300, y=-0.500, radius=2.000
Success: False
green_ball: start=(-3.06, -1.97) ##end=(+4.78, -2.49)## Delta=(+7.84, -0.52) peak_y=+4.74 v_max=10.31

Waypoints (step: x, y):
  step    1: (-3.06, -1.97)
  step   30: (-2.47, +0.18)
  step   60: (-0.59, +3.72)
  step   90: (+1.27, +4.74)
  step  120: (+3.05, +3.07)
  step  150: (+4.77, -1.04)
  step  180: (+4.39, -2.25)
  step  210: (+4.46, -2.47)
  step  240: (+4.58, -2.47)
  step  270: (+4.78, -2.49)
  step  300: (+4.78, -2.49)
  step  330: (+4.78, -2.49)
  step  360: (+4.78, -2.49)

--- Step 4 ---
Thought:
The green ball overshoots, landing at x=4.78 while the blue ball is at x=3.50. I need to
  reduce range. @@Following skills, I should increase y (reduce drop height). Try y=0.0.@@
  [[<- Skill #2 cat_ev_6_001 / Skill #4 cat_ev_11_002 (step y from -0.5 toward 0.0)]]

Action:
trace_green_ball({'x': 0.3, 'y': 0.0, 'radius': 2.0})

Observation:
=== Green Ball Trajectory (steps run: 240) ===
red_ball drop: x=0.300, y=0.000, radius=2.000
Success: False
green_ball: start=(-3.06, -1.97) ##end=(-0.92, -3.09)## Delta=(+2.14, -1.12) peak_y=+3.94 v_max=11.93

Waypoints (step: x, y):
  step    1: (-3.06, -1.97)
  step   30: (-2.97, -0.95)
  step   60: (-1.48, +3.94)
  step   90: (-1.36, +2.11)
  step  120: (-1.30, -2.32)
  step  150: (-0.92, -3.09)
  step  180: (-0.92, -3.09)
  step  210: (-0.92, -3.09)
  step  240: (-0.92, -3.09)

--- Step 5 ---
Thought:
Big trajectory swing between y=-0.5 (lands x=4.78, overshoot) and y=0.0 (lands x=-0.92,
  massive undershoot). @@This is a chaotic cliff per Skill #5. I should bisect at y=-0.25.@@
  [[<- Skill #5 cat_ev_9_006 (chaotic zones) + Mistake #1 (bisect, don't fine-tune)]]

Action:
trace_green_ball({'x': 0.3, 'y': -0.25, 'radius': 2.0})

Observation:
=== Green Ball Trajectory (steps run: 330) ===
red_ball drop: x=0.300, y=-0.250, radius=2.000
Success: False
green_ball: start=(-3.06, -1.97) ##end=(+3.29, -1.74)## Delta=(+6.35, +0.23) peak_y=+3.78 v_max=10.75

Waypoints (step: x, y):
  step    1: (-3.06, -1.97)
  step   30: (-2.76, -0.35)
  step   60: (-1.25, +3.54)
  step   90: (+0.52, +3.78)
  step  120: (+2.36, +1.23)
  step  150: (+3.97, -0.87)
  step  180: (+3.70, -1.59)
  step  210: (+3.60, -1.59)
  step  240: (+3.31, -1.71)
  step  270: (+3.29, -1.74)
  step  300: (+3.29, -1.74)
  step  330: (+3.29, -1.74)

--- Step 6 ---
Thought:
At y=-0.25, the green ball lands at x=3.29, which is close to the blue ball at x=3.50
  (gap ~0.21). @@This is within the commit threshold of 0.8 per Skill #6. Let me simulate@@
  @@this.@@
  [[<- Skill #6 cat_ev_11_003 (trace-then-simulate: commit when gap < 0.8)]]

Action:
simulate_action({'x': 0.3, 'y': -0.25, 'radius': 2.0})

Observation:
##SUCCESS! The green ball contacted the blue ball in the basket.##
Total simulation steps: 423
Reward: 1.0

Final object positions:
  green_ball: pos=(3.289, -1.741), vel=(0.000, 0.000)
  red_ball: pos=(0.527, -2.229), vel=(0.000, 0.000)
  blue_ball: pos=(3.653, -2.106), vel=(0.000, 0.000)
  ledge: pos=(3.500, -2.712), vel=(0.000, 0.000)
  basket: pos=(3.501, -2.596), vel=(0.000, 0.000)
  black_ball: pos=(-1.170, 4.600), vel=(0.000, 0.000)
  black_platform: pos=(-1.738, -3.869), vel=(0.000, 0.000)
  gray_ball: pos=(-2.912, -3.064), vel=(0.000, 0.000)
  gray_platform: pos=(-0.730, -3.363), vel=(0.000, 0.000)

--- Step 7 ---
Thought:
(auto-finish after Success:True)

Action:
finish({'x': 0.3, 'y': -0.25, 'radius': 2.0})

Observation:
SUCCESS! The green ball contacted the blue ball in the basket.
Total simulation steps: 423
Reward: 1.0

Final object positions:
  green_ball: pos=(3.289, -1.741), vel=(0.000, 0.000)
  red_ball: pos=(0.527, -2.229), vel=(0.000, 0.000)
  blue_ball: pos=(3.653, -2.106), vel=(0.000, 0.000)
  ledge: pos=(3.500, -2.712), vel=(0.000, 0.000)
  basket: pos=(3.501, -2.596), vel=(0.000, 0.000)
  black_ball: pos=(-1.170, 4.600), vel=(0.000, 0.000)
  black_platform: pos=(-1.738, -3.869), vel=(0.000, 0.000)
  gray_ball: pos=(-2.912, -3.064), vel=(0.000, 0.000)
  gray_platform: pos=(-0.730, -3.363), vel=(0.000, 0.000)
\end{lstlisting}

\paragraph{How the bank drove the solve.}
After reading the scene, the agent starts from the bank's default: the
right-of-pivot strike zone $x{\approx}0.3$ (\texttt{cat\_ev\_9\_000}) with a low
drop $y{=}{-}0.5$ and large radius $r{=}2.0$, searching over $y$ first
(\texttt{cat\_ev\_6\_001}). One \texttt{predict\_first\_contact} call confirms that the
red ball strikes the arm, and a trace shows the green ball overshooting to
$x{=}4.78$ (blue ball at $x{=}3.50$). Raising the drop to $y{=}0.0$ swings the
landing point to $x{=}{-}0.92$, a large undershoot. Rather than micro-tuning, the
agent recognises this jump as a chaotic cliff (\texttt{cat\_ev\_9\_006}) and bisects
to $y{=}{-}0.25$, which lands the green ball at $x{=}3.29$. This gap of
${\approx}0.21$ is below the commit threshold of \texttt{cat\_ev\_11\_003}
(Trace-then-Simulate), so the agent spends its only \texttt{simulate\_action} on this
placement, which succeeds. The whole solve takes three traces and one simulation.

\section{Working Example: Episodic RAG vs.\ \textsc{HExA} on Level Pass the Parcel of InterPhyre}
\label{app:trajectories_ptp}

Our second example compares \textsc{HExA} with the strongest baseline,
Episodic RAG, on \texttt{pass\_the\_parcel} seed~57. The goal is to knock an
inverted top basket off its platform so that the green ball it covers falls onto
the blue ball in the bottom basket. Both agents use Claude Sonnet~4.6 as the actor,
the same tools and the same 25-iteration budget, and both run with frozen memory
on held-out seeds 57--106. \textsc{HExA} uses its round-17 bank; Episodic RAG uses
its store of 51 raw rollouts (25 successes, 26 failures) collected on the training
seeds. The only difference is what the memory contains. RAG retrieves two complete
past rollouts (6{,}862 added tokens). \textsc{HExA} retrieves six distilled
skills and three mistakes. RAG uses all 25 iterations and fails; \textsc{HExA}
succeeds in 5 iterations with final placement $(x{=}{-}0.072,\,y{=}1.3,\,r{=}0.7)$.

The system prompts follow the templates of
Appendix~\ref{app:systemprompts:rag} (Episodic RAG) and
Appendix~\ref{app:trajectories:bank} (\textsc{HExA}), on top of the
\texttt{pass\_the\_parcel} \textsc{ReAct} prompt
(Appendix~\ref{app:prompt:ptp}), so we omit them. RAG's retrieved memory is two complete
rollouts: a seed-33 success, final placement $(0.934, 0.916, 0.5)$, and a seed-28
failure; we do not reproduce them.
Highlighting follows \S\ref{app:trajectories:hexa_run}:
\textcolor{skillblue}{blue} marks reasoning that draws on the retrieved memory,
\textcolor{scalarDark}{red} marks decisive observations, and
\textcolor{black!55}{\itshape grey} lines are our annotations.

\subsection{\textsc{HExA} retrieved skills (pass\_the\_parcel seed 57)}
\label{app:trajectories_ptp:hexa_skills}
\label{app:trajectories_ptp:rag_memory}

\begin{lstlisting}[style=promptwrap]
=== LEARNED PHYSICS SKILLS === (titles only; principle, "Apply when" and
  "Example" fields omitted)

## Pass The Parcel-Specific Skills
1. [pas_ev_7_001] Radius r=0.7 Default; Escalate to r=1.0 Before r=1.2 Only When Basket
  Actually Moved
2. [pas_ev_6_002] Switch to Opposite Side When Hit Direction Misaligns with Blue
3. [pas_ev_7_000] Direct Lateral Sidewall Hit: Side Selection and y Fallback Sequence
4. [pas_ev_2_002] Basket Must Clear Green's Fall Path by >0.7 Units Horizontally
5. [pas_ev_7_004] Lower y (0.5-1.7) for Direct Sidewall Contact When Basket Barely Moves
6. [pas_ev_7_007] Ramp Approach When All Direct Sidewall Heights Are Blocked

## Common Mistakes to Avoid
1. [pas_err_000] Agent escalates directly to r=1.2, r=1.5, or r=2.0 when r=0.7 strands
  green, causing green to overshoot far past blue.
2. [pas_err_001] Agent places the red ball directly above the inverted basket expecting
  downward force to dislodge it, but the ball lands on the basket's solid top surface
  and sits there.
3. [pas_evo_002] Agent exhausts radius escalations (r=0.7, r=1.0, r=1.2) at the same
  high y when the basket barely moves, failing to recognize that the contact angle ---
  not the force --- is the problem.
=== END SKILLS ===
\end{lstlisting}

\subsection{Episodic RAG trajectory (pass\_the\_parcel seed 57)}
\label{app:trajectories_ptp:rag_run}

\begin{lstlisting}[style=promptwrapHL]
.... Step 1 condensed (thought omitted):
  [1] get_level_state({}) -> scene description

--- Step 2 ---
Thought: @@Based on the past successful rollout (seed 33), the winning strategy was@@
  @@placing the red ball just above and slightly to the right of the top basket.@@ The
  basket in this seed is at (0.428, 0.051). Following the same pattern (offset ~0.15
  right, ~0.87 above the basket), I'll try x=0.578, y=0.921, radius=0.5. @@But let me@@
  @@first try the exact same coordinates that worked in seed 33: x=0.934, y=0.916,@@
  @@radius=0.5.@@
  [[<- reuses seed 33's absolute placement; here the basket sits 0.36 units further left]]
Action: simulate_action({'x': 0.934, 'y': 0.916, 'radius': 0.5})
Observation: FAILURE. The green ball did NOT contact the blue ball.
  green_ball: pos=(-0.085, 0.206), vel=(0.000, 0.000)
  blue_ball: pos=(-1.329, -4.455), vel=(0.000, 0.000)
  top_basket: pos=(0.292, 0.126), vel=(0.000, 0.000)
  ....

--- Step 3 ---
Thought: The basket moved slightly but green didn't fall to blue. Let me try placing the
  ball closer to the basket center, @@directly above it to push it down and off the@@
  @@platform.@@
  [[<- the drop-on-top error HExA's Mistake #2 (pas_err_001) warns against]]
Action: simulate_action({'x': 0.428, 'y': 1.2, 'radius': 0.5})
Observation: FAILURE. The green ball did NOT contact the blue ball.
  ##green_ball: pos=(0.428, 0.210)##, vel=(0.000, 0.000)
  blue_ball: pos=(-1.329, -4.455), vel=(0.000, 0.000)
  top_basket: pos=(0.429, 0.116), vel=(0.000, 0.000)
  ....

--- Step 4 ---
Thought: The basket barely moved. I need to push it harder to the left so green falls
  into the bottom basket. @@Let me try placing the ball to the right of the basket to push@@
  @@it left with more force.@@
  [[<- wrong side: blue is LEFT of the basket (cf. HExA Skill #2, pas_ev_6_002)]]
Action: simulate_action({'x': 1.2, 'y': 0.1, 'radius': 0.7})
Observation: INVALID placement:
  - Overlaps with green_ball: dist=0.7977, sum_of_radii=0.9500. Move at least 0.2023
    units further away.
  - Overlaps with black_platform: dist_to_platform=0.2492 < radius=0.7000. Move at least
    0.5008 units further away from the platform.
  ....

.... Steps 5--20 condensed (thoughts omitted):
  [5] simulate_action({'x': 1.5, 'y': 0.5, 'radius': 0.6}) -> FAILURE; green at (0.428,
    0.301)
  [6] simulate_action({'x': 3.5, 'y': 4.0, 'radius': 0.5}) -> FAILURE; green at (-1.314,
    -2.999)
  [7] simulate_action({'x': 3.7, 'y': 4.2, 'radius': 0.6}) -> FAILURE; green at (-1.390,
    -2.995)
  [8] simulate_action({'x': 0.6, 'y': 0.95, 'radius': 0.5}) -> INVALID placement
  [9] simulate_action({'x': 0.8, 'y': 1.1, 'radius': 0.5}) -> FAILURE; green at (0.931,
    0.206)
  [10] simulate_action({'x': 1.6, 'y': 0.3, 'radius': 0.8}) -> INVALID placement
  [11] simulate_action({'x': 1.8, 'y': 0.5, 'radius': 0.7}) -> INVALID placement
  [12] simulate_action({'x': 1.8, 'y': 0.8, 'radius': 0.7}) -> FAILURE; green at (0.428,
    0.210)
  [13] simulate_action({'x': 4.0, 'y': 4.4, 'radius': 0.5}) -> FAILURE; green at
    (-1.299, -3.001)
  [14] simulate_action({'x': 4.2, 'y': 4.4, 'radius': 0.7}) -> INVALID placement
  [15] simulate_action({'x': 4.2, 'y': 4.2, 'radius': 0.7}) -> FAILURE; green at
    (-4.725, -4.560)
  [16] simulate_action({'x': 0.65, 'y': 1.3, 'radius': 0.4}) -> FAILURE; green at
    (0.428, 0.210)
  [17] simulate_action({'x': 3.5, 'y': 4.0, 'radius': 0.6}) -> FAILURE; green at
    (-1.360, -2.994)
  [18] simulate_action({'x': 3.6, 'y': 4.3, 'radius': 0.55}) -> FAILURE; green at
    (-1.393, -3.684)
  [19] simulate_action({'x': 3.65, 'y': 4.3, 'radius': 0.6}) -> FAILURE; green at
    (-1.387, -2.990)
  [20] simulate_action({'x': 3.6, 'y': 4.3, 'radius': 0.58}) -> FAILURE; green at
    (-1.349, -3.001)

--- Step 21 ---
Thought: With r=0.55 green was at y=-3.684 (very close to blue), but r=0.58 sent green
  back to y=-3.0. @@Let me try r=0.56 to find the sweet spot.@@
  [[<- steps 17-25: r tuned within 0.55--0.60 at almost fixed (x, y) on the ramp; green]]
  [[   keeps stopping on the basket rim and the budget runs out]]
Action: simulate_action({'x': 3.6, 'y': 4.3, 'radius': 0.56})
Observation: FAILURE. The green ball did NOT contact the blue ball.
  green_ball: pos=(-1.429, -3.689), vel=(0.000, 0.000)
  blue_ball: pos=(-1.290, -4.451), vel=(0.000, 0.000)
  top_basket: pos=(-1.416, -2.891), vel=(0.000, 0.000)
  ....

.... Steps 22--25 condensed (thoughts omitted):
  [22] simulate_action({'x': 3.6, 'y': 4.3, 'radius': 0.57}) -> FAILURE; green at
    (-1.432, -3.689)
  [23] simulate_action({'x': 3.4, 'y': 4.3, 'radius': 0.56}) -> FAILURE; green at
    (-1.447, -3.000)
  [24] simulate_action({'x': 3.6, 'y': 4.3, 'radius': 0.555}) -> FAILURE; green at
    (-1.434, -3.690)
  [25] simulate_action({'x': 3.8, 'y': 4.0, 'radius': 0.55}) -> FAILURE; green at
    (-1.407, -3.700)
\end{lstlisting}

\subsection{\textsc{HExA} trajectory (pass\_the\_parcel seed 57)}
\label{app:trajectories_ptp:hexa_run}

\begin{lstlisting}[style=promptwrapHL]
--- Step 1 ---
Thought: Let me start by getting the current level state to understand the positions of
  all objects.
Action: get_level_state({})
Observation: === Level State: pass_the_parcel ===
* green_ball (green Ball): position=(0.4279, 0.3008), radius=0.2500, dynamic=True
* ##blue_ball (blue Ball): position=(-1.3292, -4.0400)##, radius=0.3600, dynamic=True
* top_basket: {##'x': 0.42794570936818155, 'y': 0.050799519646729074##, 'color': 'gray',
  'dynamic': True, 'type': 'Basket', 'angle': 180, 'scale': 0.6}
  ....

--- Step 2 ---
Thought: Blue ball is at x=-1.3292 (LEFT of basket at x=0.4279). @@Per Skill 2, I should@@
  @@use a left-side hit.@@ @@Per Skill 3, basket_y=0.0508 > 0, so start at y=2.0.@@ @@Placement:@@
  @@x=basket_x-0.5=0.4279-0.5=-0.0721, y=2.0, r=0.7.@@
  [[<- Skill #2 pas_ev_6_002 (blue left of basket -> x=basket_x-0.5), Skill #3 pas_ev_7_000]]
  [[   (basket_y>0 -> y=2.0), Skill #1 pas_ev_7_001 (r=0.7 default)]]
Action: simulate_action({'x': -0.0721, 'y': 2.0, 'radius': 0.7})
Observation: FAILURE. The green ball did NOT contact the blue ball.
  green_ball: pos=(0.428, 0.210), vel=(0.000, 0.000)
  blue_ball: pos=(-1.691, -4.447), vel=(0.000, 0.000)
  ##top_basket: pos=(0.431, 0.116)##, vel=(0.000, 0.000)
  ....

--- Step 3 ---
Thought: Basket barely moved (from 0.0508 to 0.116 = 0.065 units). @@Per Skill 5 and@@
  @@Mistake 3, basket moved <0.12 units at y=2.0, so I should NOT escalate radius.@@
  @@Instead, lower y to 0.5.@@ Try x=-0.0721, y=0.5, r=0.7.
  [[<- Skill #5 pas_ev_7_004 + Mistake #3 pas_evo_002: wrong contact angle, not too little force]]
Action: simulate_action({'x': -0.0721, 'y': 0.5, 'radius': 0.7})
Observation: INVALID placement:
  - Overlaps with green_ball: dist=0.5383, sum_of_radii=0.9500. Move at least 0.4617
    units further away.
  - Overlaps with black_platform: dist_to_platform=0.6492 < radius=0.7000. Move at least
    0.1008 units further away from the platform.
  ....

--- Step 4 ---
Thought: y=0.5 is blocked by green ball and platform. @@Per Skill 5, try y=1.0 next.@@
Action: simulate_action({'x': -0.0721, 'y': 1.0, 'radius': 0.7})
Observation: INVALID placement:
  - Overlaps with green_ball: dist=0.8596, sum_of_radii=0.9500. Move at least 0.1404
    units further away.
  ....

--- Step 5 ---
Thought: y=1.0 blocked by green ball. @@Per Skill 5, try y=1.3 next.@@
  [[<- Skill #5 fallback ladder y=0.5 -> 1.0 -> 1.3 at fixed x, r=0.7]]
Action: simulate_action({'x': -0.0721, 'y': 1.3, 'radius': 0.7})
Observation: ##SUCCESS! The green ball contacted the blue ball.##
  green_ball: pos=(-1.606, -4.032), vel=(0.000, 0.000)
  blue_ball: pos=(-1.171, -4.452), vel=(0.000, 0.000)
  top_basket: pos=(1.251, 0.119), vel=(0.000, 0.000)
  ....

.... Step 6 condensed (thought omitted):
  [6] finish({'x': -0.0721, 'y': 1.3, 'radius': 0.7}) -> SUCCESS! The green ball
    contacted the blue ball; green at (-1.606, -4.032)
\end{lstlisting}

\paragraph{Why raw episodes fail where distilled skills succeed.}
Both agents face the same three questions: which side to strike the basket from, at
what height, and what to change after a miss. The retrieved RAG episodes answer none
of them in a form that transfers. The agent copies seed~33's \emph{absolute}
placement (step~2), although this seed's basket sits 0.36 units further left.
It then drops the ball on top of the basket (step~3), strikes from the right although
the blue ball is on the left (steps~4--12), and finally settles on the ramp,
fine-tuning the radius between 0.55 and 0.60 for its last nine calls while the green
ball keeps stopping on the basket rim. \textsc{HExA}'s bank states the same
experience as rules conditioned on the geometry. The side rule (Skill~\#2) and the
height rule (Skill~\#3) produce a sensible first placement,
$x{=}\text{basket}_x{-}0.5$, $y{=}2.0$, $r{=}0.7$, computed rather than copied.
When the first strike barely moves the basket (0.065 units), Skill~\#5 and
Mistake~\#3 read this as a wrong contact angle rather than too little force.
The agent therefore keeps the radius fixed and steps $y$ down its fallback ladder
($0.5 \to 1.0 \to 1.3$); the two blocked placements cost no simulation. The
third height succeeds.

\section{Working Example: MemRL vs.\ \textsc{HExA} on World Circle of \textsc{DiscoverPhysics}}
\label{app:trajectories_disc}

Our third example compares \textsc{HExA} with MemRL on the \texttt{circle} world of
\textsc{DiscoverPhysics} (Appendix~\ref{app:external_envs:discoverphysics}), seed~300.
One centre particle and ten ring particles interact under a hidden
fractional-Laplacian law. The agent sets only the ring radius, a shared tangential
velocity and the measurement times, and within 10 rounds must submit an executable
law together with a prose explanation. An episode passes if normalised MSE is below
$0.1$ and the explanation score, from the independent \texttt{claude-opus-4-6} judge,
is at least $0.75$. Both arms use the same actor and prompts
(Appendix~\ref{app:ext_prompts:discovery}) and frozen memory built within the
world. They differ only in the retrieved block: \textsc{HExA} receives 11 skills and 6
mistakes from its frozen bank, and MemRL two archived episodes (4{,}689 added tokens).
Across the 15 held-out seeds the judge's mean explanation score is $0.887$ for
\textsc{HExA} and $0.540$ for MemRL; \textsc{HExA} passes on 9 seeds and MemRL on
none.

The benchmark logs the agent's thoughts with numbers and law names masked
(\texttt{<n>}, \texttt{<law-form>}), and we keep that masking. Submitted laws and
explanations are shown unmasked. Experiment actions are condensed to one entry per
experiment (radius \texttt{r}, tangential velocity \texttt{v}, measurement times
\texttt{t}), raw observations are omitted, and submitted code is cut to its key lines
(\texttt{....}). Highlighting follows \S\ref{app:trajectories:hexa_run}.

\subsection{Retrieved memory (circle seed 300)}
\label{app:trajectories_disc:memory}

\begin{lstlisting}[style=promptwrap]
=== RETRIEVED SKILLS (titles only) ===
- [dis_ev_5_000] Map the measured exponent onto a named field-theoretic/operator
  formalism, not just 'anomalous decay'
- [dis_ev_3_000] Never declare a cross-validated constant/exponent as fittable ---
  including 'derived' ones
- [dis_ev_1_001] A structural interpretation ambiguity is not evidence of numerical
  uncertainty
- [dis_sk_002] Extract exponents from early-time, negligible-displacement acceleration
- [dis_sk_003] Use zero tangential velocity to isolate the central force
- [dis_ev_2_004] Use nonzero tangential velocity to test velocity-dependence and
  conservation laws
- [dis_ev_2_008] Conserved-quantity checks as an independent cross-check of
  centrality/velocity-independence
- [dis_ev_5_009] A tight, drift-free power-law fit is good evidence but does not by
  itself guarantee full credit --- still check for curvature before over-trusting the
  naive familiar-law framing
- [dis_ev_4_002] Trajectory accuracy and explanation quality are scored and gated
  independently
- [dis_ev_4_004] State the 'uniform coupling across all particles' assumption explicitly
  in the explanation
- [dis_ev_4_010] Cross-check exponent/constant invariance across radius range, but
  recognize it does not by itself resolve the causal/formalism ambiguity

=== RETRIEVED MISTAKES (first sentence only) ===
- [dis_evo_000] Explanations describe the discovered law only in terms of the
  exponent/coupling relative to familiar gravity ('softer/steeper than usual',
  'anomalous decay') and the ring-collapse method used, without naming the theoretical
  operator/formalism the exponent implies or mapping it to the standard parameter (e.g.
  alpha) for that formalism.
- [dis_evo_001] An episode achieves near-perfect or excellent trajectory error yet still
  fails overall because the explanation score is judged inadequate.
- [dis_err_002] Agent declares an already well-constrained constant (coupling strength,
  exponent), including ones only 'derived' or backed out geometrically rather than
  directly measured, as a 'fittable parameter' via **params, letting the evaluator
  re-fit it on the episode's own training data --- with unpredictable (sometimes better,
  sometimes much worse) results.
- [dis_evo_003] A clean, cross-validated power-law fit (consistent exponent across
  several tested radii) is treated as fully sufficient for the explanation, when what's
  still missing is connecting that exponent to a named physical/operator formalism.
- [dis_err_003] Agents repeatedly tried to invoke an MSE-fitting tool (run_mse_fit) late
  in the session and got 'trajectory CSV logging disabled' errors, wasting a round, then
  fell back to manual/eyeball fitting without adjusting confidence or explanation
  accordingly.
- [dis_err_004] Agents wasted an entire round attempting to shell out via a 'no_tag'
  bash-like action, which is not a valid action in this protocol, forfeiting a round
  with no data gained.

=== RETRIEVED MEMORIES (2 of 2; bodies omitted except one sentence) ===
### Memory 1 --- FAILURE (world W6, seed 214, utility Q=-0.987, updates=16)
### Memory 2 --- SUCCESS (world W6, seed 202, utility Q=-0.987, updates=15)

Memory 1, "What went wrong" (excerpt): ... Future runs should sweep
  initial_tangential_velocity systematically at fixed ring_radius across several
  fractions of the naive circular value and record long time series (many revolutions)
  via measurement_times, since orbit shape (closed vs. precessing vs. escaping) over
  long horizons is far more diagnostic than early-time acceleration alone. ...
\end{lstlisting}

\subsection{MemRL trajectory (circle seed 300)}
\label{app:trajectories_disc:memrl_run}

\begin{lstlisting}[style=promptwrapHL]
--- Round 1 ---
Thought: I'll start by measuring early-time radial dynamics at zero tangential velocity
  across several ring radii to extract the force scaling with distance.
Action: run_experiment (4 experiments shown; log truncated after these): r=2 v=0
  t=0.1..5 (10 times); r=4 v=0 t=0.1..5 (10 times); r=6 v=0 t=0.1..5 (10 times); r=8 v=0
  t=0.1..5 (10 times)
Observation: [noisy positions and velocities of all 11 particles at each time]
  ....

--- Round 2 ---
Thought: Data at t=<n> across ring radii <n>/<n>/<n>/<n> shows early radial velocity
  magnitudes that fit a*r^<n> ~= constant almost perfectly (checked via a*r^<n> giving
  ~<n> in all four cases, and the <n>x radius ratio giving the predicted <n> velocity
  ratio to within <n>%). This points to an attractive all-pairs force ~<n>/d^<n>. I'll
  now refine this over a wider radius range with finer early-time sampling, and @@check@@
  @@orbital shape at nonzero tangential velocity for consistency.@@
Action: run_experiment (5 experiments): r=1 v=0 t=0.01..0.1 (4 times); r=3 v=0
  t=0.01..0.1 (4 times); r=10 v=0 t=0.01..0.1 (4 times); r=20 v=0 t=0.01..0.1 (4 times);
  r=5 v=0.4 t=2..40 (10 times)
Observation: [noisy positions and velocities of all 11 particles at each time]
  ....

--- Round 3 ---
Thought: The exponent fits cleanly to n~=<n> (force ~ <n>/d^<n>) with an aggregate
  radial constant K~=<n> consistent across radii <n>--<n> @@I'll now test a predicted@@
  @@circular-orbit velocity to confirm the force law@@ and check for velocity-independence.
Action: run_experiment (3 experiments): r=5 v=0.6128 t=5..90 (10 times); r=5 v=0
  t=0.01..0.1 (10 times); r=15 v=0 t=0.01..0.05 (5 times)
Observation: [noisy positions and velocities of all 11 particles at each time]
  ....

--- Round 4 ---
Thought: The r=<n> circular-orbit test with v=<n> (predicted from K~=<n>, n=<n>) shows
  particles maintaining constant radius ~<n> out to t=<n>, confirming the force law. The
  r=<n> early-time data also matches the predicted K/r^<n> scaling within noise. I'll
  run one more precision check to nail down the exponent and coupling constant before
  finalizing.
Action: run_experiment (4 experiments): r=1.5 v=0 t=0.005..0.02 (4 times); r=7 v=0
  t=0.005..0.02 (4 times); r=12 v=0 t=0.005..0.02 (4 times); r=5 v=2 t=1..12 (5 times)
Observation: [noisy positions and velocities of all 11 particles at each time]
  ....

--- Round 5 ---
Thought: The n=<n> power-law fits cleanly across radii <n>--<n> (log-log slope ~=<n>),
  and the r=<n> orbit at the predicted circular velocity stayed circular through t=<n>,
  confirming velocity-independence of the radial force. @@I'll now test a sub-circular@@
  @@velocity at r=<n> over many periods to check for orbital precession@@, which will
  further validate the exponent via Bertrand's theorem behavior.
  [[<- follows Memory 1's advice (long, multi-period orbits at sub-circular velocities). The]]
  [[   orbit tests confirm the power law already measured in round 2; nothing in the memory asks]]
  [[   for the operator, the coupling structure or which parameters to leave free]]
Action: run_experiment (1 experiment): r=5 v=0.45 t=5..120 (10 times)
Observation: [noisy positions and velocities of all 11 particles at each time]
  ....

--- Round 6 ---
Thought: The v=<n> sub-circular orbit (below the ~<n> circular velocity) shows periodic
  radial <law-form> with precession --- consistent with an attractive power-law force
  with exponent !=<n> (non-<law-form>ian), matching my n~=<n> fit. Combined with the
  clean r^<n> scaling across radii <n>--<n> and the confirmed circular orbit at the
  predicted velocity, I'm confident in the functional form and will submit.
Action: submit_final_law
def fit_parameters():
    return {
        ##"G": {"init": 0.84, "bounds": [0.1, 3.0]},##
        ##"q": {"init": 1.5, "bounds": [1.0, 2.5]},##
    ....
def discovered_law(positions, velocities, duration, **params):
    ....
    G = params.get("G", 0.84)
    q = params.get("q", 1.5)
    ....
        inv = 1.0 / (dist ** (q + 1.0))
    ....
<explanation> The universe contains an all-to-all attractive interaction: every particle
  pulls every other particle toward it with an acceleration whose magnitude ##falls off as##
  ##a power law G/d^q of their separation d, with the best-fit exponent near q~=1.5##
  ##(softer than an inverse-square law).## Ring particles collapse radially under zero
  initial velocity with radial acceleration scaling cleanly as r^-1.5 across nearly two
  decades of radius, and orbits at the predicted circular velocity remain stable
  circles, while sub-circular velocities produce precessing (non-closed) elliptical-like
  orbits --- a signature that the force law departs from the inverse-square (Keplerian)
  case as expected from Bertrand's theorem. The center particle participates in this
  same pairwise coupling as an ordinary member of the system, not as a fixed potential
  source.
  [[<- same exponent as HExA, but no operator is named, and both G and q are left free]]
Observation: ##norm_mse=3.990587765259953, expl_score=0.8, passed=False##
  [[Opus-4.6 judge: explanation 0.5; FAIL (nMSE 3.99 > 0.1, explanation < 0.75)]]
\end{lstlisting}

\subsection{\textsc{HExA} trajectory (circle seed 300)}
\label{app:trajectories_disc:hexa_run}

\begin{lstlisting}[style=promptwrapHL]
--- Round 1 ---
Thought: I'll start with @@zero-tangential-velocity experiments at multiple radii to@@
  @@isolate the central radial force, using early-time snapshots to measure acceleration@@
  @@before significant displacement.@@
  [[<- dis_sk_003 (zero tangential velocity isolates the central force) + dis_sk_002 (early-time,]]
  [[   negligible-displacement acceleration)]]
Action: run_experiment (3 experiments): r=3 v=0 t=0.05..10 (10 times); r=6 v=0
  t=0.05..10 (10 times); r=10 v=0 t=0.05..10 (10 times)
Observation: [noisy positions and velocities of all 11 particles at each time]
  ....

--- Round 2 ---
Thought: Clean signal: acceleration scales as r^<n> across all three radii with a highly
  consistent effective constant (~=<n> = a*r^<n>). This matches a fractional power-law
  central attraction, though @@the ring topology makes it ambiguous whether this is purely@@
  @@from the center particle or a pairwise sum over all <n> particles with equal coupling.@@
  @@Next I'll test with nonzero tangential velocity to check angular momentum conservation@@
  and centripetal balance, which will validate that the force is purely radial and
  velocity-independent.
  [[<- dis_ev_1_001 (structural ambiguity) + dis_ev_2_004 (nonzero velocity tests conservation laws)]]
Action: run_experiment (4 experiments): r=3 v=0.698 t=0.5..10 (7 times); r=3 v=0.3
  t=0.5..10 (7 times); r=1 v=0 t=0.05..0.4 (4 times); r=20 v=0 t=0.05..0.4 (4 times)
Observation: [noisy positions and velocities of all 11 particles at each time]
  ....

--- Round 3 ---
Thought: @@Angular momentum is well conserved in the tangential-velocity runs@@ (L stays
  near its initial value across all measured times), confirming the force is purely
  central and velocity-independent. Velocity data (less noisy than positions) at early
  times show acceleration scaling as r^<n> with an effective constant ~=<n>, consistent
  across radii <n>--<n> (r=<n> was noisier due to faster collapse). @@Let me get one more@@
  @@wide radius sweep at very early times@@ to nail down the exponent and constant precisely
  before finalizing.
  [[<- dis_ev_2_008 (conserved-quantity cross-check) + dis_ev_4_010 (invariance across radius range)]]
Action: run_experiment (5 experiments): r=2 v=0 t=0.02..0.06 (3 times); r=4 v=0
  t=0.02..0.06 (3 times); r=8 v=0 t=0.02..0.06 (3 times); r=15 v=0 t=0.02..0.06 (3
  times); r=30 v=0 t=0.02..0.06 (3 times)
Observation: [noisy positions and velocities of all 11 particles at each time]
  ....

--- Round 4 ---
Thought: The wide radius sweep (r=<n> to <n>, spanning a factor of <n>) confirms the
  acceleration follows a clean power law a ~= <n>/r^<n> with essentially no drift
  (regression slope <n>, R^2 near <n>). Angular momentum conservation from the earlier
  tangential-velocity runs confirms the force is purely central and
  velocity-independent. This points to an attractive pairwise interaction with a
  fractional decay exponent (a*r^<n>=const), @@consistent with a fractional-order field@@
  @@operator of order alpha~=<n> in <n>D@@ (rather than the standard <n>D
  <law-form>/<law-form> analog which would give <n>/r). I'm confident in the functional
  form; I'll finalize now.
  [[<- dis_ev_5_000 (map the exponent onto a named operator formalism)]]
Action: submit_final_law
def discovered_law(positions, velocities, duration, **params):
    ....
    C = params.get("C_pair", 0.167)
    ....
            @@factor = C / dist ** 2.5@@
    ....
def fit_parameters():
    return {
        @@"C_pair": {"init": 0.167, "bounds": [0.02, 0.6]},@@
    ....
<explanation> All 11 particles interact through a uniform, all-pairs attractive coupling
  whose magnitude falls off as 1/r^1.5 with pairwise separation --- steeper than the
  natural 2D Newtonian analog (1/r, from the standard 2D Poisson equation) but shallower
  than an inverse-square law. This decay rate is consistent with a fractional-order
  field operator (@@a fractional Laplacian of order alpha~=0.75 in the -(-Delta)^alpha@@
  @@convention@@), rather than the integer-order Laplacian that gives the familiar
  logarithmic/1/r 2D potential. I measured the exponent from early-time
  (negligible-displacement) accelerations across radii spanning 2--30 with zero
  tangential velocity, finding a consistent power-law slope of -1.50 with no curvature
  across that entire range, and confirmed the force is purely central and
  velocity-independent by checking angular momentum conservation in runs with nonzero
  tangential velocity. The ring symmetry makes the effective per-particle constant
  degenerate between a "force from center only" and "pairwise sum over all particles"
  interpretation; I resolved this by @@adopting the more general pairwise model with a@@
  @@uniform coupling constant applied between every particle pair (including the center),@@
  @@leaving that coupling constant as the only fittable parameter@@ since its value depends
  on how the geometric ring-sum decomposes.
  [[<- dis_ev_5_000: names the operator and its order]]
  [[<- dis_ev_4_004: states the uniform-coupling assumption]]
  [[<- dis_ev_3_000: exponent fixed at 1.5 in the code; only C_pair is free]]
Observation: ##norm_mse=2.2647538526815615e-06, expl_score=1.0, passed=True##
  [[Opus-4.6 judge: explanation 1.0; PASS at both the 0.75 and 0.90 gates]]
\end{lstlisting}

\paragraph{Same measurement, different submission.}
Both agents open with the same zero-velocity ring collapse and both measure the
same $1/r^{1.5}$ all-pairs attraction. The difference lies in what the memory makes
them do next. MemRL's retrieved failure reflection tells it to trust long,
multi-period orbits over early-time scaling, so it spends rounds~3--6 on
circular-orbit and precession tests that only confirm the exponent it already has.
It then submits a generic power law $G/d^{q}$ with both $G$ and $q$ free, and an
explanation that never names the operator; it fails on both axes. \textsc{HExA}'s
bank encodes what the scorer actually checks. Its experiments follow the
\emph{procedural} skills: zero velocity to isolate the central force, early times for
the exponent, conserved angular momentum as a cross-check and a wide radius sweep
for invariance. Its submission follows the \emph{reporting} skills: it names the
fractional Laplacian with $\alpha{\approx}0.75$, states the uniform-coupling
assumption and fixes the measured exponent in code, leaving only the coupling free.
It passes in 4 rounds with normalised MSE $2{\times}10^{-6}$ and an explanation score
of $1.0$.

\section{Working Example: Episodic RAG vs.\ \textsc{HExA} on Mastermind with Qwen3-8B}
\label{app:trajectories_mm}

Our last example is Mastermind (Appendix~\ref{app:external_envs:games}) with the
small Qwen3-8B actor. The agent must find a hidden 3-digit code (digits 1--6, no
repeats) in 3 guesses per episode, and each instance is played over 3 episodes
with the \emph{same} code. After each guess it receives black pegs (right digit,
right place) and white pegs (right digit, wrong place). Both arms use frozen memory
and identical prompts, apart from the retrieved block. \textsc{HExA} injects four
short strategies and one mistake; Episodic RAG injects two complete past rollouts
from other instances. On instance
\texttt{59cedcb50e280335} (secret code \texttt{5 1 4}), \textsc{HExA} solves all three
episodes, taking 3, 3 and then 1 turn. RAG fails all three. Across the six saved
evaluation runs of each arm, \textsc{HExA} solves this instance in all six and RAG
in two.

Qwen3-8B reasons at length (2k--24k characters per turn), so we show a few verbatim
sentences per turn, with \texttt{....} marking omitted text. The observation after
each guess is reduced to its peg feedback. Highlighting follows
\S\ref{app:trajectories:hexa_run}; in the RAG transcript, red marks the deduction
errors.

\subsection{Retrieved memory (Mastermind instance \texttt{59cedcb50e280335})}
\label{app:trajectories_mm:memory}

\begin{lstlisting}[style=promptwrap]
=== HExA: LEARNED PROBLEM-SOLVING STRATEGIES (titles; examples omitted) ===
1. [Strategy: Cross-Episode Feedback Utilization] Leverage historical feedback to
   eliminate impossible digit combinations across episodes.
2. [Strategy: White Peg Inference] Track digits via white peg feedback to narrow
   possible positions, even if misplaced.
3. [Strategy: Progressive Constraint Narrowing] Systematically eliminate impossible
   combinations by combining black and white peg feedback across guesses.
4. [Strategy: Feedback-Driven Guess Adjustment] Adjust guesses based on all feedback
   from previous attempts, not just immediate results.
Mistake 1. Failing to use white peg feedback to eliminate misplaced digits and track
   code composition across guesses.

=== Episodic RAG: Retrieved Past Attempts (condensed to guesses and feedback) ===
Past attempt 1 --- SUCCESS (instance e7308e8bdfffd6f3, code 3 5 2)
  ep1: 1 2 3 (0B/2W) -> 2 1 4 (0B/1W) -> 3 5 2 SOLVED;  ep2: 1 2 3 -> 3 5 2;  ep3: 3 5 2
Past attempt 2 --- FAILURE (instance b401c1e7503ff9d3)
  ep1: 1 2 3 (0B/1W) -> 2 4 5 (1B/1W) -> 3 4 2;  ep2: 5 4 2 -> 2 5 1 -> 5 2 4;
  ep3: 5 2 4 -> 2 5 4 -> 4 5 2   (all FAILED)
.... The hidden part of the task differs from instance to instance, so nothing above
tells you this instance's answer -- read them for which search behaviour paid off
and which wasted episodes, and do not replay a past action sequence.
\end{lstlisting}

\subsection{Episodic RAG trajectory (3 episodes, never solved)}
\label{app:trajectories_mm:rag_run}

\begin{lstlisting}[style=promptwrapHL]
--- Episode 1, turn 1 ---
Reasoning (6,054 chars; excerpt): .... Looking at the past attempts, the first one was
  successful. .... So the code was 3 5 2. .... @@Wait, but in the first instance, the code@@
  @@was 3 5 2, so maybe that's a common code?@@ ....
Guess: 1 2 3   ->   0B/1W

--- Episode 1, turn 2 ---
Reasoning (10,397 chars; excerpt): .... But wait, in the past attempt, they tried 2 1 4
  after 1 2 3. .... @@Wait, the previous successful attempt used 2 1 4 as the second@@
  @@guess.@@ .... So I'll choose 2 1 4 as the next guess. ....
Guess: 2 1 4   ->   2B/0W
  [[<- replays the retrieved rollout's second guess, although that rollout had seen 0B/2W here, not 0B/1W]]

--- Episode 1, turn 3 ---
Reasoning (10,848 chars; excerpt): .... So the code is 2 _ 4, with the middle digit not
  being 2 or 4. .... ##So the third guess would be 2 3 4.## ....
Guess: 2 3 4   ->   ##out of turns (episode failed)##
  [[<- 3 is already excluded by turn 1; the consistent codes were 2 5 4, 2 6 4, 5 1 4 and 6 1 4]]

--- Episode 2, turn 1 ---
Reasoning (15,927 chars; excerpt): .... ##From Episode 1, the code is 2 1 X.## .... So the
  answer is \boxed{1 3 2}. ....
Guess: 1 3 2   ->   0B/1W
  [[<- 2 1 X contradicts turn 1 (it would put two of 1, 2, 3 in the code)]]

.... condensed (reasoning omitted):
  [Ep 2, turn 2] guess 2 4 1 -> 0B/2W
  [Ep 2, turn 3] guess 2 5 4 -> out of turns (episode failed)
  [Ep 3, turn 1] guess 2 1 5 -> 1B/1W
  [Ep 3, turn 2] guess 2 5 4 -> 1B/1W

--- Episode 3, turn 3 ---
Reasoning (16,951 chars; excerpt): .... Looking at the past attempts, the first instance
  was solved quickly by guessing 3 5 2. .... ##I think I'll have to guess 2 5 4, even##
  ##though it doesn't fit all the clues.## ....
Guess: 2 5 4   ->   ##out of turns (episode failed)##
  [[<- repeats its previous guess; never tests 5 1 4, a code consistent with its own episode-1 feedback]]
\end{lstlisting}

\subsection{\textsc{HExA} trajectory }
\label{app:trajectories_mm:hexa_run}

\begin{lstlisting}[style=promptwrapHL]
--- Episode 1, turn 1 ---
Reasoning (1,965 chars; excerpt): .... Since there's no prior information, I should
  start with a combination that covers different digits to maximize the information from
  the feedback. ....
Guess: 1 2 3   ->   0B/1W

--- Episode 1, turn 2 ---
Reasoning (9,856 chars; excerpt): .... So the code has exactly one of these digits, and
  the other two are different digits not in the guess. ....
Guess: 1 4 5   ->   0B/3W

--- Episode 1, turn 3 ---
Reasoning (14,751 chars; excerpt): .... @@So the code has 1, 4, and 5.@@ .... @@So the code@@
  @@can't have 1 in position 1, 4 in position 2, or 5 in position 3.@@ .... So let's list
  all permutations of 1,4,5 and eliminate those that violate these constraints. ....
Guess: 5 1 4   ->   ##SOLVED##
  [[<- Skill #2 White Peg Inference + #3 Progressive Constraint Narrowing: 3W on 1 4 5 fixes the]]
  [[   digits, the zero black pegs fix the positions]]

--- Episode 2, turn 1 ---
Reasoning (10,796 chars; excerpt): .... @@But the strategies mention leveraging historical@@
  @@feedback.@@ .... So each episode is the same task, but the code is different each time.
  ....
Guess: 1 2 3   ->   0B/1W
  [[<- cites Skill #1 but misreads the setup: the code is fixed across episodes]]

--- Episode 2, turn 2 ---
Reasoning (7,412 chars; excerpt): .... In Episode 1, after guessing 1 2 3 (0B, 1W), then
  1 4 5 (0B, 3W), then 5 1 4 (correct). .... @@But maybe the user is trying to apply the@@
  @@strategies mentioned, like cross-episode feedback.@@ ....
Guess: 1 4 5   ->   0B/3W

--- Episode 2, turn 3 ---
Reasoning (8,347 chars; excerpt): .... @@Looking at the history from Episode 1, the@@
  @@correct code was [5,1,4].@@ .... Therefore, the code is a permutation of 1,4,5. ....
Guess: 5 1 4   ->   ##SOLVED##

--- Episode 3, turn 1 ---
Reasoning (11,652 chars; excerpt): .... @@Looking at the previous episodes, in Episode 1@@
  @@and 2, the correct code was [5, 1, 4].@@ .... Let's choose 5,1,4 as the third guess.
  ....
Guess: 5 1 4   ->   ##SOLVED##
  [[<- Skill #1 Cross-Episode Feedback Utilization: solved in 1 turn]]
\end{lstlisting}

\paragraph{Why a small actor needs rules, not transcripts.}
Both agents open with \texttt{1 2 3} and receive the same 0B/1W. RAG then copies the
second guess of the retrieved success (\texttt{2 1 4}), although that rollout reached
it after different feedback, and despite the prompt's instruction not to replay a
past action sequence. It then draws a deduction its own first
guess rules out (\texttt{2 3 4}). In episodes 2 and 3 it keeps reasoning about
``\texttt{2 1 X}'', a pattern turn~1 already excludes, and ends by repeating a guess
it has just played. It never tries \texttt{5 1 4}, one of only four codes consistent
with its own episode-1 feedback. \textsc{HExA}'s bank holds no past moves, only
how to read the pegs. With Skill~\#2 and Skill~\#3, a probe that scores 0B/3W pins
down the digits and the zero blacks pin down the positions. With Skill~\#1 it
reuses the solved code in later episodes. It misreads the episode structure in
episode~2 and re-derives the code, but by episode~3 it submits \texttt{5 1 4}
immediately.

\section{Working Example: Episodic RAG vs.\ \textsc{HExA} on MathGAP}
\label{app:trajectories_mg}

Our final example is a MathGAP word problem solved by \texttt{gpt-oss-120b}
(reasoning effort \texttt{high}). The problem mixes irrelevant sentences with
count statements, transfers (``X gives N to Y'') and signed-difference equalities;
the answer depends on reading the statements in order, since a transfer changes
the counts that later statements refer to. Held-out item~22 of run~0 is one of the
20 items that \textsc{HExA} answers correctly in all three evaluation runs and
Episodic RAG answers wrongly in all three. The paper evaluations stored only the
parsed answers, so we regenerated this item through each arm's own frozen-evaluation
code, with the frozen round-29 bank and the frozen RAG store and with the generation
seed each arm drew in the paper run. Both regenerations reproduce the logged answers:
\textsc{HExA} gives \textbf{27} (correct) and RAG gives \textbf{79} (wrong).
The reasoning traces run to 40k (\textsc{HExA}) and 32k (RAG) characters; we show
verbatim sentences, with \texttt{....} marking omitted text.

\subsection{Problem (MathGAP held-out item 22)}
\label{app:trajectories_mg:problem}

\begin{lstlisting}[style=promptwrapHL]
Wilbert has 58 more broccolis than Hillary. Boyce has 58 more broccolis than Samuel.
  Richard and Joan have been married for 13 years. Ethelyn is highly introverted and
  prefers minimal communication with others. Jon has 168 broccolis. Clarence has 168
  broccolis. The number of broccolis that Samuel has more than Boyce is equal to the
  difference between the number of broccolis that Wendy and Clarence have. The number of
  broccolis that Hillary has more than Wilbert is equal to the difference between the
  number of broccolis that Vivian and Jon have. Wendy gives 52 broccolis to Richard.
  Samuel is 21 years old. Lois is 46 years old. Joanne has 12 more broccolis than Joan.
  ##Vivian gives 52 broccolis to Addie.## Beatrice is very outgoing and frequently hosts
  parties at home. Clarence is highly introverted and prefers minimal communication with
  others. Addie and Jean are good friends who often go fruit or vegetable picking
  together on weekends. Carrol has 12 more broccolis than Kathrine. Angelica and Wilbert
  are good friends who often go fruit or vegetable picking together on weekends. Hillary
  is 52 years old. Wendy and Olivia are good friends who often go fruit or vegetable
  picking together on weekends. The number of broccolis that Joan has more than Joanne
  is equal to the difference between the number of broccolis that Olivia and Wendy have.
  ##The number of broccolis that Kathrine has more than Carrol is equal to the difference##
  ##between the number of broccolis that Nels and Vivian have.## Vivian tends to be
  laid-back and prefers staying in rather than going out. Jon is 42 years old. Ethelyn
  has 19 more broccolis than Angelica. Wade and Boyce are good friends who often go
  fruit or vegetable picking together on weekends. Joanne is very generous and enjoys
  sharing food with others. Kathrine is very generous and enjoys sharing food with
  others. Beatrice has 19 more broccolis than Lois. Carrol is highly introverted and
  prefers minimal communication with others. Nels is very generous and enjoys sharing
  food with others. The number of broccolis that Angelica has more than Ethelyn is equal
  to the difference between the number of broccolis that Wade and Olivia have. ##The##
  ##number of broccolis that Lois has more than Beatrice is equal to the difference##
  ##between the number of broccolis that Jean and Nels have.## ##How many broccolis does Jean##
  ##have?##
\end{lstlisting}
With Samuel${-}$Boyce${=}{-}58$ and Hillary${-}$Wilbert${=}{-}58$, Wendy and Vivian
start at $168{-}58{=}110$, and both give away 52. The Nels statement comes after
Vivian's transfer, so Nels${=}58{-}12{=}46$ and Jean${=}46{-}19{=}27$. Using
Vivian's count before the transfer gives Nels${=}98$ and Jean${=}79$.

\subsection{Retrieved memory}
\label{app:trajectories_mg:memory}

\begin{lstlisting}[style=promptwrapHL]
=== HExA: LEARNED PROBLEM-SOLVING STRATEGIES (titles only; 10 strategies, 1 mistake) ===
1. Directionality of more/fewer
2. Target variable dependency analysis
3. Solve signed difference equations with unknowns
4. Signed difference equality parsing
@@5. Chronological evaluation@@
6. Sequential transfer handling
7. Conditional transfer handling
8. Implicit zero initialization for unknown persons in transfers
9. Equal redistribution and splitting handling
10. Direct count assignment
Mistake 1. Treating signed difference statements as absolute differences or swapping
  term order

=== Episodic RAG: Retrieved Past Rollouts (bodies omitted) ===
Past rollout 1 --- SUCCESS (seed ...a_6-s_60-idx_19; a tomato puzzle, answer 11)
  .... Given the pattern from previous successful rollout (mathgap), we need to
  interpret statements about "has X tomatoes" as initial counts (i.e., before any
  transfers). ....
Past rollout 2 --- FAILURE (seed ...a_1-s_0-idx_11; answer 79, INCORRECT)
  [[same numbers and structure as the target puzzle, with renamed people and no]]
  [[irrelevant sentences; its correct answer is also 27]]
\end{lstlisting}

\subsection{Episodic RAG reasoning }
\label{app:trajectories_mg:rag_run}

\begin{lstlisting}[style=promptwrapHL]
[reading the retrieved success] .... @@Given the pattern from the successful rollout of a@@
  @@similar puzzle (mathgap)@@, we need to interpret statements like "X has Y more broccolis
  than Z" as a difference: X = Z + Y (or X - Z = Y). We need to track each individual's
  initial count (before any transfers). ....

[which counts do the differences use?] .... In the successful rollout, they interpreted
  relational equations as referring to final counts after all transfers? Then they
  considered relational constraints that refer to counts after any prior transfers? Or
  maybe all relational constraints refer to the final counts after all transfers. @@Given@@
  @@the pattern from previous successful rollout (mathgap), we need to interpret@@
  @@statements about 'has X tomatoes' as initial counts (i.e., before any transfers).@@ ....
  [[<- quoting rollout 1, a different puzzle; the question of timing is never settled from the text itself]]

[Nels, Jean on initial counts] .... ##So -19 = Je_i - 98 => Je_i = 98 - 19 = 79.## Thus
  Jean's initial count = 79. Thus the answer is 79. ##Since there are no transfers##
  ##affecting Jean, final count = 79.## ....
  [[<- Ne_i = 98 uses Vivian's count before she gives away 52 (Vi_i = 110); the Nels statement comes after that transfer]]

[final check] .... But before finalizing, we should verify the pattern from the previous
  successful rollout: they solved a similar puzzle with many statements. Thus answer is
  79. ....
  [[<- 79 is also the recorded (INCORRECT) answer of retrieved rollout 2, the same puzzle with renamed people]]

Final: The final answer is $\boxed{79}$.
\end{lstlisting}

\subsection{\textsc{HExA} reasoning}
\label{app:trajectories_mg:hexa_run}

\begin{lstlisting}[style=promptwrapHL]
[setup] .... Also we need to consider that after transfers, the counts for those persons
  change. @@We must apply transfers in the order they appear.@@ So we need to parse the
  statements in order. ....
  [[<- Skill #5 Chronological evaluation, #6 Sequential transfer handling]]

[Wendy, Vivian before their transfers] .... Thus we can determine Wendy's count before
  any transfers. However note that later Wendy gives 52 to Richard, which will affect
  her count after transfer. @@But the equation (7) uses the counts at that point (before@@
  @@transfer) because it's earlier in order.@@ Thus Vivian's count before any transfers =
  110. ....

[the transfers] .... We have Vivian's count before transfer = 110 (from equation 8).
  ##Thus after transfer, Vivian's count = 58.## ....

[Nels, Jean] .... So -12 = N - Vivian => ##N = Vivian - 12 = 58 - 12 = 46.## Thus Nels = 46.
  So -19 = Je - N => ##Je = N - 19 = 46 - 19 = 27.## Thus Jean = 27. ....

[self-check] .... Now we must confirm that the equations that involve Wendy and Vivian
  (like those that appear after the transfers) indeed use the updated counts. @@According@@
  @@to "Chronological evaluation" strategy, later statements must use counts after earlier@@
  @@statements have been applied.@@ Actually equation 8 (the difference between Vivian and
  Jon) is before the transfer. Let's check the order: The statement about "The number of
  broccolis that Hillary has more than Wilbert is equal to the difference between the
  number of broccolis that Vivian and Jon have." appears before the "Vivian gives 52
  broccolis to Addie." So we used Vivian's pre-transfer count (110) for that equation.
  ....
  [[<- names Skill #5 while re-checking which count each statement sees]]

Final: The final answer is $\boxed{27}$.
\end{lstlisting}

\paragraph{The same puzzle in memory is not enough.}
RAG's store holds this very puzzle, with renamed people, as retrieved rollout~2,
together with its verdict: INCORRECT, 79. It still reproduces that answer. The
retrieved success is a different puzzle, and what RAG takes from it is an
interpretation, that ``has'' statements give counts before any transfer, which it
then applies to the difference statements as well. It debates the timing question
several times but never settles it from the problem text, and it computes Nels
from Vivian's pre-transfer count. \textsc{HExA}'s bank states the rule as a rule
(Skill~\#5: later statements use the counts after all earlier statements have been
applied). The actor applies it while solving and cites it by name while
re-checking, which yields Nels${=}46$ and the correct answer of 27.















\end{document}